\documentclass{article} 
\PassOptionsToPackage{numbers, compress, sort}{natbib}
\usepackage{iclr2025_conference}

\usepackage{minitoc}

\usepackage{xspace}
\usepackage{selectp}
\usepackage[ruled, linesnumbered, vlined]{algorithm2e}

\iclrfinalcopy

\usepackage{graphicx}
\usepackage[acronym]{glossaries}
\newacronym{nas}{NAS}{neural architecture search}
\newacronym{we}{WE}{Weight Entanglement}
\newacronym{ws}{WS}{Weight Sharing}
\newacronym{our method}{}{\textit{tanglenas}}
\newacronym{hpo}{HPO}{hyperparameter optimization}
\usepackage{wrapfig}
\newacronym{dpn}{DPN}{Dual Path Networks}
\newacronym{we1}{WEv1}{Weight-Entanglement V1}
\newacronym{we2}{WEv2}{Weight-Entanglement V2}
\usepackage{xcolor}
\usepackage{amsmath}
\usepackage{amssymb}
\usepackage[bottom]{footmisc}
\usepackage{mathtools}
\usepackage{amsthm}
\usepackage{upgreek}
\usepackage{tipa}

\usepackage[utf8]{inputenc} 
\usepackage[T1]{fontenc}    
\usepackage{url}            
\usepackage{booktabs}       
\usepackage{amsfonts}       
\usepackage{nicefrac}       
\usepackage{microtype}      
\usepackage[backref=page]{hyperref}
\usepackage[nameinlink,capitalise,noabbrev]{cleveref} 
\usepackage{afterpage}

\usepackage{lipsum}                     
\usepackage{xargs}
\usepackage[colorinlistoftodos,prependcaption,textsize=tiny]{todonotes}

\newcommandx{\unsure}[2][1=]{\todo[linecolor=red,backgroundcolor=red!25,bordercolor=red,#1]{#2}}
\newcommandx{\change}[2][1=]{\todo[linecolor=blue,backgroundcolor=blue!25,bordercolor=blue,#1]{#2}}
\newcommandx{\info}[2][1=]{\todo[linecolor=OliveGreen,backgroundcolor=OliveGreen!25,bordercolor=OliveGreen,#1]{#2}}
\newcommandx{\improvement}[2][1=]{\todo[linecolor=Plum,backgroundcolor=Plum!25,bordercolor=Plum,#1]{#2}}
\newcommandx{\thiswillnotshow}[2][1=]{\todo[disable,#1]{#2}}

\usepackage{times}
\usepackage{graphicx}
\usepackage{subcaption}
\usepackage{amssymb}
\usepackage{mathtools}
\usepackage{amsthm}
\usepackage{xspace}
\usepackage{selectp}

\usepackage{url}
\usepackage{microtype}
\usepackage{booktabs}
\usepackage{csquotes}

\usepackage{comment}
\usepackage[utf8]{inputenc} 
\usepackage[T1]{fontenc}    
\usepackage{url}            
\usepackage{booktabs}       
\usepackage{amsfonts}       
\usepackage{nicefrac}       
\usepackage{microtype}      
\usepackage{siunitx}

\usepackage{algpseudocode}
\usepackage{multicol}
\usepackage{mathtools}
\usepackage{caption}
\usepackage{float}
\usepackage{wrapfig}
\usepackage[scaled=0.85]{beramono}
\usepackage{multirow}
\usepackage{mathrsfs}
\usepackage{newfloat}
\usepackage{relsize}
\usepackage{enumitem}
\usepackage{pifont}
\usepackage{caption}

\usepackage{atbegshi,etoolbox}
\makeatletter
\newcommand{\discardpages}[1]{
  \xdef\discard@pages{#1}
  \AtBeginShipout{
    \renewcommand*{\do}[1]{
      \ifnum\value{page}=##1\relax%
        \AtBeginShipoutDiscard
        \gdef\do####1{}
      \fi%
    }%
    \expandafter\docsvlist\expandafter{\discard@pages}
  }%
}
\newif\ifkeeppage
\newcommand{\keeppages}[1]{
  \xdef\keep@pages{#1}
  \AtBeginShipout{
    \keeppagefalse%
    \renewcommand*{\do}[1]{
      \ifnum\value{page}=##1\relax%
        \keeppagetrue
        \gdef\do####1{}
      \fi%
    }%
    \expandafter\docsvlist\expandafter{\keep@pages}
    \ifkeeppage\else\AtBeginShipoutDiscard\fi
  }%
}
\makeatother

\usepackage{floatflt}

\newcommand{\improvementimgnet}[1]{{xx}}
\newcommand{\paramsfactor}[1]{{35}}
\newcommand{\flopsfactor}[1]{{xx}}

\newcommand{\xaxis}[1]{{error}}
\newcommand{\Xaxis}[1]{{Error}}
\newcommand{\hypernetwork}{hypernetwork\xspace}
\newcommand{\metahypernet}{\texttt{MetaHypernetwork}\xspace}
\newcommand{\architect}{\texttt{Architect}\xspace}
\newcommand{\metapredictor}{\texttt{MetaPredictor}\xspace}
\newcommand{\supernet}{\texttt{Supernetwork}\xspace}

\DeclareMathOperator*{\argmin}{argmin}
\DeclareMathOperator*{\argmax}{argmax}

\newcommand{\methodname}{MODNAS \xspace}

\newcommand{\Y}{\mathcal{Y}}
\newcommand{\X}{\mathcal{X}}

\newcommand{\bigO}{\mathcal{O}}

\newcommand\normx[1]{\left\Vert#1\right\Vert}

\newcommand{\bx}{\bm{x}}

\newcommand{\bw}{\bm{w}}
\newcommand{\br}{\bm{r}}
\newcommand{\bbR}{\mathbb{R}}

\newcommand{\dataset}{\mathcal{D}}
\newcommand{\Loss}{\mathcal{L}}
\newcommand{\Lossv}{\mathbf{L}}

\newcommand{\archss}{\mathcal{A}}
\newcommand{\opss}{\mathcal{O}}

\newcommand{\pareto}{\mathcal{P}}
\newcommand{\sspace}{\mathcal{S}}
\newcommand{\oscript}{m}
\newcommand{\Oscript}{M}
\newcommand{\tscript}{t}
\newcommand{\Tscript}{T}

\theoremstyle{plain}
\newtheorem{theorem}{Theorem}[section]

\theoremstyle{definition}
\newtheorem{definition}[theorem]{Definition}

\theoremstyle{remark}

\usepackage{amsmath,amsfonts,bm}

\def\eqref#1{equation~\ref{#1}}

\def\1{\bm{1}}

\DeclareMathAlphabet{\mathsfit}{\encodingdefault}{\sfdefault}{m}{sl}
\SetMathAlphabet{\mathsfit}{bold}{\encodingdefault}{\sfdefault}{bx}{n}

\author{Rhea Sanjay Sukthanker$^1$\thanks{Equal contribution. Email to: \texttt{\{sukthank, zelaa\}@cs.uni-freiburg.de}} , ~Arber Zela$^1$\footnotemark[1] , ~Benedikt Staffler$^2$, ~Samuel Dooley$^3$, \\
    \textbf{Josif Grabocka$^4$, Frank Hutter$^{1,5}$}
    \vspace*{1mm}\\
    $^1$ University of Freiburg, $^2$ Bosch Center for AI, $^3$ Meta,\\ $^4$ University of Technology Nuremberg, $^5$ ELLIS Institute Tübingen
}

\discardpages{45}

\begin{document}

\doparttoc 
\faketableofcontents 


\title{Multi-objective Differentiable Neural \\Architecture Search}
\maketitle

\setlength{\parskip}{4pt}

\begin{abstract}
Pareto front profiling in multi-objective optimization (MOO), i.e., finding a diverse set of Pareto optimal solutions, is challenging, especially with expensive objectives that require training a neural network. Typically, in MOO for neural architecture search (NAS), we aim to balance performance and hardware metrics across devices. Prior NAS approaches simplify this task by incorporating hardware constraints into the objective function, but profiling the Pareto front necessitates a computationally expensive search for each constraint. In this work, we propose a novel NAS algorithm that encodes user preferences to trade-off performance and hardware metrics, yielding representative and diverse architectures across multiple devices in just a single search run. To this end, we parameterize the joint architectural distribution across devices and multiple objectives via a hypernetwork that can be conditioned on hardware features and preference vectors, enabling zero-shot transferability to new devices. Extensive experiments involving up to 19 hardware devices and 3 different objectives demonstrate the effectiveness and scalability of our method. Finally, we show that, without any additional costs, our method outperforms existing MOO NAS methods across a broad range of qualitatively different search spaces and datasets, including MobileNetV3 on ImageNet-1k, an encoder-decoder transformer space for machine translation and a decoder-only space for language modelling.
\end{abstract}

\section{Introduction}
\label{sec:introduction}

The ability to make good tradeoffs between predictive accuracy and efficiency (in terms of latency and/or energy consumption) has become crucial in an age of ever increasing neural networks complexity and size \citep{kaplan2020scaling,hoffmann2022training,zhai2022scaling,alabdulmohsin2023getting} and a plethora of embedded devices. However, finding the right trade-off remains a challenging task that typically requires human intervention and a lot of trial-and-error across devices. With multiple conflicting objectives, it becomes infeasible to optimize all of them simultaneously and return a single solution.
Ideally, NAS should empower users to choose from a set of diverse Pareto optimal solutions that represent their preferences regarding the trade-off between objectives.\looseness=-1

Neural Architecture Search (NAS)~\citep{white-arxiv23} provides a principled framework to search for network architectures in an automated fashion. 
Several works \citep{elsken2018efficient,cai-iclr2020,wang2020hat,chen2021autoformer} have extended NAS for multi-objective optimization (MOO), considering performance and hardware efficiency metrics like latency and energy consumption.
However, to the best of our knowledge, no existing gradient-based method returns the full Pareto front for the MOO problem at hand \emph{without repeating their search routine multiple times with different hardware constraints}. 

In this work, we propose a scalable and hardware-aware \textbf{M}ulti-\textbf{O}bjective \textbf{D}ifferentiable \textbf{N}eural \textbf{A}rchitecture \textbf{S}earch (\textbf{MODNAS}) algorithm that efficiently trains a single supernet which can be used to read off 
Pareto-optimal solutions for different user preferences and different target devices,
without any additional search steps.
To search across devices, we frame the problem as a multi-task, multi-objective optimization problem, where each task (device) has multiple (conflicting) objectives, e.g., classification accuracy and latency.
The user's preferences are modelled by a \emph{preference vector} that defines a scalarization (weighted sum), of the different objectives. This preference vector, along with features of the hardware of interest, is fed to a \hypernetwork~\citep{ha-iclr17} that outputs continuous architectural parameters $\alpha$. To search in the space of architectures, we employ a one-shot model and a bi-level optimization scheme, as is typically done in gradient-based NAS. In our case, however, the upper-level parameters are the \hypernetwork weights, optimized in expectation over different preference vectors and hardware devices via multiple gradient descent~\citep{desideri2012mgd}.

\begin{figure*}[t!]
    \includegraphics[width=\linewidth]{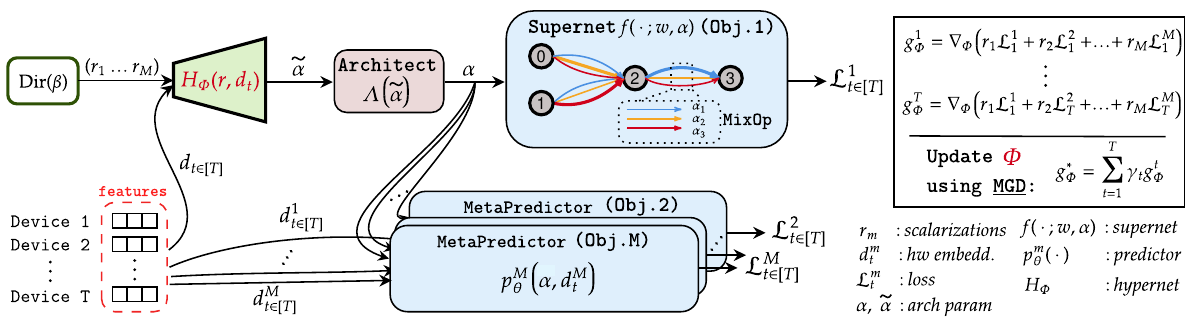}
    \caption{
    \textbf{MODNAS overview.} Given a set of $T$ devices, MODNAS seeks to optimize $M$ (potentially conflicting) objectives across these devices. To this end, it employs a \metahypernet $H_\Phi (r, d_t)$, that takes as input a scalarization $r$, representing the user preferences, and a device embedding $d_t$, to yield an un-normalized architectural distribution $\Tilde{\alpha}$. The \architect uses $\Tilde{\alpha}$ to sample differentiable discrete architectures, used in the \supernet to estimate accuracy and in the \metapredictor to estimate the other $M-1$ loss functions (e.g. latency, energy consumption) for every device. By iterating over devices and sampling scalarizations uniformly from the $M$-dimensional simplex, at each iteration we update the \metahypernet using multiple gradient descent (MGD).
    }
    \label{fig:modnas overvies}
    \vspace{-2ex}
\end{figure*}

To evaluate our method, we conduct experiments on multiple NAS search spaces, including CNN and Transformer architectures, and up to 3 objectives across 19 hardware devices. While other NAS methods that utilize hardware constraints in their search objectives require substantial search costs both for each new constraint and each new hardware, MODNAS addresses both in a zero-shot manner, without extra search cost, while yielding higher quality solutions.

Our contributions can be summarized as follows:
\begin{enumerate}[leftmargin=*]
    \item We present a principled approach for \textbf{Multi-objective Differentiable NAS}, that leverages \emph{hypernetworks} and \emph{multiple gradient descent} for fast Pareto-Front approximation across devices.
    \item This work is the \emph{first} to provide a global view of the Pareto solutions with \textbf{just a single search run}, without the need to repeat search or fine-tune on new target devices.
    \item Extensive evaluation of our method across \textbf{4 different search spaces} (NAS-Bench-201, MobileNetV3, an encoder-decoder and a decoder-only Transformer space), \textbf{3 tasks} (image classification, machine translation and language modeling), and \textbf{up to 19 hardware devices and 3 objectives}, show both improved \emph{efficiency} and \emph{performance} in comparison to previous approaches that use a constrained objective in their search.
\end{enumerate}
To facilitate reproducibility, we provide our code in \url{https://github.com/automl/modnas}.

\section{Background and Related Work}
\label{sec:background}
In this section, before describing our algorithm, we introduce some basic concepts, definitions and related work. Refer to Appendix~\ref{sec:related_work} for an extended related work.

\textbf{Multi-objective optimization (MOO) for Multi-Task Learning.} Consider a multi-task dataset $\dataset$ consisting of $N$ instances, where the feature vector of the $i-$th instance is denoted as $x_i \in \X$, and the $\Oscript$-many associated target variables as $y_i^{1} \in \Y^1, \dots, y_i^{\Oscript} \in \Y^{\Oscript}$. Moreover, consider there exists a family of parametric models $f (\bx; \bw): \X \rightarrow \{ \Y^1 \times  \dots \times \Y^\Oscript\}$, parameterized by $\bw$, that maps the input $\bx$ to the joint space of the multiple tasks. To simplify the notation, we denote the prediction of the $\oscript$-th task as $f^\oscript(\bx; \bw): \X \rightarrow \Y^\oscript$, and the respective loss $\Loss^\oscript(\bw) \triangleq \frac{1}{N}\sum_i^N \ell^\oscript(y_i^\oscript, f^\oscript(x_i; \bw))$. The vector of the values of all loss functions is denoted as $\Lossv(\bw) \triangleq ( \Loss^1(\bw),\dots , \Loss^\Oscript(\bw))$. MOO then seeks to find a set of Pareto-optimal solutions $\bw^*$ that jointly minimize $\Lossv(\bw)$\footnote{$\bw$ can be replaced with any other parameter here, also architectural ones (see Section~\ref{sec:methodology}).}:
\begin{equation}
    \bw^* \in \argmin_{\bw} \Lossv(\bw)
    \label{eq:moo}
\end{equation}

\begin{definition}{(Pareto Optimality):}
A solution $\bw_2$ dominates $\bw_1$ iff $\Loss^\oscript(\bw_2) \leq \Loss^\oscript(\bw_1)$, $\forall \oscript \in \{1,\dots,\Oscript\}$, and $\Lossv(\bw_1) \neq \Lossv(\bw_2)$. In other words, a dominating solution has a lower loss value on at least one task and no higher loss value on any task. A solution $\bw^*$ is called \emph{Pareto optimal} iff there exists no other solution dominating $\bw^*$. 
\end{definition}

\begin{definition}{(Pareto front):}
The sets of Pareto optimal points and their function values are called \emph{Pareto set} ($\pareto_{\bw}$) and \emph{Pareto front} ($\pareto_{\Lossv} = \{ \Lossv(\bw)_{\bw\in\pareto_{\bw}} \}$), respectively.
\end{definition}

\textbf{Linear Scalarization.} In MOO, a standard technique to solve the $\Oscript$-dimensional problem is using a preference vector $\br \in \sspace \triangleq \{ \bbR^{\Oscript} | \sum_{\oscript = 1}^{\Oscript} r_{\oscript} = 1, r_\oscript \geq 0, \forall \oscript \in \{1,\dots,\Oscript\}\}$ in the $\Oscript$-dimensional probability simplex~\citep{lin-neurips19, mahapatra-icml20a, ruchte2021scalable}. 
Every $\br\in\sspace$ yields a convex combination of the loss functions in Equation~\ref{eq:moo} as $\Loss_{\br}(\bw) = \br^{\mathbf{T}} \Lossv(\bw)$. Given a preference vector $\br$, one can apply standard, single-objective optimization algorithms to find a minimizer $\bw_{\br}^* = \argmin_{\bw} \Loss_{\br}(\bw)$. By sampling multiple $\br$ vectors, one can compute Pareto-optimal solutions $\bw_{\br}^*$ that profile the Pareto front. Several methods~\citep{Lin2020ControllablePM, navon2020learning, hoang2023improving, phan2022stochastic} employ a hypernetwork~\citep{ha-iclr17} to generate Pareto-optimal solutions given different preference vectors as input.
In this work, we utilize a hypernetwork conditioned on scalarizations to generate Pareto-optimal architectures. Furthermore, we also extend the hypernetwork by conditioning it on different task vectors. 

\textbf{Multiple Gradient Descent (MGD).} MOO can be solved to local optimality via MGD~\citep{desideri2012mgd}, as a natural extension of single-objective gradient descent, which iteratively updates $\bw$ towards a direction that ensures that all tasks improve simultaneously (called \emph{Pareto improvement}): $\bw^{\prime} \leftarrow \bw - \xi g_{\bw}^*$, where $g_{\bw}^*$ is a vector field that needs to be determined. If we denote by $g_{\bw}^\oscript = \nabla_{\bw} \Loss^\oscript(\bw)$ the gradient of the $\oscript$-th scalar loss function, via Taylor approximation, the decreasing direction of $\Loss^\oscript$ when we update $\bw$ towards $g_{\bw}^*$ is given by $\langle g_{\bw}^\oscript , g_{\bw}^* \rangle \approx - (\Loss^\oscript (\bw^{\prime}) - \Loss^\oscript (\bw)) / \xi$. In MGD $g_{\bw}^*$ is chosen to maximize the slowest update rate among all objectives:
\begin{equation}
    g_{\bw}^* \propto \argmax_{g_{\bw}\in \bbR^d, ||g_{\bw}|| \leq 1} \Big\{ \min_{\oscript\in [\Oscript]} \langle g_{\bw}, g_{\bw}^\oscript \rangle\Big\}.
    \label{eq:mgd_opt}
\end{equation}
The early work of \citet{desideri2012mgd} has been extended in various settings, particularly multi-task learning, with great promise \citep{Sener2018MultiTaskLA, lin-neurips19, mahapatra-icml20a, Liu2019TheSM}, but these approaches are applied to mainly a fixed architecture and extending them to a supernet subsuming a search space of multiple architectures is non-trivial. 

\textbf{One-shot NAS and Bi-Level optimization.}
With the architecture space being intrinsically discrete, large (often consisting of upto $10^{36}$ architectures) and hence expensive to search on, most existing differentiable NAS approaches leverage the weight sharing paradigm and continuous relaxation to enable gradient descent~\citep{liu-iclr19, pham-icml18, bender-icml18a, xie-iclr18, xu-arxiv19, dong-cvpr19,Chen2020DrNASDN,liu2023bridging,movahedi2022lambda, Zhang2021iDARTSDA}. Typically, in these approaches, architectures are stacks of cells, where the cell structure is represented as a directed acyclic graph (DAG) with $N$ nodes and $E$ edges. Every transition from node $i$ to $j$, i.e. edge $(i,j)$, is associated with an operation $o^{(i,j)}\in\opss$, where $\opss$ is a predefined candidate operation set. \citet{liu-iclr19} proposed a continuous relaxation of the search space by parameterizing the discrete operation choices in the DAG edges via a learnable vector $\alpha$.
This enables framing the NAS problem as a bi-level optimization one, with differentiable objectives w.r.t. all variables:
\begin{equation}
  \begin{aligned}
    &\argmin_\alpha \Loss^{val}(\bw^*(\alpha), \alpha)
    &s.t.\quad \bw^*(\alpha) = \argmin_{\bw}\Loss^{train}(\bw, \alpha),
  \end{aligned}
  \label{eq:bilevel}
\end{equation}
where $\Loss^{train}$ and $\Loss^{val}$ are the empirical losses on the training and validation data, respectively, $\bw$ are the supernetwork parameters, 
$\alpha \in \archss$ are the continuous architectural parameters, and $\bw^*(\alpha): \archss \rightarrow \bbR^d$ is a best response function that maps architectures to their optimal weights.

\textbf{Comparison to single-objective constrained NAS.}
Early NAS methods predominantly targeted high accuracy, whereas contemporary hardware-aware differentiable NAS approaches~\citep{wu2019fbnet,wan2020fbnetv2,cai2018proxylessnas,wu2021trilevel,fu2020autogan, xu2020latency, jiang2021eh, wang2021attentivenas} are designed to identify architectures optimized for target hardware efficiency. Typically, these methods integrate hardware constraints within their objectives, yielding a single optimal solution and necessitating multiple search iterations to construct the Pareto front. Our proposed algorithm addresses this by profiling the entire Pareto front in a \emph{single search iteration}. While single-objective constrained optimization is advantageous in scenarios demanding optimization of one objective under a specific constraint, practical applications often require a suite of models adaptable to varying user preferences even on a single device. Our efficient Pareto-front approximation algorithm provides such a suite of optimal models to choose from. 

\section{Hardware-aware Multi-objective Differentiable Neural Architecture Search}
\label{sec:methodology}

We first formalize the multi-objective bi-level optimization NAS problem across multiple hardware devices, and then introduce a scalable and differentiable method that combines MGD with linear scalarizations to efficiently solve this problem.
\vspace{-0.5ex}
\subsection{Problem Definition \& Sketch of Solution Approach} 
In multi-objective NAS, the bi-level problem described in Equation~\ref{eq:moo} becomes more difficult, since we are not only concerned with finding $\bw^*$ given a fixed architecture, but we want to optimize in the space of architectures $\archss$ as well.
Assuming we have $\Tscript$ hardware devices (target functions) and $\Oscript$ objectives (e.g. accuracy, latency, etc.), similar to (\ref{eq:bilevel}), for every $t\in \{1\dots T \}$, the Pareto set can be obtained by solving the following bi-level optimization problem:
\begin{equation}
\begin{aligned}
    &\argmin_\alpha \Lossv_{\tscript}^{valid} (\bw^*(\alpha), \alpha)
    &s.t.\quad \bw^*(\alpha) = \argmin_{\bw}\Lossv_{\tscript}^{train} (\bw, \alpha),
\end{aligned}
\label{eq:bilevel_moo}
\end{equation}
where the $\Oscript$-dimensional loss vector $\Lossv_{\tscript}(\bw^*(\alpha), \alpha) \triangleq \big( \Loss_\tscript^1(\bw^*(\alpha),\alpha), \dots , \Loss_{\tscript}^\Oscript(\bw^*(\alpha),\alpha) \big)$ is evaluated $\forall \tscript\in \{1, \dots, \Tscript \}$. $\Lossv_{\tscript}^{train}$ and $\Lossv_{\tscript}^{valid}$ are the vectors with all $\Oscript$ loss functions evaluated on the train and validation splits of $\dataset$, used in the lower- and upper-level problems of (\ref{eq:bilevel_moo}), respectively.

\begin{wrapfigure}[18]{R}{.36\textwidth}
    \centering
    \vspace{-12pt}
    \includegraphics[width=.99\linewidth]{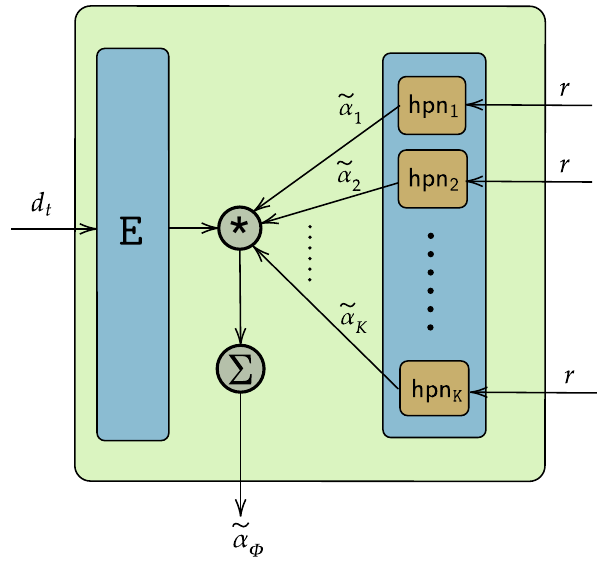}
    \vspace{-4ex}
    \caption{Architecture overview of the \metahypernet, which gets as input a device embedding $d_t$ (input to an embedding layer $\mathtt{E}$) and a scalarization $\br$ (input to K hypernetworks) and yields an architecture encoding $\Tilde{\alpha}$. 
    }
    \label{fig:hpn_arch}
\end{wrapfigure}

Our goal is to find Pareto-optimal architectures for each target device, covering diverse and representative preferences for different objectives. However, naively solving ($\ref{eq:bilevel_moo}$) for each device $\tscript$ requires $\Tscript$ independent bi-level searches, making this very inefficient for large models. To overcome this, we incorporate a single \hypernetwork within the one-shot model (supernetwork) commonly used in conventional NAS~\citep{bender-icml18a, pham-icml18, liu-iclr19}. This allows us to generate architectures based on device embeddings and preference vectors in just \emph{one search run}, reducing the search cost from $\bigO$(T) to $\bigO$(1).

\subsection{Algorithm Design and Components}

Our search procedure is composed of four core modular components (see Figure \ref{fig:modnas overvies}): (1) a \metahypernet that generates the architectural distribution; (2) an \texttt{Architect} that samples discrete architectures from this distribution; (3) a \texttt{Supernetwork} that exploits the weight-sharing paradigm for search efficiency and serves as a proxy for the network accuracy; and (4) a \texttt{MetaPredictor} that predicts hardware metrics and enables gradient propagation back to the \metahypernet. We now discuss each of these in detail.

\textbf{\metahypernet.}
In order to generate architectures across multiple devices, inspired by \citet{wang2022learning} and \citet{Lin2020ControllablePM}, we propose a \metahypernet that can meta-learn across different hardware devices (see Figure \ref{fig:hpn_arch}). 
Hypernetworks are a class of neural networks that generate the
parameters of another model. They were initially proposed for model compression~\citep{ha-iclr17} and were later adopted for NAS~\citep{Brock2017SMASHOM} and MOO~\citep{navon2020learning, Lin2020ControllablePM}. Here, given a preference vector $\br = (r_1,\dots , r_M)$ and a hardware device feature vector $d_\tscript$, for device $\tscript\in \{1,\dots,\Tscript \}$, we use the \metahypernet $H_\Phi(\br, d_\tscript)$, parameterized by $\Phi$, to generate an un-normalized architecture distribution $\Tilde{\alpha}_\Phi$ that is later used to compute the upper-level updates in ($\ref{eq:bilevel_moo}$). Similar to \citet{lee2021help}, $d_\tscript$ is a fixed-size feature vector that is obtained by evaluating a fixed set of reference architectures on device $\tscript$.
The \metahypernet is composed of 2 main components (see Figure~\ref{fig:hpn_arch}): 
\begin{enumerate}[leftmargin=*]
    \item \emph{A bank of $K$ independent hypernetworks}: $\mathtt{hpn}_1,\dots , \mathtt{hpn}_K$, that parse the preference vector $\br$ and generate the architectural parameters $\Tilde{\alpha}_1, \dots, \Tilde{\alpha}_K$, respectively.
    \item \emph{A linear layer} $\mathtt{E}$, that learns a similarity map from 
    device feature vectors to the bank of $\mathtt{hpn}$s. $\mathtt{E}$ takes as input the device feature vector $d_\tscript$ and outputs an attention vector of size $K$.
\end{enumerate}
The final output, $\Tilde{\alpha}_\Phi$, of the \metahypernet is computed as a weighted sum of the outputs of the $K$ hypernetworks, where the vector of weights is the output of the linear layer $\mathtt{E}$. For a more detailed description of the \metahypernet we refer the reader to Appendix~\ref{sec:hpn_details}.

In all experiments, we initialize the \metahypernet to yield a uniform probability mass over all architectural parameters for all scalarizations and device embeddings. 
By using the preference vector $\br$ to create a linear scalarization of $\Lossv_\tscript$ and the \metahypernet to model the architectural distribution across $\Tscript$ devices, the bi-level problem in ($\ref{eq:bilevel_moo}$) reduces to:
\begin{equation}
    \begingroup\small
    \begin{aligned}
        &\argmin_\Phi \mathbb{E}_{\br \sim \sspace} \big [ \br^{\mathbf{T}} \Lossv_{\tscript}^{valid} (\bw^*(\alpha_\Phi), \alpha_\Phi) \big ] 
        &s.t.\quad \bw^*(\alpha_\Phi) = \argmin_{\bw} \mathbb{E}_{\br \sim \sspace} \big [ \br^{\mathbf{T}} \Lossv_{\tscript}^{train} (\bw, \alpha_\Phi) \big ],
    \end{aligned}
    \endgroup
    \label{eq:bilevel_moo_ls}
\end{equation}
where $\alpha_\Phi$ are the normalized architectural parameters obtained from the \architect $\Lambda (\Tilde{\alpha}_\Phi)$ and $\br^{\mathbf{T}} \Lossv_{\tscript}(\cdot, \alpha_\Phi) = \sum_{\oscript=1}^\Oscript r_m \Loss_\tscript^\oscript(\cdot,\alpha_\Phi)$ is the scalarized loss for device $t$. Conditioning the \metahypernet on the hardware embeddings allows us to generate architectures on new test devices without extra finetuning or meta-learning steps. Iniutively, the \metahypernet, learns to map the new test device, to the most similar device, in its learnt bank of embeddings (see also Figure~\ref{fig:tsne_combined} in the appendix). We use the \emph{Dirichlet} distribution $Dir(\beta)$, $\beta = (\beta_1, \dots, \beta_\Oscript)$, to sample the preference vectors and approximate the expectation over the scalarizations using Monte Carlo sampling. In our experiments, we set $\beta_1 = \dots = \beta_\Oscript = 1$, for a uniform sampling over the $(M-1)$-simplex, however, one can set these differently based on user priors or make it a learnable parameter~\citep{Chen2020DrNASDN}. 

\begin{wrapfigure}[24]{R}{0.63\textwidth} 
\vspace{-2.7ex}
\hspace{0.9em}
\resizebox{0.95\linewidth}{!}{%
\begin{algorithm}[H]
\KwData{$\dataset_{train}$; $\dataset_{valid}$; \supernet; device features $\{ d_t \}_{t=1}^T$; \metahypernet $H_{\Phi}$; nr. of objectives $M$; \architect $\Lambda$; learning rates $\xi_1$, $\xi_2$.}
\While{$not\ converged$}{
    \For{$\tscript\in \{1,\dots,\Tscript \}$}{
        Sample scalarization $\br \sim Dir(\beta)$\\
        Set arch params $\Tilde{\alpha}_\Phi \gets H_{\Phi} (\br, d_t)$ \\
        Sample $\alpha_\Phi \sim \Lambda (\Tilde{\alpha}_\Phi)$ from \architect \\
        $g_{\Phi}^t \gets \sum_{\oscript=1}^\Oscript r_m \nabla_\Phi \Loss_\tscript^\oscript(\dataset_{valid}; \bw,\alpha_\Phi)$ \\
    }
    $\gamma \gets \mathtt{FrankWolfeSolver}(g_\Phi^1, \dots, g_\Phi^T)$ \tcp*{Alg.\ref{alg:fw_algo}}
    $g_{\Phi}^* \gets \sum_{t=1}^T \gamma_t \cdot g_{\Phi}^t$ \\
    $\Phi \gets \Phi - \xi_1 \cdot g_{\Phi}^*$  \tcp*{update $\metahypernet$}
    \For{$\tscript\in \{1,\dots,\Tscript \}$}{
        Sample scalarization $\br \sim Dir(\beta)$\\
        Set arch params $\Tilde{\alpha}_\Phi \gets H_{\Phi} (\br, d_t)$ \\
        Sample $\alpha_\Phi \sim \Lambda (\Tilde{\alpha}_\Phi)$ from \architect \\
        $g_{\bw}^t \gets \sum_{\oscript=1}^\Oscript r_m \nabla_{\bw} \Loss_\tscript^\oscript(\dataset_{train}; \bw,\alpha_\Phi)$ \\
    }
    $g_{\bw}^* \gets \frac{1}{T} \sum_{t=1}^T g_{\bw}^t$ \\
    $\bw \gets \bw - \xi_2 \cdot g_{\bw}^*$ \tcp*{update $\supernet$}  \label{lst:line14}
}
\Return{$H_{\Phi}$}
\caption{\texttt{MODNAS}}
\label{alg:modnas_algo}
\end{algorithm}
}
\end{wrapfigure}

\textbf{\metapredictor.}
For the cheap-to-evaluate hardware objectives, such as latency, energy consumption, we employ a regression model 
$p_\theta^\oscript (\alpha, d_\tscript^\oscript)$ that predicts the target labels $y_\tscript^\oscript$ for objective $\oscript$ and device $\tscript$, given an architecture $\alpha$ and device embedding $d_\tscript^\oscript$. We use the same predictors as \citet{lee2021help} and optimize the MSE loss: $\min_\theta \mathbb{E}_{\alpha\sim\archss, t\sim [T]} \big(y_\tscript^\oscript - p_\theta^\oscript (\alpha, d_\tscript^\oscript)\big)^2$, as done in \citet{Lee2021RapidNA} for meta-learning performance metrics across datasets.
In our experiments, we pretrain a separate \metapredictor for every hardware objective $m$ (e.g. latency, energy, etc.) on a subset of $(\alpha, y_\tscript^\oscript)$ pairs, and use its predicted value directly in ($\ref{eq:bilevel_moo_ls}$) as $\Loss_\tscript^\oscript(\cdot,\alpha_\Phi) = p_\theta^\oscript (\alpha_\Phi, d_\tscript^\oscript)$. Since the \metapredictor is in principle a small neural network this pretraining step is inexpensive. During search, we freeze and do not further update the $\metapredictor$ parameters $\theta$. 

\texttt{\textbf{Supernetwork}}.
For expensive objectives like neural network classification accuracy, we use a \supernet that encodes the architecture space and shares parameters between architectures, providing a best response function $\bw^* (\alpha_\Phi)$ for the scalarized loss in (\ref{eq:bilevel_moo_ls}). While any parametric model could estimate this function, such as performance predictors~\citep{Lee2021RapidNA}, this requires an expensive prior step of creating the training dataset for the predictor, by training architectures from scratch. To reduce memory costs of \supernet training, we: (1) use a one-hot encoding of $\alpha_\Phi$ for differentiable architecture sampling~\citep{dong-cvpr19, cai2018proxylessnas, xie-iclr18}, activating only one architecture per step, and (2) entangle operation choice parameters in the \supernet, further increasing memory efficiency of the supernetwork beyond weight sharing~\citep{sukthanker2023weight}.

\textbf{\architect.}
The \architect $\Lambda (\Tilde{\alpha})$ samples discrete architectural configurations from the un-normalized distribution $\Tilde{\alpha}_\Phi = H_\Phi(\br, d_\tscript)$ and enables gradient estimation through discrete variables for $\nabla_\Phi \Lossv_t(\cdot, \alpha_\Phi)$. Methods such as GDAS~\citep{dong-cvpr19} utilize the Straight-Through Gumbel-Softmax (STGS) estimator~\citep{jang-iclr17a}, that integrates the Gumbel reparameterization trick to approximate the gradient. Here we employ the recently proposed \emph{ReinMax} estimator~\citep{liu2023bridging}, that yields second-order accuracy without the need to compute second-order derivatives. See Appendix~\ref{sec:discrete_samplers} for more details on these discrete samplers.
Similar to the findings in \citet{liu2023bridging}, in our initial experiments, ReinMax outperforms the GDAS' STGS estimator (see Figure~\ref{fig:radar_plot_gdas} in the Appendix), therefore, we use ReinMax in all experiments that follow. 

\subsection{Optimizing the \metahypernet via MGD}

We denote the gradient of the scalarized loss in (\ref{eq:bilevel_moo_ls}) with respect to the \metahypernet parameters $\Phi$, shared across all devices $\tscript\in {1, \dots, \Tscript }$, as: $g_\Phi^\tscript = \br^{\mathbf{T}} \nabla_\Phi \Lossv_{\tscript}(\cdot, \alpha_\Phi) = \sum_{\oscript=1}^\Oscript r_m \nabla_\Phi \Loss_\tscript^\oscript(\cdot,\alpha_\Phi)$, where $\alpha_\Phi$ is the discrete architectural sample from the \architect $\Lambda (\Tilde{\alpha}\Phi)$.
{
\parfillskip=0pt
\parskip=0pt
\par}
\begin{wrapfigure}[25]{R}{.49\textwidth}
\centering
\vspace{-3ex}
\includegraphics[width=.999\linewidth]{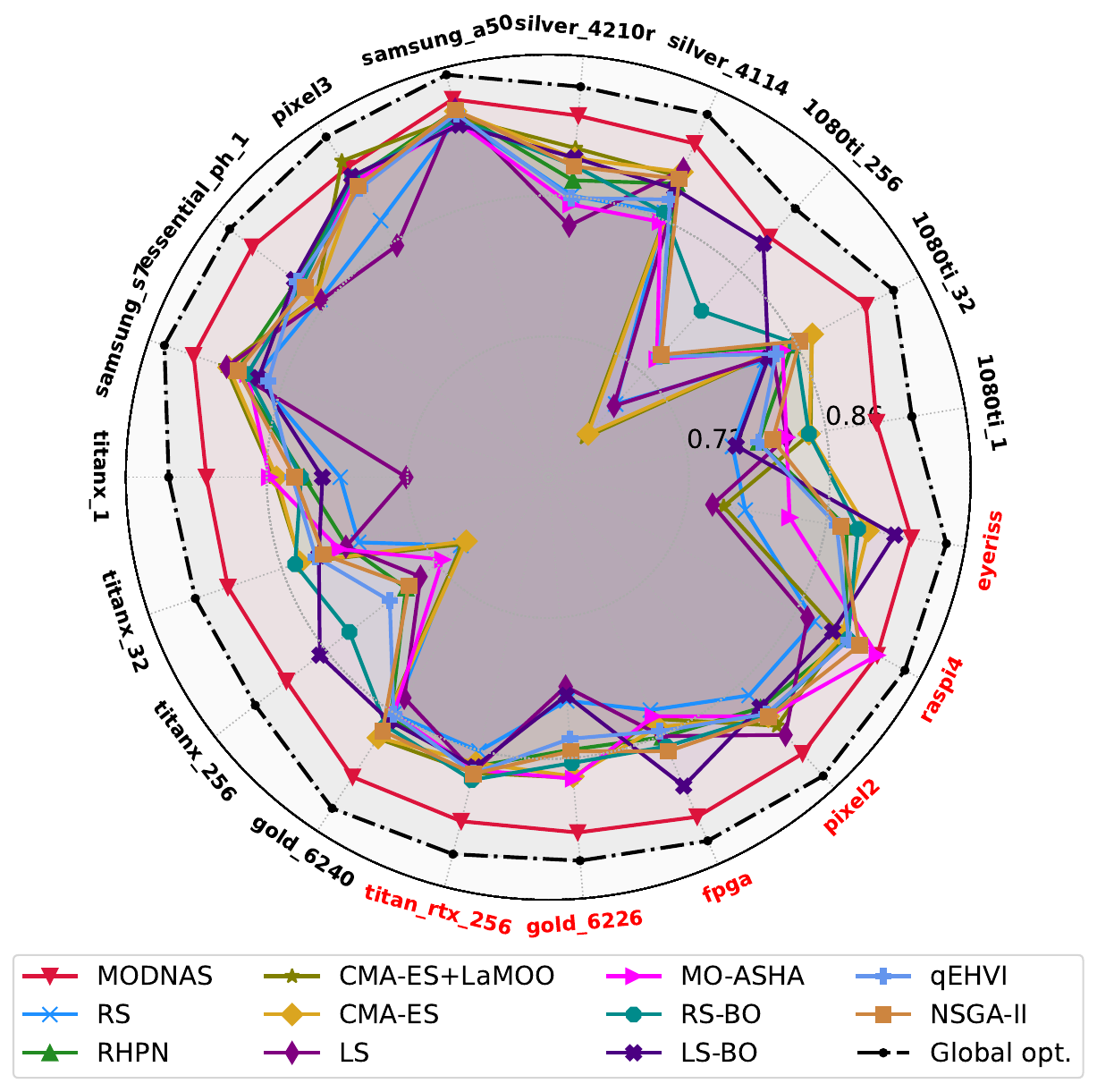}
\vspace{-3ex}
\caption{Hypervolume (HV) of MODNAS and baselines across 19 devices on NAS-Bench-201. For every device, we optimize for 2 objectives, namely \textit{latency (ms)} and \textit{test accuracy} on CIFAR-10. For each method, metric and device we report the mean of 3 independent search runs. Higher area in the radar plot indicates better HV. 
Test devices are colored in red around the plot.} 
\label{fig:radar_plot_hv}
\vspace{-6ex}
\end{wrapfigure}
\noindent
Multiple Gradient Descent (MGD)~\citep{desideri2012mgd, Sener2018MultiTaskLA} provides a plausible approach to estimate the update directions for every task simultaneously by maximizing (\ref{eq:mgd_opt}). Via the Lagrangian duality, the optimal solution to equation~\ref{eq:mgd_opt} is $g_\Phi^* \propto \sum_{\tscript=1}^\Tscript \gamma_\tscript^* g_\Phi^\tscript$, where $\{ \gamma_\tscript^* \}_{\tscript=1}^\Tscript$ is the solution of the following minimization problem:
\begin{equation}
    \min_{\gamma_1, \dots, \gamma_\Tscript} \Big\{ \normx{\sum_{\tscript=1}^\Tscript\gamma_\tscript g_{\Phi}^\tscript}_2^2 \bigg| \sum_{\tscript=1}^\Tscript \gamma_\tscript = 1, \gamma_\tscript \geq 0, \forall \tscript \Big\}
    \nonumber
    \label{eq:mgd_modnas}
\end{equation}
The solution to this problem is either 0 or, given a small step size $\xi$, a descent direction that monotonically decreases all objectives at the same time and terminates when it finds a Pareto stationary point, i.e. $g_{\Phi}^\tscript = 0, \forall \tscript \in \{1,\dots,\Tscript\}$. When $T=2$, the problem above simplifies to $\min_{\gamma\in [0,1]} \normx{ \gamma g_{\Phi}^1 + (1-\gamma) g_{\Phi}^2 }_2^2$, which is a quadratic function of $\gamma$ with a closed form solution:
\begin{equation}
    \gamma^* = \max \Bigg( \min \Big( \frac{(g_{\Phi}^2 - g_{\Phi}^1)^{\mathbf{T}} g_{\Phi}^2}{\normx{g_{\Phi}^1 - g_{\Phi}^2}_2^2}  , 1\Big), 0 \Bigg).
    \nonumber
    \label{eq:mgd_2T}
\end{equation}
When $T>2$, we utilize the \textit{Frank-Wolfe} solver~\citep{Jaggi2013RevisitingFP} as in \citet{Sener2018MultiTaskLA}, where the analytical solution in for $T=2$ is used inside the line search. We provide the full algorithm to compute $\gamma^*$ in Algorithm~\ref{alg:fw_algo} in Appendix~\ref{app:fwsolver}.

In Algorithm~\ref{alg:modnas_algo} and Figure~\ref{fig:modnas overvies} we provide the pseudocode and an illustration of the overall search phase of MODNAS. For every mini-batch sample from $\dataset_{valid}$, we iterate over the device features $d_t$ (line 2), sample one scalarization $\br$ and condition the \metahypernet on both $\br$ and $d_t$ to generate the un-normalized architectural distribution $\Tilde{\alpha}_\Phi$ (lines 3-4). We then compute the device-specific gradient in line 6 which is used to estimate the $\gamma$ coefficients (line 7) used from MGD to update $\Phi$ (lines 8-9).
Similarly to \citet{liu-iclr19}, we use the first-order approximation to obtain the best response function in the lower level (lines 10-14) and repeat the same procedure for the upper-level (lines 2-6), except now the \supernet weights $\bw$ are updated with the mean gradient (line 15), over devices.

\section{Experiments}
\label{sec:experiments}

In this section, we firstly demonstrate the scalability and generalizability of our MODNAS approach on a NAS tabular benchmark (Section~\ref{sec:nb201_experiments}). 
Then, we validate MODNAS on larger search spaces for Machine Translation (Section~\ref{sec:hat}), Image Classification and Language Modeling (Section~\ref{sec: ofa}).

\begin{wrapfigure}[9]{R}{.6\textwidth}
\centering
\vspace{-3.5ex}
\includegraphics[width=0.99\linewidth]{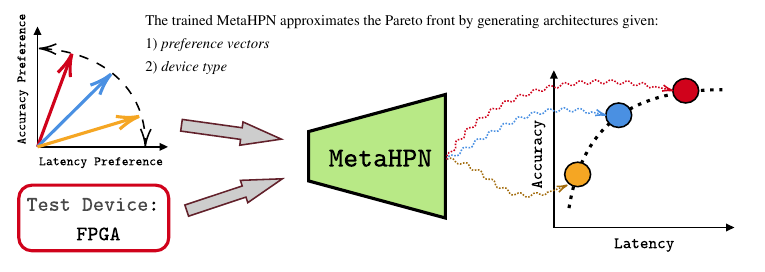}
\vspace{-1.9ex}
\caption{Illustration of MODNAS inference.} 
\label{fig:modnas_inference}
\end{wrapfigure}

\textbf{Search Spaces and Datasets.}
We evaluate MODNAS on 4 search spaces: (1) \textbf{NAS-Bench-201}~\citep{dong-iclr20a, li2021hw} with 19 devices and CIFAR-10 dataset; (2) \textbf{MobileNetV3} from Once-for-All (OFA)~\citep{cai-iclr2020} with 12 devices and ImageNet-1k dataset; (3) \textbf{Hardware-Aware-Transformer (HAT)}~\citep{wang2020hat} on the machine translation benchmark WMT'14 En-De across 3 different hardware devices; (4) \textbf{HW-GPT-Bench}~\citep{Sukthanker2024HWGPTBench} -- a GPT-2 based search space used for language modeling on the OpenWebText~\citep{Gokaslan2019OpenWeb} across 8 devices. We refer to Appendices \ref{app:search_spaces} and \ref{sec:datasets_devices} for more details on these search spaces.

\textbf{Evaluation.}
At test time, in order to profile the Pareto front with MODNAS on unseen devices, we sample 24 equidistant preference vectors $\br$ from the $\Oscript$-dimensional probability simplex and pass them through the pretrained \metahypernet $H_\Phi (\br, d_t)$ to get 24 architectures. Here the test device feature $d_t$ is obtained similarly as for the train devices. See Figure~\ref{fig:modnas_inference} or an illustration.

\begin{wrapfigure}[16]{R}{.39\textwidth}
\centering
\vspace{-3ex}
\includegraphics[width=.99\linewidth]{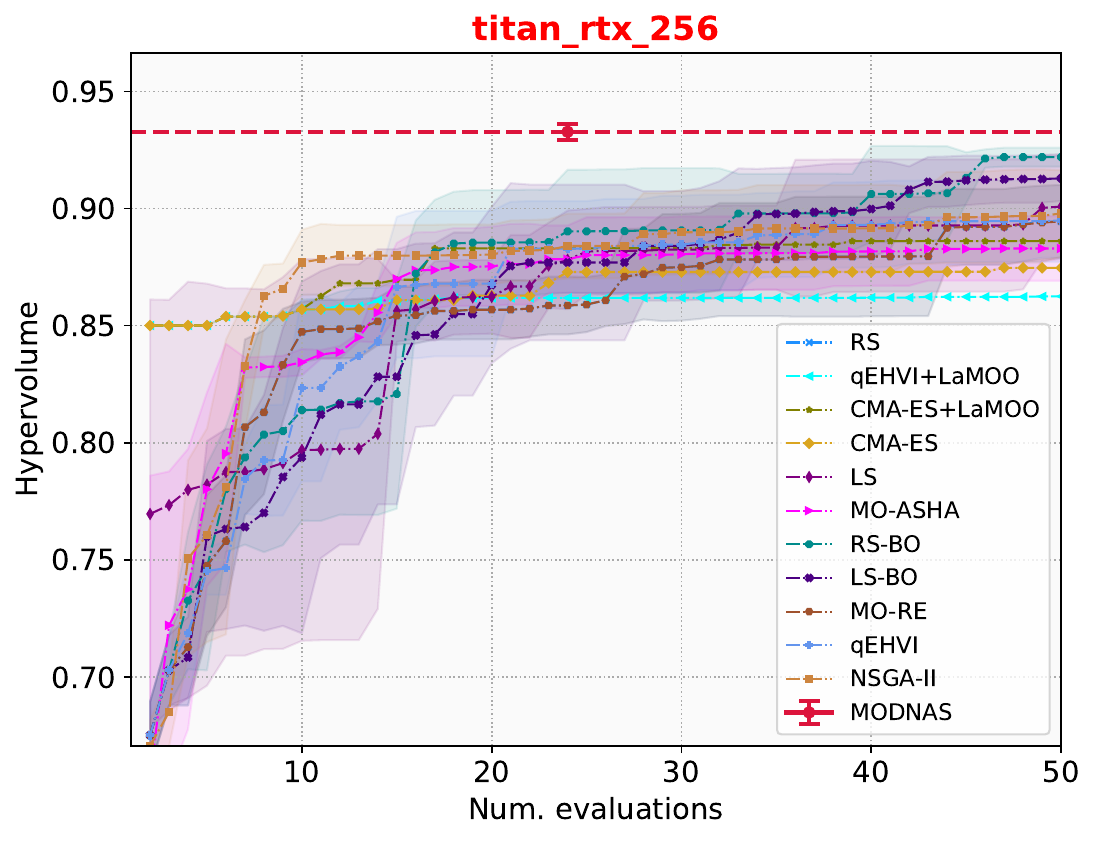}
\vspace{-2ex}
\caption{HV over number of evaluated architectures on NAS-Bench-201 of MODNAS and the blackbox MOO baselines on a test device. For MODNAS we only do 24 full evaluations.} 
\vspace{-3ex}
\label{fig:hv_budget}
\end{wrapfigure}

\textbf{Baselines.} 
We compare MODNAS against several baselines\footnote{For baselines, we use SyneTune~\citep{salinas2022syne}: \url{https://github.com/awslabs/syne-tune}}, such as Random Search (RS), Local Search (LS) and various Evolutionary Strategy and Bayesian Optimization MOO methods. Please refer to Appendix~\ref{sec:algos_full} for a more comprehensive description of each of them. Furthermore, we also evaluate the \metahypernet with randomly initialized weights (RHPN).

\textbf{Metrics.} To assess the quality of the Pareto set solutions, we use the \emph{hypervolume (HV)} indicator, which is a standard metric in MOO.
Given a \emph{reference point} $\rho = [ \rho^1,\dots, \rho^m ] \in \bbR_{+}^M$ that is an upper bound for all objectives $\{f^m (\cdot;\bw, \alpha)\}_{m=1}^M$, i.e. $sup_\alpha f^m (\cdot;\bw, \alpha) \leq \rho^m$, $\forall m\in [M]$, and a Pareto set $\mathcal{P}_\alpha \subset \archss$, HV($\mathcal{P}_\alpha$) measures the region of non-dominated points bounded above from $\rho$:
$$\lambda \Big( \big\{   q\in \bbR_{+}^M \ | \ \exists \alpha\in\mathcal{P}_\alpha : q \in \prod_{m=1}^M [f^m (\cdot;\bw, \alpha), \rho^m]    \big\} \Big),$$
where $\lambda(\cdot)$ is the Euclidean volume. HV can be interpreted as the total volume of the union of the boxes created by the Pareto front.

\subsection{Simultaneous Pareto Set Learning across 19 devices and Ablations}
\label{sec:nb201_experiments}

\begin{wrapfigure}[13]{R}{.39\textwidth}
    \centering
    \vspace{-3.5ex}
    \centering
    \includegraphics[width=.99\linewidth]{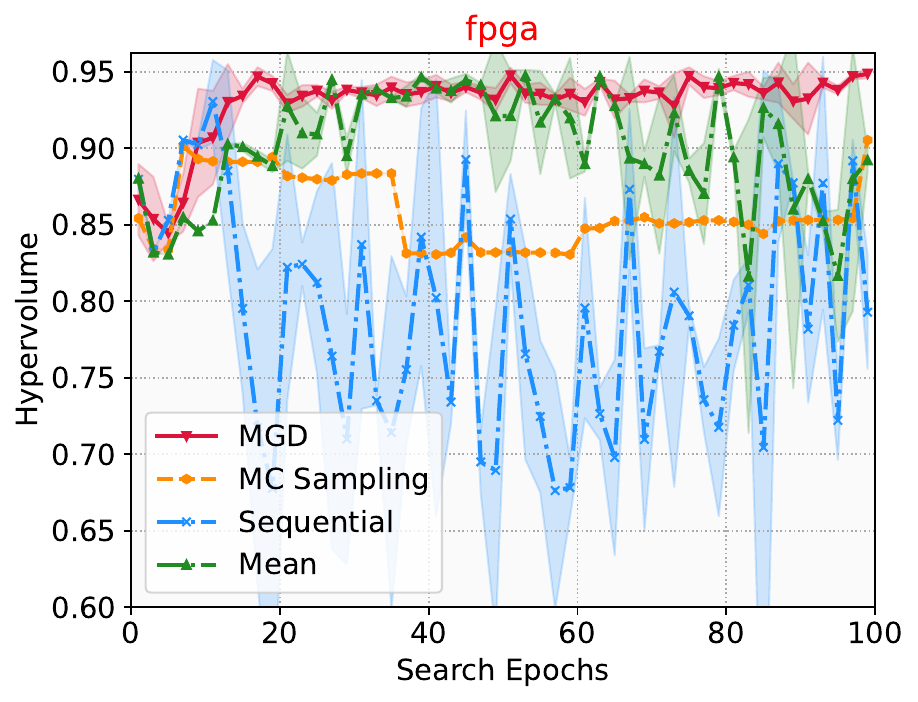}
    \vspace{-4ex}
    \caption{HV over search epochs of different gradient schemes in MODNAS.}
    \label{fig:gradschemes_hv}
    \vspace{-5ex}
\end{wrapfigure}

We firstly validate the scalability and learning capability of MODNAS by evaluating on the NAS-Bench-201~\citep{dong-iclr20a} cell-based convolutional space. Here we want to optimize both latency and classification accuracy on all devices. We utilize the same set of 19 heterogeneous devices as \citet{lee2021help}, from which we use 13 for search and 6 at test time. For the latency predictor, we use the one from HELP, namely a graph convolutional network (GCN), which we pretrain for 3 GPU hours on the ground truth latencies on the 13 search devices as described in Section~\ref{sec:methodology}.
We run the MODNAS search (see Appendix~\ref{sec: hyper_configs} for more details on the search hyperparameters), as described in Algorithm~\ref{alg:modnas_algo}, for 100 epochs (22 GPU hours on a single NVidia RTX2080Ti) and show the HV in Figure~\ref{fig:radar_plot_hv} of the evaluated Pareto front in comparison to the baselines, for which we allocate the same search time budget across all devices equivalent to the MODNAS search + evaluation.

Most notably, MODNAS consistently outperforms all other baselines across every device. For the baselines, we conduct 19 separate search runs (one for each device), whereas MODNAS leverages meta-learning to generate the Pareto set on each device using the same \metahypernet in a single search run. Interestingly, the trained MODNAS attention-based \metahypernet significantly outperforms the RHPN baseline in profiling the Pareto front, demonstrating its \emph{effectiveness in optimizing across multiple devices and conflicting objectives simultaneously}. In Figure~\ref{fig:radar_hv_50} in the Appendix, we compare MODNAS with additional baselines, running them at double the budget used for the experiments in Figure~\ref{fig:radar_plot_hv}. Figure~\ref{fig:hv_budget} (see Figure~\ref{fig:hv_budget_all} in the appendix for all devices) shows that most baselines require more than twice the number of architecture evaluations to reach the same HV as MODNAS. Results show that MODNAS remains the top performer across hardware devices on average. Furthermore, in the appendix,  Figure~\ref{fig:radar_plot_gd_igd} presents radar plots for four additional metrics, and Figure~\ref{fig:pixel3-c100} and \ref{fig:edgegpu-c100} results on NB201 when optimizing CIFAR-100 accuracy and device latency.

\textbf{Reliably learnt embeddings for hardware devices.} To demonstrate the effectiveness of our \metahypernet in learning hardware device similarities, Figure~\ref{fig:tsne_combined} in the appendix shows K-means clustering of original and \metahypernet embeddings, reduced via t-SNE. The \metahypernet successfully clusters similar devices, confirming its efficacy.

\textbf{\metahypernet update schemes: robustness of MGD.}
We compare the MGD update scheme for the \metahypernet $\Phi$ (line 9 in Alg.~\ref{alg:modnas_algo}) against (1) the \textbf{mean} gradient over tasks: $\Phi \gets \Phi - \xi \frac{1}{\Tscript}\sum_{\tscript=1}^\Tscript g_\Phi^\tscript$; (2) \textbf{sequential} updates with all single tasks' gradients: $\Phi \gets \Phi - \xi g_\Phi^\tscript$, $\forall \tscript$; (3) single updates using gradients of \textbf{MC samples} over tasks: $\Phi \gets \Phi - \xi g_\Phi^\tscript$, $t\sim \{1,\dots\Tscript\}$. Figure~\ref{fig:gradschemes_hv} (see Figure~\ref{fig:gradschemes_hv_full} in Appendix~\ref{sec:additional_esperiments} for more results) shows the HV over search epochs for these schemes. MGD, by accounting for inter-task dependencies, achieves higher final HV, better anytime performance, and faster convergence than the other schemes.

\begin{figure}[t]
\vspace{-5mm}
\begin{minipage}{0.52\linewidth}
    \centering
    \begin{minipage}{0.47\linewidth}
      \centering
      \includegraphics[width=\linewidth]{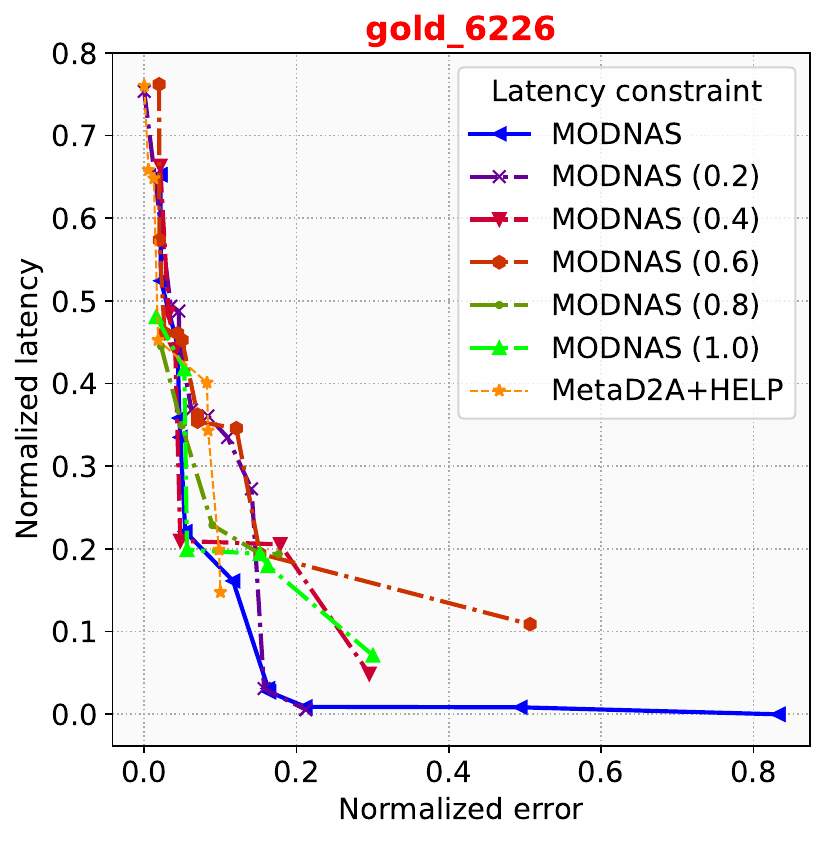}
    \end{minipage}%
    \begin{minipage}{0.52\linewidth}
      \centering
      \includegraphics[width=\linewidth]{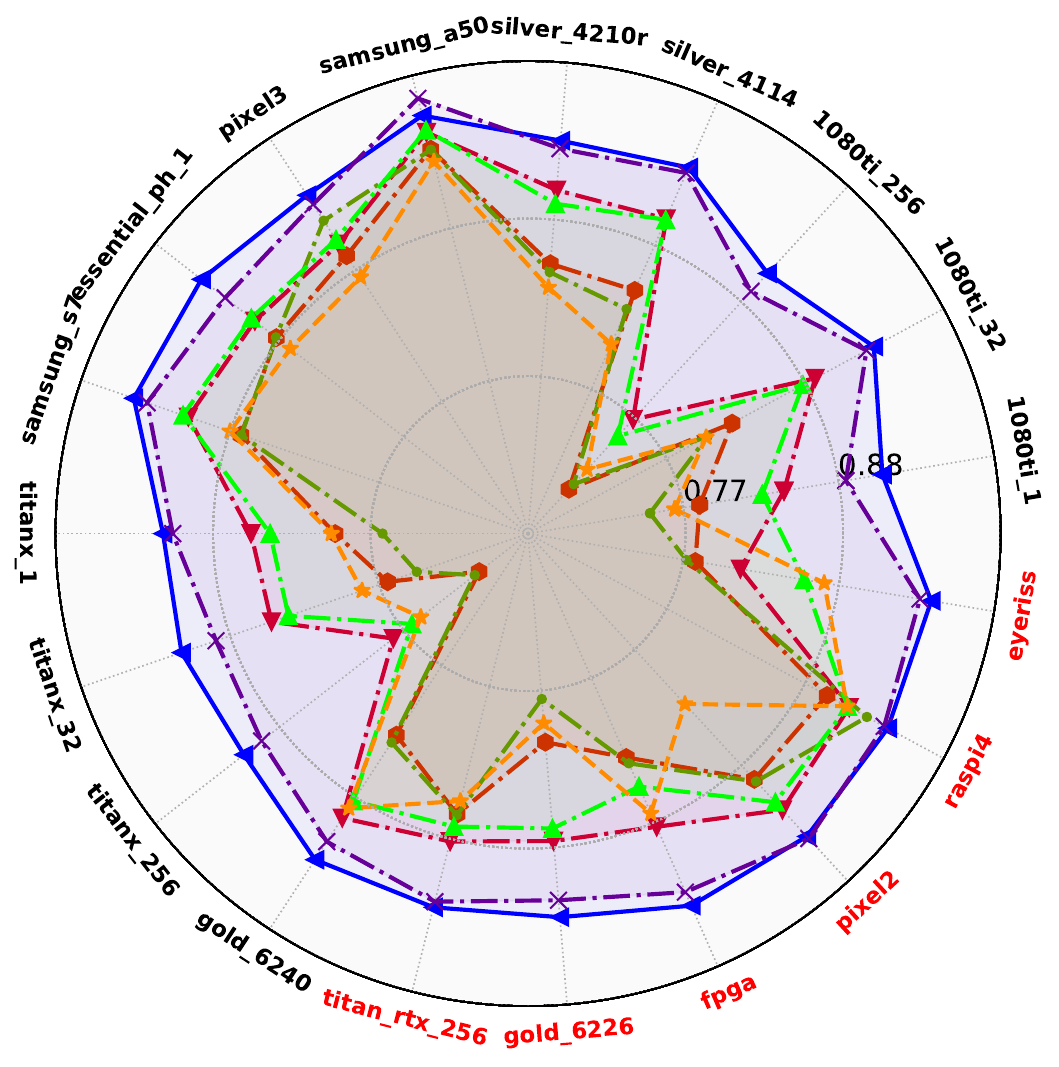}
    \end{minipage}%
    \vspace{-1ex}
    \caption{Pareto front on Eyeriss (\emph{left}) and HV across devices (\emph{right}) of MODNAS ran with various latency constraints on NAS-Bench-201. See Fig.~\ref{fig:fronts_constraints_full} in Appendix~\ref{sec:additional_esperiments} for all results.}
    \label{fig:fronts_constr}
\end{minipage}\hspace{5pt}
\begin{minipage}{0.47\linewidth} 
    \vspace{-1ex}
    \centering
    \begin{minipage}{0.49\linewidth}
    \centering
    \includegraphics[width=\linewidth]{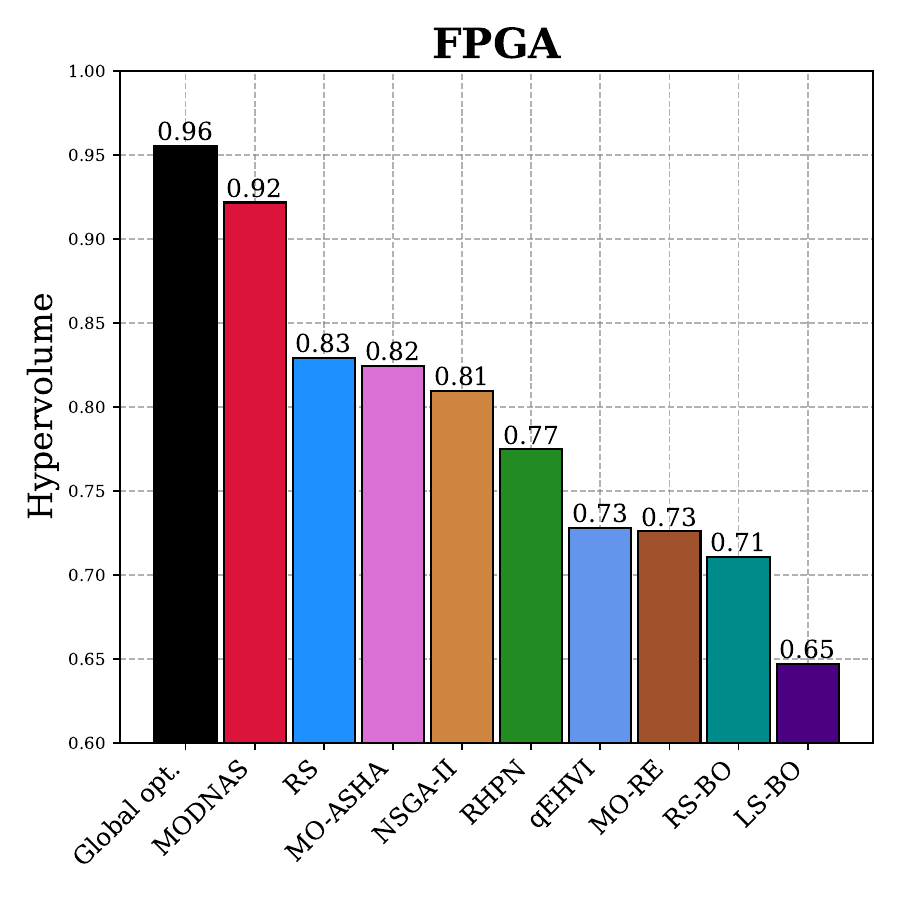}
    \end{minipage}%
    \begin{minipage}{0.51\linewidth}
    \centering
    \includegraphics[width=\linewidth]{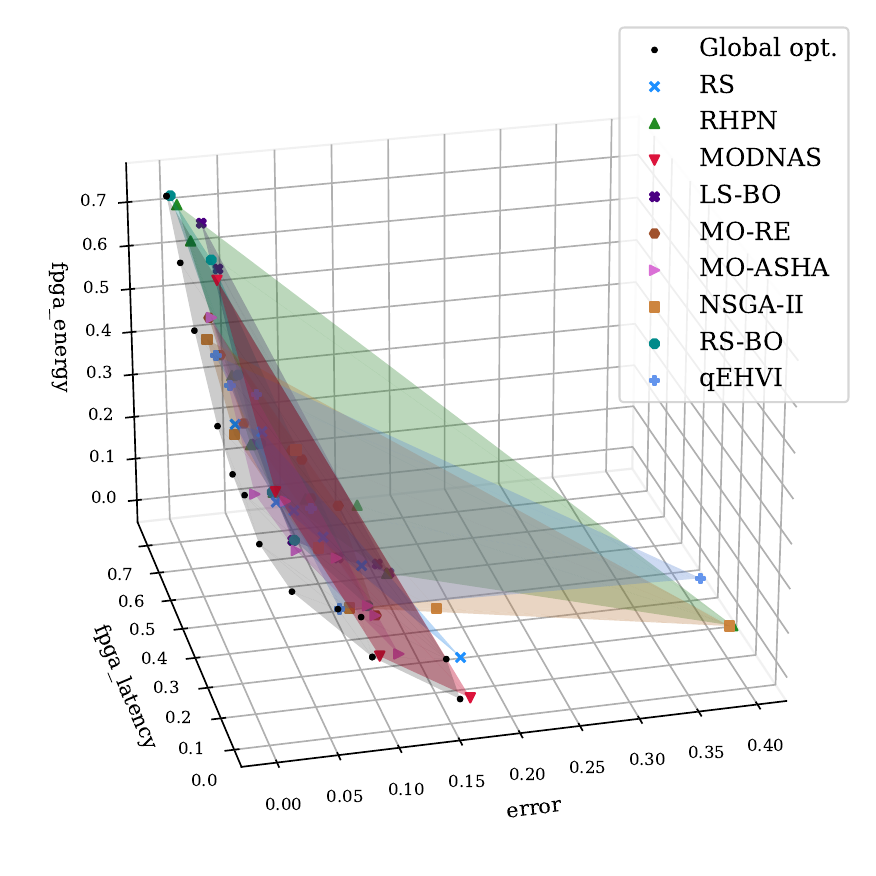}
    \end{minipage}%
    \vspace{-2ex}
    \caption{HV (\textit{left}) and Pareto front (\textit{right}) of MODNAS and baselines on FPGA with 3 normalized objectives: error, latency and energy usage. HV is computed using the $(1,1,1)$ reference point on the right 3D plot. See Fig.~\ref{fig:3d_plot_eyeriss} for results on Eyeriss.}
    \label{fig:3d_plot}
\end{minipage}
\vspace{-4ex}
\end{figure}

\textbf{Scalability to three objectives.}
We show the scalability of MODNAS to 3 objectives, namely, accuracy, latency and energy consumption. For this experiment we use the FPGA and Eyeriss tabular energy usage values from HW-NAS-Bench~\citep{li2021hw}. In addition to the \metapredictor for latency, we pretrain a second predictor on the energy usage objective. We then run MODNAS and the MOO baselines with the same exact settings as for 2 objectives. Results shown in Figure~\ref{fig:3d_plot} indicate that MODNAS can scale to $\Oscript>2$ without additional search costs or hyperparameter tuning and yet achieves HV close to the global optimum front of the NAS-Bench-201 space.

\begin{wrapfigure}[12]{R}{.53\textwidth}
\vspace{-3ex}
    \centering
    \begin{minipage}{0.5\linewidth}
      \centering
      \includegraphics[width=\linewidth]{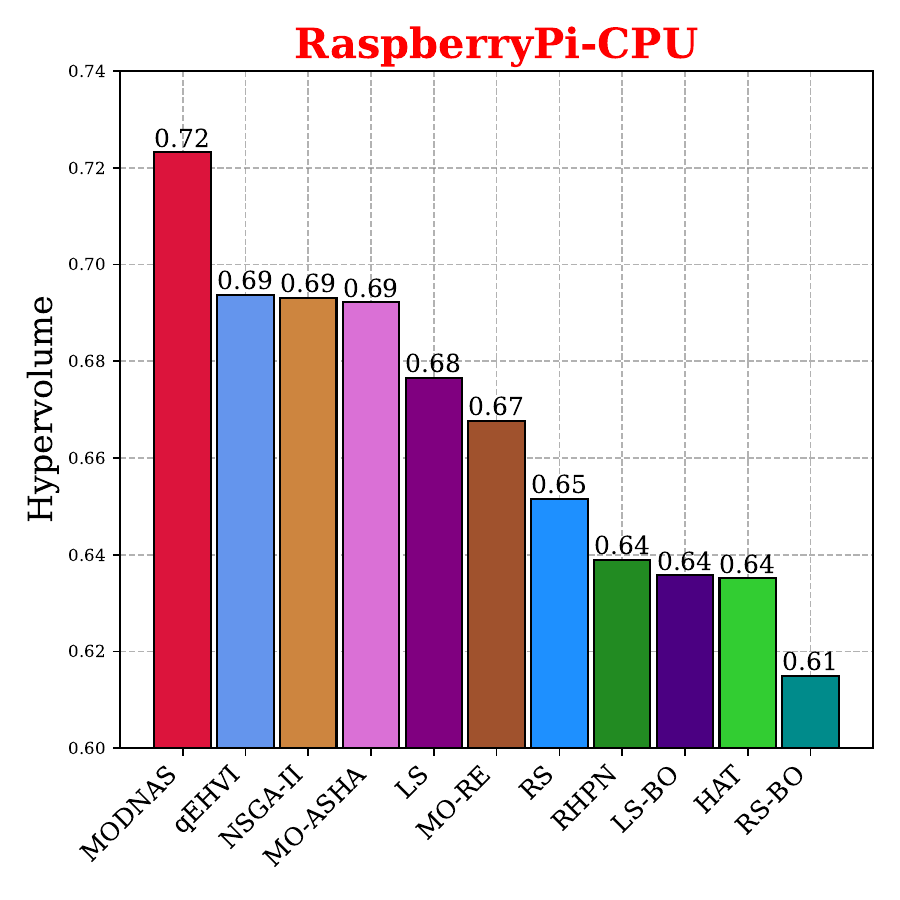}
    \end{minipage}%
    \begin{minipage}{0.5\linewidth}
      \centering
      \vspace{-2ex}
      \includegraphics[width=\linewidth]{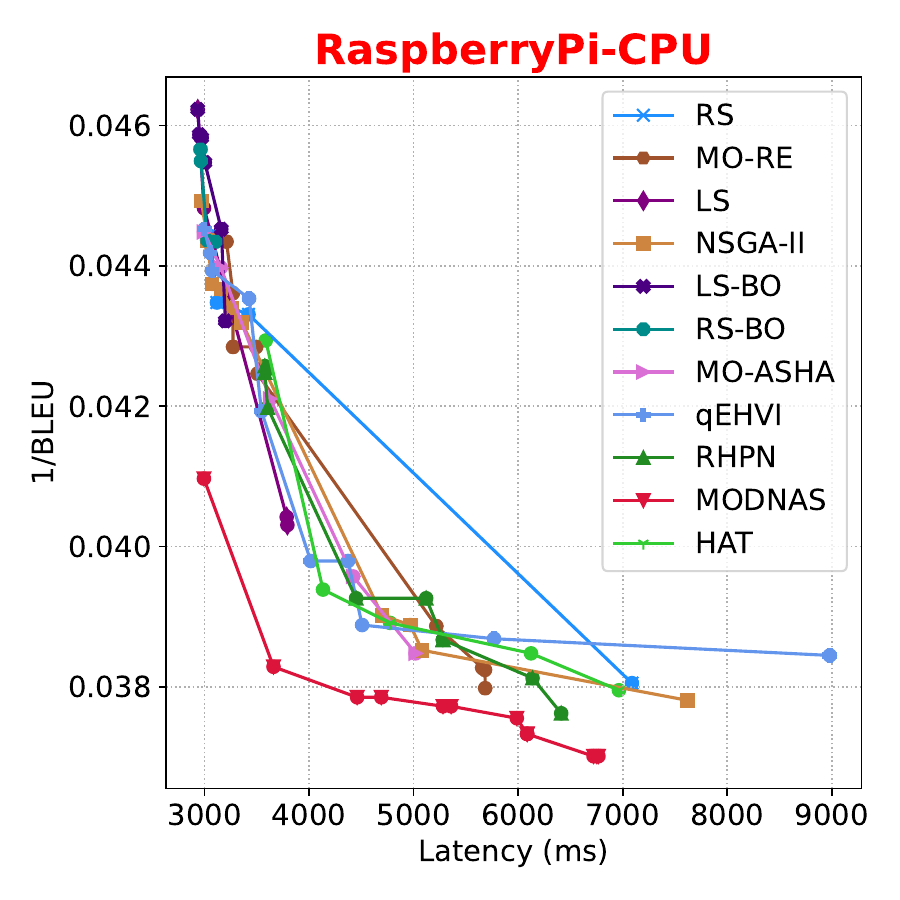}
    \end{minipage}%
    \vspace{-2ex}
    \caption{HV and Pareto fronts of MODNAS and baselines across devices on the HAT space.}
    \label{fig:hat_barplot_pareto}
\end{wrapfigure}
\textbf{MODNAS vs. constrained single-objective optimization.}
To compare against single-objective NAS with hardware constraints in the objective, we run \textbf{MetaD2A+HELP}~\citep{lee2021help}. Since MetaD2A + HELP is not able to profile the Pareto front directly, we run the NAS search 24 times with different constraints, which we compute by denormalizing the same 24 equidistant preference vectors we use to evaluate MODNAS. 
We also extend MODNAS to incorporate user prior constraints over the multiple objectives being optimized during search. Namely, we add a normalized constraint $c^m$, such that if the predicted value from the \metapredictor during search satisfies this constraint, i.e. $p_\theta^m (\alpha_\Phi, d_t^m) \leq c^m$, we remove the gradient w.r.t. to that objective in lines 6 and 14 of Algorithm~\ref{alg:modnas_algo}. In Figure~\ref{fig:fronts_constr} (other devices in Figure~\ref{fig:fronts_constraints_full}) we can see that when increasing the latency constraint to 1 (only cross-entropy optimized), though the HV decreases, MODNAS returns Pareto sets with more performant architectures. MetaD2A+HELP, despite multiple search runs, prioritizes performance over diversity, resulting in less varied solutions.

\subsection{Pareto Front Profiling on Transformer Space}
\label{sec:hat}

To demonstrate its effectiveness beyond image classification and CNN spaces, we apply MODNAS to the hardware-aware Transformer (HAT) search space from \citet{wang2020hat} on the WMT'14 En-De~\citep{jean2015montreal,machacek-bojar-2014-results} machine translation task. We pretrain the \metapredictor (details in Appendix~\ref{app:meta_predictor_details}) for 5 GPU hours on 2000 architecture samples from the search space and then conduct the search for 110 epochs (6 days on 8 NVIDIA RTX A6000 GPUs) using 2 search devices, adhering to the same hyperparameters as \citet{wang2020hat} to optimize for \emph{latency} and \emph{validation cross entropy loss}. 
We allocate to each baseline 2.5$\times$ more runtime budget than MODNAS, resulting in 1300 (RS-BO) to 6000 (MO-ASHA) total architecture evaluations, whereas MODNAS evaluates only 24 generated architectures. Details on the HAT search space and search hyperparameters are in Appendix~\ref{app:search_spaces}. We evaluate MODNAS on all 3 devices (2 search and 1 test) using the BLEU score, and results in Figure~\ref{fig:hat_barplot_pareto} show that MODNAS outperforms all baselines, achieving a higher hypervolume (left plot) of the generated Pareto fronts (right plot). For HAT, we evaluate the architectures provided in their paper. Additional results on other training devices and evaluation metrics are presented in Figures~\ref{fig:pareto_hat_full}, \ref{fig:hat_barplot} and \ref{fig:hat_barplot_sacre} in the Appendix.

\subsection{Efficient Differentiable MOO starting from Pretrained Supernetworks}
\label{sec: ofa}

\begin{wrapfigure}[19]{R}{.3\textwidth}
    \vspace{-4ex}
    \centering
    \includegraphics[width=.97\linewidth]{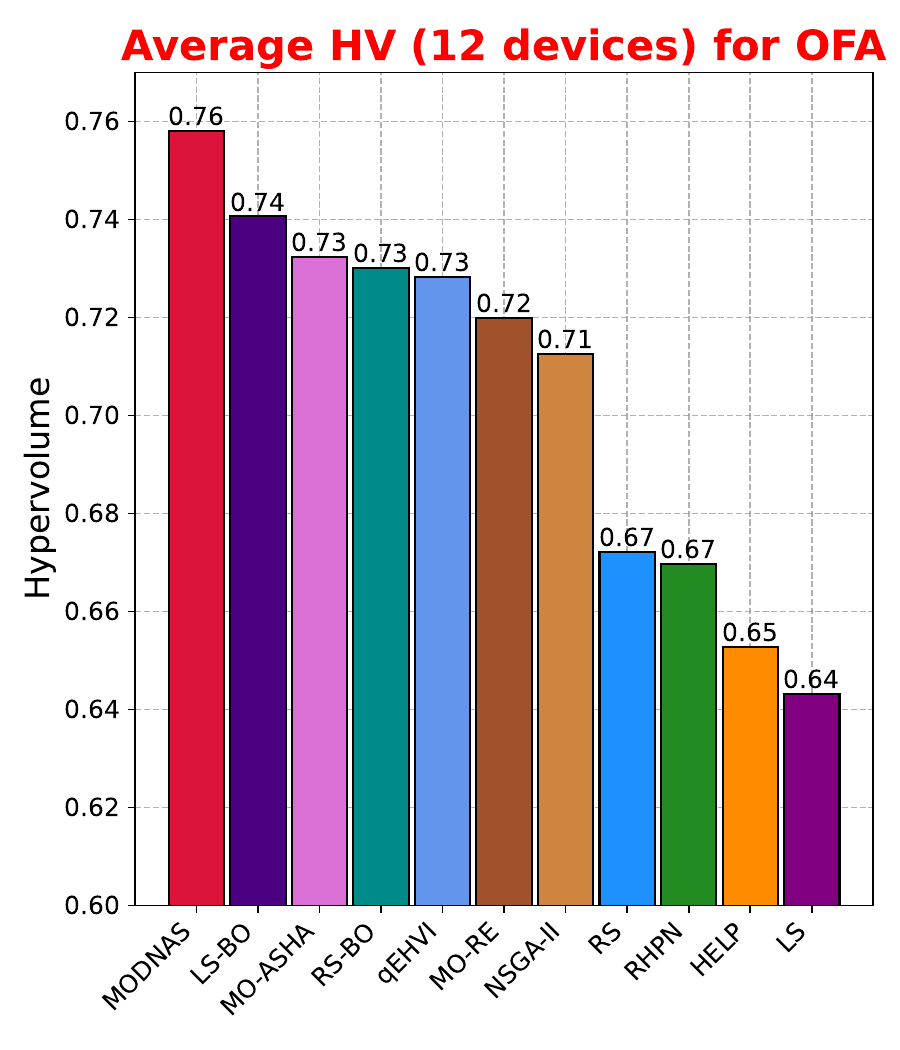}
    \vspace{-2ex}
    \caption{Average HV of MODNAS and baselines across 12 devices on OFA space. For every device we optimize for 2 objectives, namely \textit{latency (ms)} and \textit{test accuracy} on ImageNet-1k.} 
    \label{fig:ofa_radar}
\end{wrapfigure}

\textbf{Image Classification on ImageNet-1k.} We now evaluate MODNAS on ImageNet-1k using the MovileNetV3 search space from Once-for-All (OFA)~\citep{cai-iclr2020}.
For this experiment, we run MODNAS using 11 search (and 1 test) devices starting with the pretrained OFA supernetwork and run the search further for 1 day on 8 RTX2080Ti GPUs. During the search, we only update the \metahypernet weights and keep the pretrained \supernet weights frozen. Details on the search space and hyperparameters are in Appendices~\ref{app:search_spaces} and \ref{sec:modnas_hyperparameters}. We use the simple MLP from \citet{lee2021help} as our \metapredictor, pretraining it for 6 hours on 5000 sampled architecture-latency pairs. To evaluate the 24 points generated by our \metahypernet and baselines, we use the OFA pretrained \supernet. Results in Figure~\ref{fig:ofa_radar} show that MODNAS achieves a higher average HV across all devices compared to baselines, which we run for 192 hours using the OFA pretrained accuracy predictor (see Figure~\ref{fig:ofa_barplot_all} for all results and Figure~\ref{fig:pareto_ofa_full} for the Pareto fronts).

\textbf{Comparison to Zero-Cost Proxies.} We also compare the HV of the Pareto front obtained by MODNAS to that produced by NSGA-II~\citep{deb2002fast}, which uses a zero-cost proxy (ZCP)~\citep{abdelfattah-iclr21} (we chose Zico~\citep{li2023zico}) for performance estimation instead of the actual accuracy. Table~\ref{tab:ofa-table-zico} in the appendix presents the results of this experiment on two devices. As shown, despite its improved runtime efficiency, the ZCP-guided search underperforms compared to both the existing baselines and MODNAS, which optimize for accuracy directly.

\begin{wrapfigure}[14]{R}{.33\textwidth}
    \vspace{-2ex}
    \centering
    \includegraphics[width=.97\linewidth]{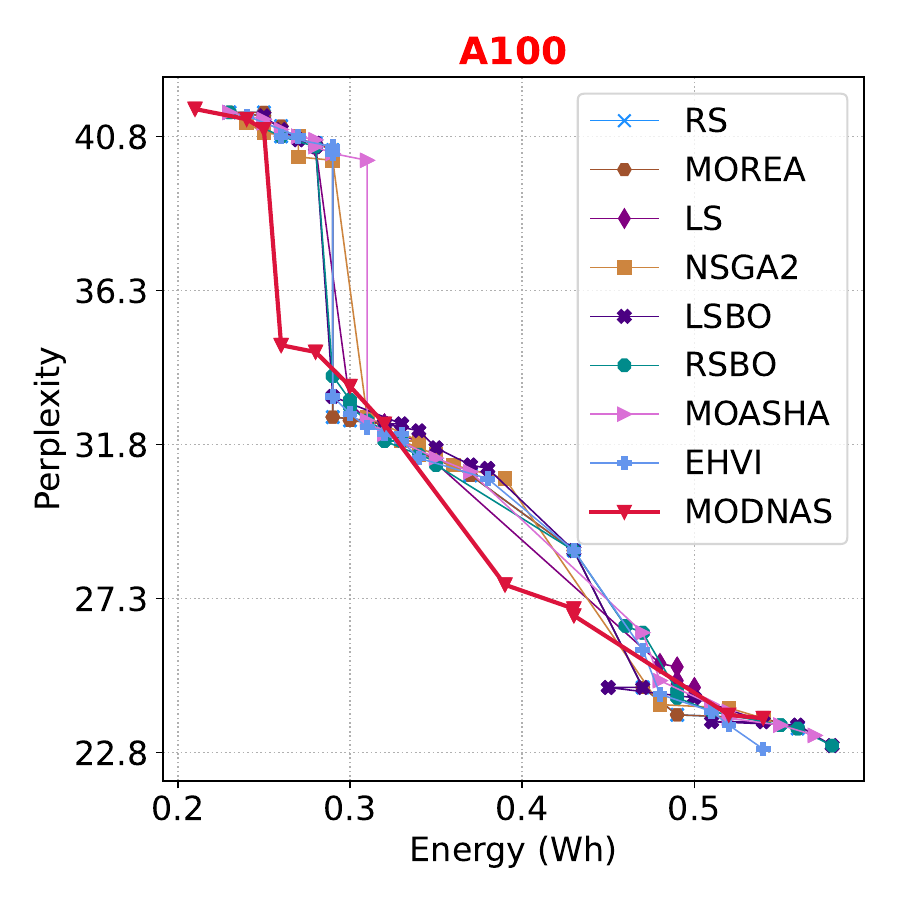}
    \vspace{-2.5ex}
    \caption{Pareto front of MODNAS and baselines on the HW-GPT-Bench, A100 GPU.} 
    \label{fig:gpt_pareto}
\end{wrapfigure} 
\textbf{Language Modeling with GPT-2.} 
With the rapid growth of language model sizes, it is crucial to identify transformer variants that are efficient during inference (latency) while maintaining competitive performance. We apply MODNAS to the GPT-S space from HW-GPT-Bench~\citep{Sukthanker2024HWGPTBench}, which features a non-convex Pareto front between perplexity and hardware metric objectives. Using pretrained \supernet weights from HW-GPT-Bench, we conduct a single 6-hour search on 4 Nvidia A100 GPUs, optimizing for energy consumption (Wh) and perplexity across 8 different GPU devices. See Appendix~\ref{sec: hyper_configs} for details on the \metahypernet architecture and search hyperparameters. The \supernet weights are kept frozen while updating the \metahypernet. Figure~\ref{fig:gpt_pareto} shows that, with the same time budget, MODNAS matches or surpasses other MOO baselines, demonstrating its effectiveness in optimizing beyond convex Pareto fronts.

\subsection{Computational Complexity}

\begin{wraptable}[9]{R}{.55\textwidth} 
\vspace{-2.8ex}
\caption{Cost of MODNAS and other methods. N is the number of trained architectures during search, T the number of devices and C the number of constraints.}
\vspace{-1.9mm}
\label{tab:modnas_cost}
\resizebox{0.999\linewidth}{!}{%
\begin{tabular}{l|c|c}
\toprule
\textbf{Method}         & \textbf{Search Cost} & \textbf{Pareto Set Build Cost} \\ \midrule\midrule
LEMONADE~\citep{elsken-iclr19a}          &      $\bigO$(NT)     &          $\bigO$(1)          \\ 
Blackbox MOO~\citep{daulton-neurips2020a, zhao2022multiobjective}          &      $\bigO$(NT)     &          $\bigO$(1)          \\ 
ProxylessNAS~\citep{cai2018proxylessnas} &      $\bigO$(CT)    &          $\bigO$(1)          \\ 
MetaD2A + HELP~\citep{Lee2021RapidNA,lee2021help}       &      $\bigO$(N)     &          $\bigO$(CT)          \\
OFA~\citep{cai-iclr2020} + HELP~\citep{lee2021help}  &      $\bigO$(1)     &          $\bigO$(CT)          \\
\textbf{MODNAS (Ours)}                   & $\bigO$(1) &      $\bigO$(1) \\ \bottomrule
\end{tabular}
}
\end{wraptable}

Ignoring the cost to train final architectures in the Pareto set, methods like MetaD2A + HELP~\citep{Lee2021RapidNA,lee2021help} have a worst-case time complexity of $\bigO$(CT) to build the Pareto set, where T is the number of devices and C is the number of constraints. MODNAS reduces this to $\bigO$(1) by conditioning a single \metahypernet on both device types and constraints. Methods like LEMONADE~\citep{elsken-iclr19a} and ProxylessNAS~\citep{cai2018proxylessnas} apply constraints during the search phase, requiring an independent search per device. Black-box methods such as LEMONADE, NSGA-II~\citep{deb2002fast}, or qEHVI~\citep{daulton-neurips2020a} train $\bigO$(NT) architectures or a surrogate based on $\bigO$(N) architectures in the case of MetaD2A + HELP. In contrast, MODNAS and OFA have a cost of $\bigO$(1) as they train a single supernetwork. Although MODNAS iterates over T devices to compute $g_{\Phi}^*$ and $g_{\bw}^*$, Figure~\ref{fig:ndevices_hv_full} in Appendix~\ref{sec:additional_nb201_results} shows that MODNAS generalizes well on 17 test devices with only 2 search devices due to its meta-learning capabilities. See Tables~\ref{tab:modnas_cost} and \ref{tab:search_times} in the Appendix for more details.

\section{Conclusions, Broader Impact and Limitations}
\label{sec:conclusion}

In this paper, we propose a novel hardware-aware differentiable NAS algorithm for profiling the Pareto front in multi-objective problems. In contrast to constraint-based NAS methods, ours can generate Pareto optimal architectures across multiple devices with a single hypernetwork that is conditioned on preference vectors encoding the trade-off between objectives. Experiments across various hardware devices (up to 19), objectives (accuracy, latency and energy usage), search spaces (CNNs and Transformers), and applications (classification, machine translation, language modeling) demonstrate the effectiveness and efficiency of our method.

\textbf{Broader Impact.} In an era of large-scale models (e.g. foundation models), speeding up the search and training cost for inference-optimal neural architectures is an important aspect of responsible research \citep{muralidharan2024compact,cai2024flextron,zhang2024laptop}. 
The main goal of this work is to improve the search costs, as well as the efficiency of the found architectures in terms of various hardware metrics, therefore reducing the energy consumption and CO$_2$ footprint.
The energy savings of these architectures will be amplified as they might be deployed on a large number of devices.

\textbf{Limitations.} While our differentiable multi-objective search method shows promising results, there are potential limitations. MODNAS inherits challenges common to gradient-based search, such as the risk of failure without proper tuning or regularization~\citep{zela-iclr20a}. For example, gradients may favor one objective, leading to local optima that hinder exploration of the full Pareto front. Additionally, the method relies on differentiable proxies for objectives, which may not always align with ground truth values.

\clearpage
\newpage

\section*{Acknowledgments}
This research was partially supported by the following sources:
TAILOR, a project funded by EU Horizon 2020 research and innovation programme under GA No
952215; the Deutsche Forschungsgemeinschaft (DFG, German Research Foundation) under
grant number 417962828; the European Research Council (ERC) Consolidator Grant “Deep Learning
2.0” (grant no. 101045765). Robert Bosch GmbH is acknowledged for financial support. The authors acknowledge support from ELLIS and ELIZA. The authors gratefully acknowledge the Gauss Center for Supercomputing eV (www.gauss-centre.eu) for funding this project by providing computing time on the GCS supercomputer JUWELS at Jülich Supercomputing Center (JSC). Funded by
the European Union. Views and opinions expressed are however those of the author(s) only and do
not necessarily reflect those of the European Union or the ERC. Neither the European Union nor the
ERC can be held responsible for them.
\begin{center}\includegraphics[width=0.3\textwidth]{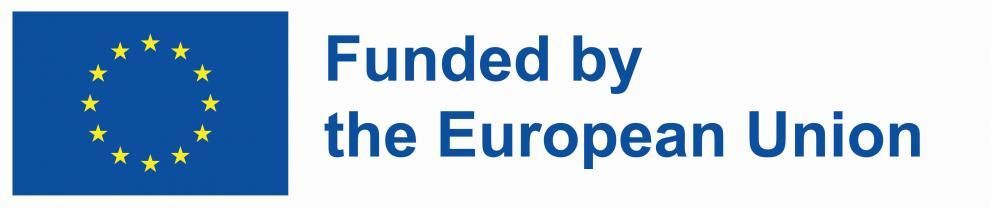}\end{center}. 
\vspace{-1cm}
\bibliographystyle{iclr2025_conference}
\bibliography{lib,proc,shortproc,strings,shortstrings}
\clearpage\newpage


\appendix
\addcontentsline{toc}{section}{Appendix} 
\part{Appendix} 
\parttoc 
\newpage
\section{Algorithmic components}
\label{app:algo_pseudocodes}
In this section, we provide the pseudocodes for some of the algorithmic components we use in MODNAS.

\subsection{Discrete Samplers}
\label{sec:discrete_samplers}

Given the architecture parameters $\Tilde{\alpha}_\Phi$ from the \metahypernet, we obtain a differentiable discrete architecture sample from the \architect as $\mathtt{\alpha_\Phi \gets \pi - stop\_g(\pi) + \alpha_\Phi }$, where $\mathtt{\alpha_\Phi \sim Cat \big(\mathtt{softmax}_1(\Tilde{\alpha}_\Phi) \big)}$ and 
\begin{equation}
    \begin{aligned}
        &\mathtt{\pi \gets 2 \cdot \mathtt{softmax}_1 \Big(stop\_g \big(ln ( \frac{\alpha_\Phi + \mathtt{softmax}_\tau(\Tilde{\alpha}_\Phi)}{2})-\Tilde{\alpha}_\Phi \big) +\Tilde{\alpha}_\Phi \Big) - \frac{\mathtt{softmax}_1(\Tilde{\alpha}_\Phi)}{2}}.\\
    \end{aligned}
    \nonumber
    \label{eq:reinmax}
\end{equation}
Here, $\mathtt{Cat}$ is the categorical distribution, $\tau$ is the temperature in the tempered softmax $\mathtt{softmax}_\tau(\alpha)_i = \frac{exp(\alpha_i / \tau)}{\sum_{j=1}^{|\opss|} exp(\alpha_j / \tau) }$, and $\mathtt{stop\_g(\cdot)}$ duplicates its input and detaches it from backpropagation. Refer to the ReinMax paper~\citep{liu2023bridging} for more details. The algorithm pseudocode on how a one-hot encoded (discrete) architecture is sampled given an unnormalized architectural distribution $\Tilde{\alpha}$ is given in Algorithm~\ref{alg:st_algo} and Algorithm~\ref{alg:reinmax_algo}, for the Straight-Through~\citep{jang-iclr17a} and ReinMax~\citep{liu2023bridging} gradient estimators, respectively.

\begin{minipage}{.48\textwidth}
\begin{algorithm}[H] 
\KwData{$\Tilde{\alpha}$: softmax input, $\tau$ : temperature}
\KwResult{$\alpha$: one-hot samples}
$\pi_0\gets \mathtt{softmax}_1(\Tilde{\alpha})$ \\
$\alpha \sim \mathtt{Cat} (\pi_0)$ \\
$\pi_1 \gets  \mathtt{softmax}_\tau(\Tilde{\alpha})$ \\
$\alpha \gets \pi_1 - \mathtt{stop\_g}(\pi_1) + \alpha$\\
\Return{$\alpha$}
\caption{$\mathtt{Straight-Through}$~\citep{jang-iclr17a}}
\label{alg:st_algo}
\end{algorithm}
\vspace{5.5mm}
\end{minipage}
\hfill
\begin{minipage}{.48\textwidth}
\begin{algorithm}[H] 
\KwData{$\Tilde{\alpha}$: softmax input, $\tau$ : temperature}
\KwResult{$\alpha$: one-hot samples}
$\pi_0\gets \mathtt{softmax}_1(\Tilde{\alpha})$ \\
$\mathtt{\alpha \sim Cat (\pi_0)}$ \\
$\pi_1 \gets \frac{\alpha + \mathtt{softmax}_\tau(\Tilde{\alpha})}{2}$ \\
$\pi_1 \gets \mathtt{softmax}_1\big(\mathtt{stop\_g}\big(\mathtt{ln}(\pi_1)-\Tilde{\alpha}\big)+\Tilde{\alpha}\big)$ \\
$\pi_2 = 2 \cdot \pi_1 - \frac{1}{2} \cdot \pi_0$ \\
$\alpha \gets \pi_2 - \mathtt{stop\_g}(\pi_2) + \alpha$\\
\Return{$\alpha$}
\caption{$\mathtt{ReinMax}$~\citep{liu2023bridging}}
\label{alg:reinmax_algo}
\end{algorithm}    
\end{minipage}

\subsection{Details on the \architect Gradient Computation}

In this section, we provide additional details on how the \architect utilizes the Straight-Through Estimator (STE) to backpropagate through the sampling of discrete architectural parameters.

\textbf{Forward pass:}
\begin{enumerate}[leftmargin=*]
    \item The \metahypernet parameterizes the unnormalized architectural distribution: $\Tilde{\alpha} = H_{\Phi}$, where $\Phi$ are the \metahypernet parameters.
    \item $\Tilde{\alpha}$ is passed to \architect and it does the following steps:
    \begin{enumerate}
        \item Normalizes $\Tilde{\alpha}$ and samples a one-hot (discrete) $\alpha$: $\alpha \sim \mathtt{Cat}(\mathtt{softmax}(\Tilde{\alpha}))$.
        \item Sets the \supernet architectural parameters to the one-hot $\alpha$, i.e. resulting in a single subnetwork by masking the \supernet.
        \item Passes $\alpha$ as input to \metapredictor.
    \end{enumerate}
    \item The \supernet and \metapredictor do a forward pass using the training data (e.g., images) and hardware embedding, respectively.
    \item Compute the scalarized loss function.
\end{enumerate}

The main problem now is that we cannot directly backpropagate the gradient computation through the \architect to update the \metahypernet parameters $\Phi$. This is due to the sampling from the Categorical distribution in step 2/(a) above being non-differentiable. The STE approximates the gradient for the discrete architectural parameters by ignoring this actual non-differentiable sampling operation.

\newpage
\textbf{Backward pass:}
\begin{enumerate}[leftmargin=*]
    \item Calculate the gradient of the scalarized loss with respect to the discrete architectural parameters $\alpha$: $\partial \mathcal{L} / \partial \alpha$.
    \item Propagate this gradient back to $\Phi$ (\metahypernet parameters) via the probability distribution:
    $$\nabla_{\Phi} \mathcal{L} = \frac{\partial \mathcal{L}}{\partial \alpha}\frac{\partial\alpha}{\partial \mathtt{softmax}}\nabla_{\Phi} \mathtt{softmax}(H_{\Phi}).$$

    STE backpropagates "through" a proxy that treats the non-differentiable function (sampling of $\alpha$) as an identity function (as a result $\frac{\partial\alpha}{\partial \mathtt{softmax}} = 1$) and computes the gradient w.r.t. to the \metahypernet parameters:
    $$\nabla_{\Phi} \mathcal{L} = \frac{\partial \mathcal{L}}{\partial \alpha} \nabla_{\Phi} \mathtt{softmax}(H_{\Phi})$$.
\end{enumerate}
To recap, during the forward pass the \architect samples a discrete architecture from an architecture distribution parameterized by the \metahypernet, and during backpropagation the STE is utilized to propagate back through the sampling operation and update the \metahypernet parameters, hence the distribution from which the discrete architectures in the next iteration will be sampled.

\subsection{Frank-Wolfe Solver}
\label{app:fwsolver}
In this section, we provide the pseudocode of the Frank-Wolfe solver~\citep{Jaggi2013RevisitingFP} used to compute the gradient coefficients used for the MGD updates. To solve the constrained optimization problem, the Frank-Wolfe solver uses analytical solution for the line search with $T=2$ (Algorithm~\ref{alg:argminalg}).

\begin{algorithm}[ht] 
\KwData{$g_\Phi^1, \dots, g_\Phi^T$}
\KwResult{$\mathbf{\gamma} =$ ($\gamma_1$, \dots, $\gamma_T$)}
$\mathtt{Initialize}$ $\mathbf{\gamma} \gets$ ($\frac{1}{T}$, \dots, $\frac{1}{T}$) \\
$\mathtt{Precompute}$ $ \mathcal{M}$ s.t. $\mathcal{M}_{i,j}=(g_\Phi^i)^{\mathbf{T}}(g_\Phi^j)$ \\
\Repeat{$\hat{\delta}\sim 0$ or Number of Iterations Limit}
{
$\hat{t} \gets \argmin_{r}\sum_{t=1}^{T}\gamma_{t}\mathcal{M}_{rt}$ \\
$e_{\hat{t}} \gets \mathcal{M}_{\hat{t}, \cdot}$ \tcp*{$\hat{t}$-th row of $\mathcal{M}$}
$\hat{\delta} \gets \argmin_{\delta}\big((1-\delta)\gamma + \delta e_{\hat{t}}\big)^{\mathbf{T}}\mathcal{M}\big((1-\delta)\gamma+\delta e_{\hat{t}}\big)$ \tcp*{using \cref{alg:argminalg}} 
$\gamma \gets (1-\hat{\delta})\gamma+\hat{\delta}e_{\hat{t}}$
}
\Return{$\gamma$}
\caption{$\mathtt{FrankWolfeSolver}$~\citep{Jaggi2013RevisitingFP}}
\label{alg:fw_algo}
\end{algorithm}

\begin{algorithm}[ht] 
\uIf{$\theta^{\mathbf{T}}\Bar{\theta}\ge\theta^{\mathbf{T}}\theta$}{$\delta\gets1$}\uElseIf{$\theta^{\mathbf{T}}\Bar{\theta}\ge\Bar{\theta}^{\mathbf{T}}\bar{\theta}$}{$\delta\gets0$}\Else{$\delta \gets \frac{(\Bar{\theta}-\theta)^{\mathbf{T}}\Bar{\theta}}{\vert\vert\theta-\Bar{\theta}\vert\vert^2_2}$}
\Return{$\delta$}
\caption{Solver $\min_{\delta\in[0,1]}{\vert\vert\delta\theta+(1-\delta)\Bar{\theta}\vert\vert}_2^2$}
\label{alg:argminalg}
\end{algorithm}

\newpage
\section{Multi-objective NAS algorithms}
\label{sec:algos_full}

This section elaborates on the multi-objective NAS methods we utilize as baselines in Section~\ref{sec:experiments}.

\begin{itemize}[leftmargin=*]
    \item \textbf{Random Search (RS)} is a robust baseline for both single-objective \citep{bergstra-jmlr12a, li-uai20} and multi-objective \citep{cai-iclr2020,chen2021autoformer} architecture searches. This baseline involves randomly sampling architectures from the search space and computing the Pareto front from these samples. While RS is computationally efficient and often effective, it may not always find the optimal architectures, especially in larger search spaces.
    \item \textbf{Local Search (LS)} is adapted to refine solutions near Pareto-optimal points in multi-objective optimization, iteratively improving solutions within defined neighborhoods.
    \item \textbf{Multi-objective Asynchronous Successive Halving (MO-ASHA)}~\citep{Schmucker2021MultiobjectiveAS} is a multi-fidelity method that utilizes an asynchronous successive halving scheduler~\citep{Li2018MassivelyPH} and non-dominating sorting for budget allocation. MO-ASHA uses the NSGA-II selection mechanism and the $\epsilon$-net~\citep{Salinas2021AMP} exploration strategy that ranks candidates in the same Pareto set by iteratively selecting the one with the largest Euclidian distance from the previous set of candidates.
    \item \textbf{Multi-Objective Regularized Evolution (MO-RE)} builds on Regularized Evolution (RE)~\citep{real2019regularized}, which evolves a population of candidates through mutation and periodically removes the oldest individuals, thus regularizing the population. MO-RE adapts this by using multi-objective non-dominated sorting to score candidates, with parents sampled based on these scores.
    \item \textbf{Non-dominated Sorting Genetic Algorithm II (NSGA-II)}~\citep{deb2002fast} is a multi-objective evolutionary algorithm designed to find a Pareto set of architectures. It ranks architectures using non-dominated sorting and maintains diversity with crowding distance. Through selection, crossover, and mutation, NSGA-II evolves populations towards the Pareto front, although it is known for being sample inefficient.
    \item \textbf{Covariance Matrix Adaptation Evolution Strategy (CMA-ES)}~\citep{Igel2007CovarianceMA} is an evolutionary algorithm particularly effective in continuous optimization problems. In a multi-objective context, it adapts its covariance matrix to the shape of the search space, iteratively updating its sampling distribution to favor promising regions. This method efficiently handles complex, non-linear optimization landscapes and can be adapted to multi-objective scenarios by using techniques such as Pareto-based selection to maintain a diverse set of solutions.
    \item \textbf{Latent Action MOO (LaMOO)}~\citep{zhao2022multiobjective} uses a parametric model and Monte Carlo Tree Search (MCTS) to learn to partition the objective space based on the dominance number, which indicates the vicinity of a point to the Pareto front relative to the other samples. qEHVI+LaMOO and CMA-ES+LaMOO use the original qEHVI and CMA-ES, respectively, as an inner routine in the learned subspaces.
    \item \textbf{Bayesian Optimization with Random Scalarizations (RS-BO)}~\citep{paria2020flexible} uses an acquisition function based on random linear scalarizations of objectives across multiple points to find the Pareto-optimal set that minimizes Bayesian regret.
    \item \textbf{Bayesian Optimization with Linear Scalarizations (LS-BO)} is similar to RS-BO but optimizes a single objective derived from a fixed linear combination of two objectives instead of using randomized linear scalarizations.
    \item \textbf{Expected Hypervolume Improvement (qEHVI)}~\citep{daulton-neurips2020a} is a Bayesian optimization acquisition function that explores the Pareto front by quantifying potential hypervolume improvement. This approach measures the volume dominated by Pareto-optimal solutions and guides the search towards regions likely to offer better trade-offs, aiding in the discovery of diverse Pareto-optimal solutions.
\end{itemize}

\section{Evaluation Details}
\label{sec:metrics}

\subsection{Other Metrics}
For NAS-Bench-201, in addition, we evaluate the \emph{generational distance} (GD) and \emph{inverse generational distance} (IGD) (see Appendix~\ref{sec:metrics}). See Figure~\ref{fig:radar_plot_gd_igd} for the results complementary to the hypervolume radar plot in Figure~\ref{fig:radar_plot_hv} of the main paper.

\paragraph{Generational Distance ($GD$) and Inverse Generational Distance ($IGD$).}
Given a \emph{reference set} $\mathcal{S} \subset \archss$ and a Pareto set $\mathcal{P}_\alpha \subset \archss$ with $dim(\archss) = K$, the GD indicator is defined as the distance between every point $\alpha \in \mathcal{P}_\alpha$ and the closest point in $s \in \mathcal{S}$, averaged over the size of $\mathcal{P}_\alpha$:
\begin{equation*}
    GD(\mathcal{P}_\alpha, \mathcal{S}) = \frac{1}{\vert \mathcal{P}_\alpha \vert} \bigg( \sum_{\alpha\in\mathcal{P}_\alpha} \min_{s\in\mathcal{S}} d(\alpha, s)^2 \bigg)^{1/2},
\end{equation*}
where $d(\alpha, s) = \sqrt{\sum_{k=1}^K (\alpha_k - s_k)^2}$ is the Euclidean distance from $\alpha$ to its nearest reference point in $\mathcal{S}$.

The inverted generational distance (IGD) is computed as $IGD(\mathcal{P}_\alpha, \mathcal{S}) = GD(\mathcal{S}, \mathcal{P}_\alpha)$.

\paragraph{Generational Distance Plus ($GD^+$) and Inverse Generational Distance Plus ($IGD^+$).}
$GD^+(\mathcal{P}_\alpha, \mathcal{S}) = IGD^+(\mathcal{S}, \mathcal{P}_\alpha)$ replaces the euclidean distance $d(\alpha, s)$ in GD with:
\begin{equation*}
    d^+ (\alpha, s) = \sqrt{\sum_{k=1}^K (\max \{ \alpha_k - s_k, 0  \} )^2}
\end{equation*}

\subsection{MODNAS-SoTL}
\label{app:sotl}
On the NAS-Bench-201 search space, since the architectures evaluated with the supernetwork weights are not highly correlated to the ones trained independently from scratch, we employ the Sum of Training Losses (SoTL) proxy from \citet{ru2021speedy}.
To profile the Pareto front with SoTL, we firstly evaluate the 24 architectures using the exponential moving average of the sum of training losses for the initial 12 epochs of training as $\sum_{e=1}^{12} 0.9^{12-e} \Loss^{train} (\bw, \alpha)$, and then train from scratch only the subset of architectures in the Pareto set built using the SoTL evaluations. We present the results of MODNAS-SoTL in Figure~\ref{fig:radar_plot_gd_igd}, where we compare to the other baselines as well. As we see, we can further decrease the evaluation cost via MODNAS-SoTL, by trading off the number of solutions in the Pareto set with HV.

\section{Experimental Details}
\label{sec: hyper_configs}
\subsection{\texttt{MetaPredictor} Architectures}
\label{app:meta_predictor_details}
For all search spaces we set the dimensionality of the hardware embedding to 10. This corresponds to latency evaluations on a set of 10 reference architectures, which are the same used by \citet{lee2021help}. 

\begin{figure}[ht]
    \centering
    \begin{subfigure}[b]{0.48\textwidth}
        \centering
        \includegraphics[width=\textwidth]{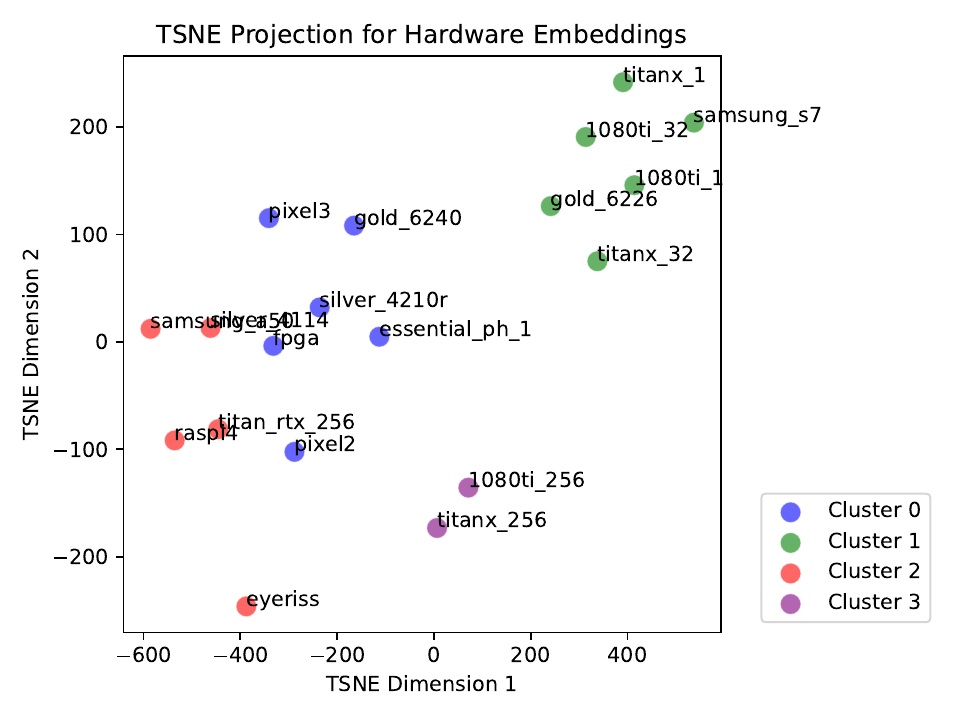}
        \caption{Original hardware embeddings}
        \label{fig:tsnehw}
    \end{subfigure}
    \hfill
    \begin{subfigure}[b]{0.48\textwidth}
        \centering
        \includegraphics[width=\textwidth]{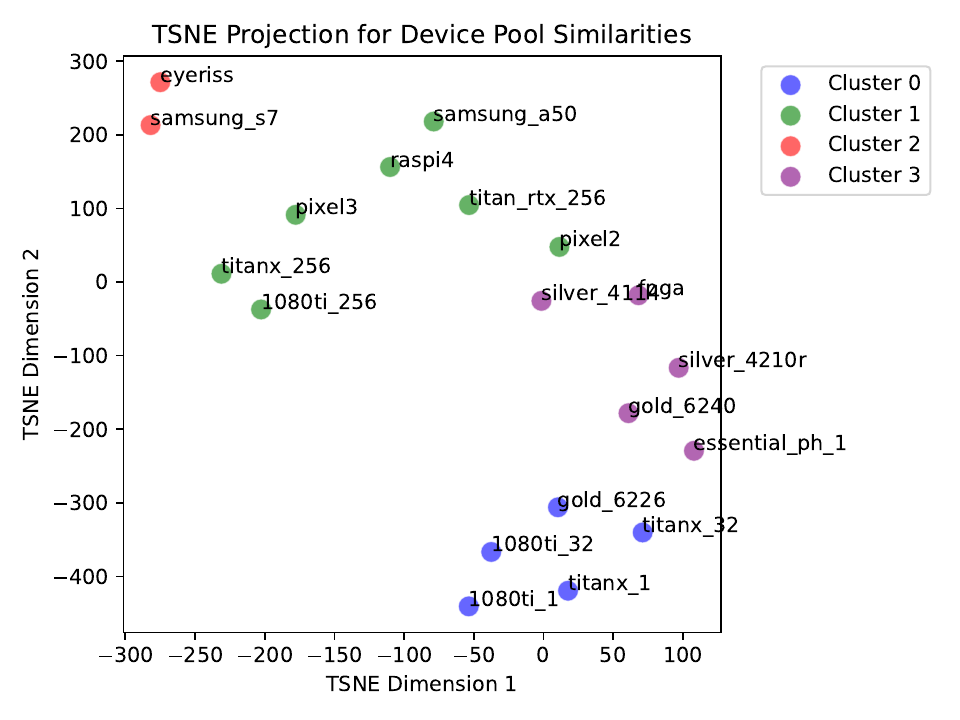}
        \caption{\metahypernet embeddings}
        \label{fig:tsnepool}
    \end{subfigure}
    \caption{K-means clustering on the t-SNE projections of the original hardware device embeddings and learned embeddings from the \metahypernet on NB201.}
    \label{fig:tsne_combined}
\end{figure}

\paragraph{NAS-Bench-201.} For the NAS-Bench-201~\citep{dong-iclr20a} search space we use a Graph Convolutional Network (GCN) as proposed in \citet{dudziak2020brp}. Furthermore, in addition to the \textit{one-hot operation encoding} and \textit{adjacency matrix} corresponding to the architecture cells, we also input the hardware embedding to this predictor, as done by \citet{lee2021help}. The  number of nodes in the GCN is 8 and the dimensionality of the layers is set to 100 following HELP~\citep{lee2021help}. 
In order to show the effectiveness of our \metahypernet to learn the hardware device similarities, in Figure~\ref{fig:tsne_combined} we cluster the original device embedding vectors and the learned \metahypernet embeddings using K-means clustering after reducing their dimensionality using t-SNE. As we can see, the \metahypernet learns to cluster similar devices together in latent space, demonstrating the efficacy of our algorithm.

\paragraph{MobileNetV3 (OFA).}
Following HELP \citep{lee2021help}, we employ a simple feedforward neural network in the MobileNetV3 search space. The input dimension of the \metapredictor is set to 160, matching the concatenated architecture encoding dimension. We set the size of the hidden layers to 100. Specifically, the \metapredictor comprises 2 linear layers with ReLU activation for processing the 160-dimensional one-hot architecture encoding and 2 linear layers for processing the hardware embedding. The outputs from these two paths are concatenated and passed through a final linear layer to predict the latency.


\paragraph{Seq-Seq Transformer (HAT).} HELP~\footnote{\url{https://github.com/HayeonLee/HELP}} does not release the architecture or the meta-learned pretrained predictor for HAT\citep{wang2020hat}. However, HAT~\footnote{\url{https://github.com/mit-han-lab/hardware-aware-transformers}} releases code and pretrained models for each of the devices and tasks trained independently. Hence, we build our single per-task \metapredictor based on the architecture of the HAT predictor, i.e. a simple feedforward neural network. 
The input dimension corresponds to the one-hot architecture encoding of the candidate Transformer architecture. Additionally, to condition on the hardware embedding, we include 2 extra linear layers for processing the hardware embedding, which is then concatenated with the processed architecture encoding to produce the final latency prediction. The hidden dimension of the \metahypernet is set to 400, with 6 hidden layers. The predictor's input feature dimension is 130.

\paragraph{HW-GPT-Bench.} 
We utilize the raw energy observations released in \citep{Sukthanker2024HWGPTBench} to train a single hardware-aware meta-predictor across energy observations from eight GPU types. Our meta-predictor is a simple MLP, similar to the one in HAT, with 4 hidden layers, 2 layers for processing the hardware embedding (which the network is conditioned on). The MLP's hidden dimension is 256, and the input feature dimension matches the one-hot encoded architecture feature map for this space, i.e., 80.

\subsection{\texttt{MetaHypernetwork} Architecture}
\label{sec:hpn_details}

\begin{wrapfigure}[41]{R}{.5\textwidth}
    \centering
    \vspace{-2ex}
    \includegraphics[width=.99\linewidth]{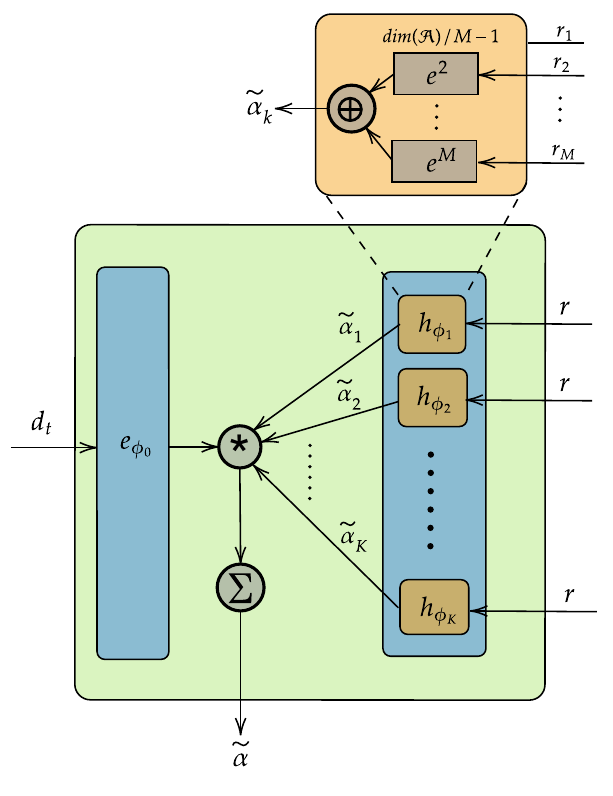}
    \vspace{-5ex}
    \caption{\metahypernet architecture overview in the case of $M$ objectives. Note that $m=1$ is reserved for the accuracy objective, which we model through the cross-entropy loss in the \supernet. The initial linear layer $e_{\phi_0}$ gets the $d_t$ hardware embedding and outputs a weight that scales each of the $K$ hypernetworks' (orange boxes) outputs from the hypernetwork bank. The scaled architectural parameters are then summed up element-wise. All individual hypernetwork $h_{\phi_k}$ get as input the same scalarization $\br$. Each of them has $M-1$ embedding layers with dimensions $n_m \times \frac{dim(\archss)}{M-1}$, $\forall m \in \{ 2, \dots, M \}$ that gets as input the scalarizations for objectives $m=2,\dots,m=M$, and yields a vector of size $\frac{dim(\archss)}{M-1}$. The output from the $M-1$ embedding layers are concatenated to give the architecture encoding $\Tilde{\alpha}$.
    }
    \label{fig:hpn_arch_2}
\end{wrapfigure}

Given a preference vector $\br\in \bbR^\Oscript$, we use the \hypernetwork $h_\phi (\br): \bbR^M \rightarrow \archss$, parameterized by $\phi\in \bbR^n$, to generate an un-normalized architecture distribution $\Tilde{\alpha}$ that is later used to compute the upper-level updates in ($\ref{eq:bilevel_moo}$). 
In our experiments, $h_{\phi}$ is composed of $\Oscript-1$~\footnote{$m=1$ (CE loss) does not have an hardware embedding.} embedding layers $e^m$, $m\in \{2,\dots , \Oscript \}$ with $n_m$ possible learnable vectors of size $\frac{dim(\archss)}{\Oscript - 1}$. The output of $h_\phi$ is the concatenation of all $\Oscript -1$ outputs of $e^m$, such that its size matches $dim(\archss)$. See Figure~\ref{fig:hpn_arch_2} for details.

In order to enable the hypernetwork to generate architectures across multiple devices, inspired by \citet{wang2022learning} and \citet{Lin2020ControllablePM}, we propose a \metahypernet $H_\Phi(\br, d_\tscript): \bbR^M \times \mathcal{H}^{\Oscript -1} \rightarrow \archss$ that can meta-learn across $\Tscript$ different hardware devices (see Figure~\ref{fig:modnas overvies}). The input to $H_\Phi$ is a concatenation of device feature vectors across all metrics, i.e. $d_t = \oplus_{\oscript=2}^{\Oscript} d_t^m$. Similar to \citet{lee2021help}, $d_\tscript^\oscript \in \mathcal{H}$ is a fixed-size feature vector representative of device $\tscript \in \{1,\dots ,\Tscript \}$ and objective $\oscript \in \{2,\dots ,\Oscript \}$, that is obtained by evaluating a fixed set of reference architectures for a given metric.
The \metahypernet, with $\Phi = \cup_{k=0}^K \phi_k$ parameters, contains a bank of $K > \Tscript$ hypernetworks $\{ h_{\phi_k} (\br) \}_{k=1}^K$ and an additional linear layer $e_{\mathcal{\phi}_0} (d_\tscript): \mathcal{H}^{\Oscript -1} \rightarrow  \bbR^K$ at the beginning, that learns a similarity map for every device feature to the hypernetworks' bank.
If we denote by $h_{\phi_{1:k}} = (h_{\phi_1} \cdots h_{\phi_k} )^{\mathbf{T}}$ the vector of all hypernetworks in the bank, then, given a preference vector $\br$, to obtain $\Tilde{\alpha}$ for device $\tscript$, we compute a weighted mixture of predictions of all $h_{\phi}$ in the hypernetwork bank as follows:
\begin{equation*}
\begin{aligned}
    \Tilde{\alpha}_\Phi = H_\Phi(\br, d_\tscript) &= \sum_{k=1}^K e_{\phi_0} (d_\tscript)[k] \cdot h_{\phi_{k}}(\br) \\
    &= e_{\phi_0} (d_\tscript) \cdot h_{\phi_{1:k}}(\br).
\end{aligned}
\end{equation*}

We keep the \texttt{MetaHypernetwork} architecture similar across search spaces. The only thing we adapt is the output dimensionality of the \hypernetwork (in the hypernetwork bank of \texttt{MetaHypernetwork}), which corresponds to the dimensionality of the architecture parameters of the respective search space. We set the size of the initial hardware embedding layer and the hypernetwork bank to 50 for all search spaces. Furthermore, each hypernetwork has 100 possible learnable embeddings $e^m$, for every objective $m\in \{ 2,\dots,M \}$, to map the scalarization vector to an architecture. We quantize the continuous sampled $r_m \in [0, 1]$ to the discrete $[0, 1, … 100]$ interval before indexing the respective embedding layers. See Figure~\ref{fig:hpn_arch} for an illustration of the \metahypernet architecture.

For the \textbf{NAS-Bench-201} search space, we use a single embedding layer of dimensionality 30, corresponding to the dimensionality of the architecture space: $6\times5$ (6 edges and 5 operation choices on each edge). For the 3-objective experiment, we include an additional embedding for the energy usage objective, concatenated with the latency embedding before passing it to the \metahypernet. The individual hypernetworks in the \metahypernet bank have 2 embedding layers with dimensionality 15, whose outputs are concatenated to match the architecture space dimensions.

In the \textbf{MobileNetV3} space, we use 4 embedding layers -- for depth, expansion ratio, kernel size, and resolution. The space comprises 5 blocks, each with 3 depth choices, making the depth embedding layer dimensionality $5 \times 3$. The kernel and expansion embedding layers have dimensions $5 \times 4 \times 3$, corresponding to 5 blocks with a maximum depth of 4 and 3 possible kernel size or expansion ratio choices. The resolution embedding layer has a dimension of 25, representing 25 possible resolution choices.

In the \textbf{Seq-Seq Transformer (HAT)} space, the individual hypernetworks of the \metahypernet utilize 9 embedding layers (the encoder layer count is fixed; see Table~\ref{tab:hat_search_space}):
\begin{itemize}[leftmargin=4ex]
\item 2 embedding layers of size 2 for the encoder and decoder blocks to map the scalarization to the embedding dimension architecture parameter, held constant throughout the encoder or decoder block.
\item 2 embedding layers with dimensions $6 \times 3$ (6 encoder/decoder layers, 3 choices) for the linear layer size in every attention block for both encoder and decoder.
\item 2 embedding layers with dimensions $6 \times 2$ for the number of heads in each attention block.
\item 1 embedding layer of size 6 to encode the 6 possible choices for the number of layers in the decoder.
\item 1 embedding layer of size $6 \times 3$ (6 encoder layers, 3 choices) for the arbitrary encoder layer choice for attention.
\item 1 embedding layer of size $6 \times 2$ (6 encoder layers, 2 choices) for the number of heads in the encoder-decoder attention.
\end{itemize}

For the \textbf{HW-GPT-Bench} space, the individual hypernetworks of the \metahypernet contain 5 embedding layers:
\begin{itemize}[leftmargin=4ex]
\item 1 embedding layer of dimension $1 \times 3$ for mapping the scalarization to the embedding dimension architecture parameter of the language model, with 3 choices.
\item 1 embedding layer of dimension $1 \times 3$ for mapping the scalarization to the layer number dimension architecture parameter of the language model, with 3 choices.
\item 1 embedding layer of dimension $12 \times 3$ for mapping the scalarization to the mlp\_ratio dimension architecture parameter of the language model, with 12 layers and 3 mlp\_ratio choices per layer.
\item 1 embedding layer of dimension $12 \times 3$ for mapping the scalarization to the num\_heads dimension architecture parameter of the language model, with 12 layers and 3 choices per layer.
\item 1 embedding layer of dimension 2 for toggling the bias in linear layers on or off.
\end{itemize} 

\subsection{MODNAS Hyperparameter Configurations}
\label{sec:modnas_hyperparameters}

In Table~\ref{tab:hps}, we show the search hyperparameters and their corresponding values we use to conduct our experiments with MODNAS. For the convolutional spaces we subtract a cosine similarity penalty from the scalarized loss following \citep{ruchte2021scalable}:
\begin{equation}
g_{\Phi}^t \gets \br^{\mathbf{T}} \nabla_\Phi \Lossv_{\tscript}(\dataset_{valid}, \bw, \alpha_\Phi) - \lambda \nabla_\Phi \frac{\br^{\mathbf{T}} \Lossv_{\tscript}(\dataset_{valid}, \bw, \alpha_\Phi)}{\vert\vert \br \vert\vert \hspace{1mm} \vert\vert \Lossv_{\tscript}(\dataset_{valid}, \bw, \alpha_\Phi) \vert\vert },
\label{eqn:cosine_sim}
\end{equation}
where $\vert\vert \cdot \vert\vert$ is the $l_2$ norm. We set $\lambda$ to 0.001. Empirically we did not observe significant differences on disabling the cosine penalty term. 

\subsection{Normalization of objectives}
\label{sec:obj_norm}
Since our method relies on a \textit{scalarization} of different objectives, it is important that the objectives being optimized are on the same scale. For simplicity, lets consider the scenario where the two objectives of interest are the \textit{cross-entropy} loss and \textit{latency}. Since we pretrain and freeze our \metapredictor, the latency-scale remains constant throughout the search, while the cross-entropy loss of the \supernet (likely) decreases over time. To this end, we use the following max-min normalization to normalize the objectives:
\begin{equation}
\Loss_\tscript^\oscript(\cdot,\alpha_\Phi) = \frac{\Loss_\tscript^\oscript(\cdot,\alpha_\Phi) -\min(\mathbf{\Bar{L}})}{\max(\mathbf{\Bar{L}})-\min(\mathbf{\Bar{L}})},
\label{eq:normalization}
\end{equation}
where $\mathbf{\Bar{L}} = \bigcup_{i=1}^N \mathtt{stop\_g}\big(\Loss_\tscript^\oscript(\cdot,\alpha_i)^i\big)$ is the set of losses evaluated on $N$ architectures and potentially $N$ previous steps. For the latency objective, we precompute these sample-statistics using N samples (ground-truth for NAS-Bench-201 and predicted for OFA and HAT spaces) from the search space, whilst for the cross-entropy loss we compute them throughout the search. Furthermore, to take into account the decreasing cross-entropy, we reset the cross-entropy loss statistics after every epoch.

\clearpage\newpage

\begin{table}[ht]
\centering
\caption{Hyperparameters used on different search spaces}
\label{tab:hps}
\resizebox{0.8\linewidth}{!}{%
\begin{tabular}{lllc}
\toprule\toprule
\multicolumn{1}{l|}{\textbf{Search Space}}                    & \multicolumn{2}{c|}{\textbf{Hyperparameter Type}}                                                                                                     & \textbf{Value}       \\ \midrule
\multicolumn{1}{l|}{\multirow{18}{*}{\textbf{NAS-Bench-201}}} & \multicolumn{1}{l|}{\multirow{6}{*}{\metahypernet}}  & \multicolumn{1}{l|}{learning rate}                                                     & 3e-4                 \\ \cline{3-4} 
\multicolumn{1}{l|}{}                                         & \multicolumn{1}{l|}{}                                        & \multicolumn{1}{l|}{weight decay}                                                      & 1e-3                 \\ \cline{3-4} 
\multicolumn{1}{l|}{}                                         & \multicolumn{1}{l|}{}                                        & \multicolumn{1}{l|}{\begin{tabular}[c]{@{}l@{}}embedding layer size\end{tabular}}   & 100                  \\ \cline{3-4} 
\multicolumn{1}{l|}{}                                         & \multicolumn{1}{l|}{}                                        & \multicolumn{1}{l|}{\begin{tabular}[c]{@{}l@{}}hypernetwork bank size\end{tabular}}  & 50                   \\ \cline{3-4} 
\multicolumn{1}{l|}{}                                         & \multicolumn{1}{l|}{}                                        & \multicolumn{1}{l|}{optimizer}                                                         & Adam                 \\ \cline{3-4} 
\multicolumn{1}{l|}{}                                         & \multicolumn{1}{l|}{}                                        & \multicolumn{1}{l|}{\begin{tabular}[c]{@{}l@{}}ReinMax temperature\end{tabular}}    & 1                    \\ \cline{2-4}\cline{2-4}
\multicolumn{1}{l|}{}                                         & \multicolumn{1}{l|}{\multirow{12}{*}{\supernet}} & \multicolumn{1}{l|}{learning rate}                                                     & 0.025                \\ \cline{3-4} 
\multicolumn{1}{l|}{}                                         & \multicolumn{1}{l|}{}                                        & \multicolumn{1}{l|}{momentum}                                                          & 0.9                  \\ \cline{3-4} 
\multicolumn{1}{l|}{}                                         & \multicolumn{1}{l|}{}                                        & \multicolumn{1}{l|}{weight decay}                                                      & 0.0027               \\ \cline{3-4} 
\multicolumn{1}{l|}{}                                         & \multicolumn{1}{l|}{}                                        & \multicolumn{1}{l|}{\begin{tabular}[c]{@{}l@{}}learning rate scheduler\end{tabular}} & cosine               \\ \cline{3-4} 
\multicolumn{1}{l|}{}                                         & \multicolumn{1}{l|}{}                                        & \multicolumn{1}{l|}{epochs}                                                            & 100                  \\ \cline{3-4} 
\multicolumn{1}{l|}{}                                         & \multicolumn{1}{l|}{}                                        & \multicolumn{1}{l|}{batch size}                                                        & 256                  \\ \cline{3-4} 
\multicolumn{1}{l|}{}                                         & \multicolumn{1}{l|}{}                                        & \multicolumn{1}{l|}{gradient clipping}                                                 & 5                    \\ \cline{3-4} 
\multicolumn{1}{l|}{}                                         & \multicolumn{1}{l|}{}                                        & \multicolumn{1}{l|}{cutout}                                                            & true                 \\ \cline{3-4} 
\multicolumn{1}{l|}{}                                         & \multicolumn{1}{l|}{}                                        & \multicolumn{1}{l|}{cutout length}                                                     & 16                   \\ \cline{3-4} 
\multicolumn{1}{l|}{}                                         & \multicolumn{1}{l|}{}                                        & \multicolumn{1}{l|}{initial channels}                                                  & 16                   \\ \cline{3-4} 
\multicolumn{1}{l|}{}                                         & \multicolumn{1}{l|}{}                                        & \multicolumn{1}{l|}{optimizer}                                                         & SGD                  \\ \cline{3-4} 
\multicolumn{1}{l|}{}                                         & \multicolumn{1}{l|}{}                                        & \multicolumn{1}{l|}{train portion}                                                     & 0.5                  \\ \midrule\midrule

\multicolumn{1}{l|}{\multirow{18}{*}{\textbf{MobileNetV3 (OFA)}}} & \multicolumn{1}{l|}{\multirow{6}{*}{\metahypernet}}  & \multicolumn{1}{l|}{learning rate}                                                     & 1e-5                 \\ \cline{3-4} 
\multicolumn{1}{l|}{}                                         & \multicolumn{1}{l|}{}                                        & \multicolumn{1}{l|}{weight decay}                                                      & 1e-3                 \\ \cline{3-4} 
\multicolumn{1}{l|}{}                                         & \multicolumn{1}{l|}{}                                        & \multicolumn{1}{l|}{\begin{tabular}[c]{@{}l@{}}embedding layer size\end{tabular}}   & 100                  \\ \cline{3-4} 
\multicolumn{1}{l|}{}                                         & \multicolumn{1}{l|}{}                                        & \multicolumn{1}{l|}{\begin{tabular}[c]{@{}l@{}}hypernetwork bank size\end{tabular}}  & 50                   \\ \cline{3-4} 
\multicolumn{1}{l|}{}                                         & \multicolumn{1}{l|}{}                                        & \multicolumn{1}{l|}{optimizer}                                                         & Adam                 \\ \cline{3-4} 
\multicolumn{1}{l|}{}                                         & \multicolumn{1}{l|}{}                                        & \multicolumn{1}{l|}{\begin{tabular}[c]{@{}l@{}}ReinMax temperature\end{tabular}}    & 1                    \\ \cline{2-4}\cline{2-4}
\multicolumn{1}{l|}{}                                         & \multicolumn{1}{l|}{\multirow{12}{*}{\supernet}} & \multicolumn{1}{l|}{learning rate}                                                     & 1e-3                \\ \cline{3-4} 
\multicolumn{1}{l|}{}                                         & \multicolumn{1}{l|}{}                                        & \multicolumn{1}{l|}{momentum}                                                          & 0.9                  \\ \cline{3-4} 
\multicolumn{1}{l|}{}                                         & \multicolumn{1}{l|}{}                                        & \multicolumn{1}{l|}{weight decay}                                                      & 3e-5               \\ \cline{3-4} 
\multicolumn{1}{l|}{}                                         & \multicolumn{1}{l|}{}                                        & \multicolumn{1}{l|}{\begin{tabular}[c]{@{}l@{}}learning rate scheduler\end{tabular}} & cosine               \\ \cline{3-4} 
\multicolumn{1}{l|}{}                                         & \multicolumn{1}{l|}{}                                        & \multicolumn{1}{l|}{epochs}                                                            & 50                  \\ \cline{3-4} 
\multicolumn{1}{l|}{}                                         & \multicolumn{1}{l|}{}                                        & \multicolumn{1}{l|}{batch size}                                                        & 32                  \\ \cline{3-4} 
\multicolumn{1}{l|}{}                                         & \multicolumn{1}{l|}{}                                        & \multicolumn{1}{l|}{bn\_momentum}                                                 & 0.1                    \\ \cline{3-4} 
\multicolumn{1}{l|}{}                                         & \multicolumn{1}{l|}{}                                        & \multicolumn{1}{l|}{bn\_eps}                                                            & 1e-5                 \\ \cline{3-4} 
\multicolumn{1}{l|}{}                                         & \multicolumn{1}{l|}{}                                        & \multicolumn{1}{l|}{dropout}                                                     & 0.1                   \\ \cline{3-4} 
\multicolumn{1}{l|}{}                                         & \multicolumn{1}{l|}{}                                        & \multicolumn{1}{l|}{width}                                                  & 1.2                 \\ \cline{3-4} 
\multicolumn{1}{l|}{}                                         & \multicolumn{1}{l|}{}                                        & \multicolumn{1}{l|}{optimizer}                                                         & SGD                  \\ \cline{3-4} 
\multicolumn{1}{l|}{}                                         & \multicolumn{1}{l|}{}                                        & \multicolumn{1}{l|}{train portion}                                                     & 1.0                 \\ \midrule\midrule

\multicolumn{1}{l|}{\multirow{18}{*}{\textbf{Seq-Seq Transformer (HAT)}}} & \multicolumn{1}{l|}{\multirow{6}{*}{\metahypernet}}  & \multicolumn{1}{l|}{learning rate}                                                     & 3e-4                 \\ \cline{3-4} 
\multicolumn{1}{l|}{}                                         & \multicolumn{1}{l|}{}                                        & \multicolumn{1}{l|}{weight decay}                                                      & 1e-3                 \\ \cline{3-4} 
\multicolumn{1}{l|}{}                                         & \multicolumn{1}{l|}{}                                        & \multicolumn{1}{l|}{\begin{tabular}[c]{@{}l@{}}embedding layer size\end{tabular}}   & 100                  \\ \cline{3-4} 
\multicolumn{1}{l|}{}                                         & \multicolumn{1}{l|}{}                                        & \multicolumn{1}{l|}{\begin{tabular}[c]{@{}l@{}}hypernetwork bank size\end{tabular}}  & 50                   \\ \cline{3-4} 
\multicolumn{1}{l|}{}                                         & \multicolumn{1}{l|}{}                                        & \multicolumn{1}{l|}{optimizer}                                                         & Adam                 \\ \cline{3-4} 
\multicolumn{1}{l|}{}                                         & \multicolumn{1}{l|}{}                                        & \multicolumn{1}{l|}{\begin{tabular}[c]{@{}l@{}}ReinMax temperature\end{tabular}}    & 1                    \\ \cline{2-4}\cline{2-4}
\multicolumn{1}{l|}{}                                         & \multicolumn{1}{l|}{\multirow{12}{*}{\supernet}} & \multicolumn{1}{l|}{learning rate}                                                     & 1e-7               \\ \cline{3-4} 
\multicolumn{1}{l|}{}                                         & \multicolumn{1}{l|}{}                                        & \multicolumn{1}{l|}{momentum}                                                          & 0.9                  \\ \cline{3-4} 
\multicolumn{1}{l|}{}                                         & \multicolumn{1}{l|}{}                                        & \multicolumn{1}{l|}{weight decay}                                                      & 0.0              \\ \cline{3-4} 
\multicolumn{1}{l|}{}                                         & \multicolumn{1}{l|}{}                                        & \multicolumn{1}{l|}{\begin{tabular}[c]{@{}l@{}}learning rate scheduler\end{tabular}} & cosine               \\ \cline{3-4} 
\multicolumn{1}{l|}{}                                         & \multicolumn{1}{l|}{}                                        & \multicolumn{1}{l|}{epochs}                                                            & 110                  \\ \cline{3-4} 
\multicolumn{1}{l|}{}                                         & \multicolumn{1}{l|}{}                                        & \multicolumn{1}{l|}{batch size/max-tokens}                                                        & 4096                  \\ \cline{3-4} 
\multicolumn{1}{l|}{}                                         & \multicolumn{1}{l|}{}                                        & \multicolumn{1}{l|}{criterion}                                                 & smoothed\_cross\_entropy                    \\ \cline{3-4} 
\multicolumn{1}{l|}{}                                         & \multicolumn{1}{l|}{}                                        & \multicolumn{1}{l|}{attention-dropout}                                                            & 0.1                \\ \cline{3-4} 
\multicolumn{1}{l|}{}                                         & \multicolumn{1}{l|}{}                                        & \multicolumn{1}{l|}{dropout}                                                     & 0.3                   \\ \cline{3-4} 
\multicolumn{1}{l|}{}                                         & \multicolumn{1}{l|}{}                                        & \multicolumn{1}{l|}{precision}                                                  & float32                   \\ \cline{3-4} 
\multicolumn{1}{l|}{}                                         & \multicolumn{1}{l|}{}                                        & \multicolumn{1}{l|}{optimizer}                                                         & Adam                 \\ \cline{3-4} 
\multicolumn{1}{l|}{}                                         & \multicolumn{1}{l|}{}                                        & \multicolumn{1}{l|}{train portion}                                                     & 1.0                  \\ 
\midrule\midrule
\multicolumn{1}{l|}{\multirow{18}{*}{\textbf{HW-GPT-Bench}}} & \multicolumn{1}{l|}{\multirow{6}{*}{\metahypernet}}  & \multicolumn{1}{l|}{learning rate}                                                     & 1e-5                \\ \cline{3-4} 
\multicolumn{1}{l|}{}                                         & \multicolumn{1}{l|}{}                                        & \multicolumn{1}{l|}{weight decay}                                                      & 1e-3                 \\ \cline{3-4} 
\multicolumn{1}{l|}{}                                         & \multicolumn{1}{l|}{}                                        & \multicolumn{1}{l|}{\begin{tabular}[c]{@{}l@{}}embedding layer size\end{tabular}}   & 100                  \\ \cline{3-4} 
\multicolumn{1}{l|}{}                                         & \multicolumn{1}{l|}{}                                        & \multicolumn{1}{l|}{\begin{tabular}[c]{@{}l@{}}hypernetwork bank size\end{tabular}}  & 50                   \\ \cline{3-4} 
\multicolumn{1}{l|}{}                                         & \multicolumn{1}{l|}{}                                        & \multicolumn{1}{l|}{optimizer}                                                         & Adam                 \\ \cline{3-4} 
\multicolumn{1}{l|}{}                                         & \multicolumn{1}{l|}{}                                        & \multicolumn{1}{l|}{\begin{tabular}[c]{@{}l@{}}ReinMax temperature\end{tabular}}    & 1                    \\ \cline{2-4}\cline{2-4}
\multicolumn{1}{l|}{}                                         & \multicolumn{1}{l|}{\multirow{12}{*}{\supernet}} & \multicolumn{1}{l|}{learning rate}                                                     & 0.000316               \\ \cline{3-4} 
\multicolumn{1}{l|}{}                                         & \multicolumn{1}{l|}{}                                        & \multicolumn{1}{l|}{momentum}                                                          & -                \\ \cline{3-4} 
\multicolumn{1}{l|}{}                                         & \multicolumn{1}{l|}{}                                        & \multicolumn{1}{l|}{weight decay}                                                      & 0.1              \\ \cline{3-4} 
\multicolumn{1}{l|}{}                                         & \multicolumn{1}{l|}{}                                        & \multicolumn{1}{l|}{\begin{tabular}[c]{@{}l@{}}learning rate scheduler\end{tabular}} & cosine               \\ \cline{3-4} 
\multicolumn{1}{l|}{}                                         & \multicolumn{1}{l|}{}                                        & \multicolumn{1}{l|}{steps}                                                            & 800k                  \\ \cline{3-4} 
\multicolumn{1}{l|}{}                                         & \multicolumn{1}{l|}{}                                        & \multicolumn{1}{l|}{batch size/max-tokens}                                                        & 32768                  \\ \cline{3-4} 
\multicolumn{1}{l|}{}                                         & \multicolumn{1}{l|}{}                                        & \multicolumn{1}{l|}{criterion}                                                 & cross\_entropy                    \\ \cline{3-4} 
\multicolumn{1}{l|}{}                                         & \multicolumn{1}{l|}{}                                        & \multicolumn{1}{l|}{attention-dropout}                                                            & 0.0                \\ \cline{3-4} 
\multicolumn{1}{l|}{}                                         & \multicolumn{1}{l|}{}                                        & \multicolumn{1}{l|}{dropout}                                                     & 0.0                   \\ \cline{3-4} 
\multicolumn{1}{l|}{}                                         & \multicolumn{1}{l|}{}                                        & \multicolumn{1}{l|}{precision}                                                  & bfloat16                  \\ \cline{3-4} 
\multicolumn{1}{l|}{}                                         & \multicolumn{1}{l|}{}                                        & \multicolumn{1}{l|}{optimizer}                                                         & AdamW                 \\ \cline{3-4} 
\multicolumn{1}{l|}{}                                         & \multicolumn{1}{l|}{}                                        & \multicolumn{1}{l|}{train portion}                                                     & 1.0                  \\ \bottomrule\bottomrule
\end{tabular}
}
\end{table}

\clearpage
\newpage

\section{Details on Search Spaces}
\label{app:search_spaces}
\begin{wraptable}[12]{R}{.5\textwidth} 
\vspace{-3ex}
\centering
\caption{Encoder-Decoder Search Space for HAT.}
\label{tab:hat_search_space}
\resizebox{\linewidth}{!}{\begin{tabular}{lll}
\toprule
\textbf{Module} & \textbf{Searchable Dim}            & \textbf{Choices}               \\
\midrule
\textbf{Encoder} & No. of Layers               & {[}6{]} (fixed)              \\
        & Embedding dim               & {[}640, 512{]}         \\
        & No. of heads                & {[}8, 4{]}             \\
        & FFN dim                     & {[}3072, 2048, 1024{]} \\
        \midrule
\textbf{Decoder} & No. of layers               & {[}6, 5, 4, 3, 2, 1{]} \\
        & Embedding dim               & {[}640, 512{]}         \\
        & No. of heads                & {[}8, 4{]}             \\
        & FFN dim                     & {[}3072, 2048, 1024{]} \\
        & Arbitrary-Encoder-Layer     & {[}-1, 1, 2{]}         \\
        & Enc-Dec attention num heads & {[}8, 4{]}     \\
        \bottomrule
\end{tabular}}
\end{wraptable}

\textbf{NAS-Bench-201}~\citep{dong-iclr20a} is a convolutional, cell-based search space. The search space consists of 3 stages, each with number of channels 16, 32 and 64, respectively. Each stage contains a convolutional cell repeated 5 times. Here, every cell is represented as a directed acyclic graph (DAG) which has 4 nodes, densely connected with 6 edges. Each edge has 5 possible operation choices: a skip connection, a zero operation, a 3$\times$3 convolution, a 5$\times$5 convolution or an average pooling operation. NAS-Bench-201 is a tabular benchmark exhaustively constructed, where the objective is finding the optimal cell for the given macro skeleton. 

\textbf{MobileNetV3} proposed in OFA \citep{cai-iclr2020} is a macro convolutional search space. The different searchable dimensions in the search space are the depth (per block), the kernel size (for every layer in every block) and the channel expansion ratio (for every layer in every block). There are a total of 5 blocks, each with 3 possible depth choices and every layer in this block has 3 possible kernel sizes and channel expansion ratio choices. This amounts to a total search space size of $((3\times3)^2+(3\times3)^3+(3\times3)^4)^5\approx 2\times 10^{19}$. Additionally, every architecture has 25 possible choices for the size of the input resolution. The 3 possible choices for depth, kernel size and expansion ratio are $\{ 2, 3, 4\}$, $\{3, 5, 7\}$ and $\{3,4,6\}$, respectively. The input resolution choices are $\{$128, 132, 136, 140, 144, 148, 152, 156, 160, 164, 168, 172, 176, 180, 184, 188, 192, 196, 200, 204, 208, 212, 216, 220, 224$\}$. We use a width factor of 1.2 similar to OFA \citep{cai-iclr2020}. 

\textbf{Seq-Seq Encoder-Decoder Transformer (HAT)}~\citep{wang-arxiv20b} for the En-De machine translation task has a searchable number of layers, embedding dimension, feedforward expansion layer dim per-layer, number of heads per-layer for both the encoder and the decoder sub-modules. In addition to this, the number of encoder layers the decoder attends to, and the number of attention heads in the encoder-decoder attention is also searchable. We present the details of the search space in Table~\ref{tab:hat_search_space}.

\textbf{HW-GPT-Bench} \citep{Sukthanker2024HWGPTBench} is a decoder-only transformer space designed for autoregressive language modeling. The search space includes choices for embedding dimensions \{768, 384, 192\}, the number of layers from \{10, 11, 12\}, the MLP expansion ratio per layer from \{2,3,4\}, the number of heads per layer from \{12,8,4\}, and the option to toggle the bias parameter on or off in the layers.

\begin{table}[ht]
\caption{Search-test split for hardware devices and datasets for different search spaces.}
\label{tab:devicesanddatasets}
\resizebox{\textwidth}{!}{
\begin{tabular}{llll}
\toprule
\textbf{Search Space}                      & \textbf{Train-devices}                                          & \textbf{Test devices }    & \textbf{Dataset}          \\
\midrule
\multirow{3}{*}{\textbf{NAS-Bench-201}  }     & 1080ti\_1, 1080ti\_32, 1080ti\_256, silver\_4114,           & titan\_rtx\_256, gold\_6226, & \multirow{3}{*}{CIFAR10}\\
                                   & silver\_4210r, samsung\_a50, pixel3, essential\_ph\_1,      & fpga, pixel2,                \\
                                   & samsung\_s7, titanx\_1, titanx\_32, titanx\_256, gold\_6240 & raspi4, eyeriss             \\
                                   \midrule
\multirow{3}{*}{\textbf{MobileNetV3 (OFA)}} & 2080ti\_1, 2080ti\_32, 2080ti\_64, titan\_xp\_1,            & titan\_rtx\_64             & \multirow{3}{*}{ImageNet-1k}\\
                                   & titan\_xp\_32, titan\_xp\_64, v100\_1, v100\_32,            &                            \\
                                   & v100\_64, titan\_rtx\_1, titan\_rtx\_32                  &                            \\
                                   \midrule
\textbf{Seq-Seq Transformer (HAT) }        & titanxp gpu, cpu xeon                                   & cpu raspberrypi & WMT14.en-de      \\
\midrule
\textbf{HW-GPT-Bench} & a40, v100, rtx2080, rtx3080 &  a100, h100, P100, a6000 & OpenWebText \\
\bottomrule
\end{tabular}
}
\end{table}

\section{Datasets and Devices}
\label{sec:datasets_devices}
This section describes the hardware devices and tasks used to evaluate MODNAS and the MOO baselines throughout the paper. We assess our methods across small- and large-scale image classification datasets, including CIFAR-10 and ImageNet-1K. For the machine translation task, we evaluate our method on the WMT'14 En-De dataset \citep{machacek-bojar-2014-results}, and we use the OpenWebText \citep{Gokaslan2019OpenWeb} dataset for language modeling. Furthermore, we evaluate MODNAS across 19 devices on NAS-Bench-201, 12 devices on MobileNetV3, three devices on Seq-Seq Transformer, and eight devices from HW-GPT-Bench \citep{Sukthanker2024HWGPTBench}, with zero-shot generalization to test devices. Table~\ref{tab:devicesanddatasets} lists the devices used. For more details on the devices, we refer readers to \citet{lee2021help}, \citet{cai-iclr2020}, \citet{wang2020hat}, \citet{li2021hw}, and \citet{Sukthanker2024HWGPTBench}.

\section{Runtime Comparison}
\label{sec:runtime_comparison}

In Table~\ref{tab:search_times} we provide the number of GPU hours we ran MODNAS and baselines on every search space. We ran the search on NAS-Bench-201, OFA, together with the evaluations on Nvidia RTX2080Ti, while for HAT we used NVidia A6000. For both OFA and HAT, we used 8 GPUs in parallel. Similar as in \citet{Sukthanker2024HWGPTBench}, on the HW-GPT-Bench space we ran the MODNAS search and evaluations on 4 Nvidia A100 GPUs.

\begin{table}[ht]
    \centering
    \caption{Total amount of GPU hours required to run MODNAS' and baselines' search on every search space.}
    \label{tab:runtimes}
    \resizebox{\linewidth}{!}{
    \begin{tabular}{llccccc}
    \toprule
        \textbf{Search Spaces} & \textbf{Method} & \textbf{Lat/En/Mem Pred.} & \textbf{Supernet} & \textbf{Acc./Ppl Pred.} & \textbf{Search} & \textbf{Total Time} \\ \midrule\midrule
        \multirow{3}{*}{NASBench201}
         & MetaD2A+HELP & 25 & - & 8629 & 0.3 & 8654.3 \\ \cmidrule{2-7}
         & MOO Baselines & - & - & - & 370.5 & 370.5 \\ \cmidrule{2-7}
         & MODNAS & 3 & 22 & - & 0.05 & \textbf{25.25} \\ \midrule\midrule
        \multirow{3}{*}{Once-For-All} 
         & OFA+HELP & 6 & 1200 & 356 & 10 & 1572 \\ \cmidrule{2-7}
         & MOO Baselines & 6 & 1200 & 356 & 192 & 1754 \\ \cmidrule{2-7}
         & MODNAS & 6 & 1392 & - & 0.05 & \textbf{1398.25} \\ \midrule\midrule
        \multirow{4}{*}{HAT} 
         & HAT & 15 & 346.7 & - & 210.9 & 572.6 \\ \cmidrule{2-7}
         & MOO Baselines & 15 & 346.7 & - & 576 & 937.7 \\ \cmidrule{2-7}
         & MODNAS & 5 & 576 & - & 0.05 & \textbf{581.25} \\ \midrule\midrule
        \multirow{2}{*}{HW-GPT-Bench}
         & MOO Baselines & 1 & 192 & - & 48 & 241 \\ \cmidrule{2-7}
         & MODNAS & 1 & 216 & - & 0.05 & \textbf{217.25} \\ \bottomrule
    \end{tabular}}
\label{tab:search_times}
\end{table}

\section{Additional Experiments}
\label{sec:additional_esperiments}

\begin{figure}[t!]
    \centering
    \includegraphics[width=.24\linewidth,height=4cm]{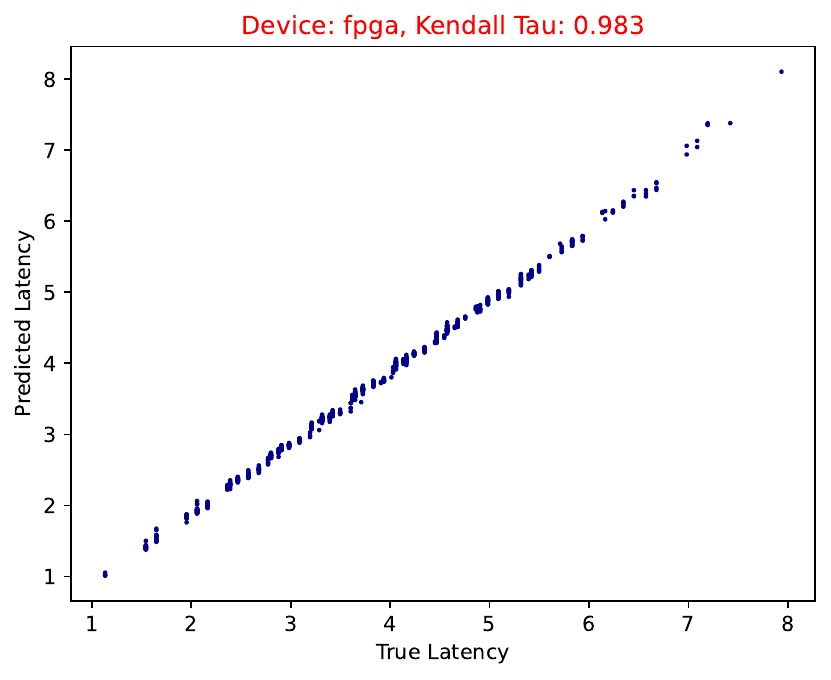}
    \includegraphics[width=.24\linewidth,height=4cm]{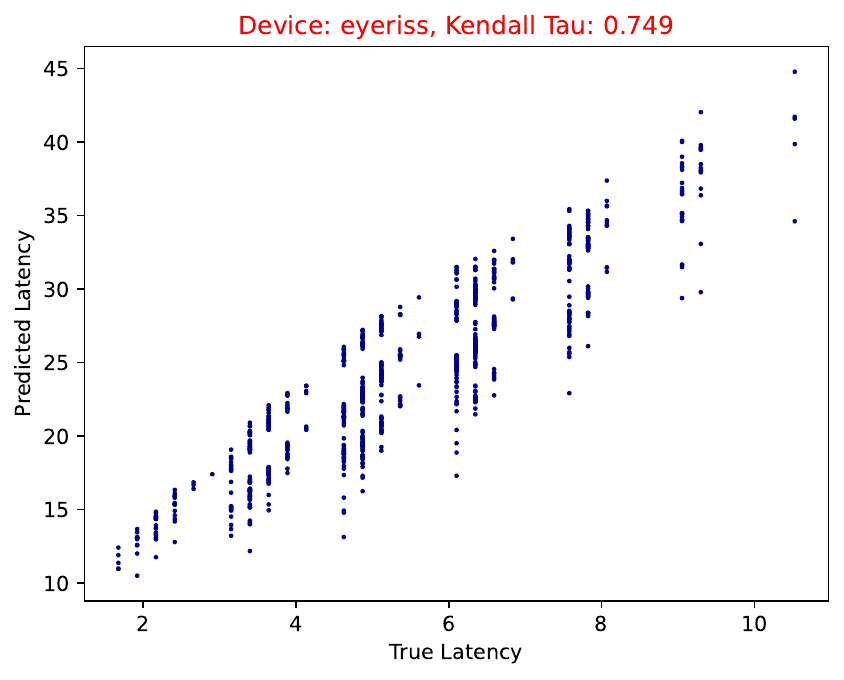}
    \includegraphics[width=.24\linewidth,height=4cm]{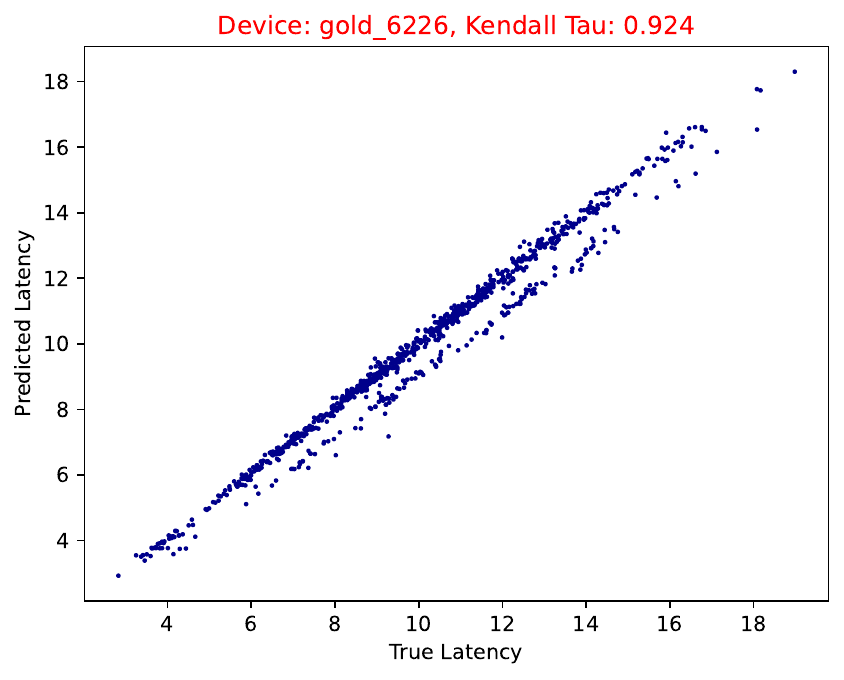}
    \includegraphics[width=.24\linewidth,height=4cm]{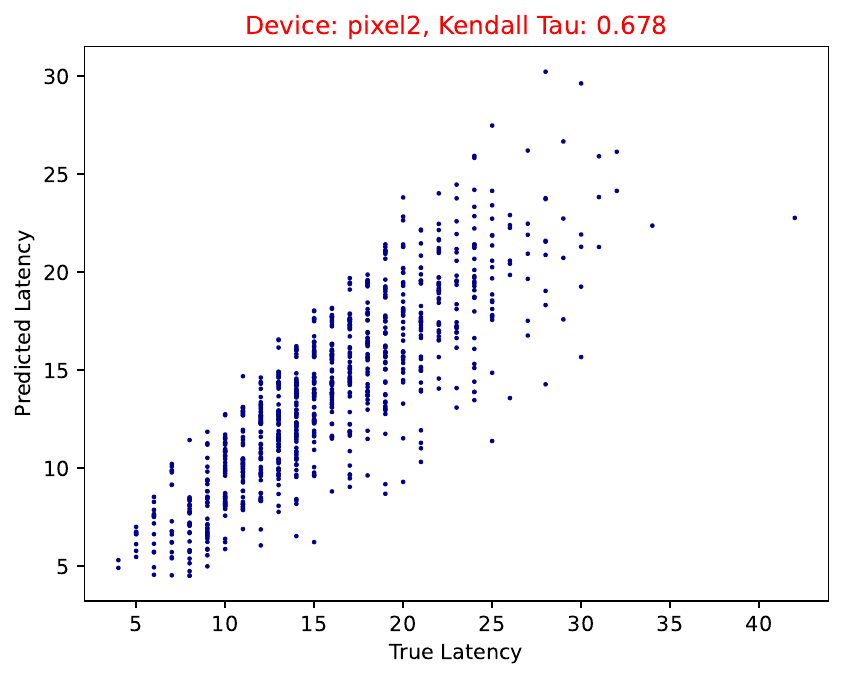} \\
    \includegraphics[width=.24\linewidth,height=4cm]{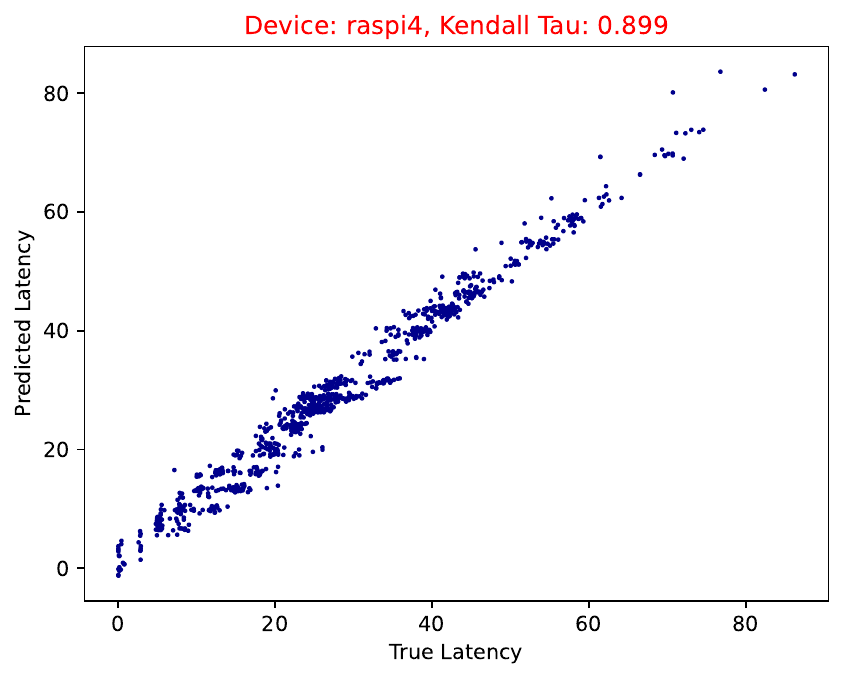}
    \includegraphics[width=.24\linewidth,height=4cm]{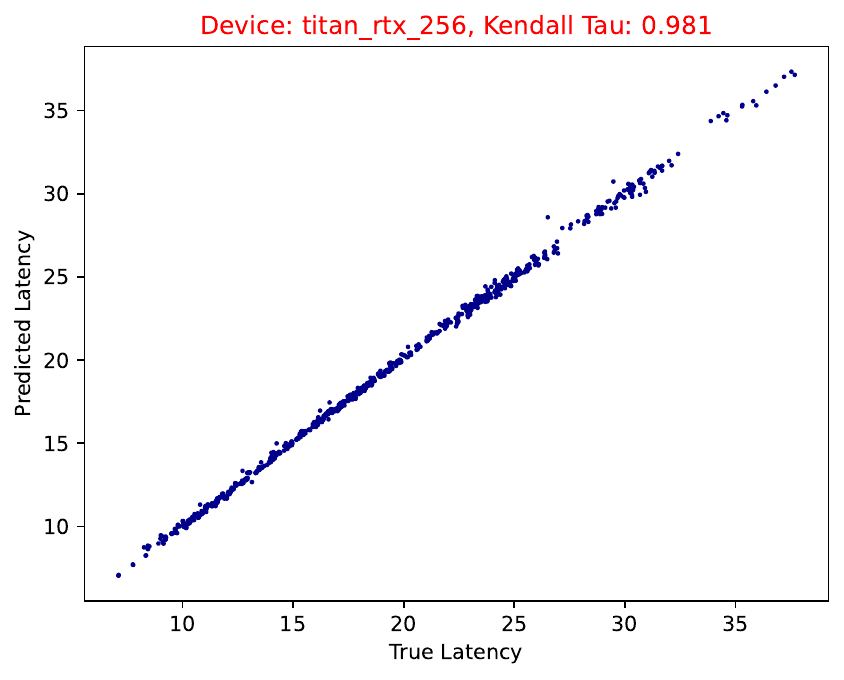}
    \includegraphics[width=.24\linewidth,height=4cm]{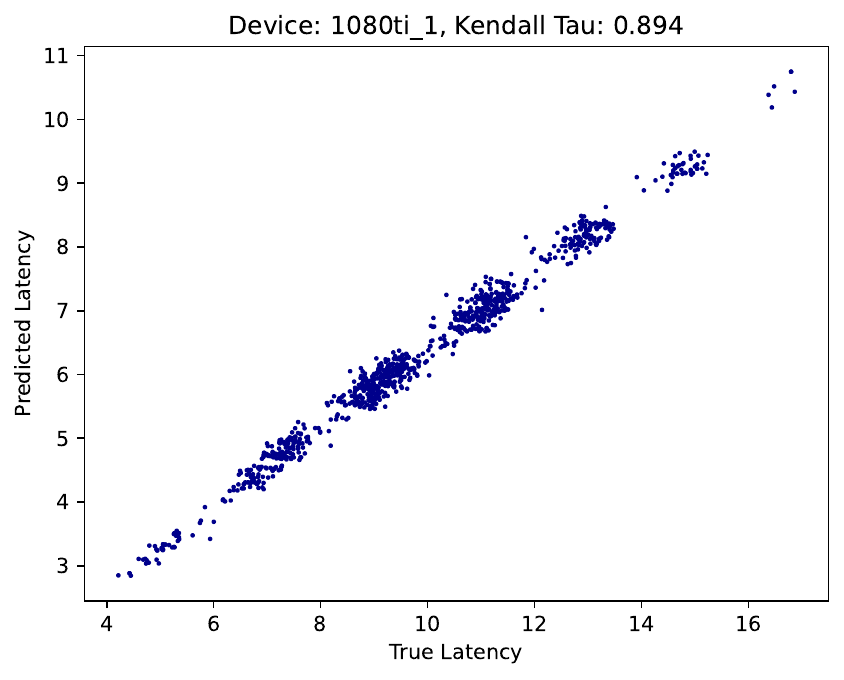}
    \includegraphics[width=.24\linewidth,height=4cm]{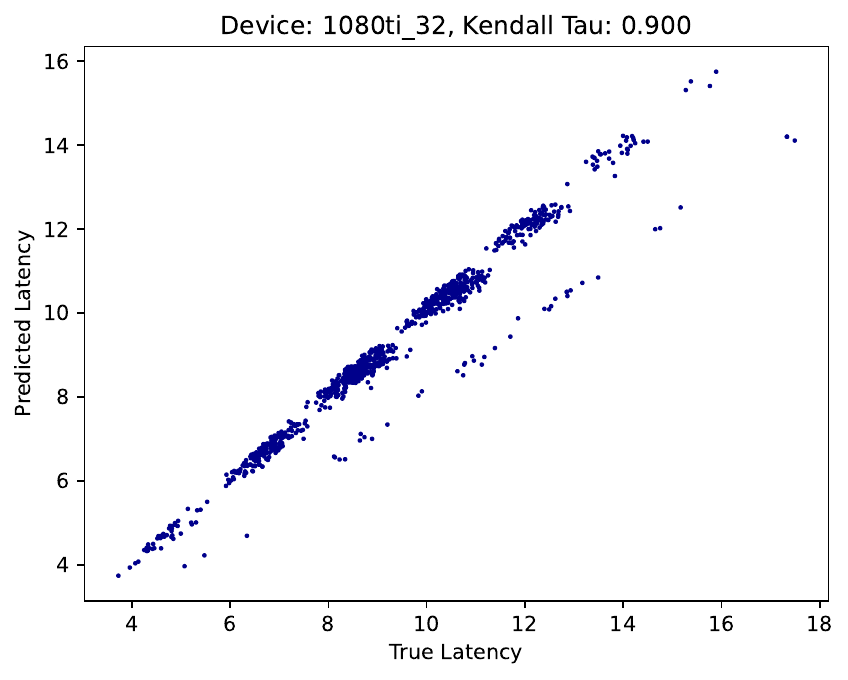} \\
    \includegraphics[width=.24\linewidth,height=4cm]{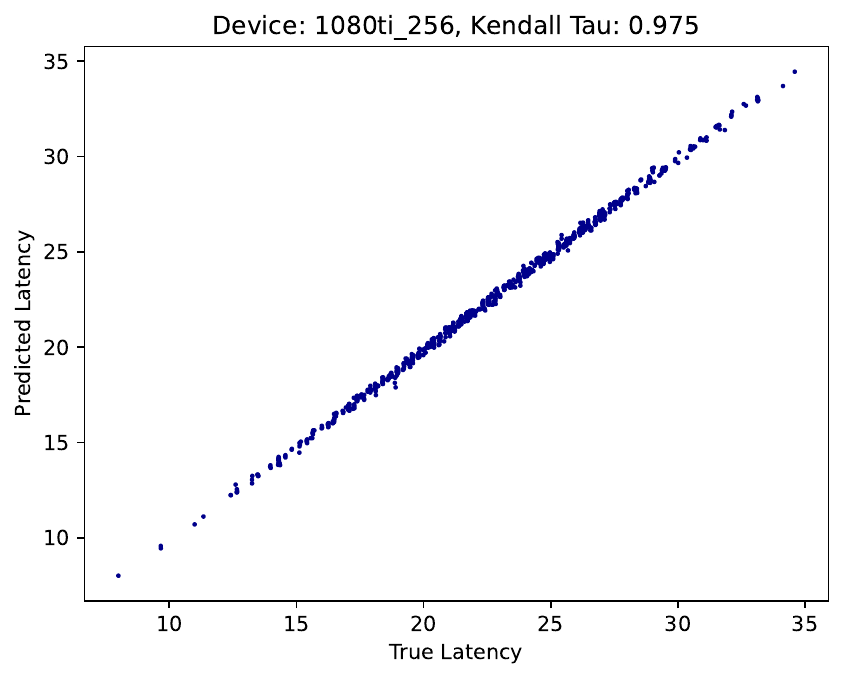}
    \includegraphics[width=.24\linewidth,height=4cm]{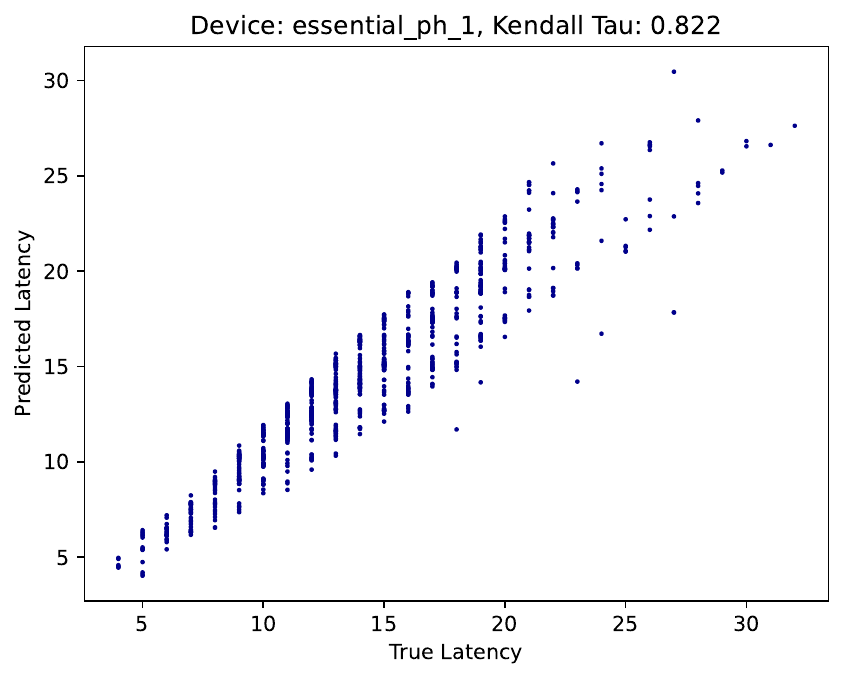}
    \includegraphics[width=.24\linewidth,height=4cm]{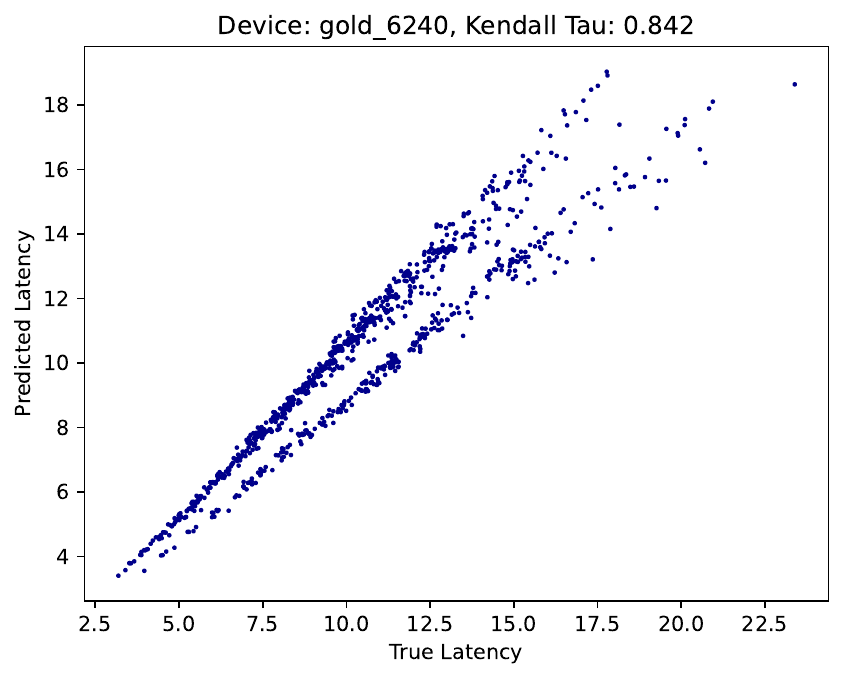}
    \includegraphics[width=.24\linewidth,height=4cm]{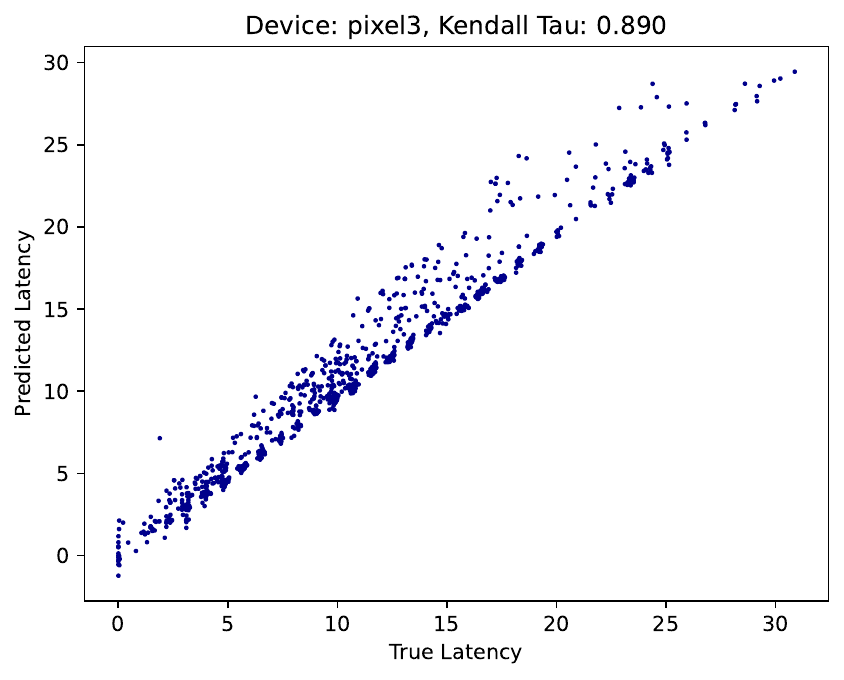} \\
    \includegraphics[width=.24\linewidth,height=4cm]{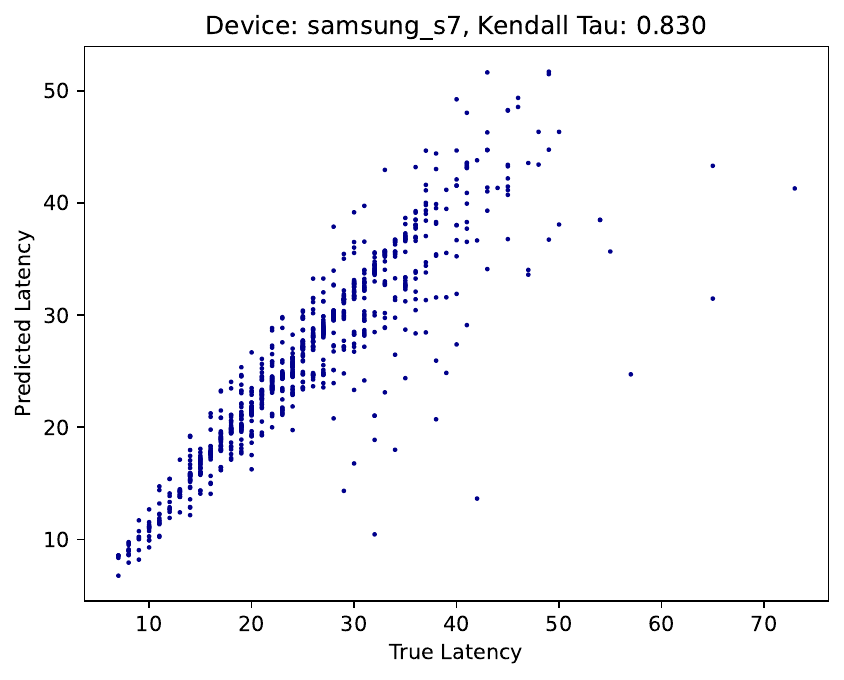}
    \includegraphics[width=.24\linewidth,height=4cm]{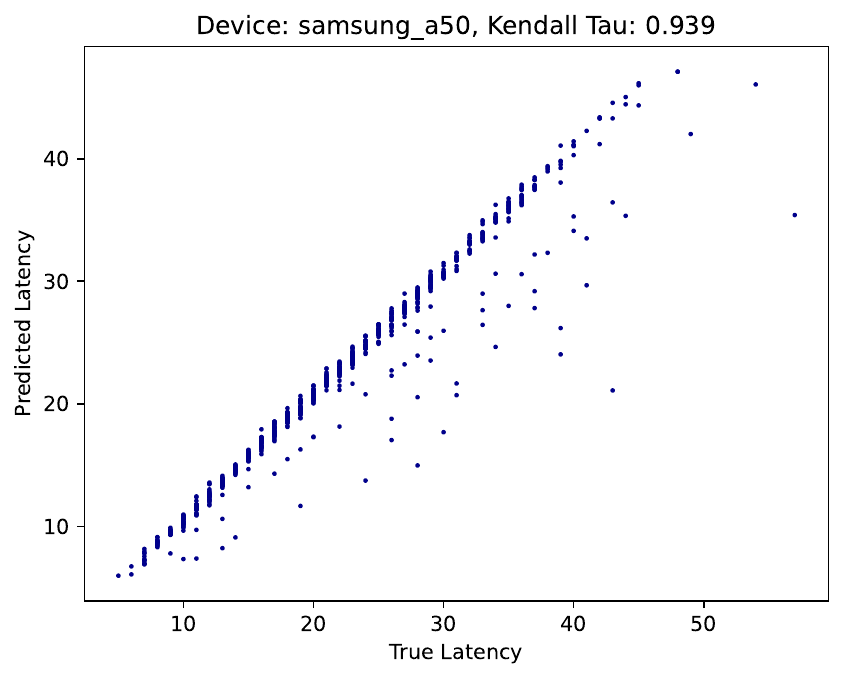}
    \includegraphics[width=.24\linewidth,height=4cm]{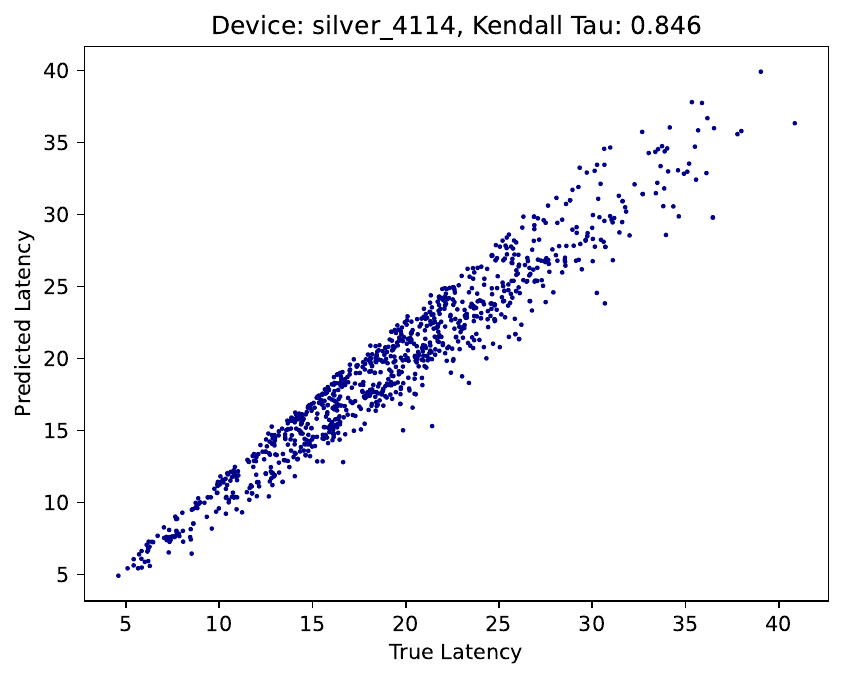}
    \includegraphics[width=.24\linewidth,height=4cm]{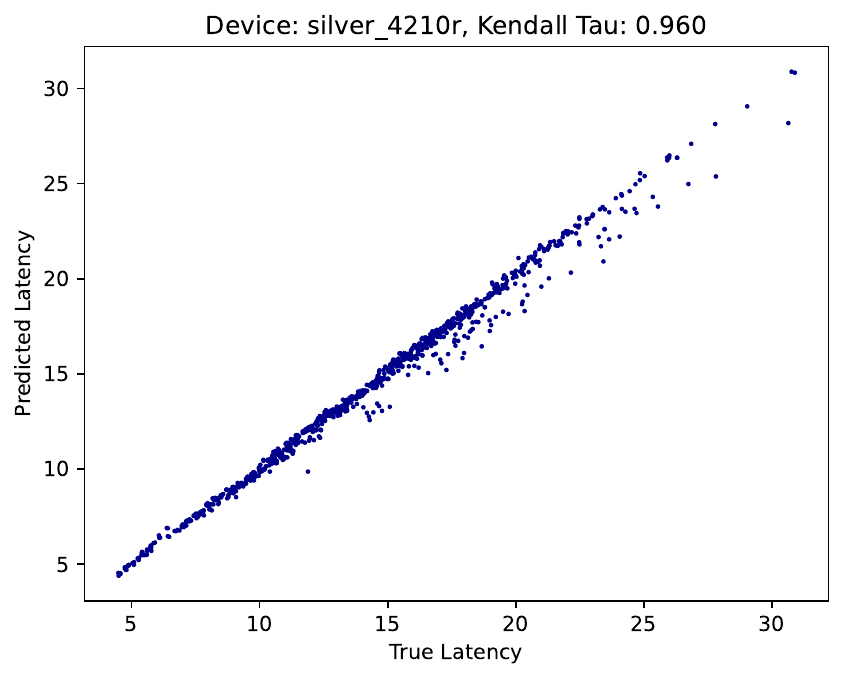} \\
    \includegraphics[width=.24\linewidth,height=4cm]{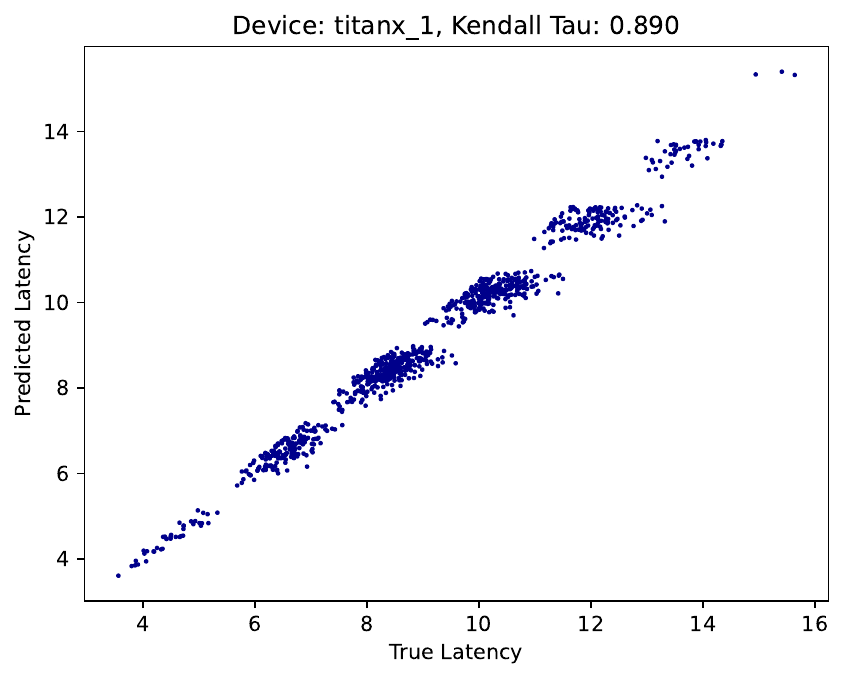}
    \includegraphics[width=.24\linewidth,height=4cm]{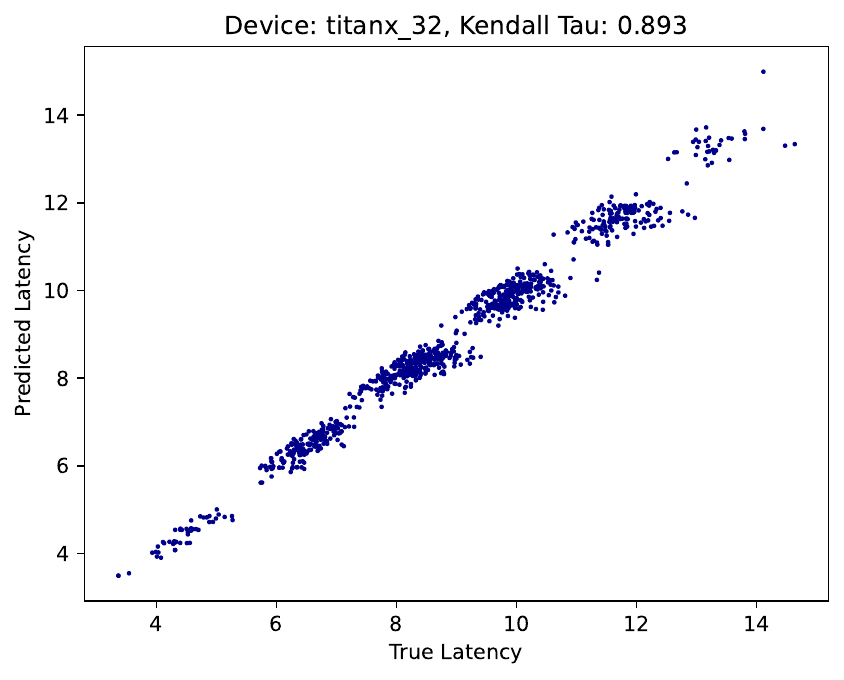}
    \includegraphics[width=.24\linewidth,height=4cm]{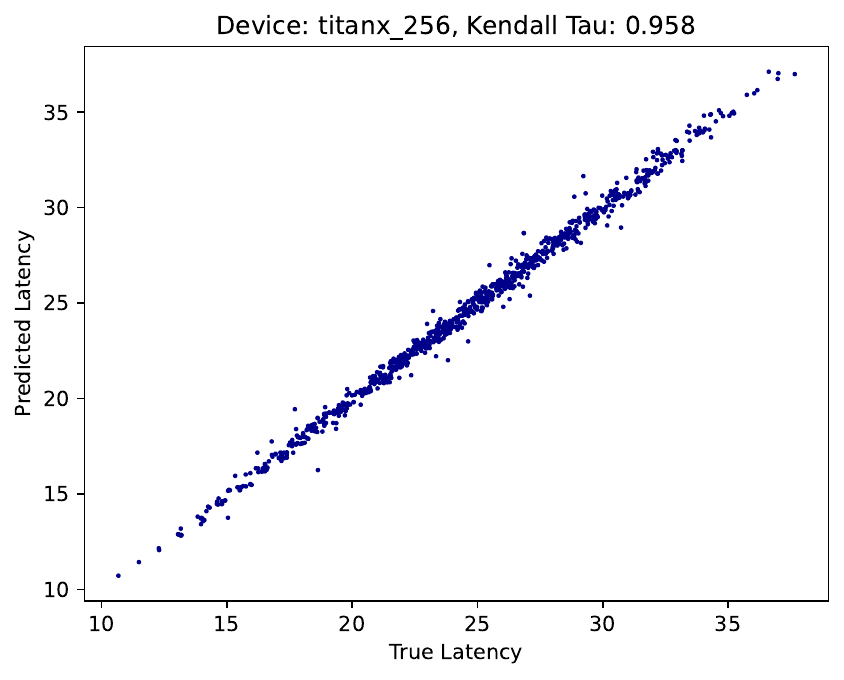}
    \caption{Scatter plots of predicted latencies from our pretrained \metapredictor vs. ground-truth latencies (test devices in red).}
    \label{fig:lat_pred}
\end{figure}

\begin{figure}[ht]
\centering
\begin{minipage}{0.33\textwidth}
  \centering
\includegraphics[width=\textwidth]{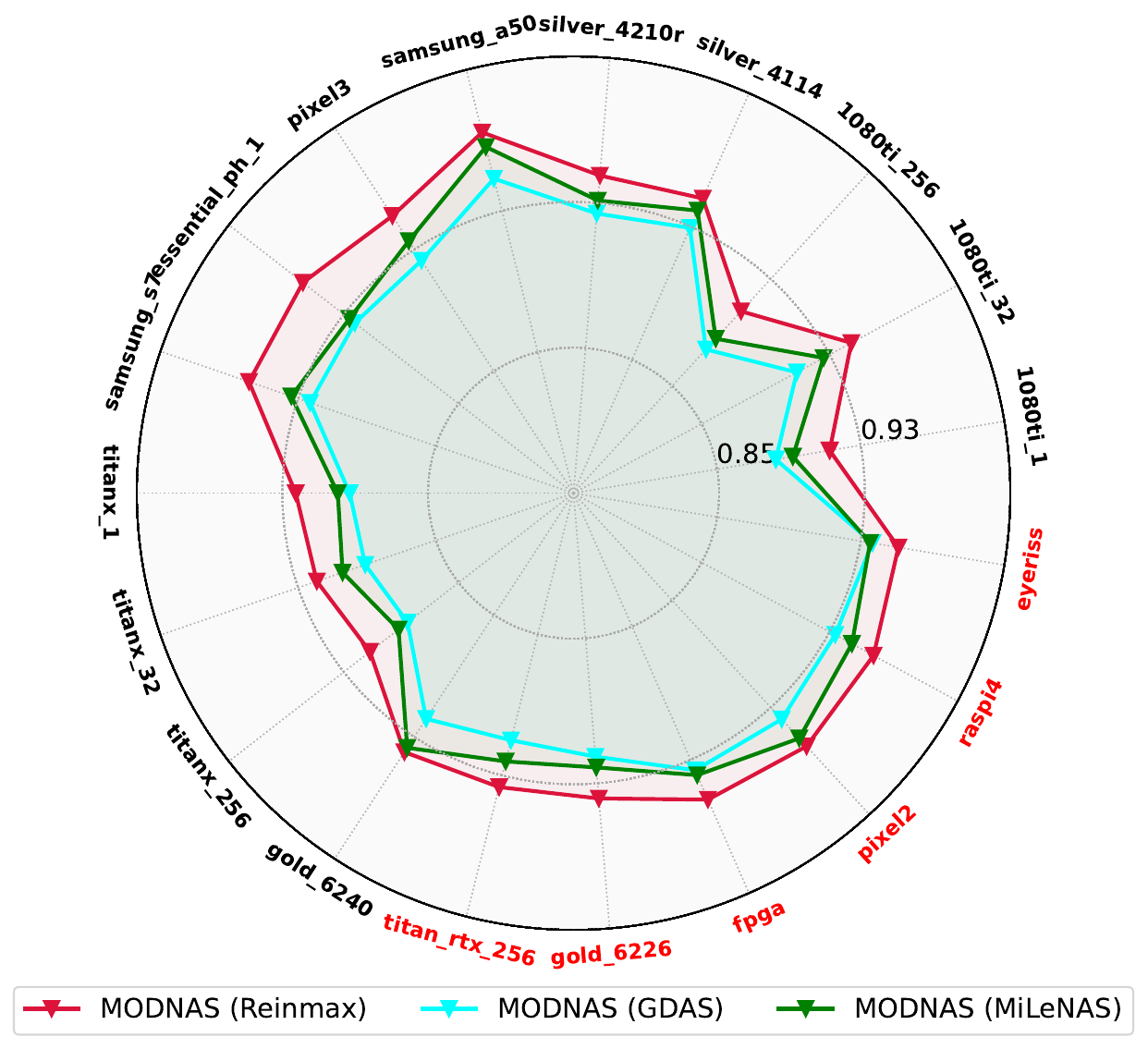}
\subcaption[hv_gdas]{$\text{Hypervolume}$}\label{fig:radar_hv_gdas}
\end{minipage}%
\begin{minipage}{0.33\textwidth}
  \centering
\includegraphics[width=\textwidth]{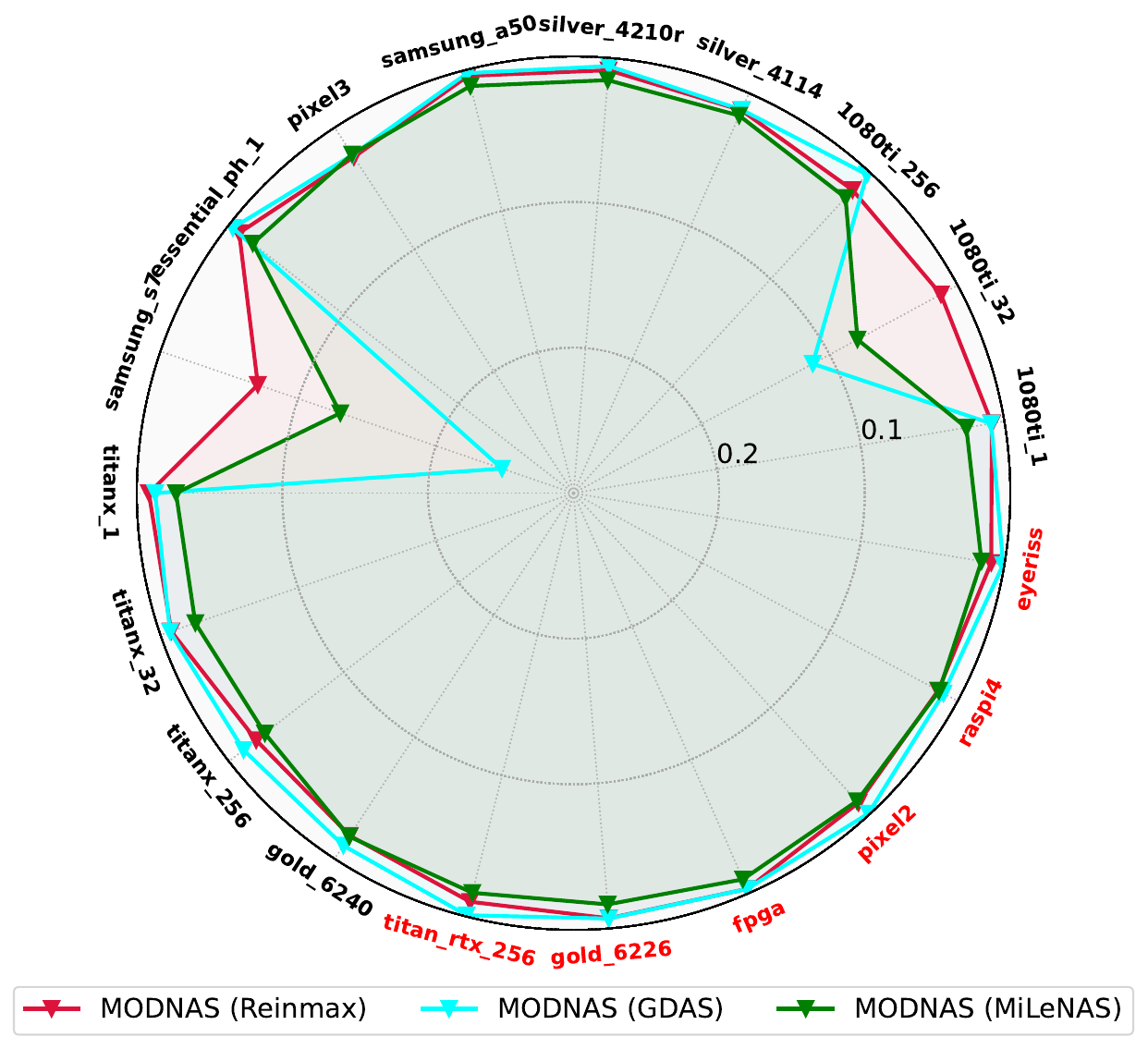}
\subcaption[gd_gdas]{$\text{GD+}$}\label{fig:radar_gd_gdas}
\end{minipage}
\begin{minipage}{0.33\textwidth}
  \centering
\includegraphics[width=\textwidth]{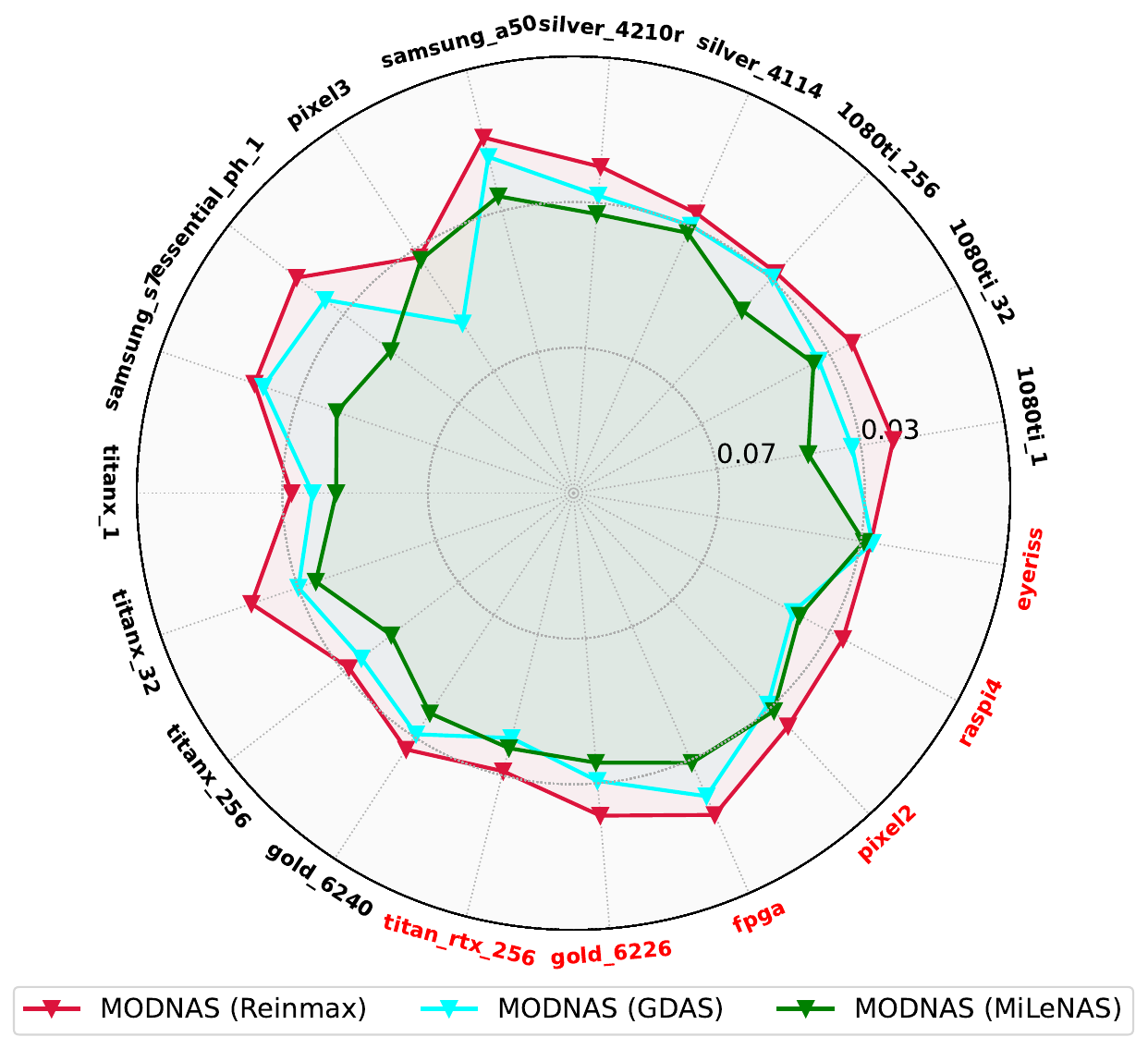}
\subcaption[igd_gdas]{$\text{IGD+}$}\label{fig:radar_igd_gdas}
\end{minipage}
\caption{Hypervolume, GD+ and IGD+ of MODNAS with Reinmax as gradient estimator in the \architect vs. the one from GDAS~\citep{dong-cvpr19} {\color{blue}and MiLeNAS~\citep{He2020MiLeNASEN}} across 19 devices on NAS-Bench-201. Higher area in the radar indicates better performance for every metric. Test devices are colored in red around the radar plot.} 
\label{fig:radar_plot_gdas}
\end{figure}

\begin{figure}[ht]
    \centering
    \begin{minipage}{0.37\linewidth}
    \centering
    \includegraphics[width=\linewidth]{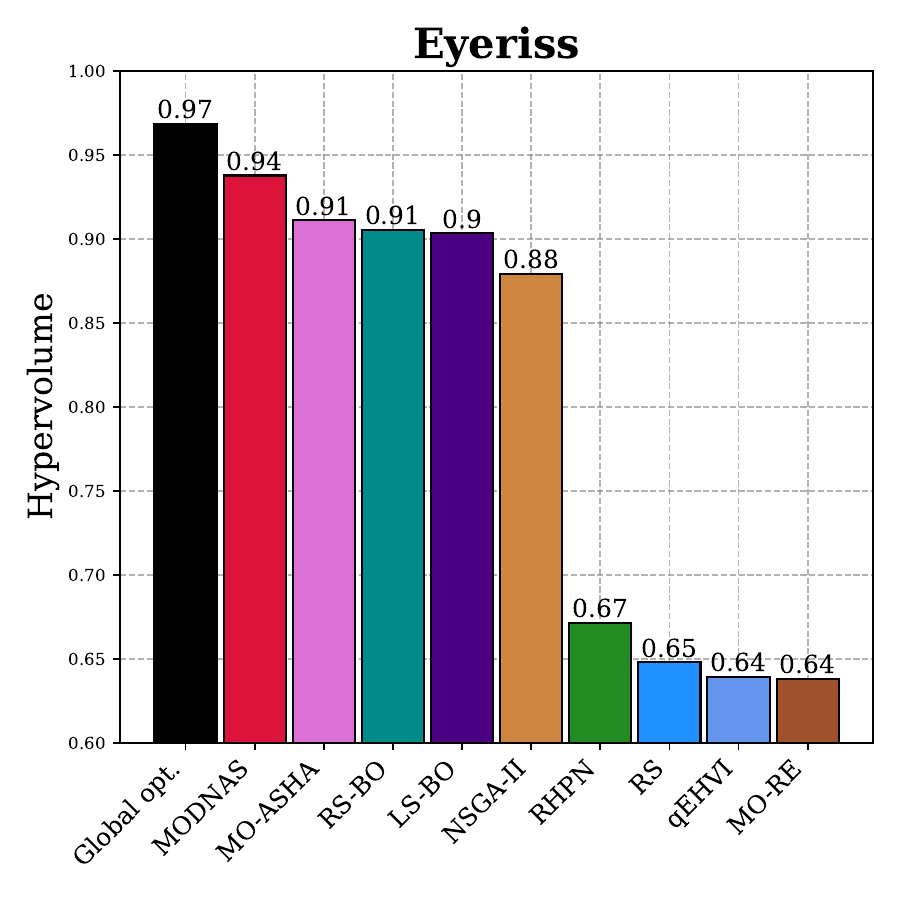}
    \end{minipage}%
    \begin{minipage}{0.5\linewidth}
    \centering
    \includegraphics[width=.89\linewidth]{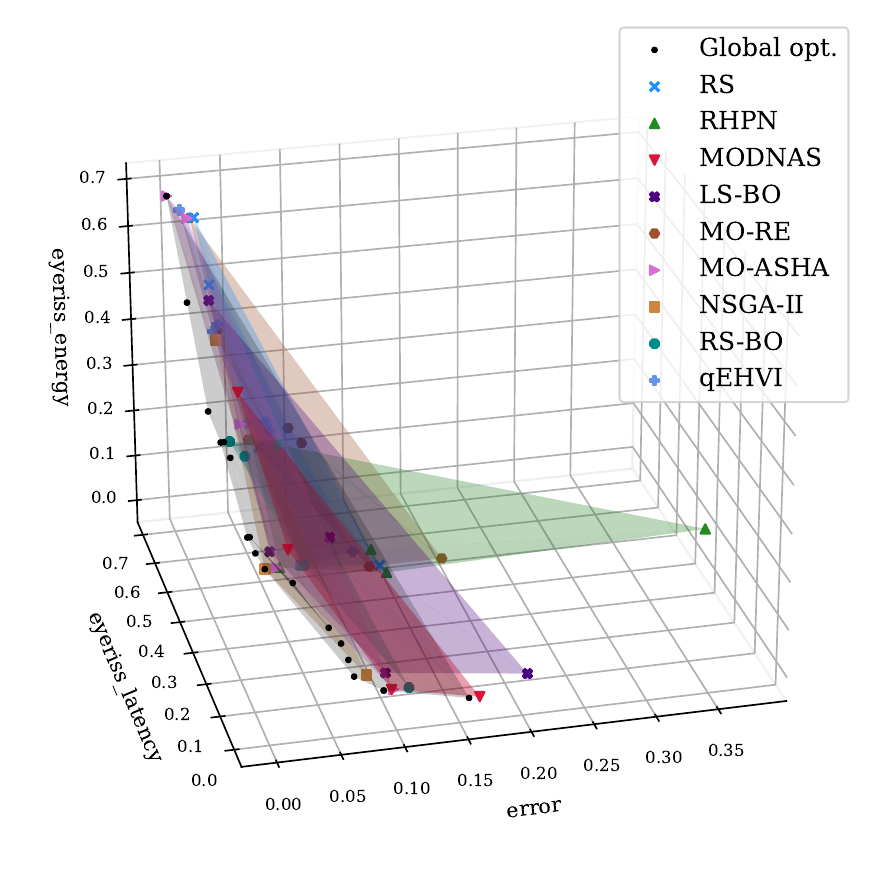}
    \end{minipage}
    \caption{HV (\textit{left}) and Pareto front (\textit{right}) of MODNAS and baselines on Eyeriss with 3 normalized objectives: error, latency and energy usage. HV was computed using the $(1,1,1)$ reference point on the right 3D plot.}
    \label{fig:3d_plot_eyeriss}
    \vspace{-3ex}
\end{figure}

\begin{figure}[ht]
    \centering
    \hspace{-7mm}
    \begin{minipage}[b]{0.39\textwidth}
        \centering
        \includegraphics[width=\textwidth]{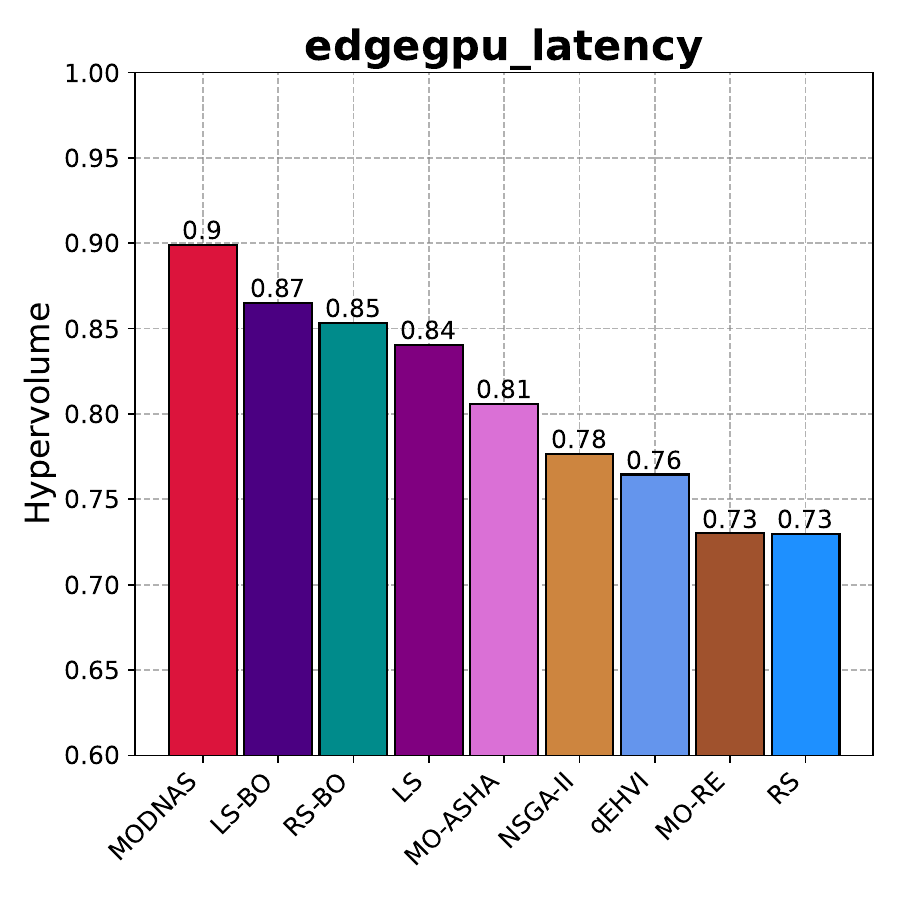}
        \caption{Hypervolume on CIFAR-100 and edgegpu device.}
        \label{fig:edgegpu-c100}
    \end{minipage}
    \hspace{7mm}
    \begin{minipage}[b]{0.39\textwidth}
        \centering
        \includegraphics[width=\textwidth]{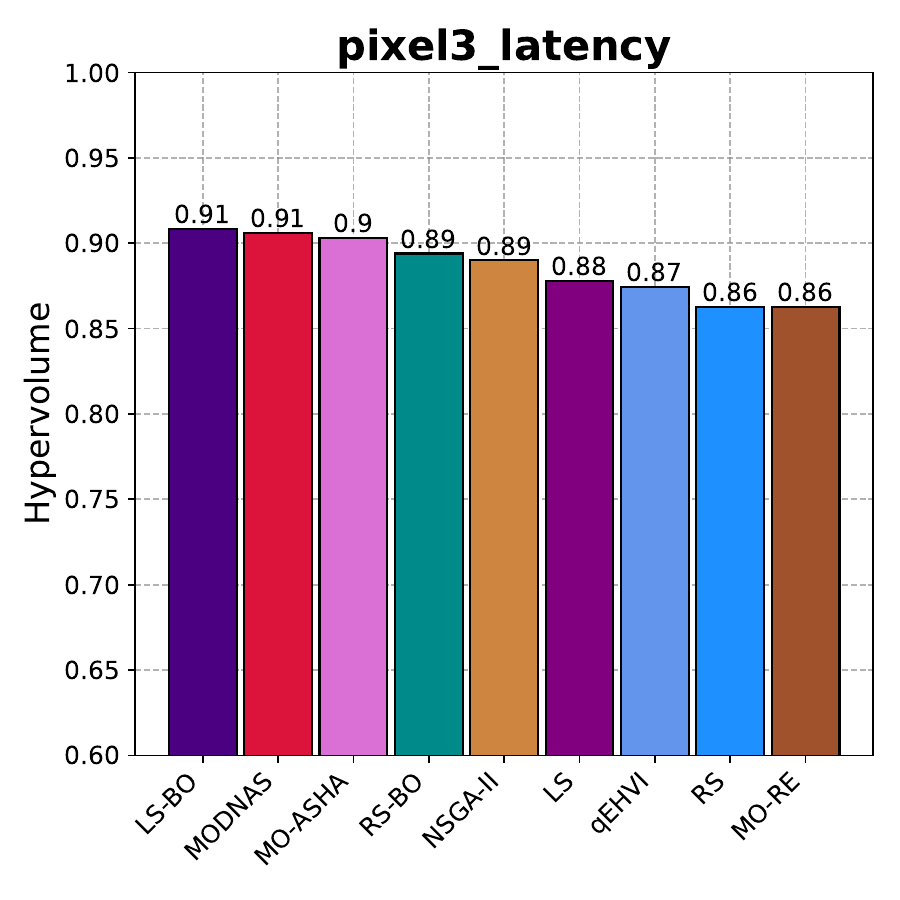}
        \vspace{-11pt}
        \caption{Hypervolume on CIFAR-100 and Pixel3 device.}
        \label{fig:pixel3-c100}
    \end{minipage}
\end{figure}

\subsection{Additional Results on NAS-Bench-201}
\label{sec:additional_nb201_results}
\begin{figure}[t!]
    \centering
    \includegraphics[width=.24\linewidth]{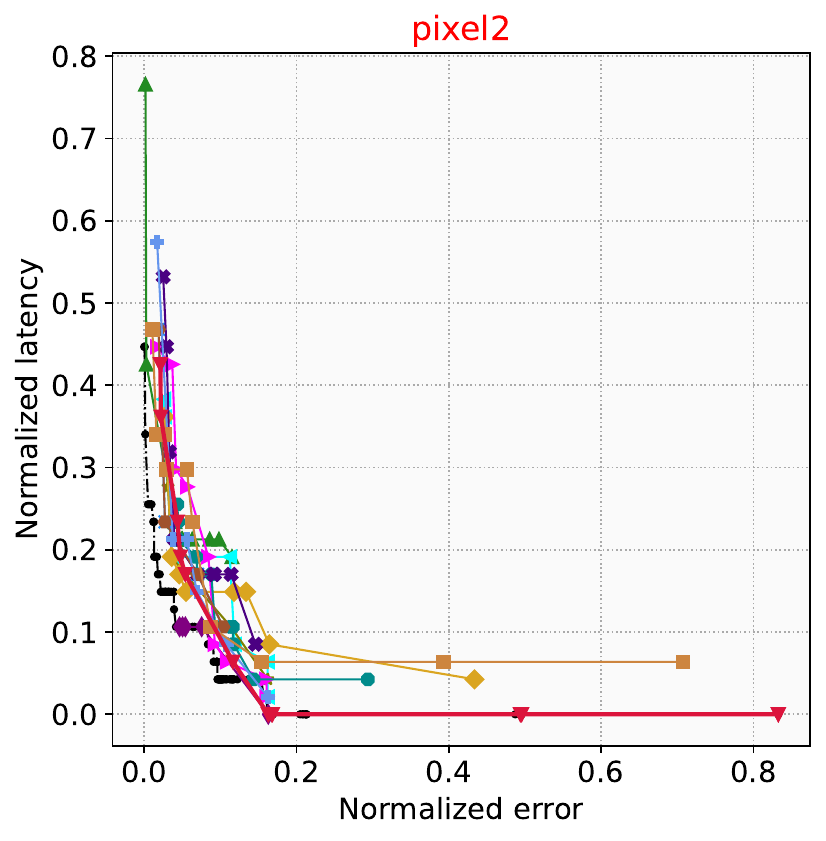}
    \includegraphics[width=.24\linewidth]{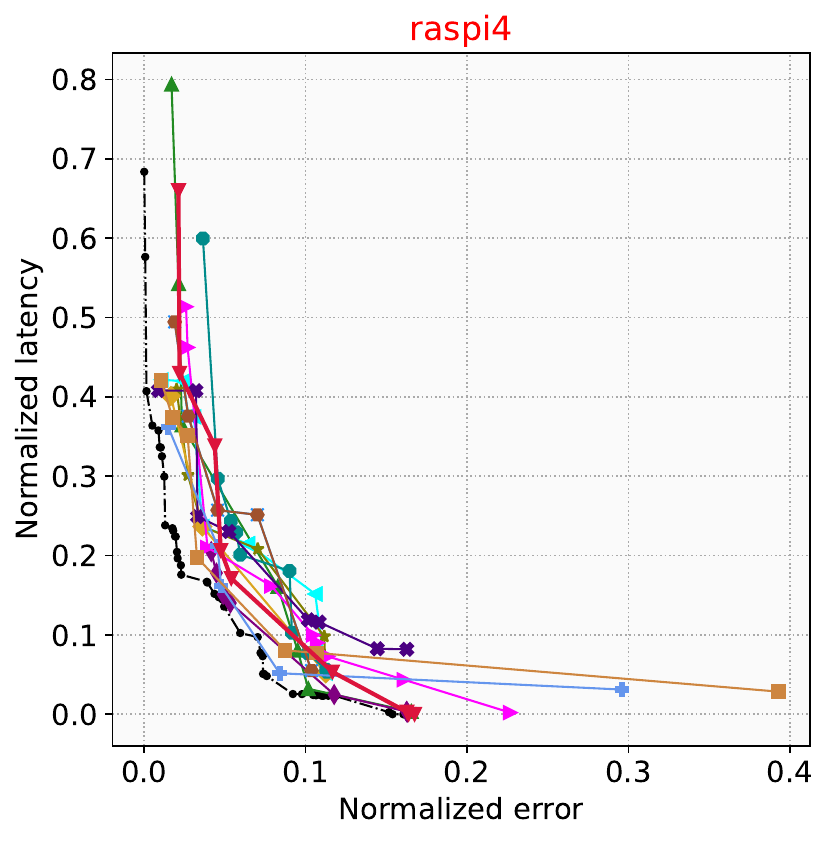}
    \includegraphics[width=.24\linewidth]{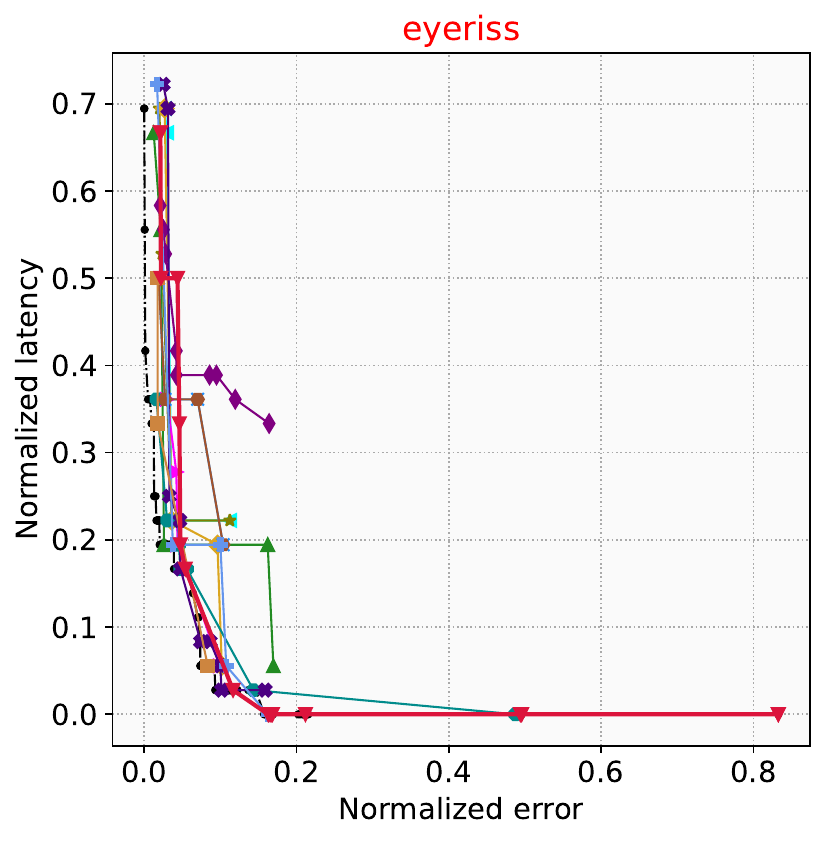}
    \includegraphics[width=.24\linewidth]{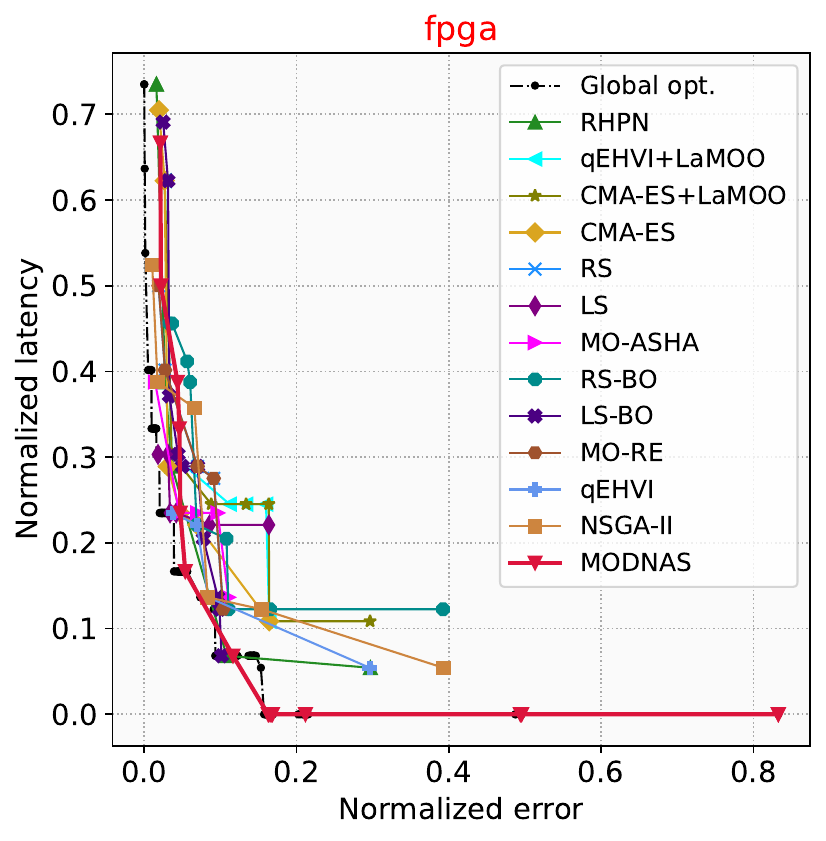}\\
    \includegraphics[width=.24\linewidth]{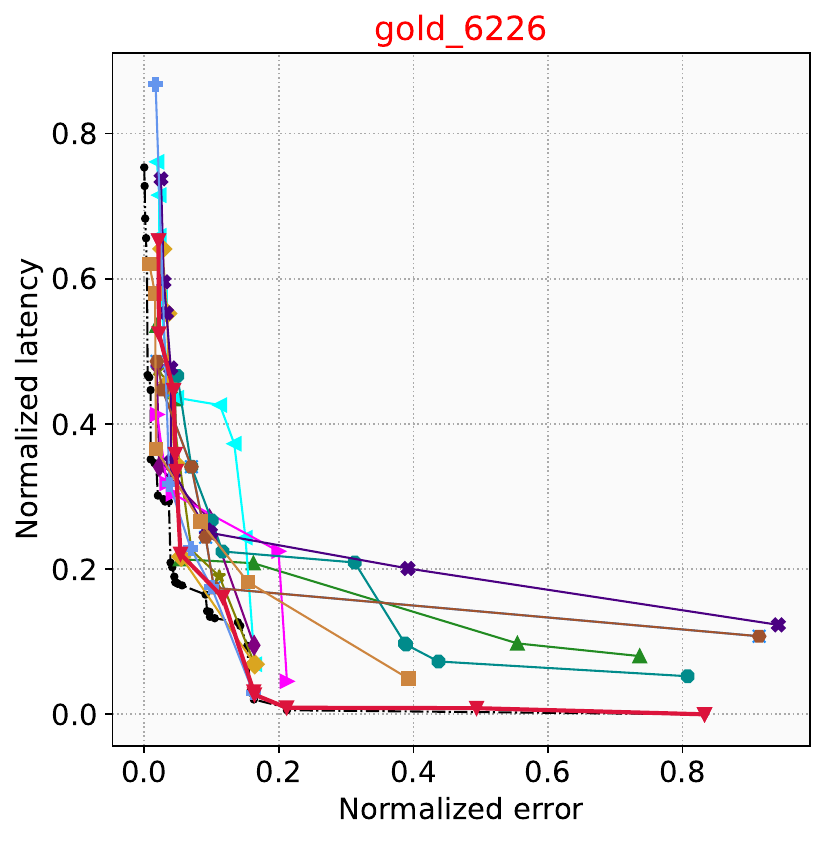}
    \includegraphics[width=.24\linewidth]{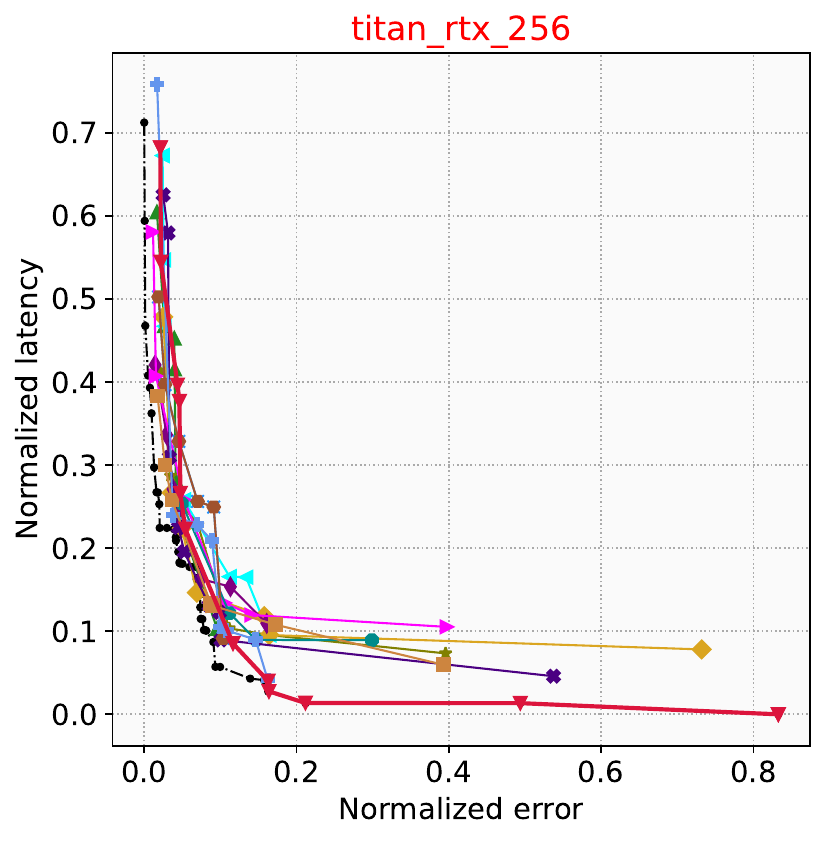}
    \includegraphics[width=.24\linewidth]{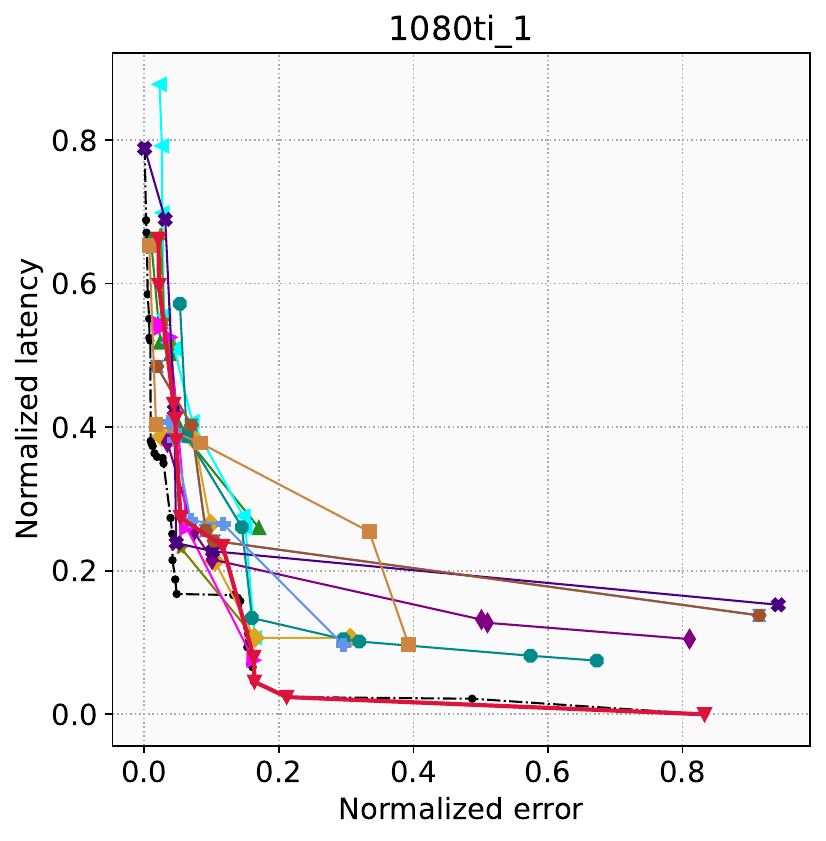}
    \includegraphics[width=.24\linewidth]{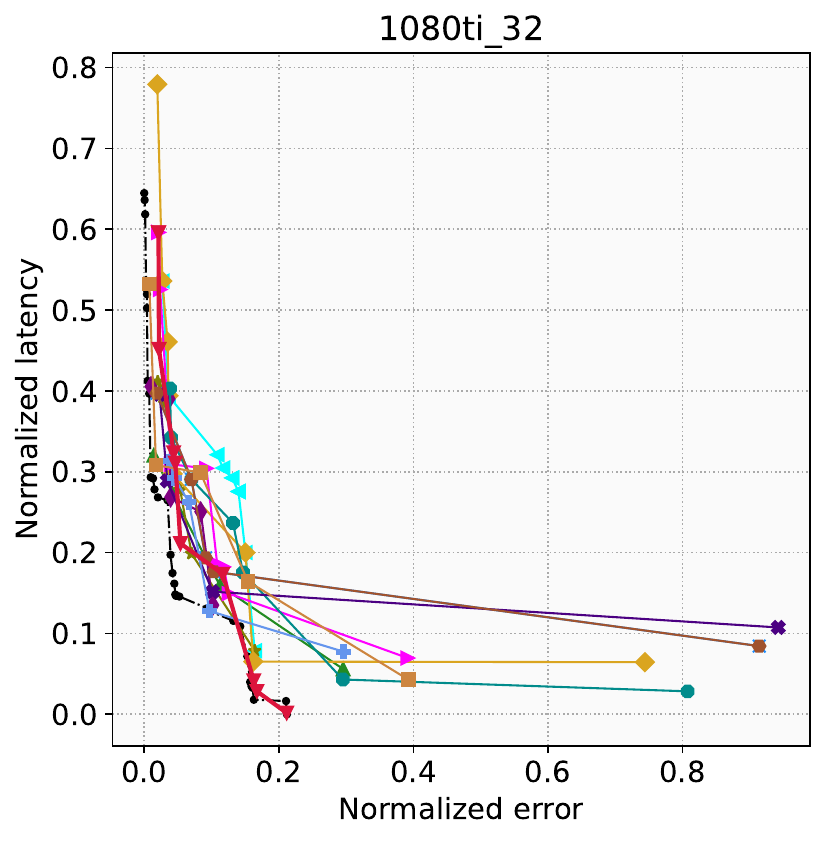}\\
    \includegraphics[width=.24\linewidth]{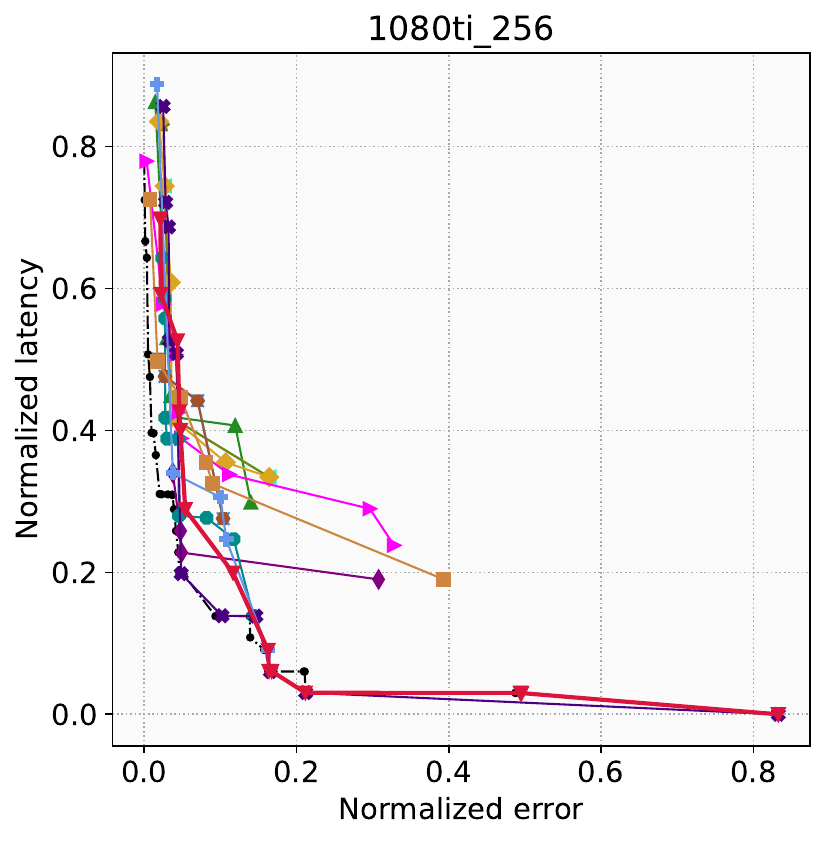}
    \includegraphics[width=.24\linewidth]{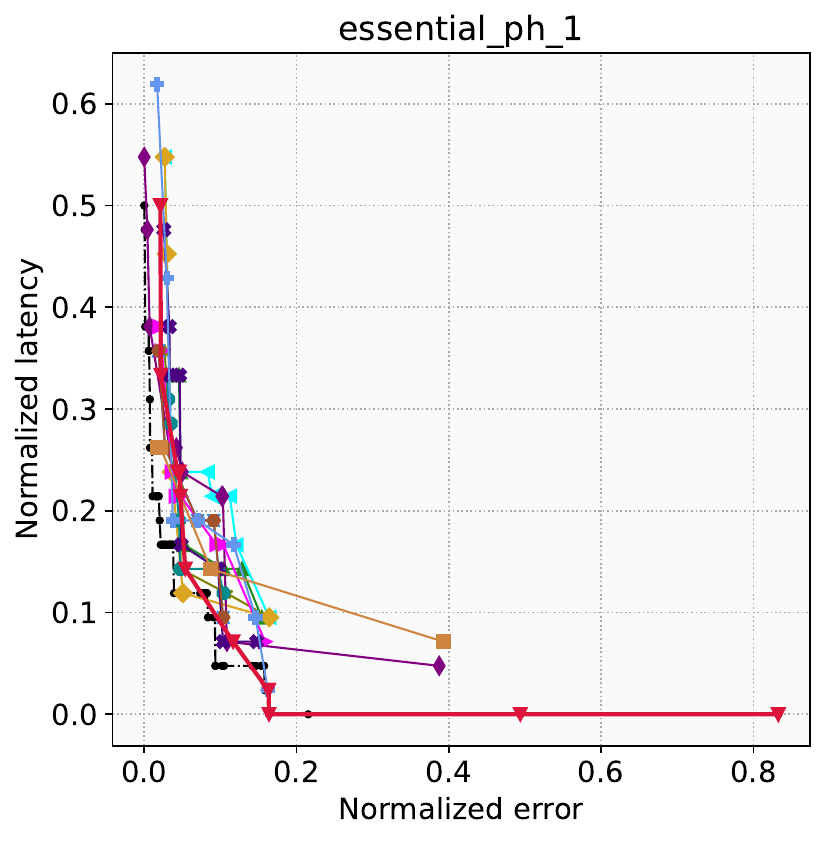}
    \includegraphics[width=.24\linewidth]{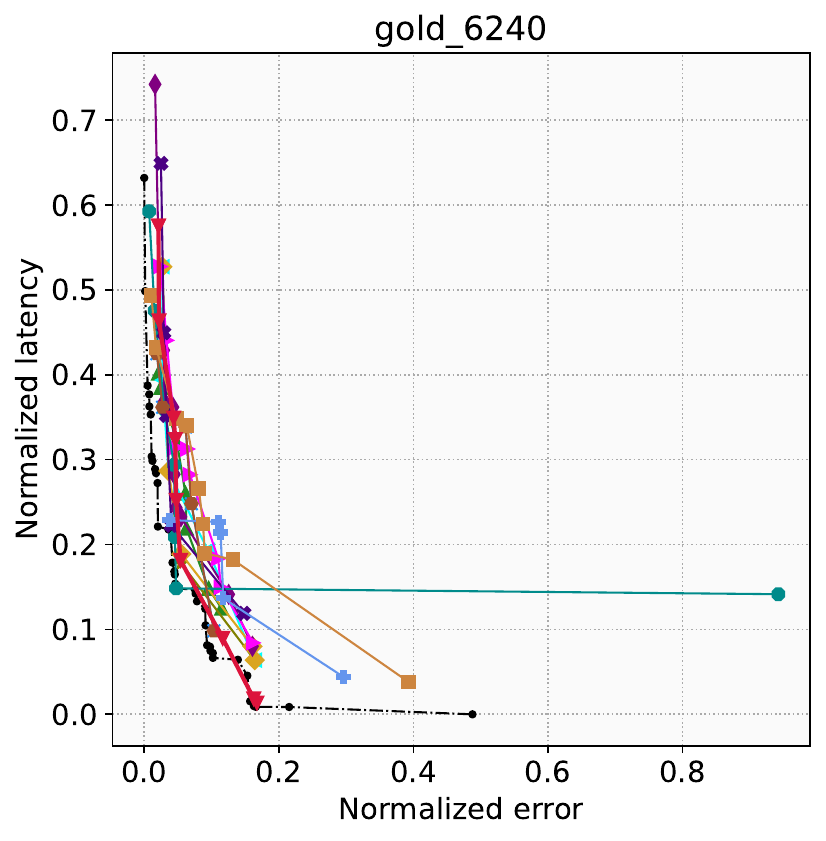}
    \includegraphics[width=.24\linewidth]{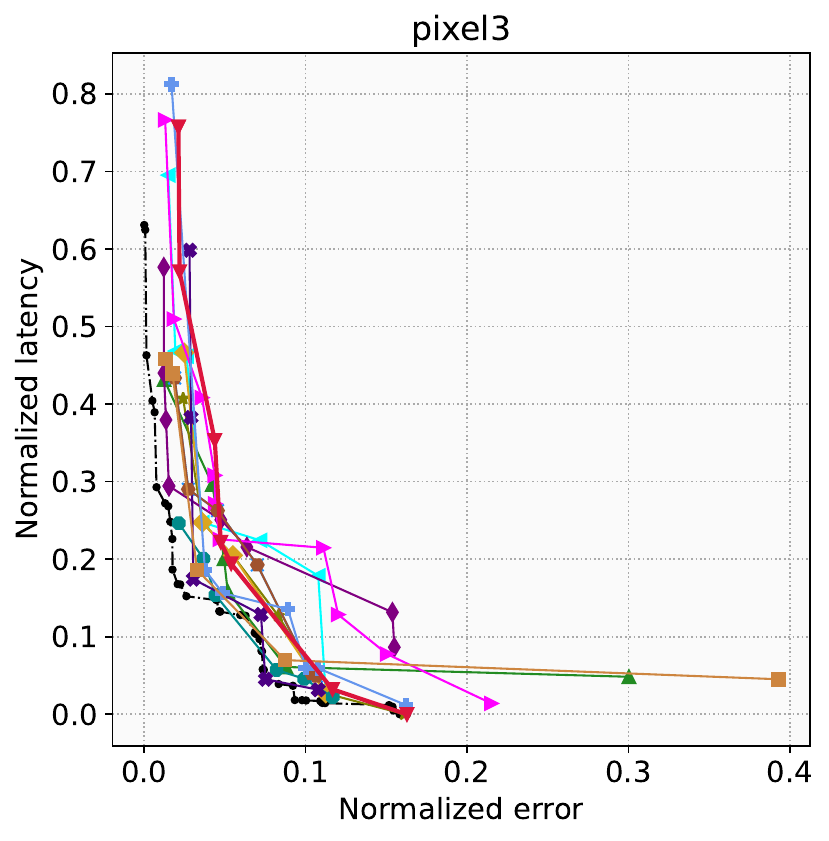}\\
    \includegraphics[width=.24\linewidth]{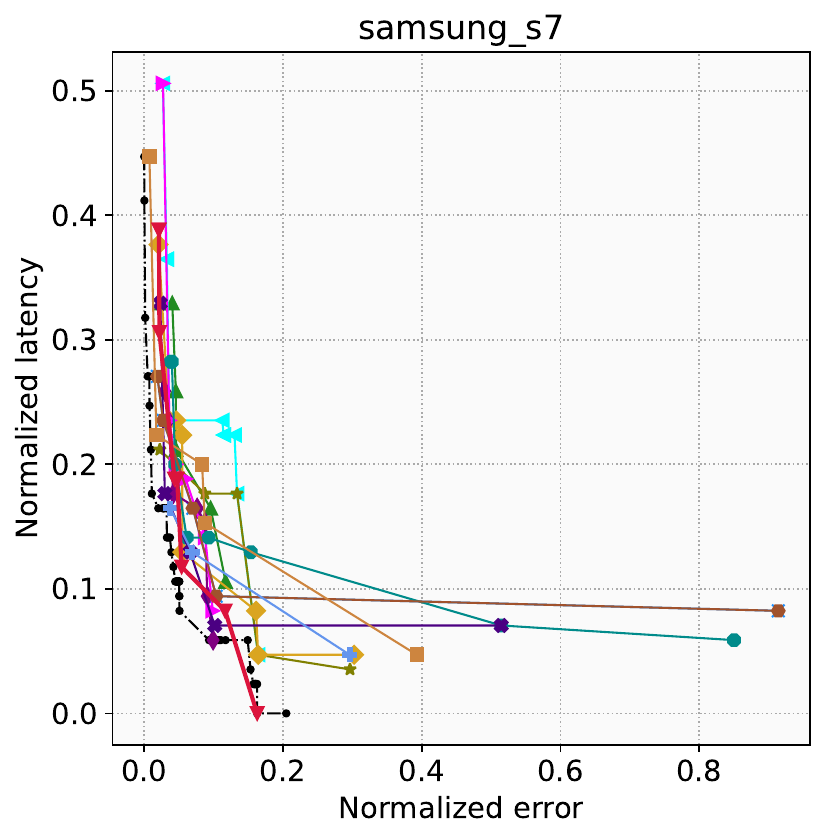}
    \includegraphics[width=.24\linewidth]{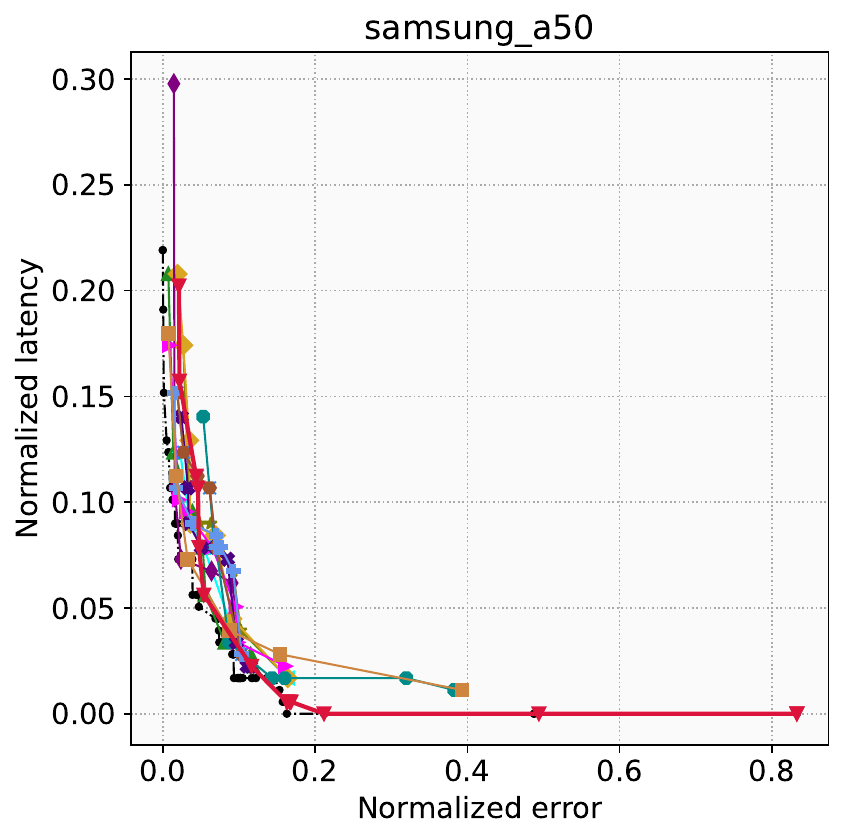}
    \includegraphics[width=.24\linewidth]{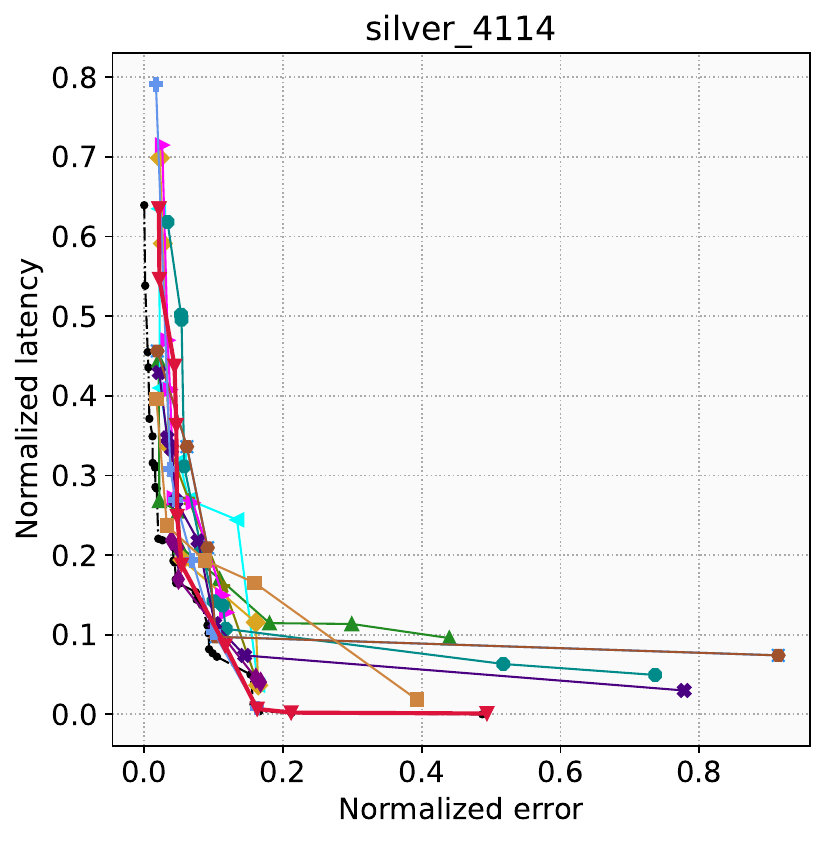}
    \includegraphics[width=.24\linewidth]{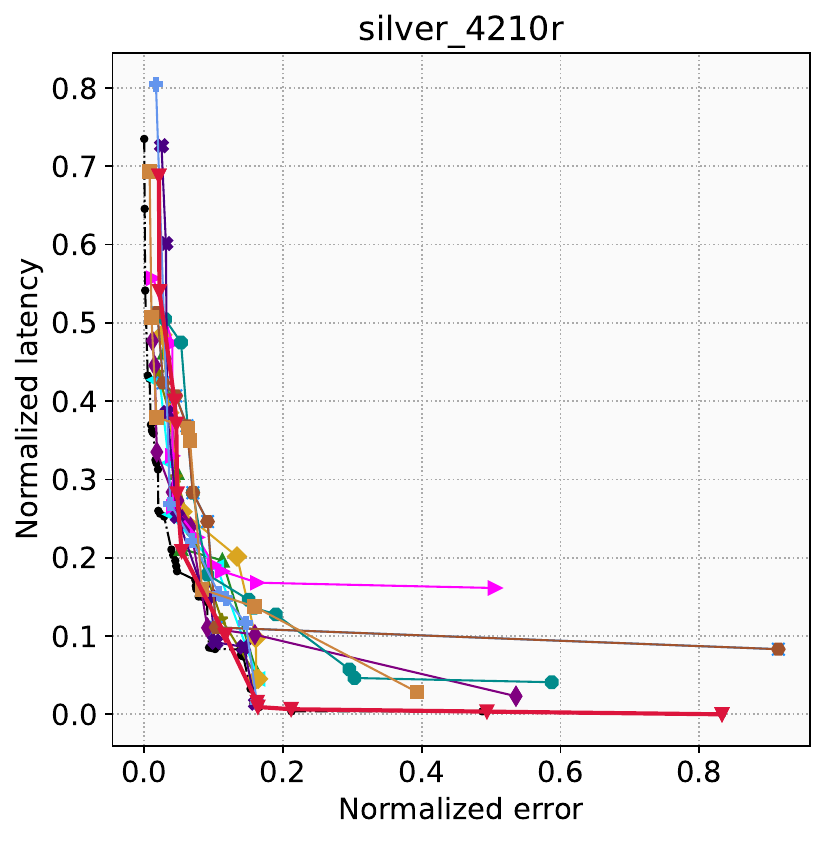}\\
    \includegraphics[width=.24\linewidth]{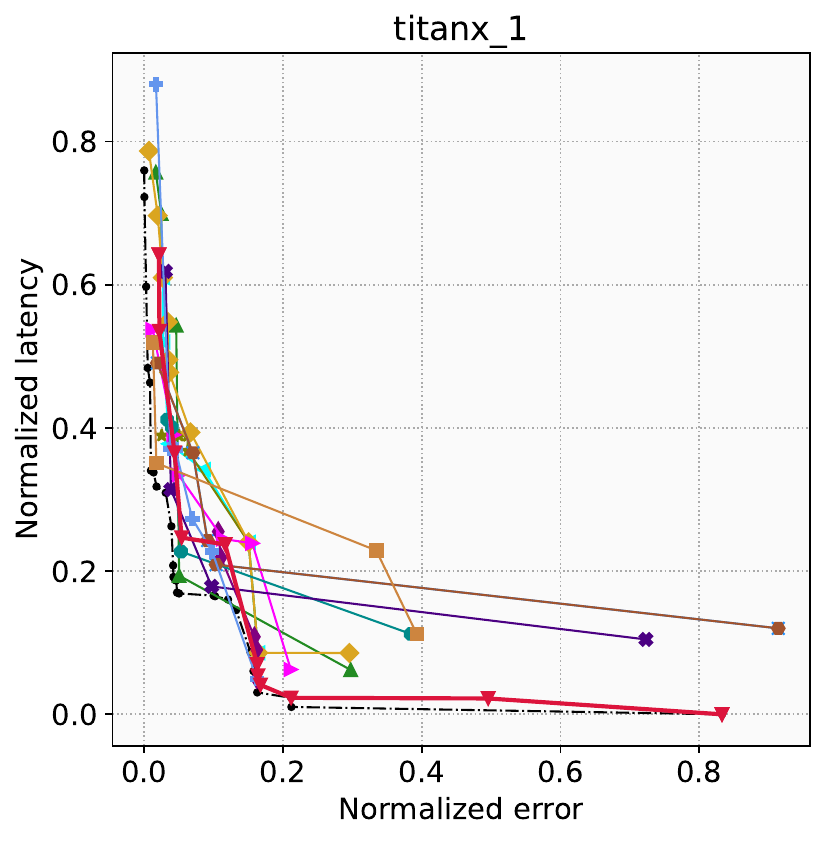} 
    \includegraphics[width=.24\linewidth]{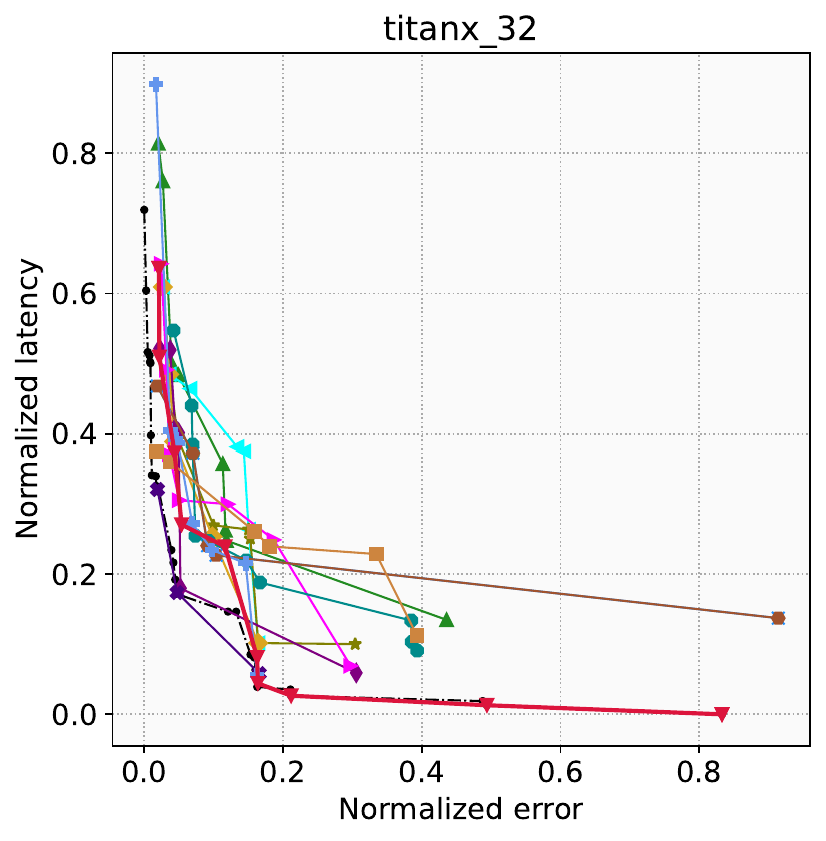}
    \includegraphics[width=.24\linewidth]{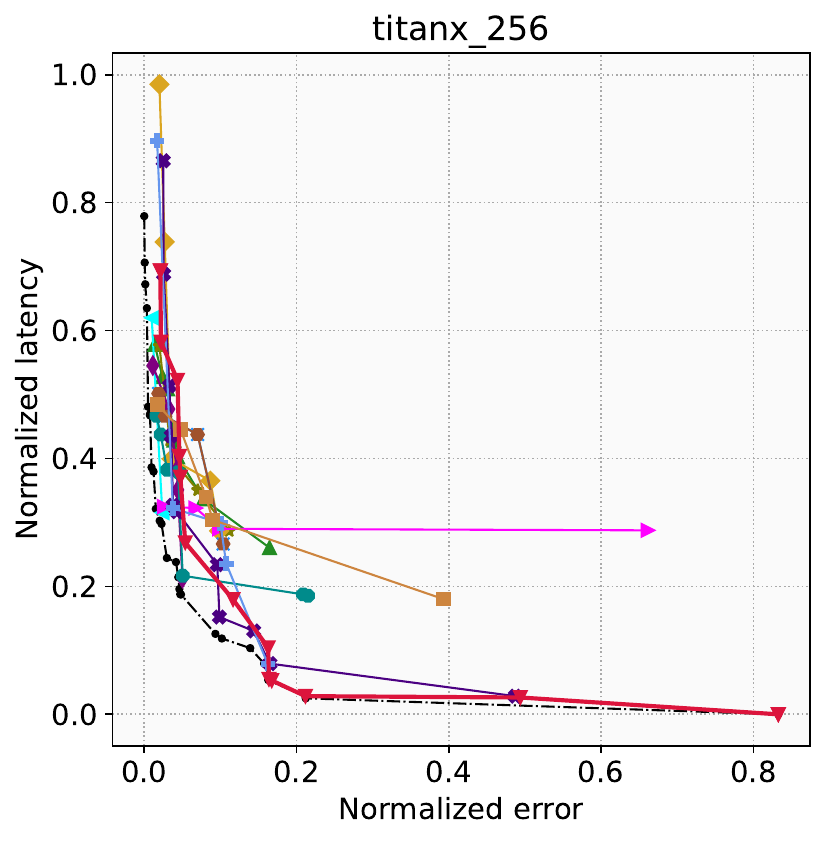}
    \caption{Pareto fronts of MODNAS and baselines on NAS-Bench-201. MODNAS-SoTL is not shown for better visibility.}
    \label{fig:pareto_nb201_full}
\end{figure}

\begin{figure}[t!]
\centering
\begin{minipage}{0.53\textwidth}
  \centering
\includegraphics[width=\textwidth]{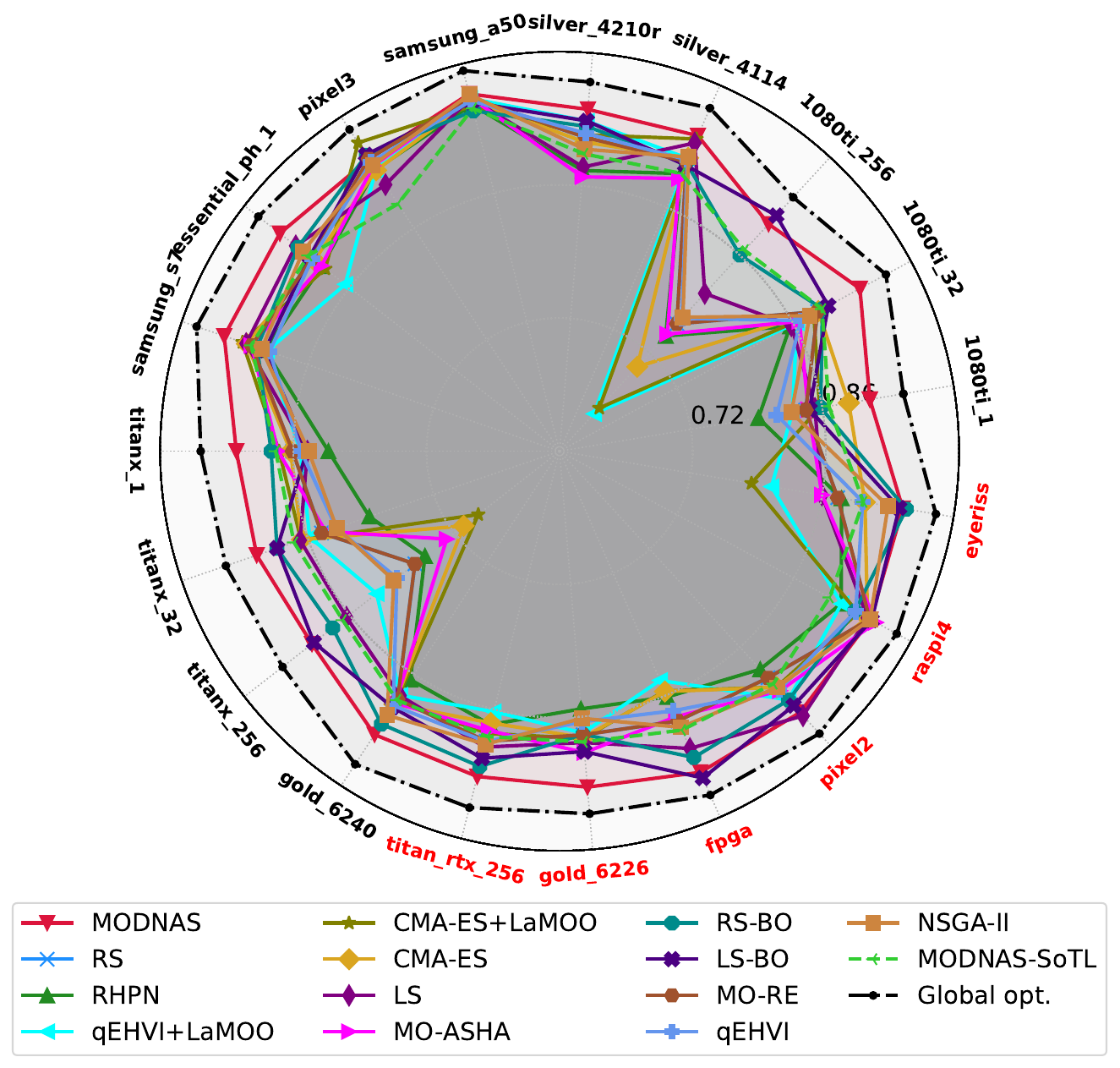}
\subcaption[hv]{HV}\label{fig:radar_hv_50}
\end{minipage}\\
\begin{minipage}{0.39\textwidth}
  \centering
\includegraphics[width=\textwidth]{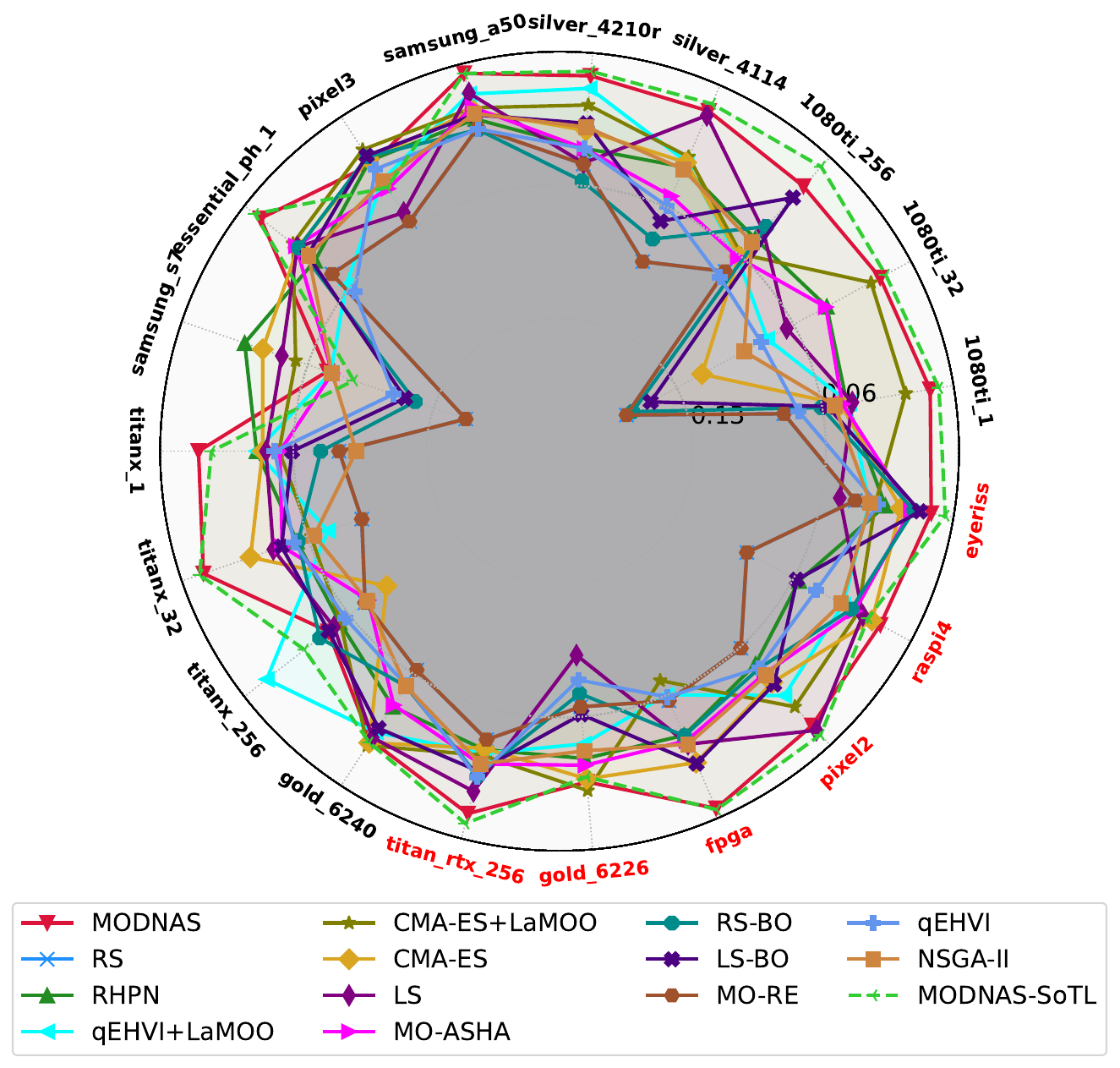}
\subcaption[gd]{GD}\label{fig:radar_gd}
\end{minipage}%
\hspace{1cm}
\begin{minipage}{0.39\textwidth}
  \centering
\includegraphics[width=\textwidth]{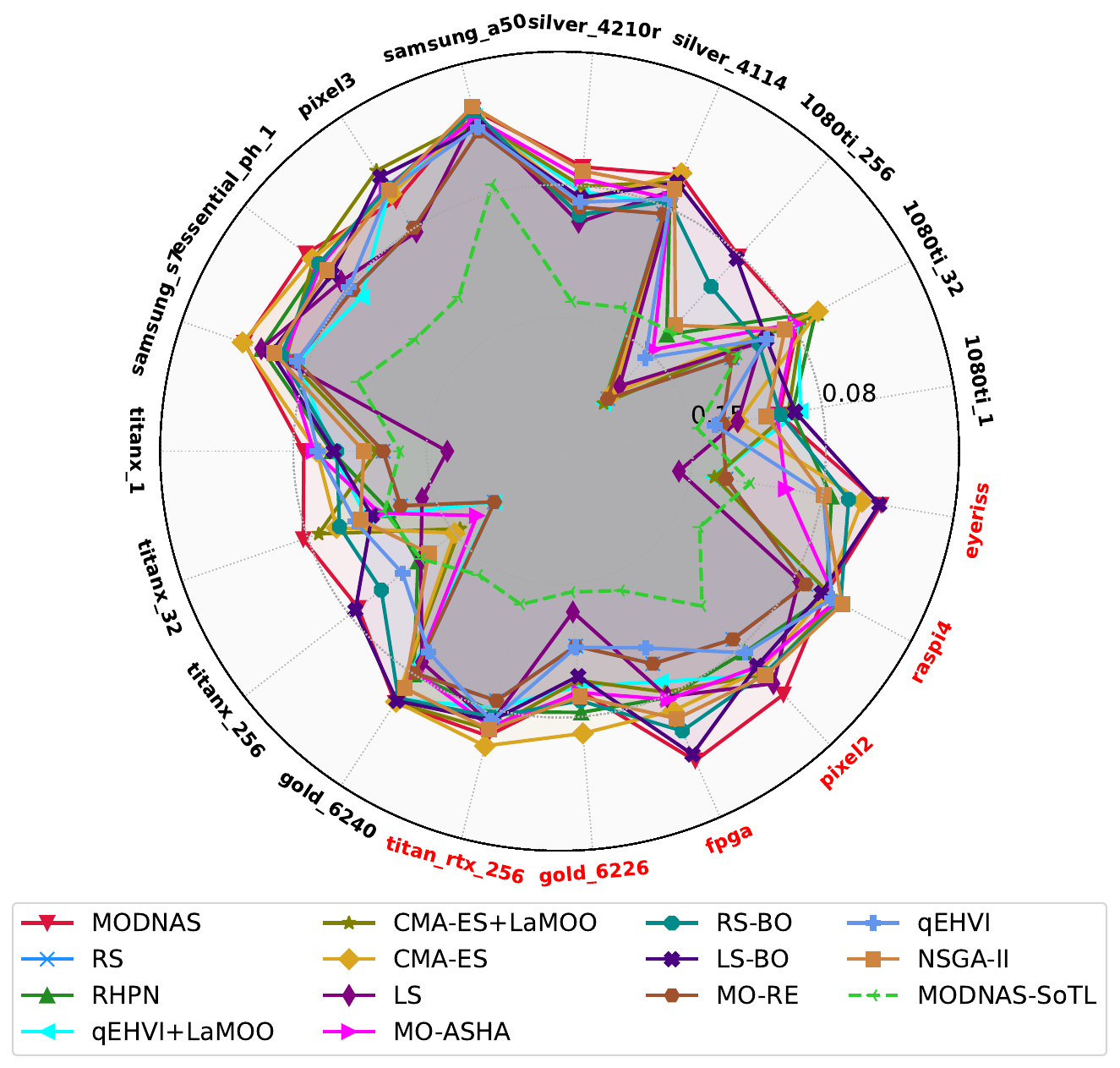}
\subcaption[igd]{$\text{IGD}$}\label{fig:radar_igd}
\end{minipage}\\
\begin{minipage}{0.39\textwidth}
  \centering
\includegraphics[width=\textwidth]{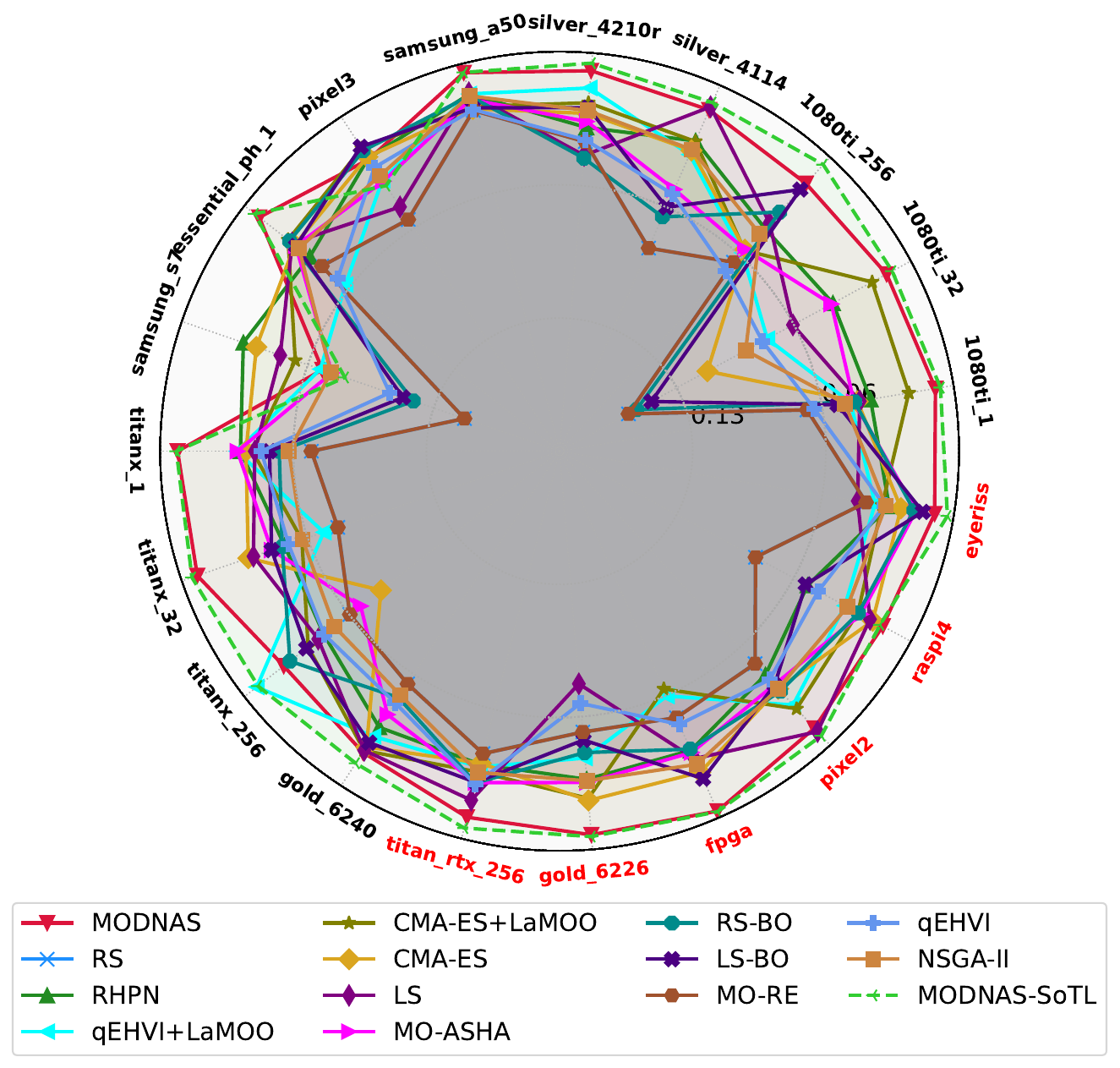}
\subcaption[gd_plus]{GD+}\label{fig:radar_gd_plus}
\end{minipage}%
\hspace{1cm}
\begin{minipage}{0.39\textwidth}
  \centering
\includegraphics[width=\textwidth]{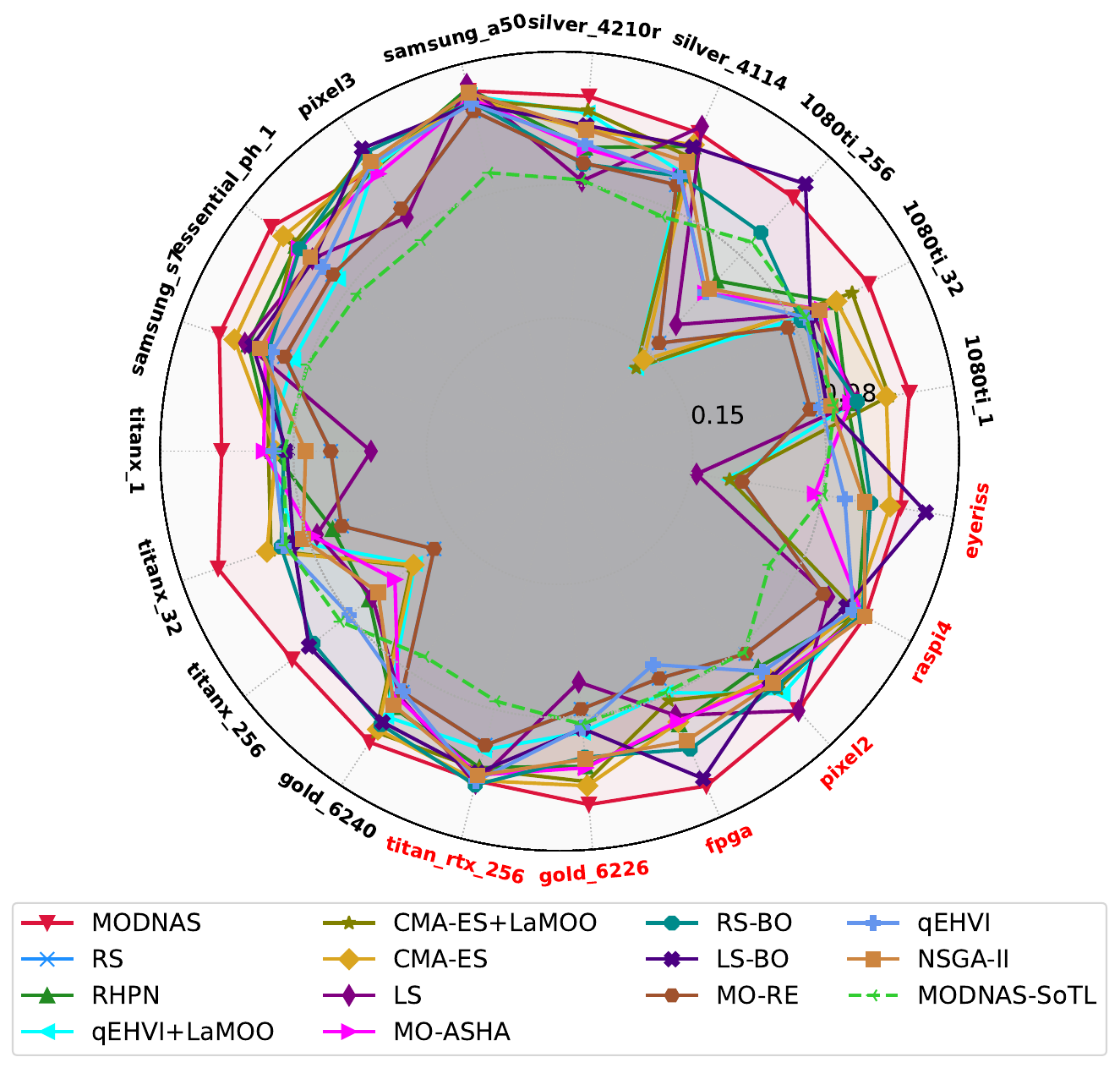}
\subcaption[igd_plus]{$\text{IGD+}$}\label{fig:radar_igd_plus}
\end{minipage}
\caption{HV, GD, GD+, IGD and IGD+ of MODNAS and baselines across 19 devices on NAS-Bench-201. For every device we optimize for 2 objectives, namely \textit{latency (ms)} and \textit{test accuracy} on CIFAR-10. For method, metric and device we report the mean of 3 independent search runs. Higher area in the radar indicates better performance for every metric. Test devices are colored in red around the radar plot. Here we allocate double the budget to baselines, i.e. we run all baselines for 50 function evaluations.} 
\label{fig:radar_plot_gd_igd}
\end{figure}

\begin{figure}[t!]
    \centering
    \includegraphics[width=.24\linewidth]{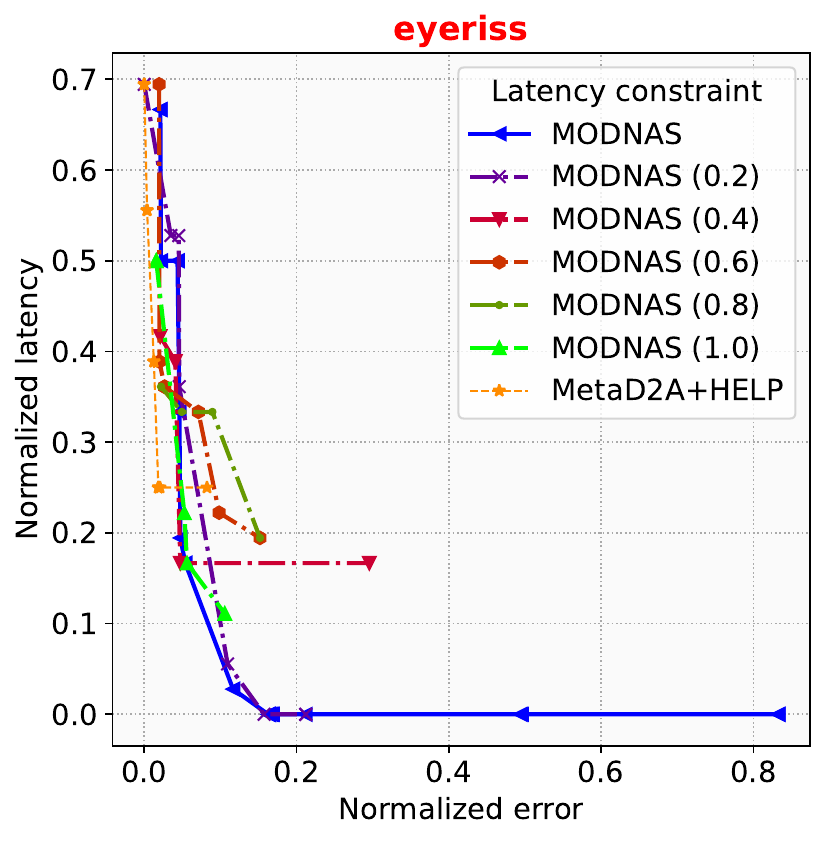}
    \includegraphics[width=.24\linewidth]{figures/constraints_pareto/gold_6226.pdf}
    \includegraphics[width=.24\linewidth]{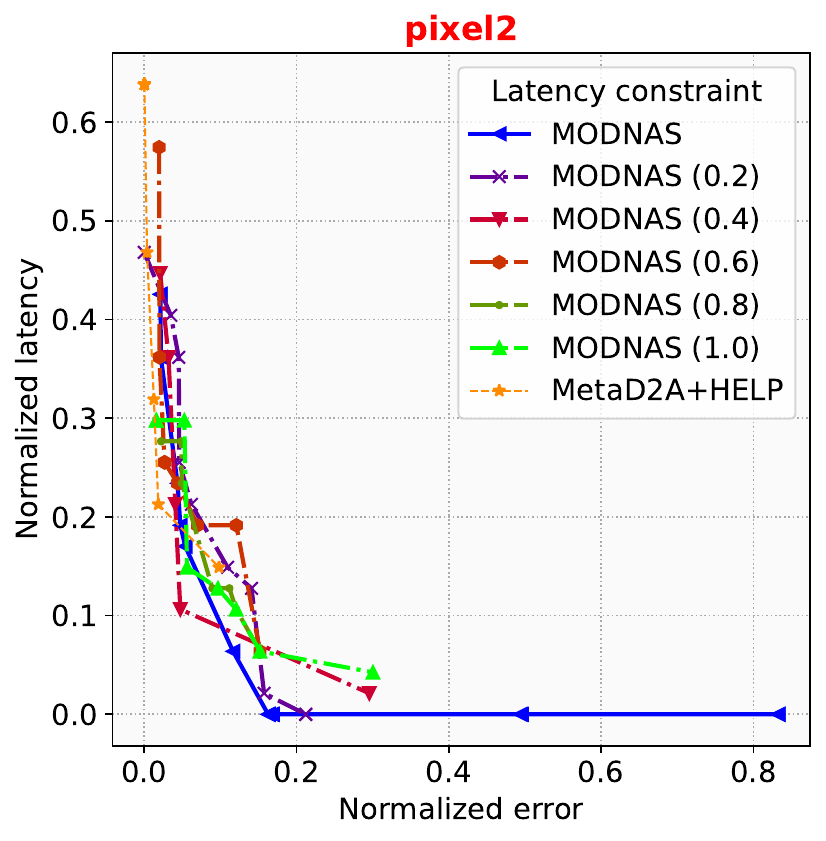}
    \includegraphics[width=.24\linewidth]{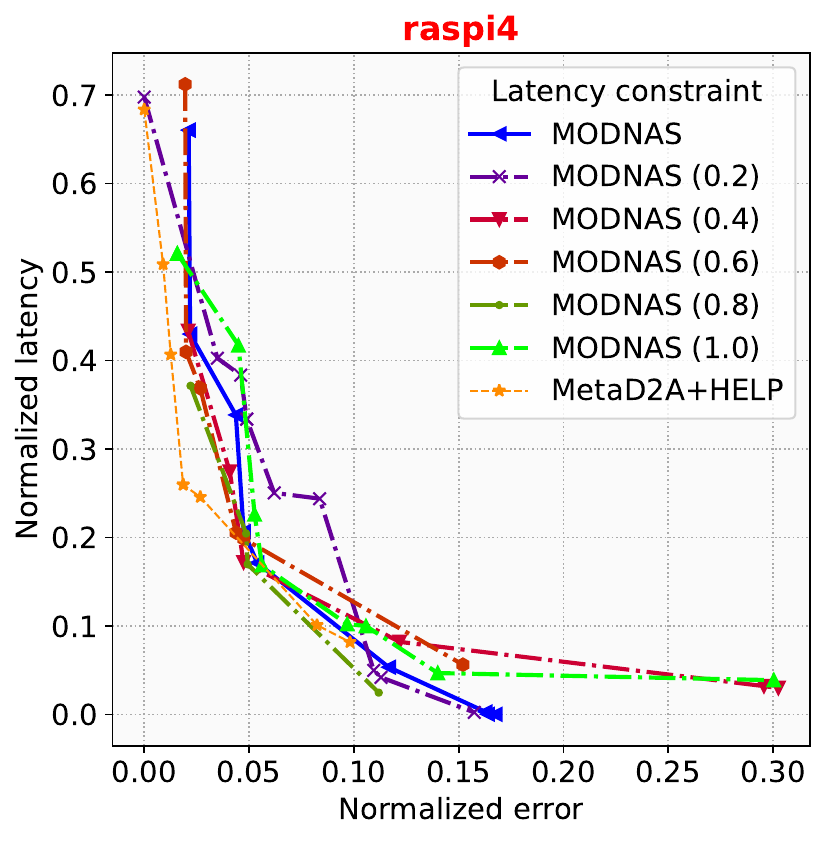}\\
    \includegraphics[width=.24\linewidth]{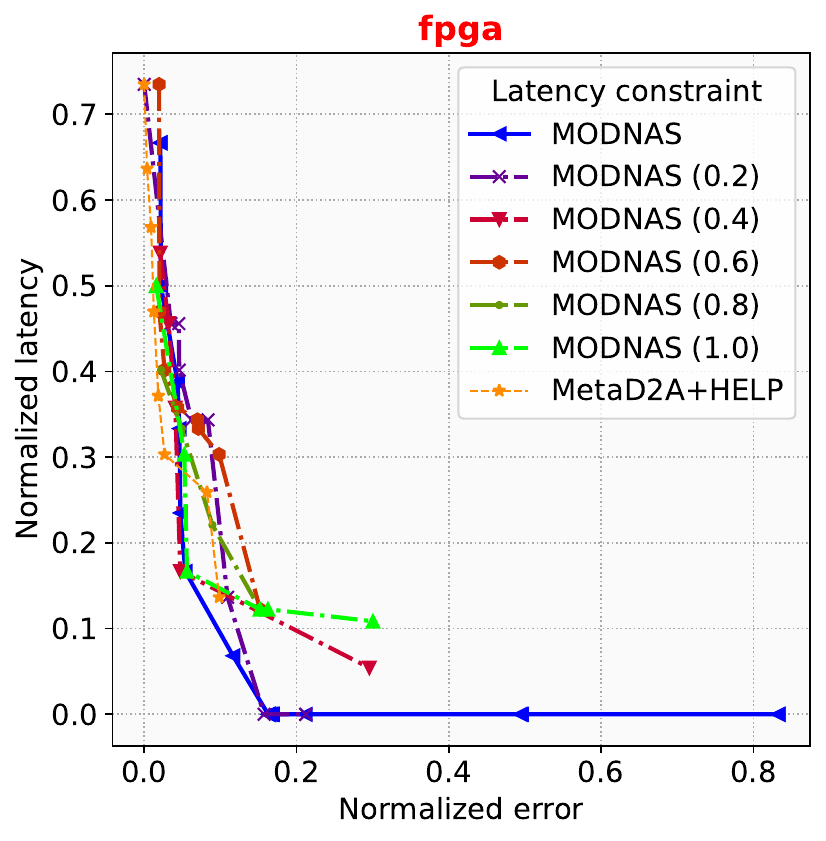}
    \includegraphics[width=.24\linewidth]{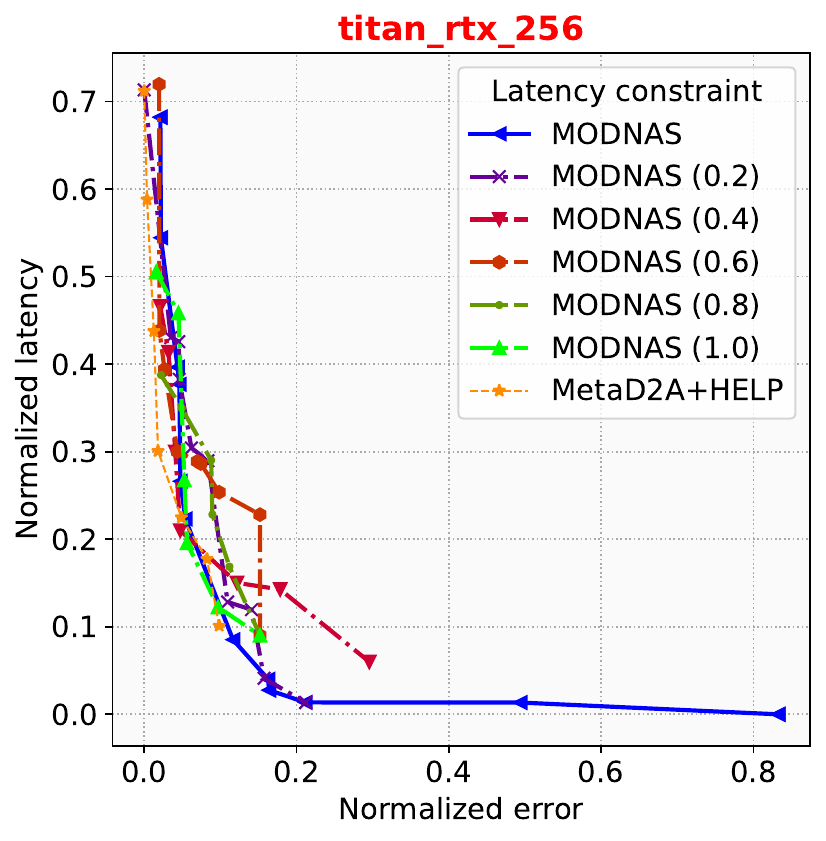}
    \includegraphics[width=.24\linewidth]{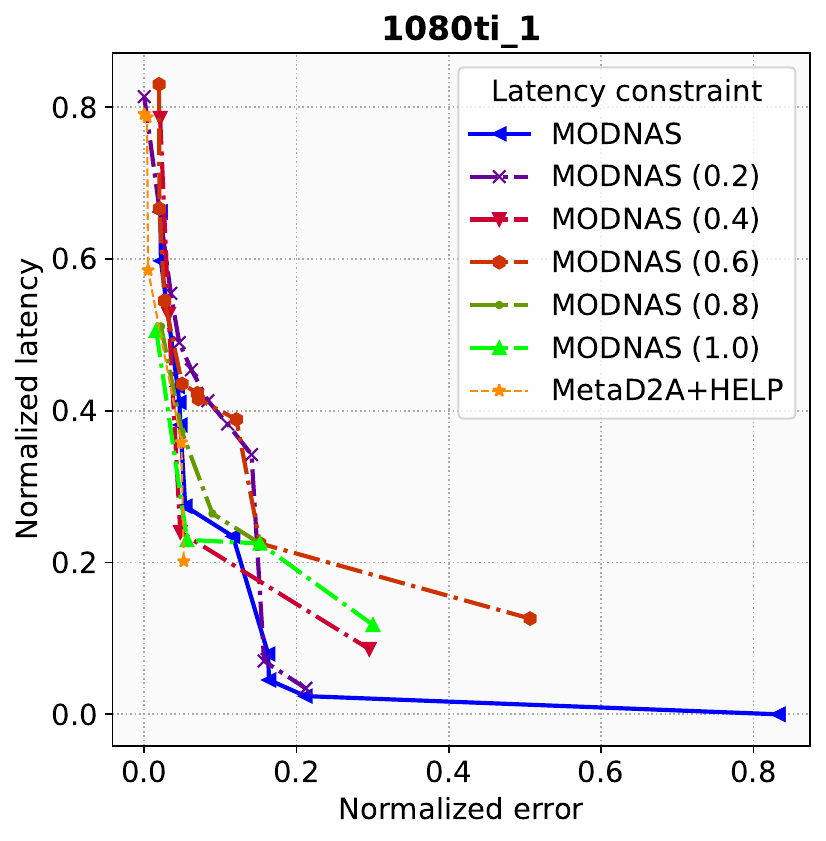}
    \includegraphics[width=.24\linewidth]{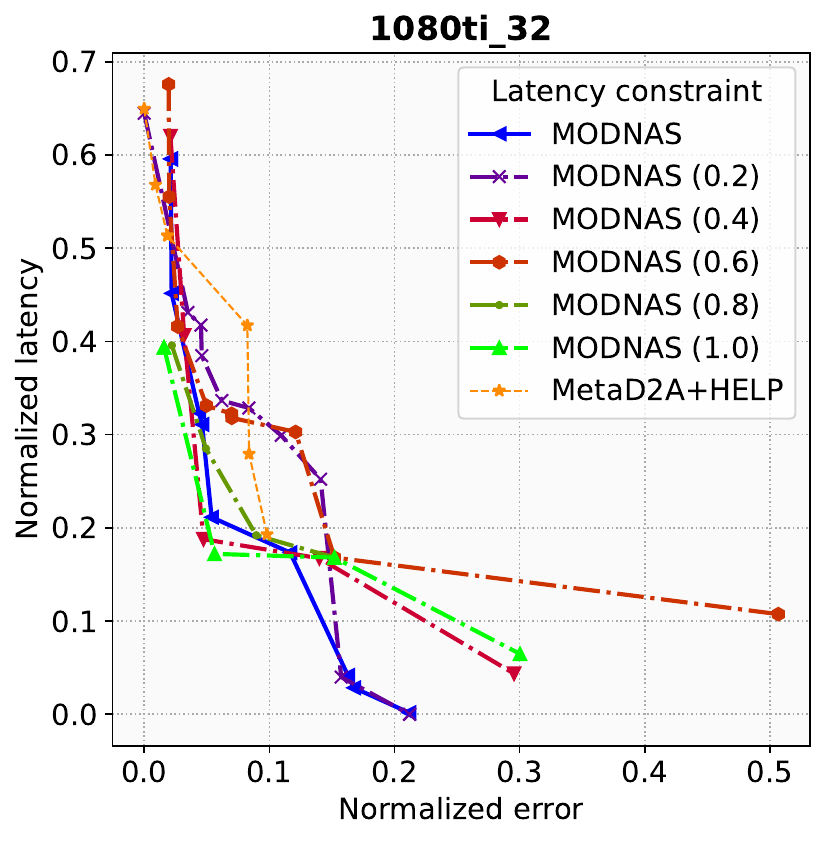}\\
    \includegraphics[width=.24\linewidth]{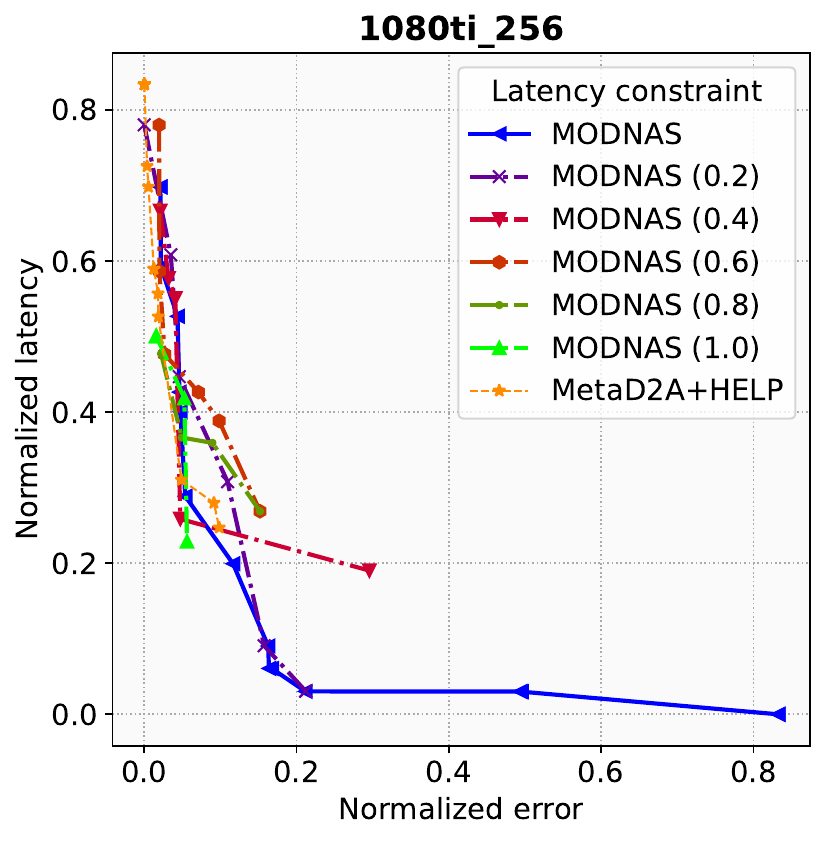}
    \includegraphics[width=.24\linewidth]{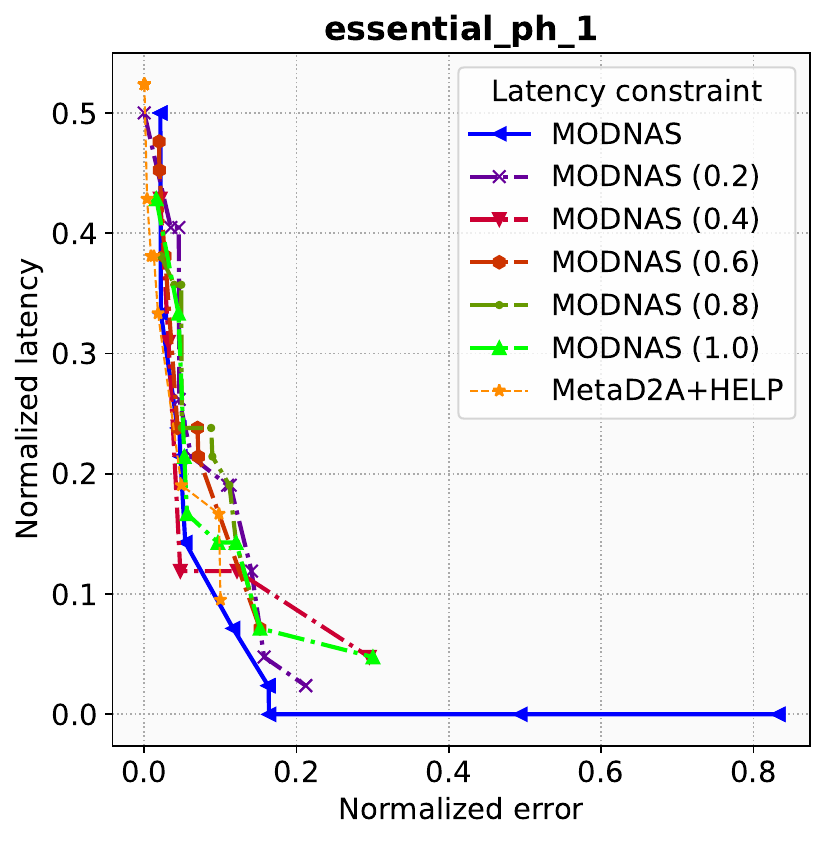}
    \includegraphics[width=.24\linewidth]{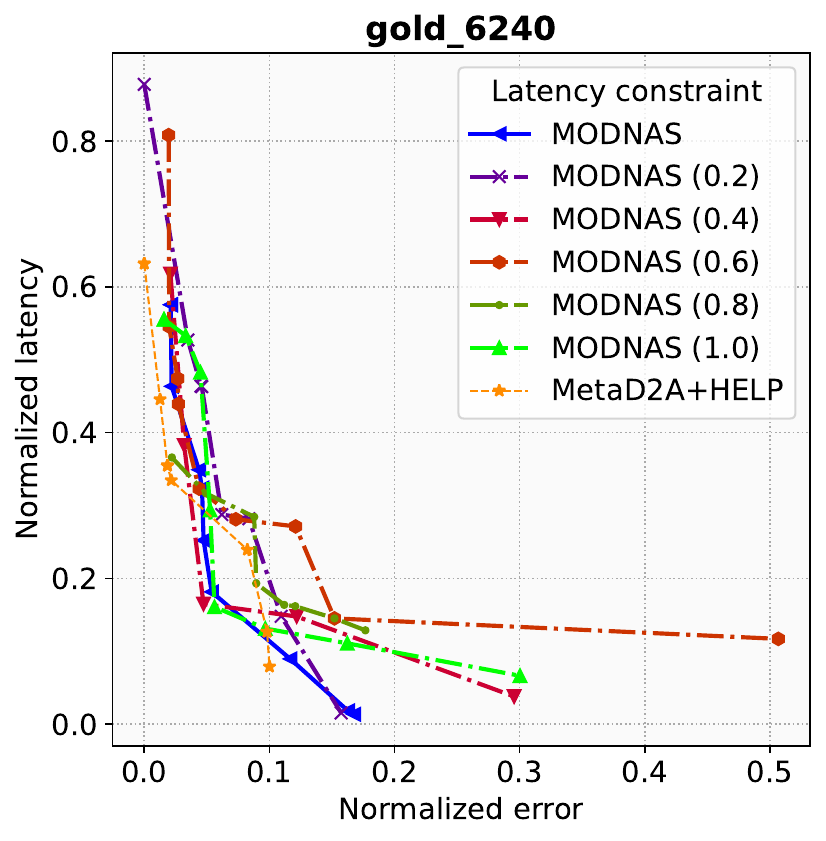}
    \includegraphics[width=.24\linewidth]{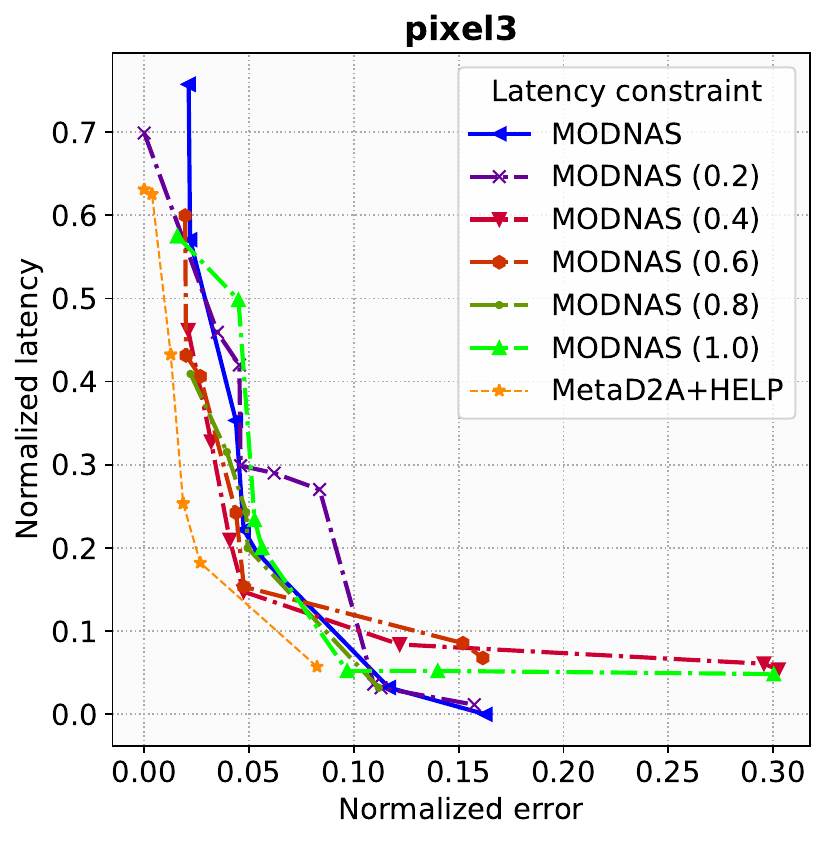}\\
    \includegraphics[width=.24\linewidth]{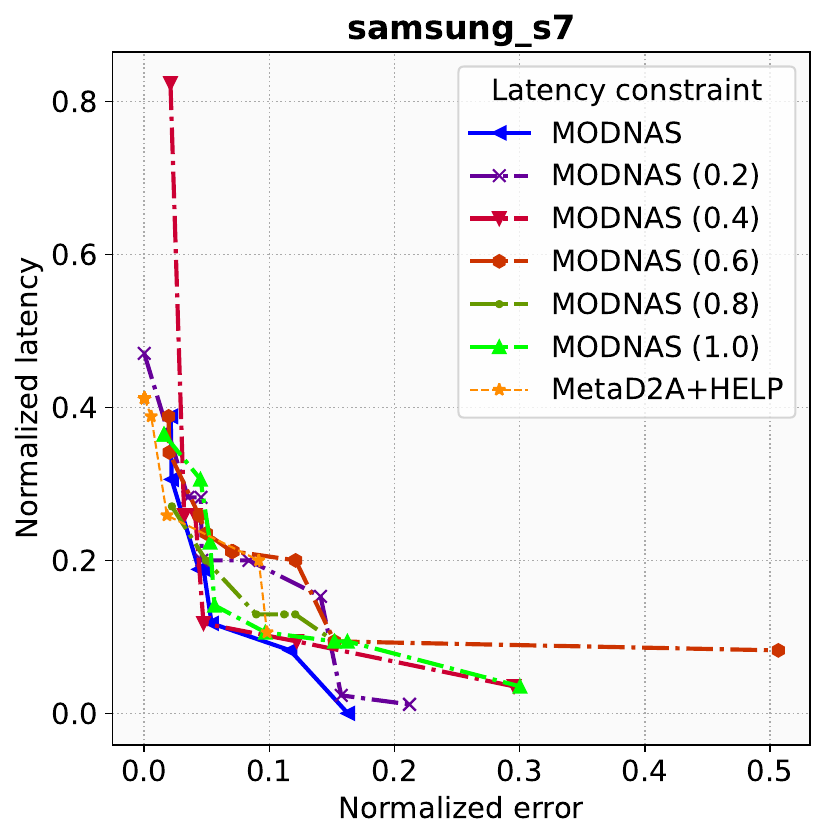}
    \includegraphics[width=.24\linewidth]{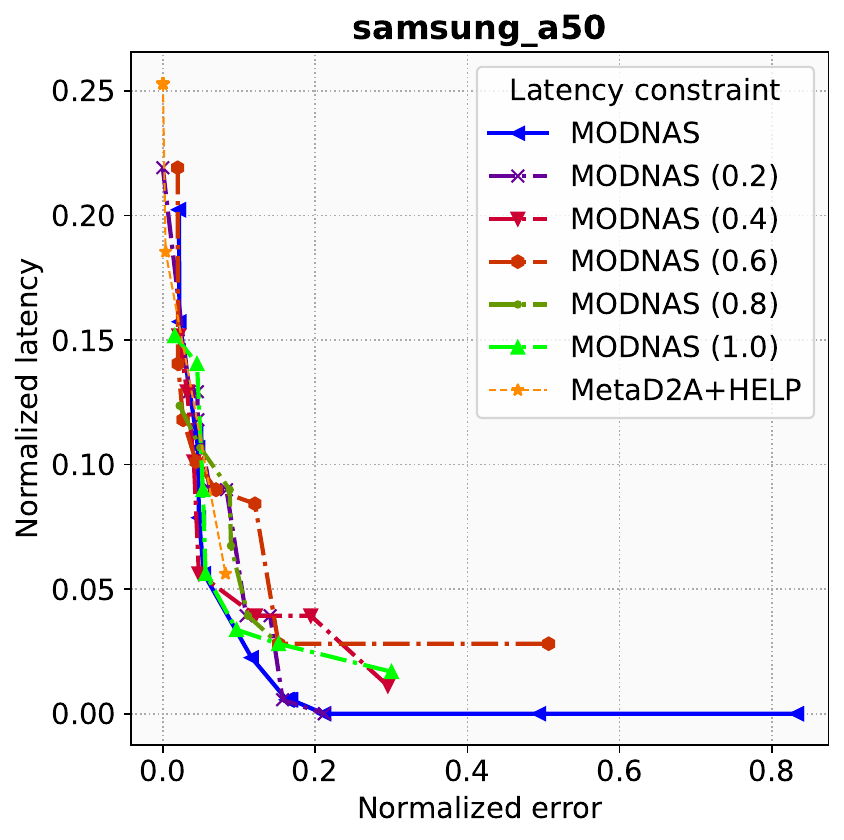}
    \includegraphics[width=.24\linewidth]{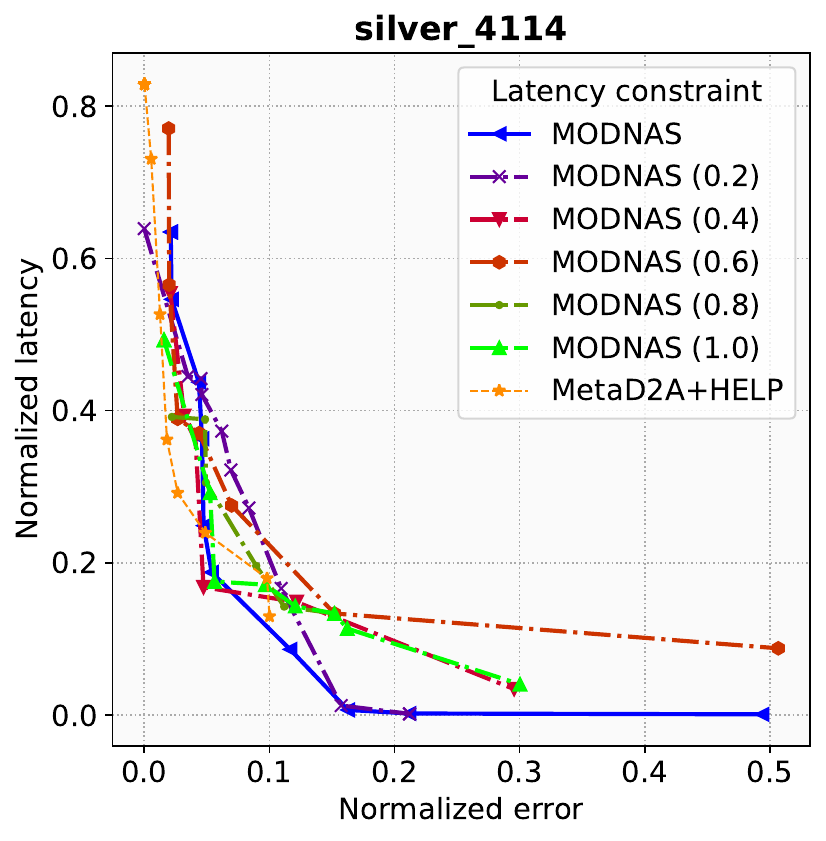}
    \includegraphics[width=.24\linewidth]{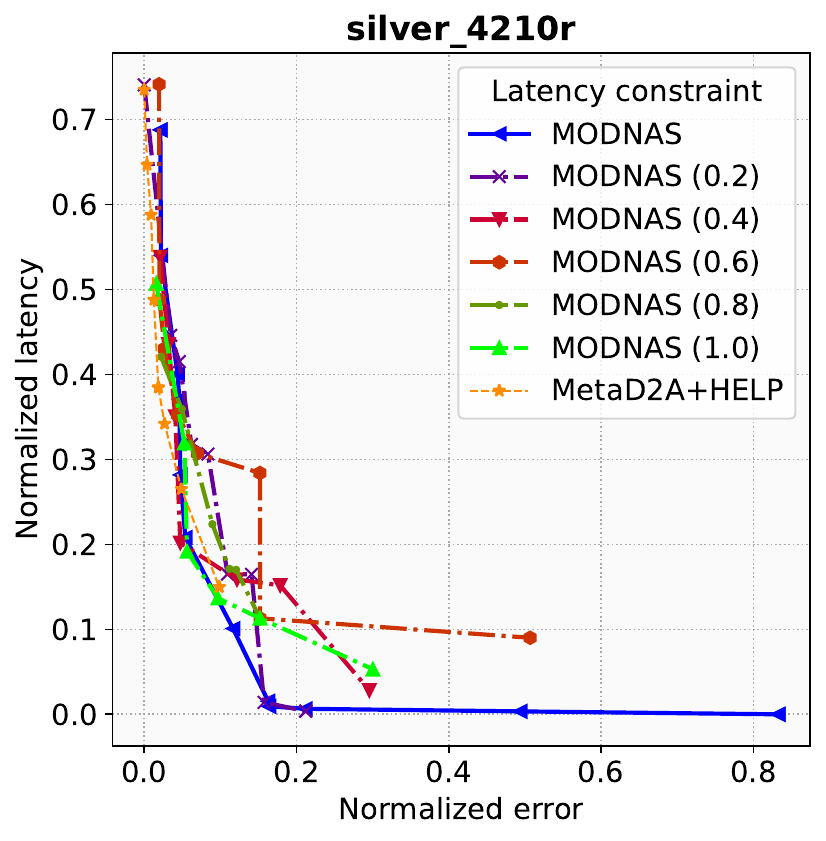}\\
    \includegraphics[width=.24\linewidth]{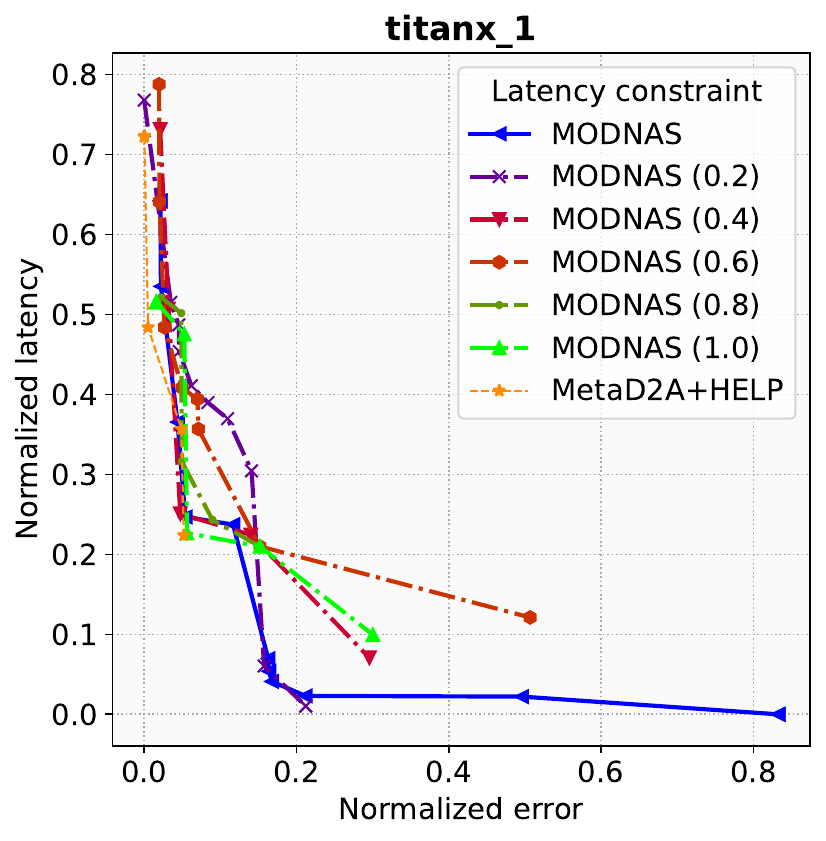}
    \includegraphics[width=.24\linewidth]{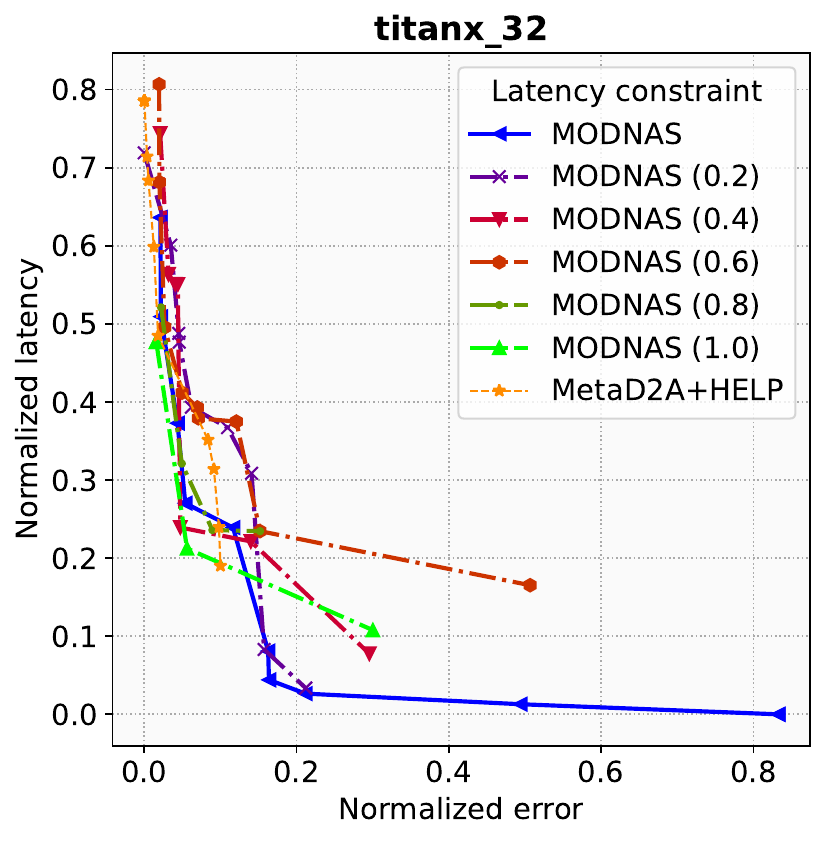}
    \includegraphics[width=.24\linewidth]{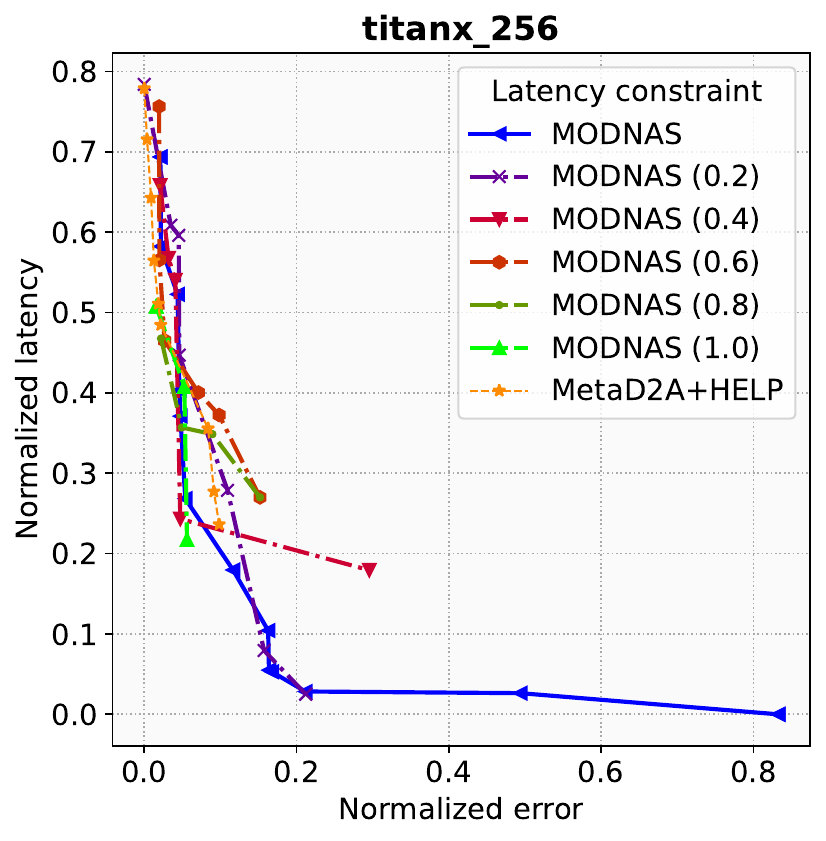}
    \caption{Pareto fronts of MODNAS ran with different latency constraints during search.}
    \label{fig:fronts_constraints_full}
\end{figure}

\begin{figure}[t!]
    \centering
    \includegraphics[width=.24\linewidth]{figures/gradchemes_hv/hv_fpga.pdf}
    \includegraphics[width=.24\linewidth]{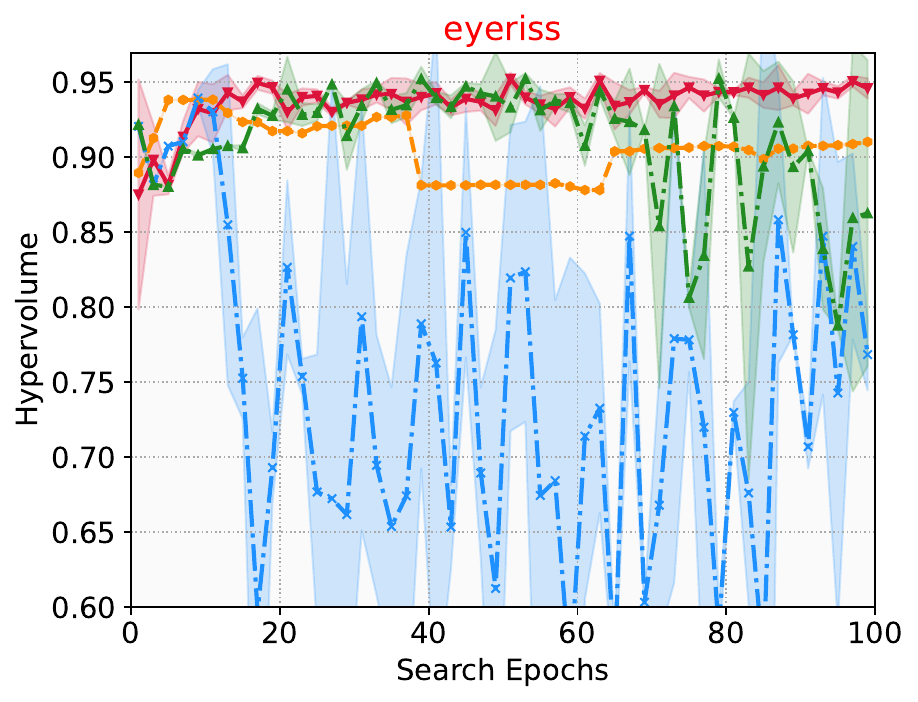}
    \includegraphics[width=.24\linewidth]{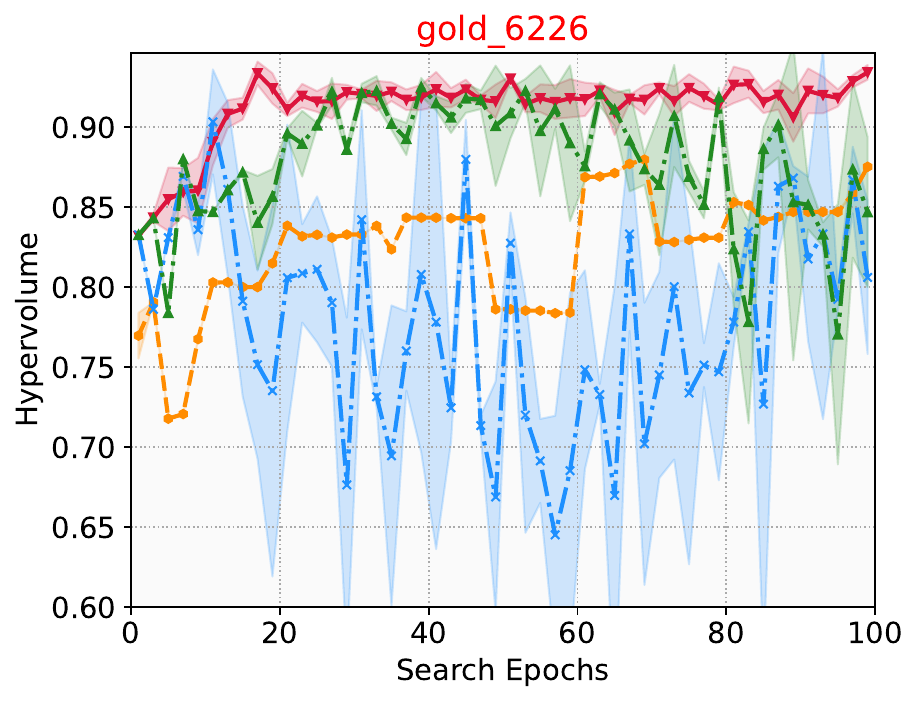}
    \includegraphics[width=.24\linewidth]{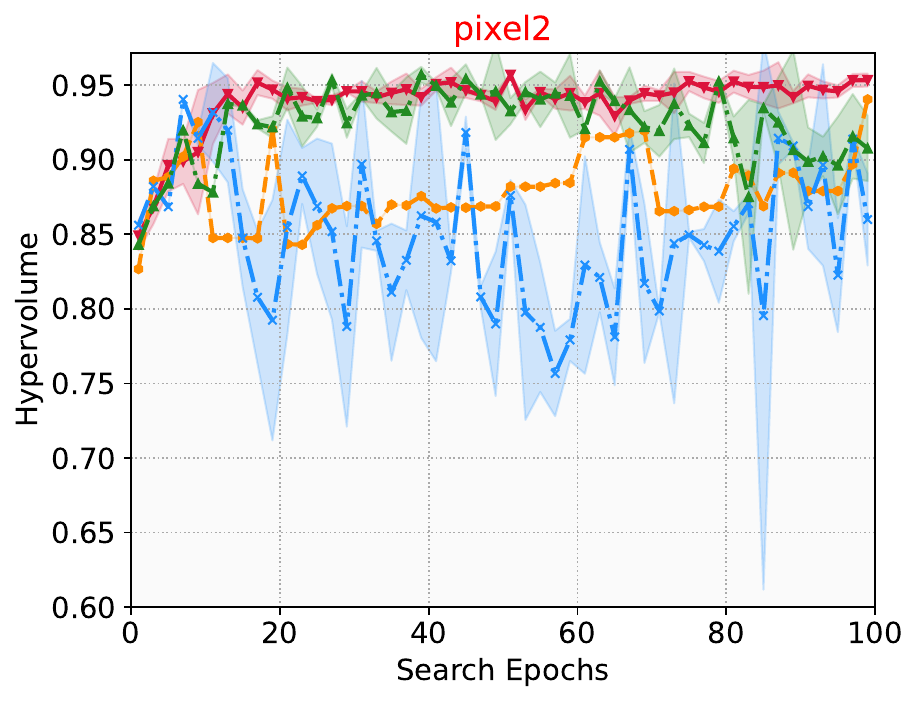} \\
    \includegraphics[width=.24\linewidth]{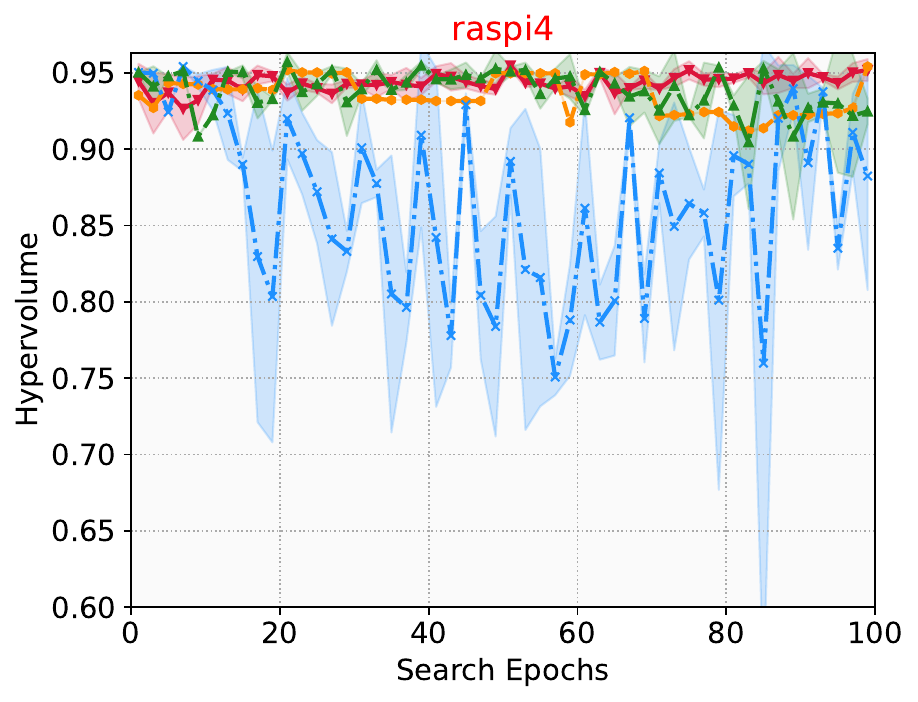}
    \includegraphics[width=.24\linewidth]{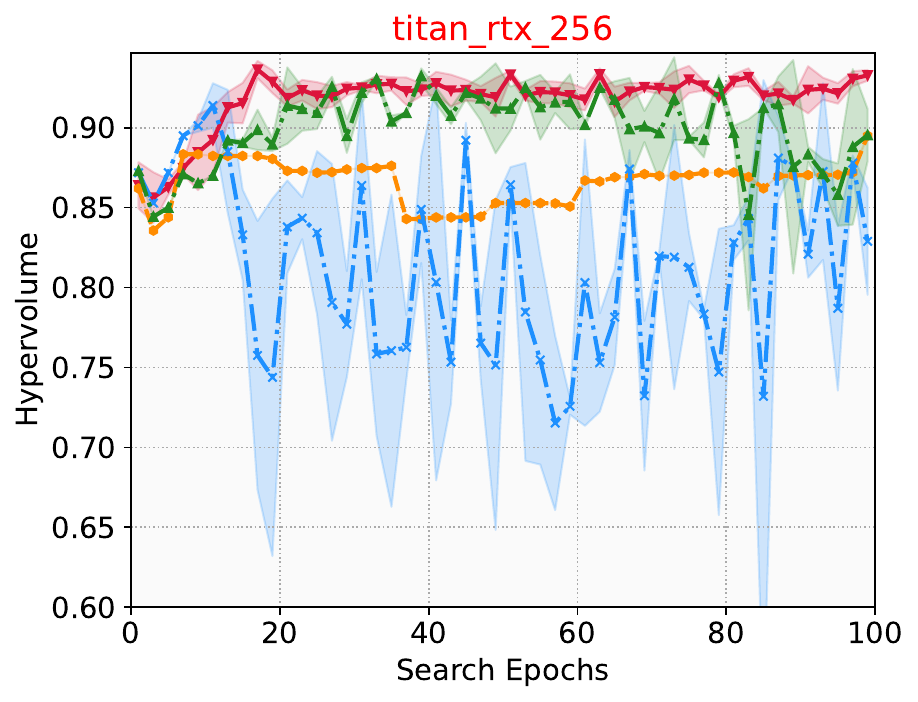}
    \includegraphics[width=.24\linewidth]{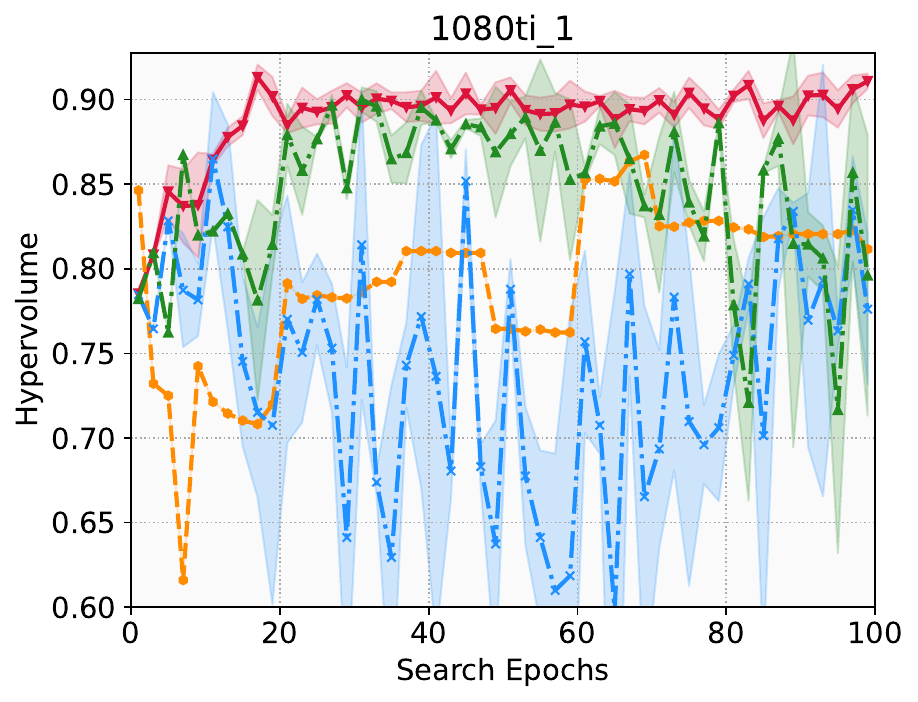}
    \includegraphics[width=.24\linewidth]{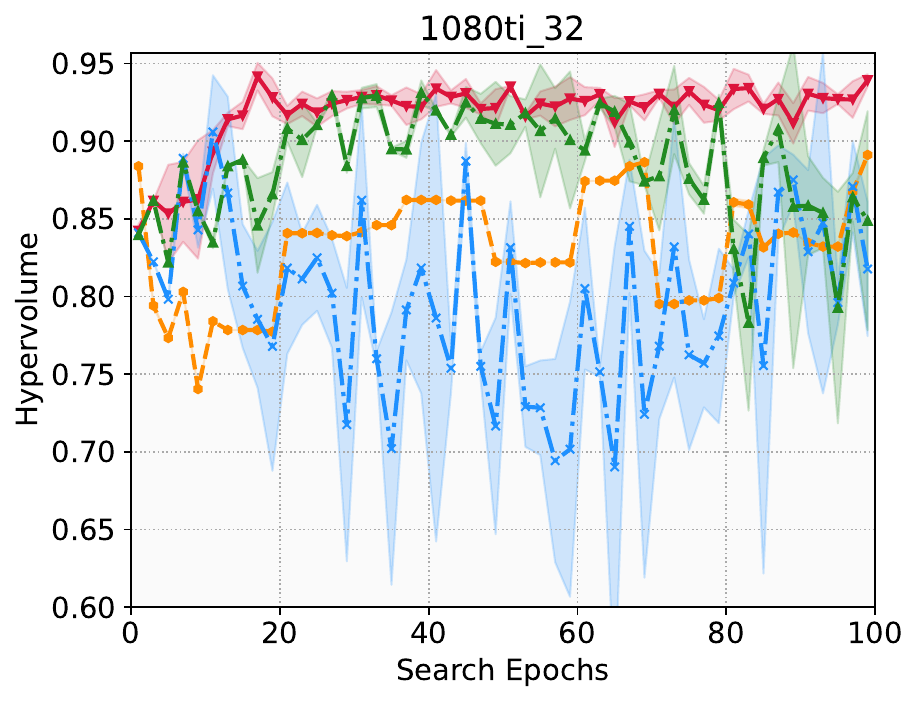}\\
    \includegraphics[width=.24\linewidth]{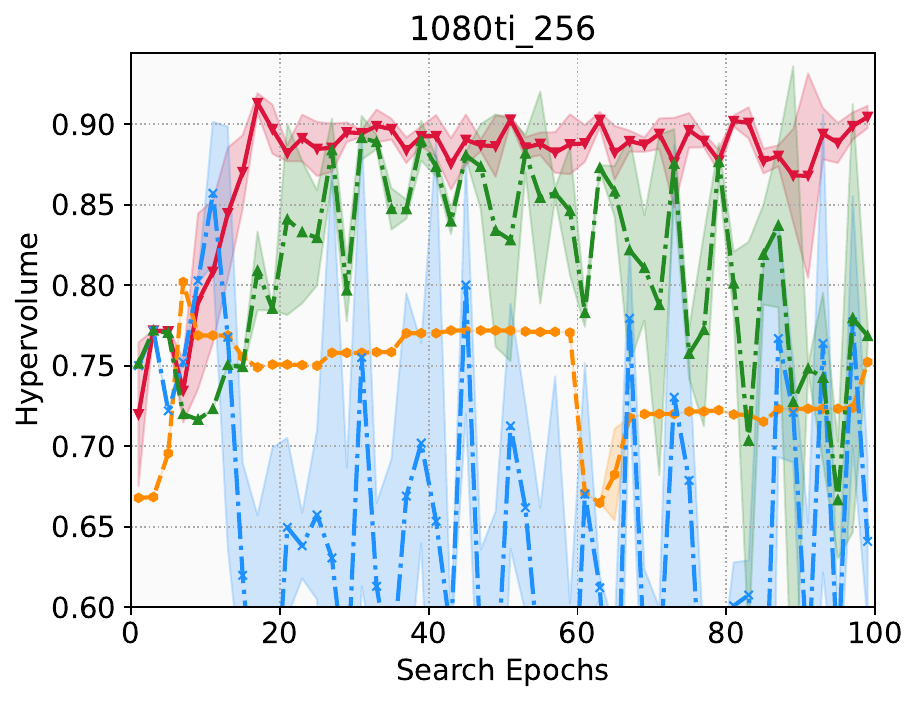}
    \includegraphics[width=.24\linewidth]{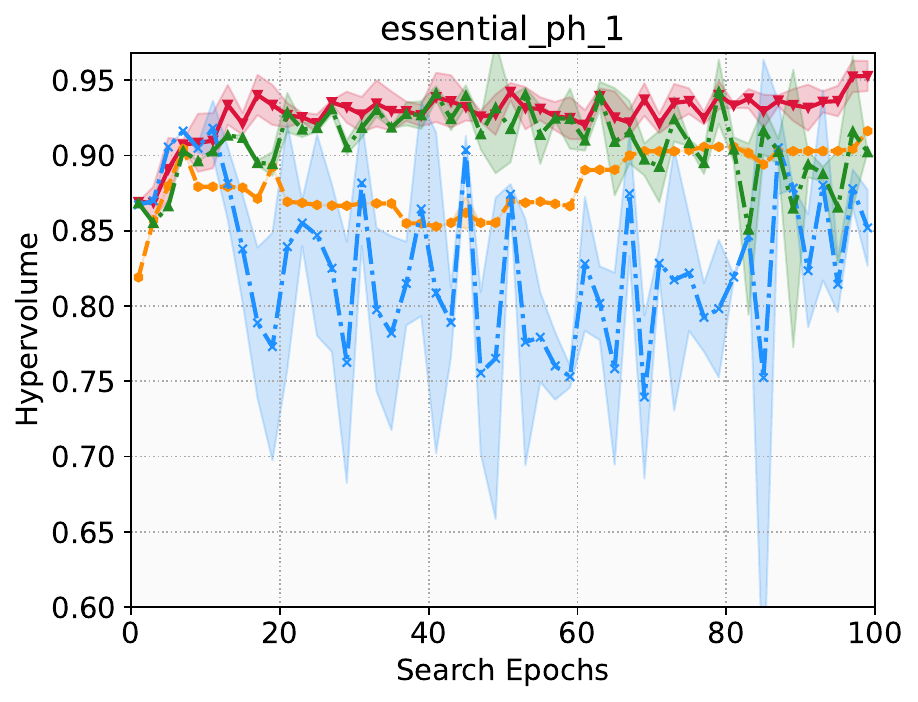}
    \includegraphics[width=.24\linewidth]{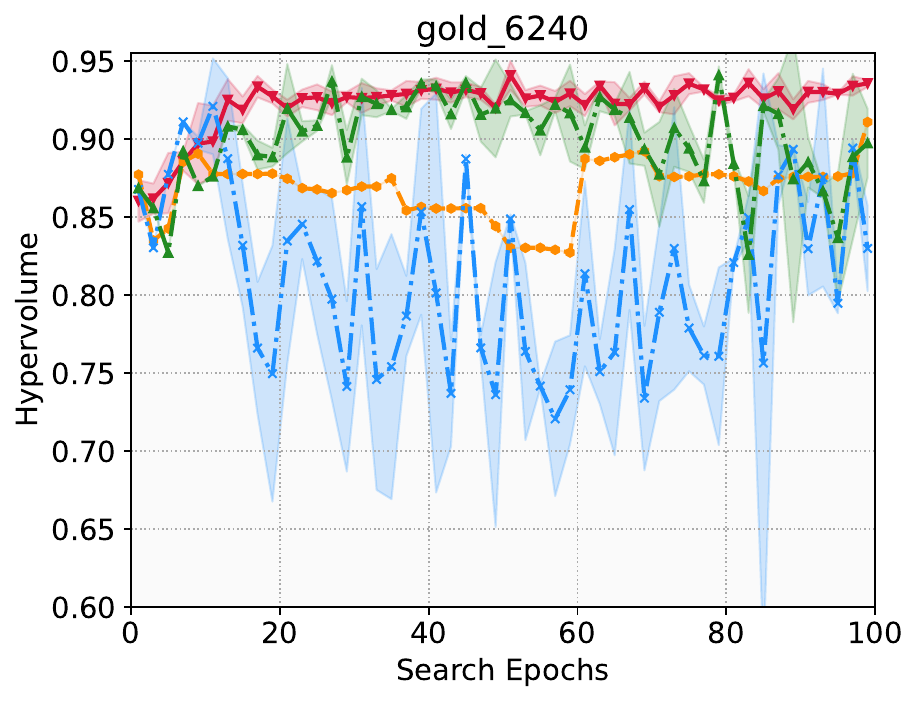}
    \includegraphics[width=.24\linewidth]{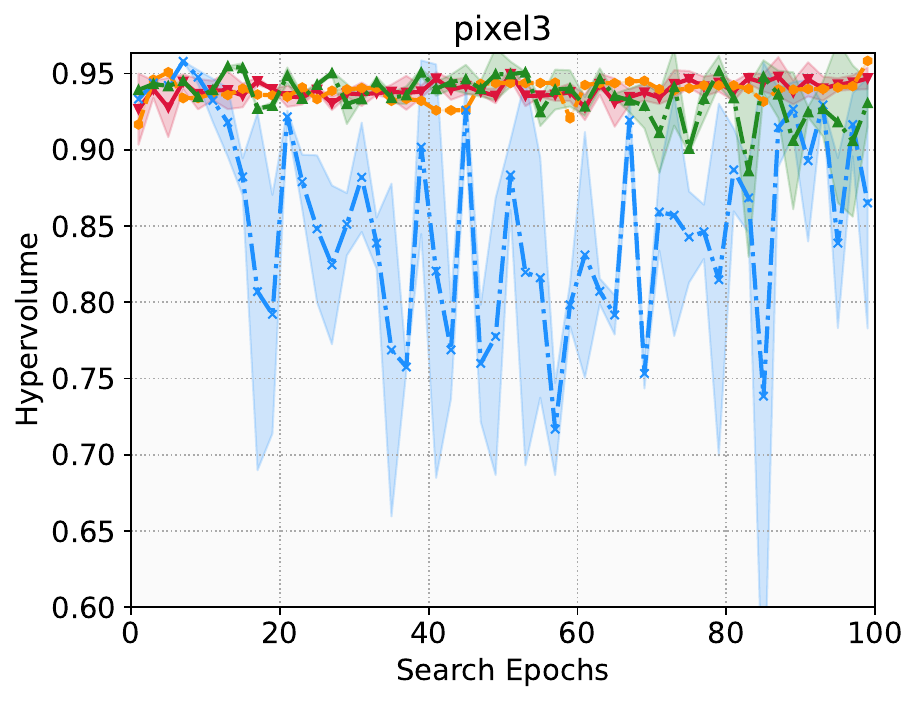} \\
    \includegraphics[width=.24\linewidth]{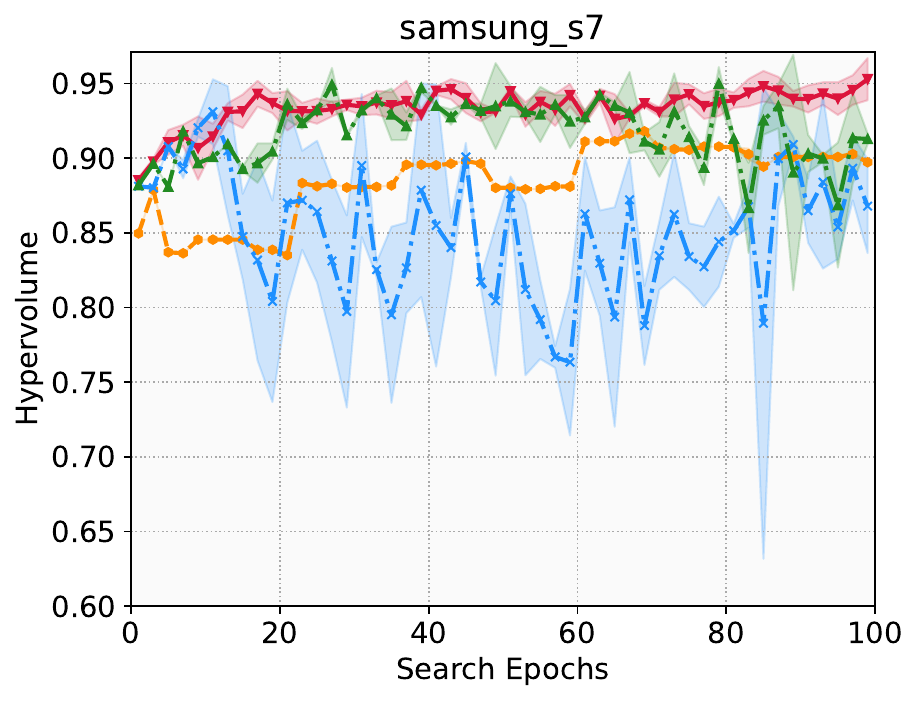}
    \includegraphics[width=.24\linewidth]{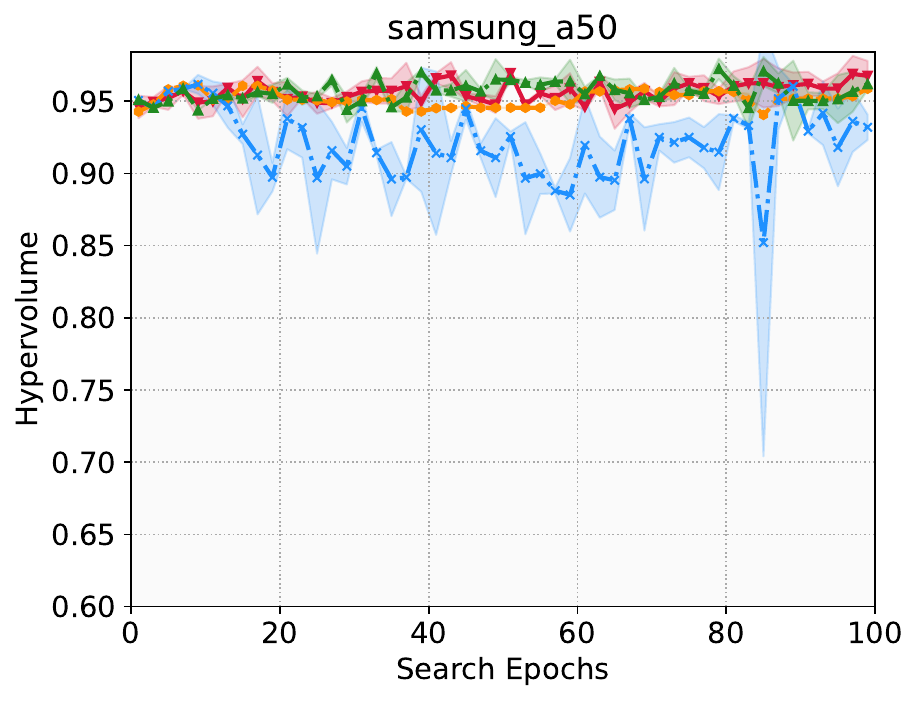}
    \includegraphics[width=.24\linewidth]{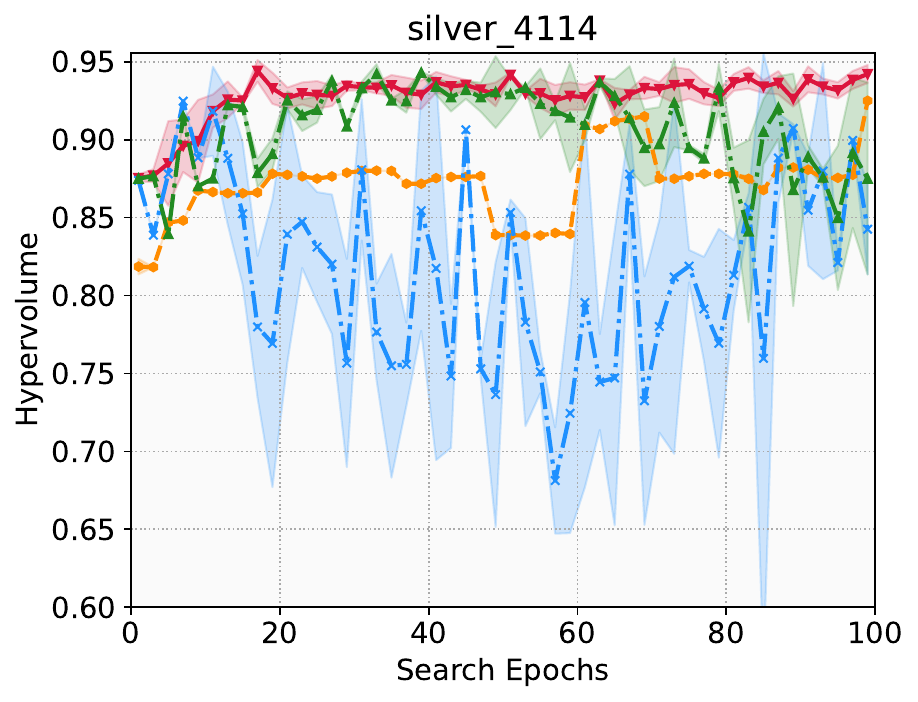}
    \includegraphics[width=.24\linewidth]{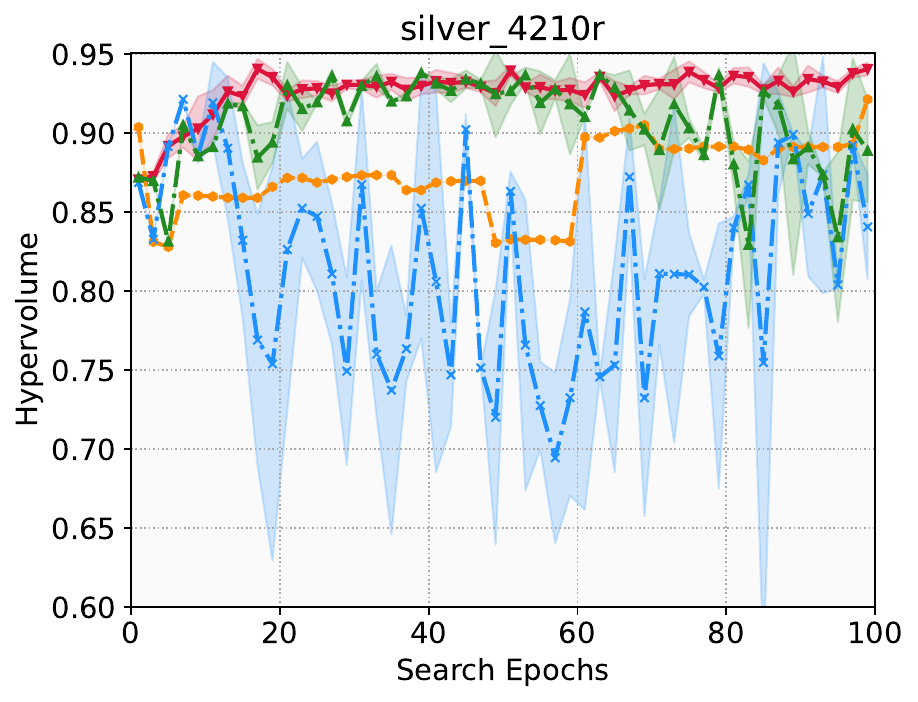}\\
    \includegraphics[width=.24\linewidth]{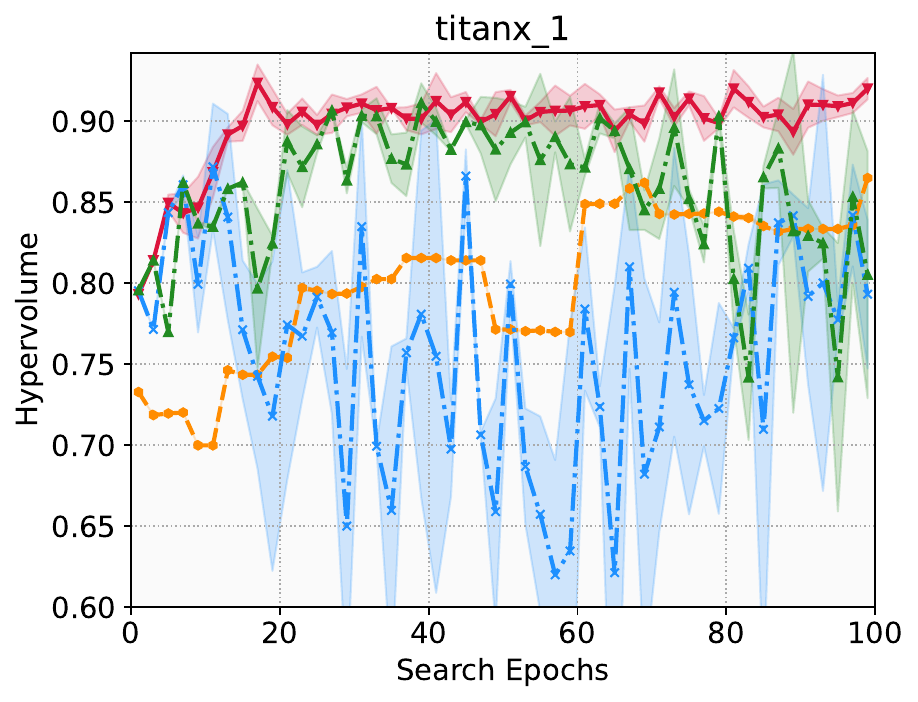}
    \includegraphics[width=.24\linewidth]{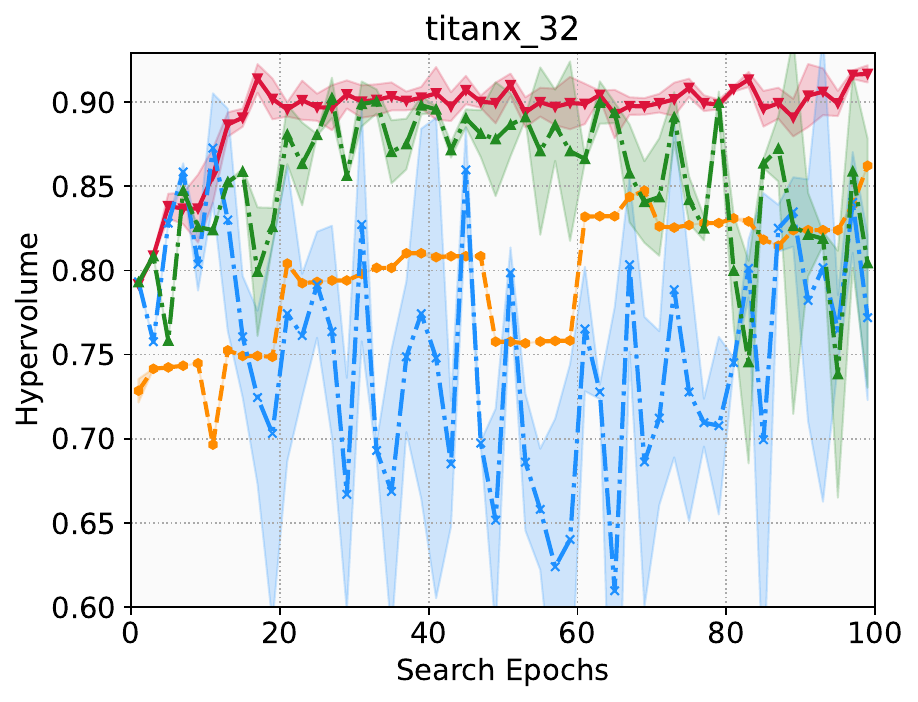}
    \includegraphics[width=.24\linewidth]{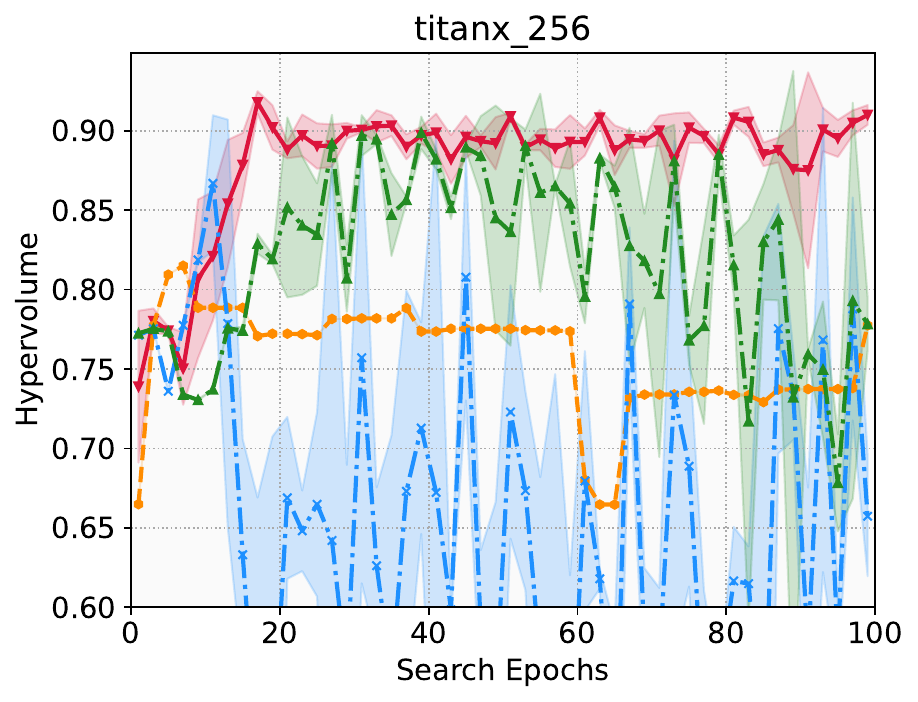}
    \caption{HV over time on NAS-Bench-201 of MODNAS with different gradient update schemes.}
    \label{fig:gradschemes_hv_full}
\end{figure}

\begin{figure}[t!]
    \centering
    \includegraphics[width=.24\linewidth]{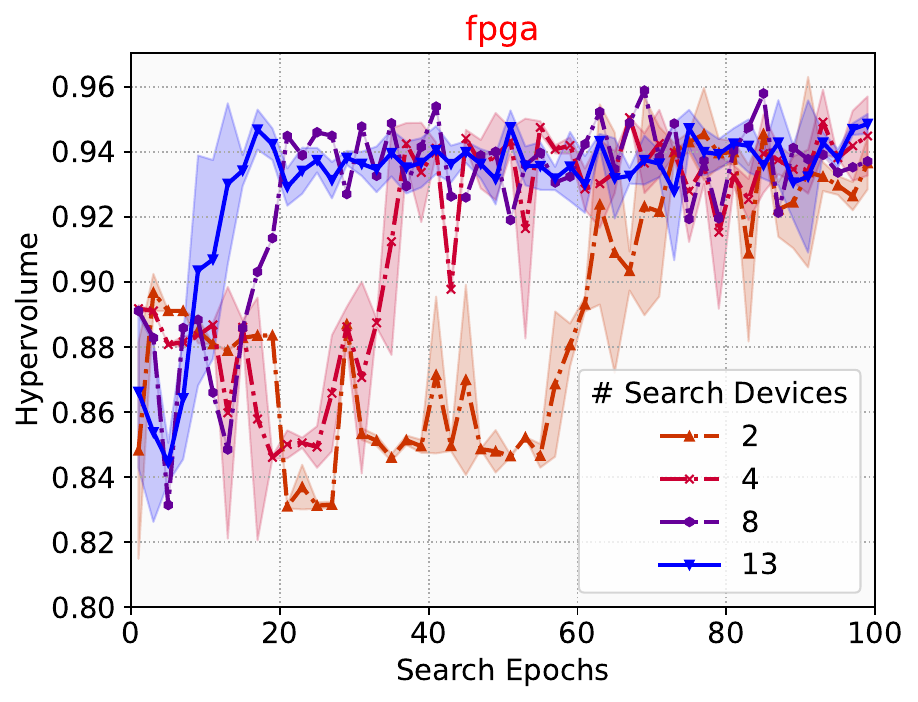}
    \includegraphics[width=.24\linewidth]{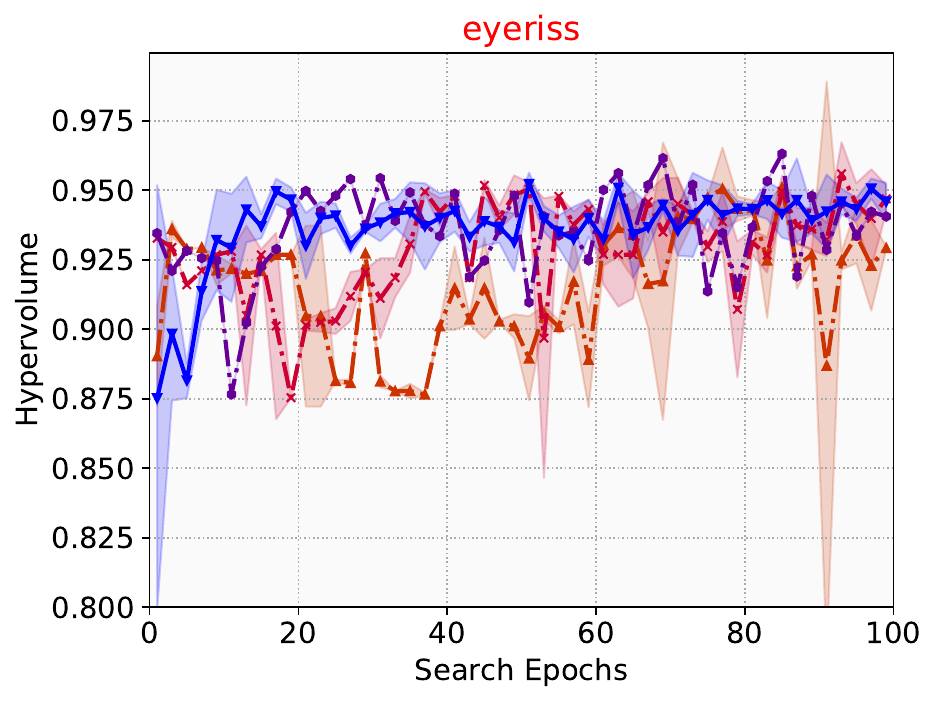}
    \includegraphics[width=.24\linewidth]{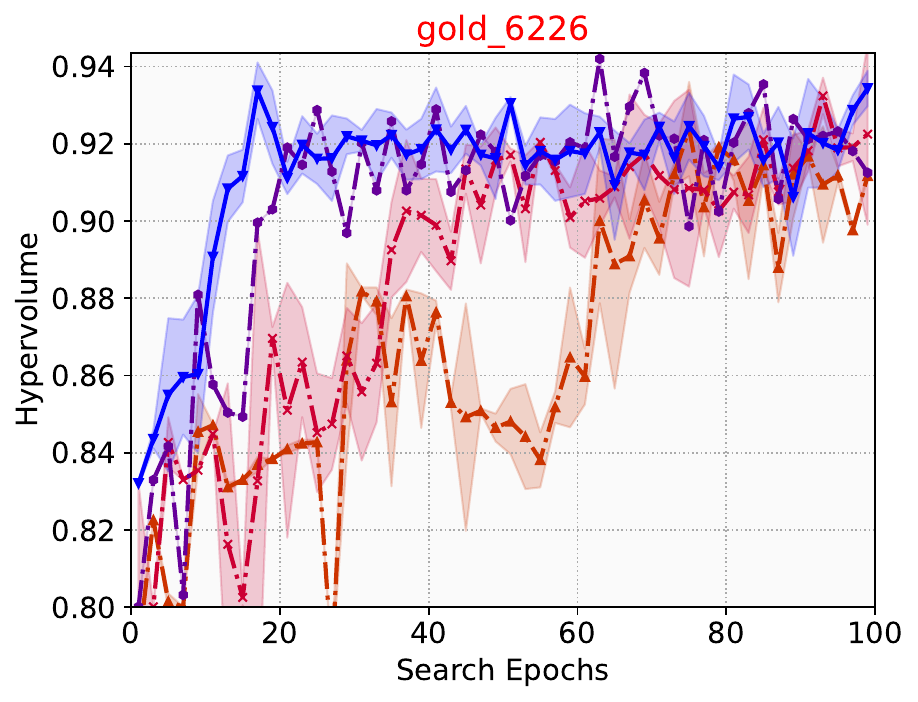}
    \includegraphics[width=.24\linewidth]{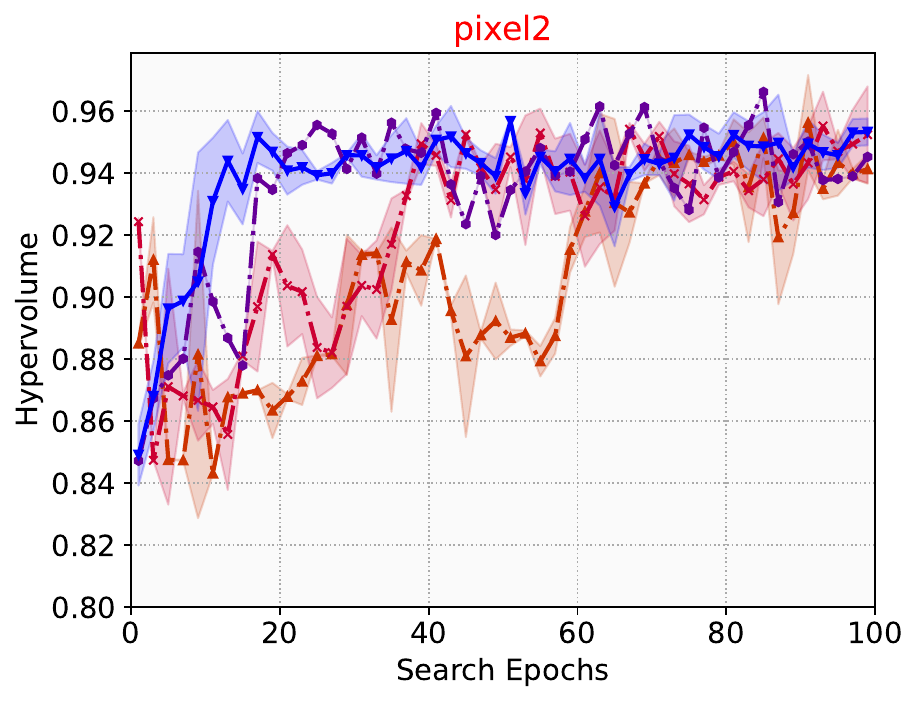} \\
    \includegraphics[width=.24\linewidth]{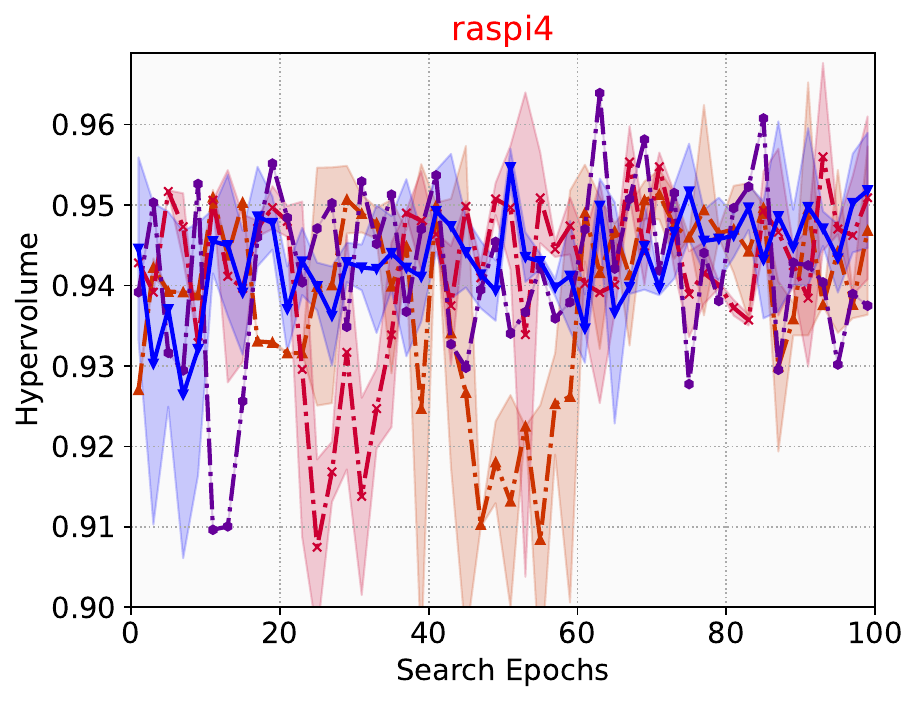}
    \includegraphics[width=.24\linewidth]{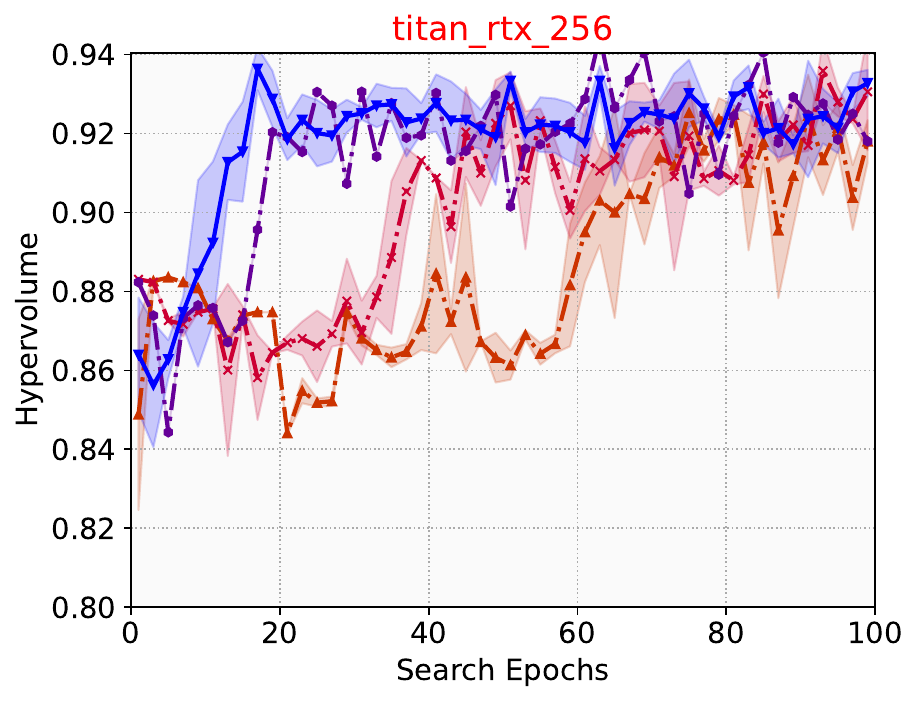}
    \includegraphics[width=.24\linewidth]{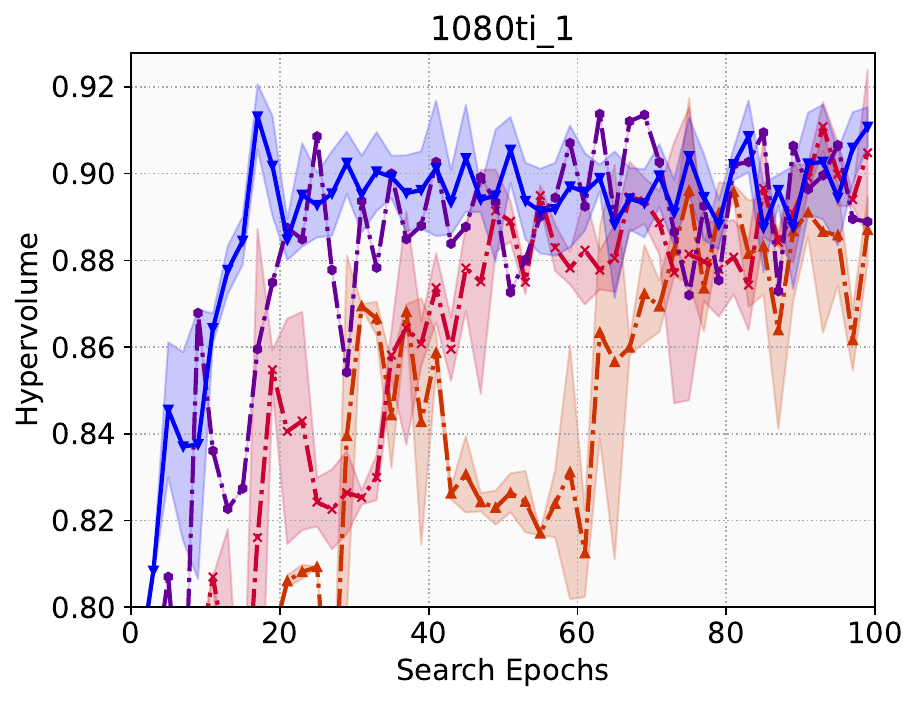}
    \includegraphics[width=.24\linewidth]{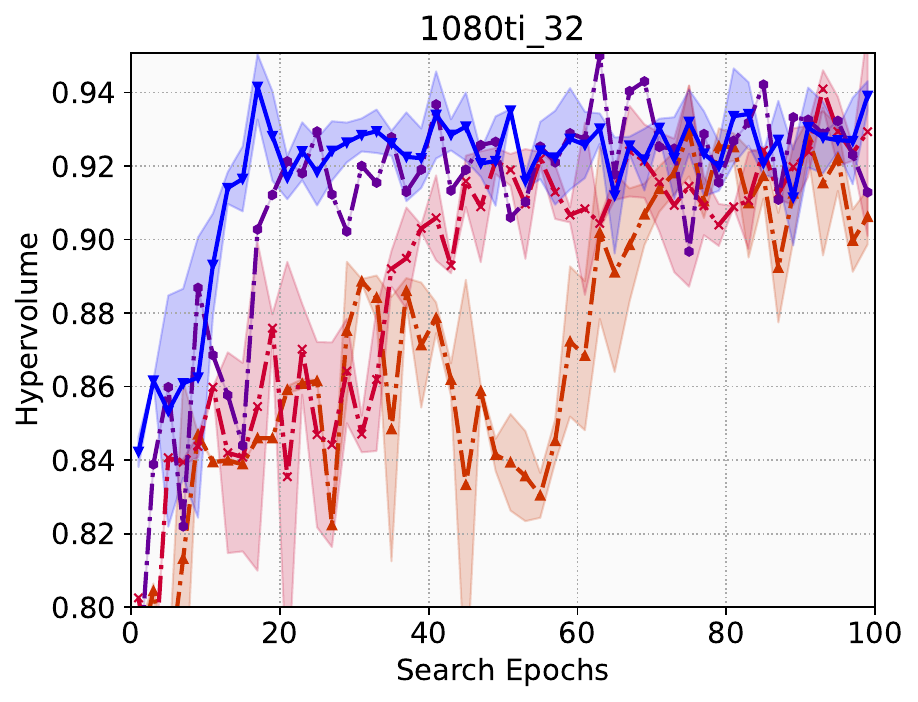}\\
    \includegraphics[width=.24\linewidth]{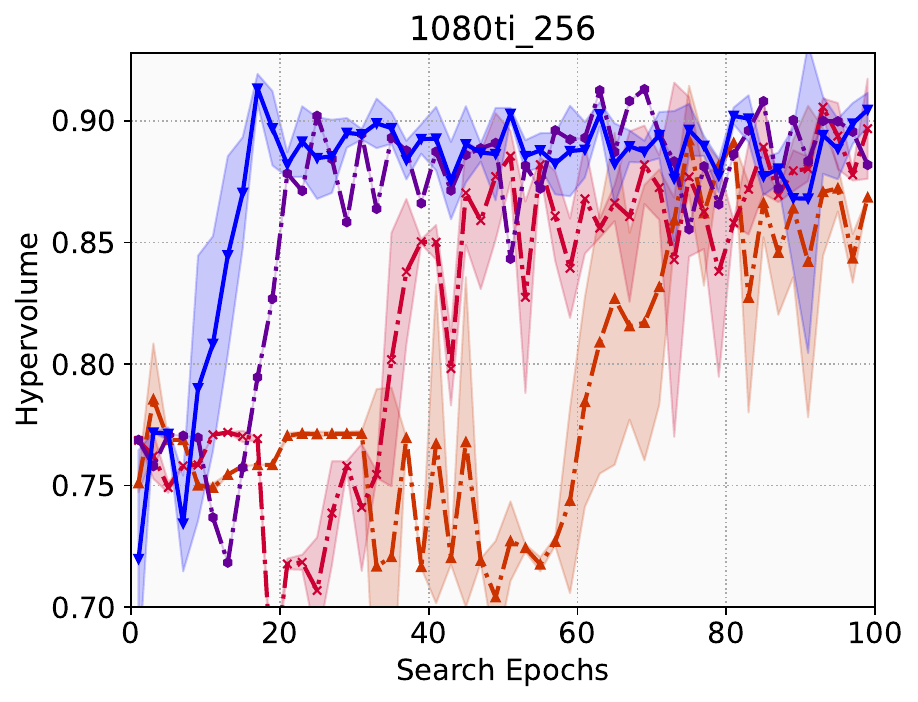}
    \includegraphics[width=.24\linewidth]{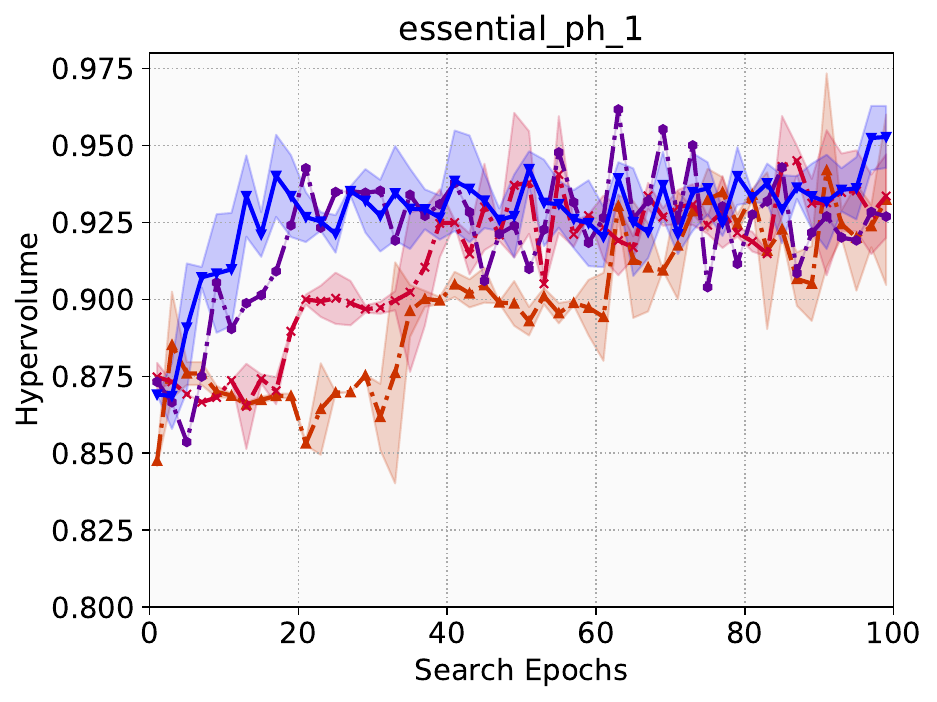}
    \includegraphics[width=.24\linewidth]{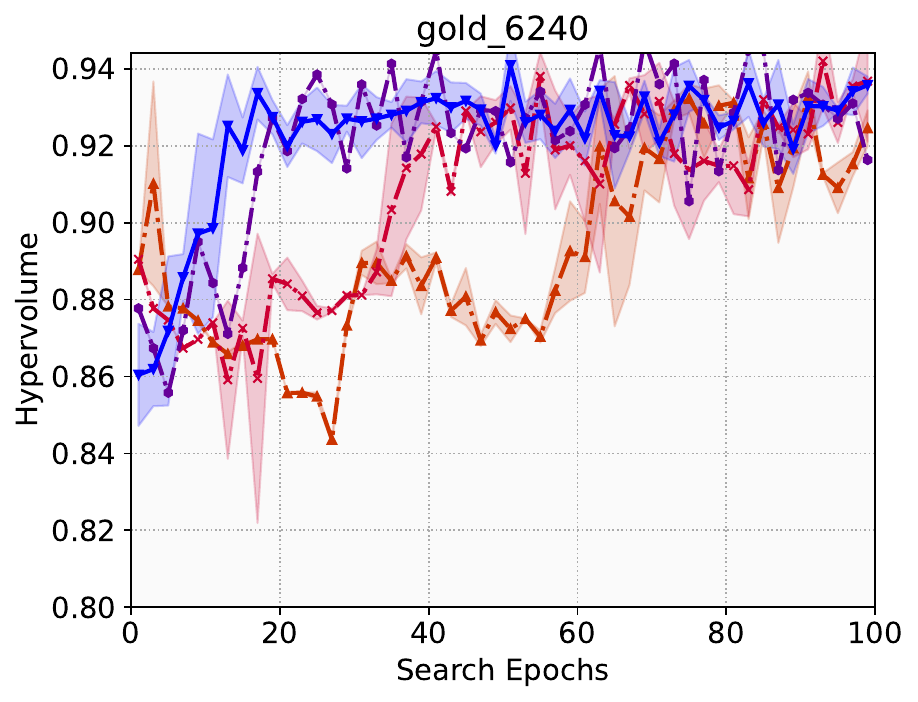}
    \includegraphics[width=.24\linewidth]{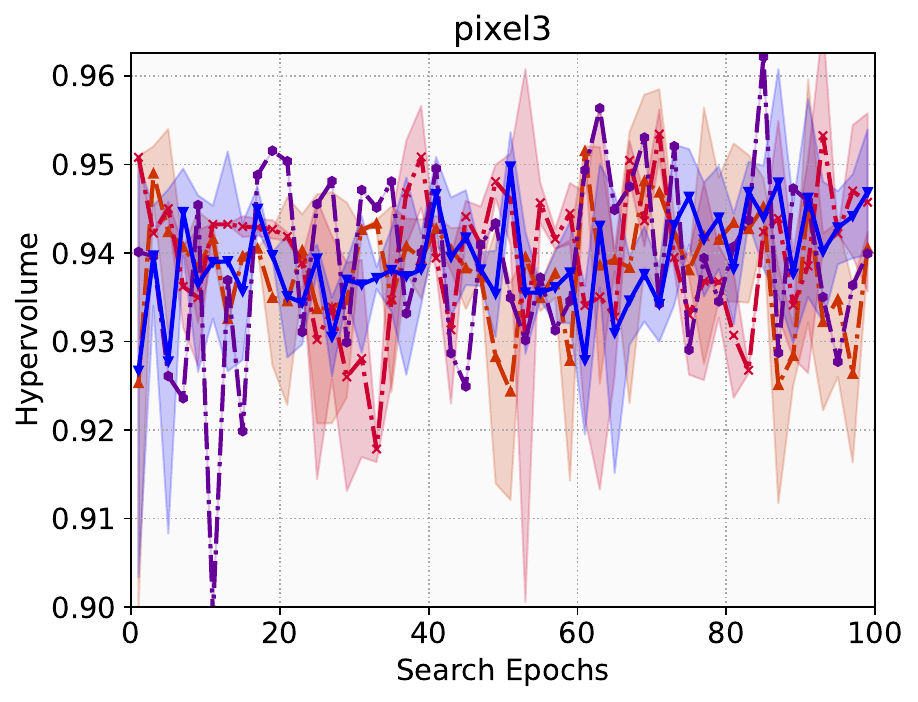} \\
    \includegraphics[width=.24\linewidth]{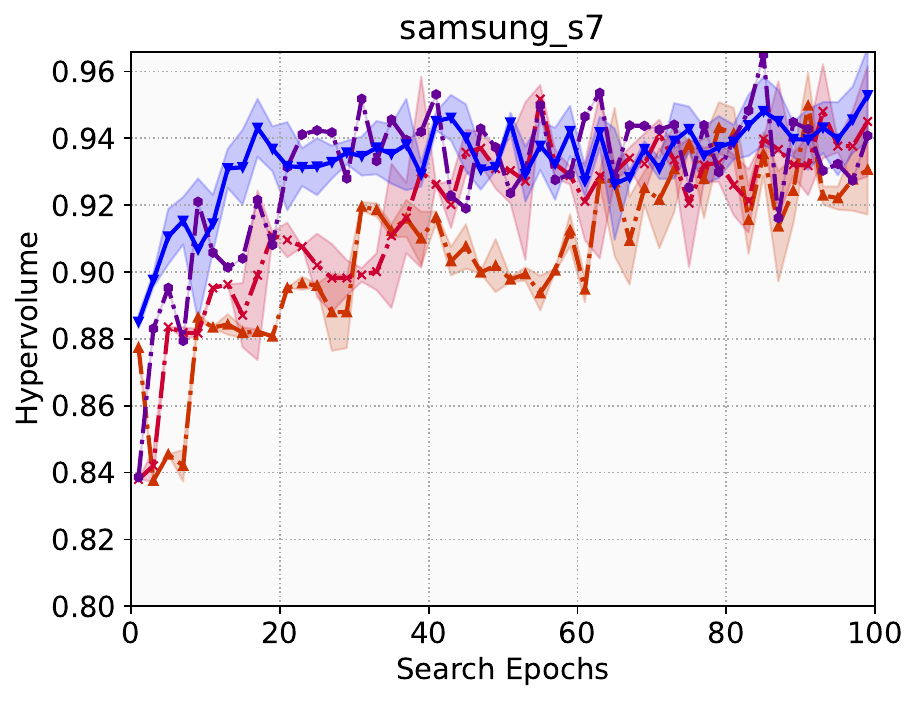}
    \includegraphics[width=.24\linewidth]{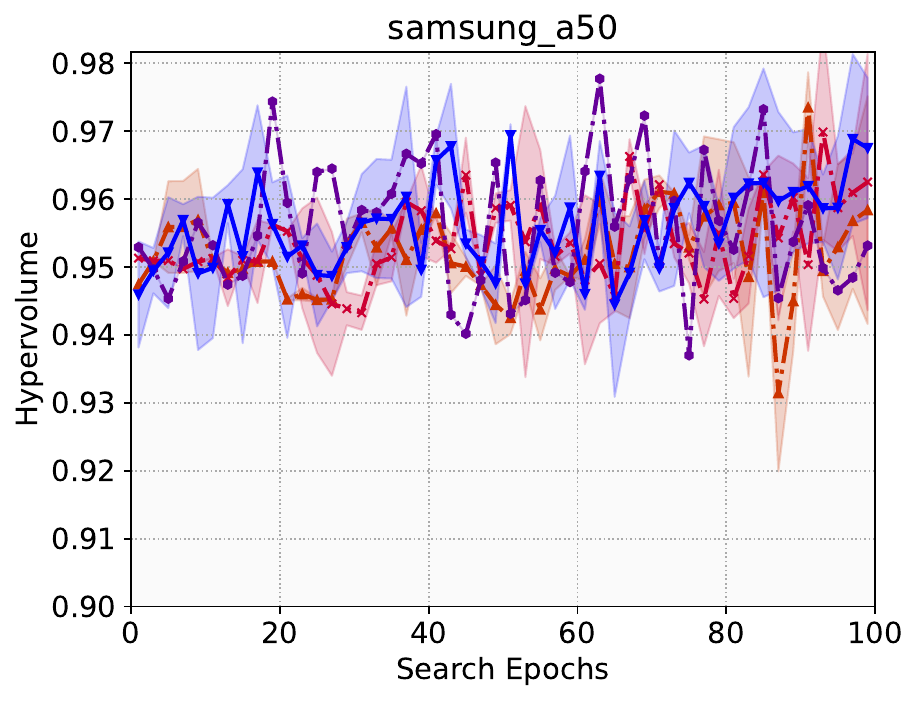}
    \includegraphics[width=.24\linewidth]{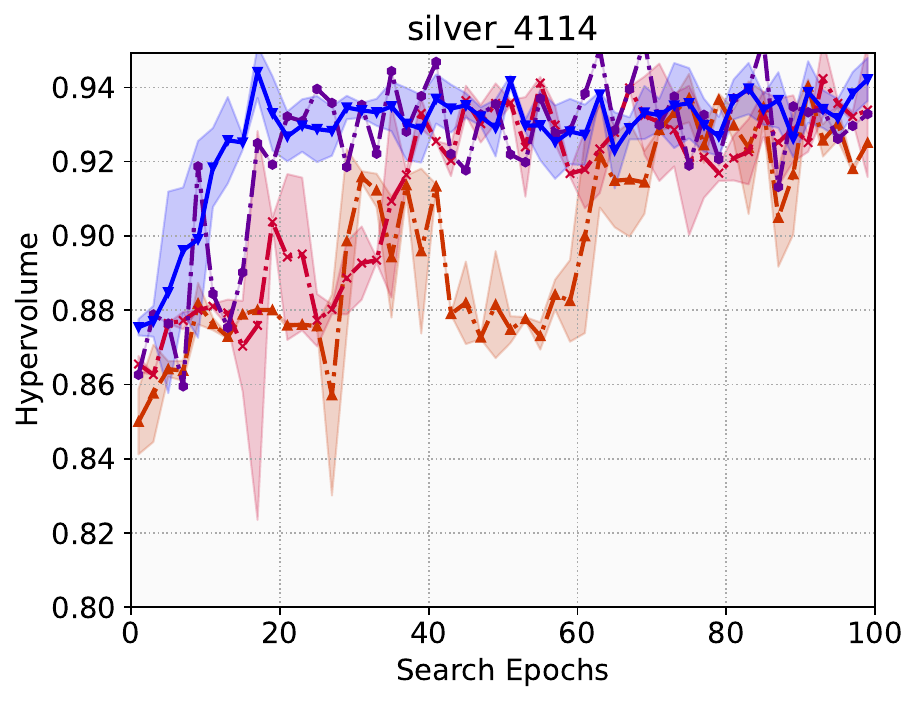}
    \includegraphics[width=.24\linewidth]{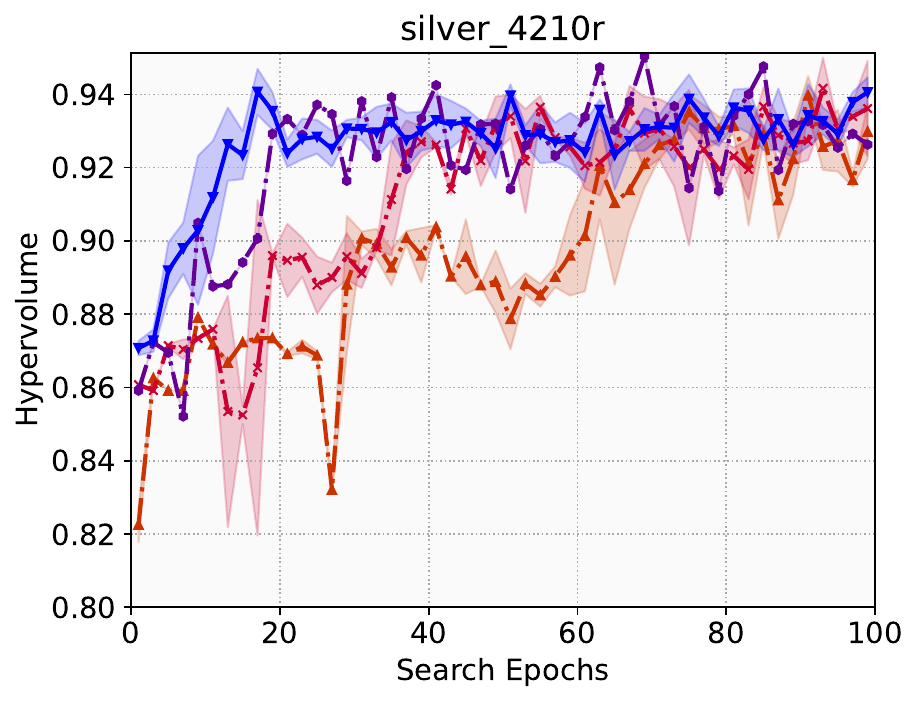}\\
    \includegraphics[width=.24\linewidth]{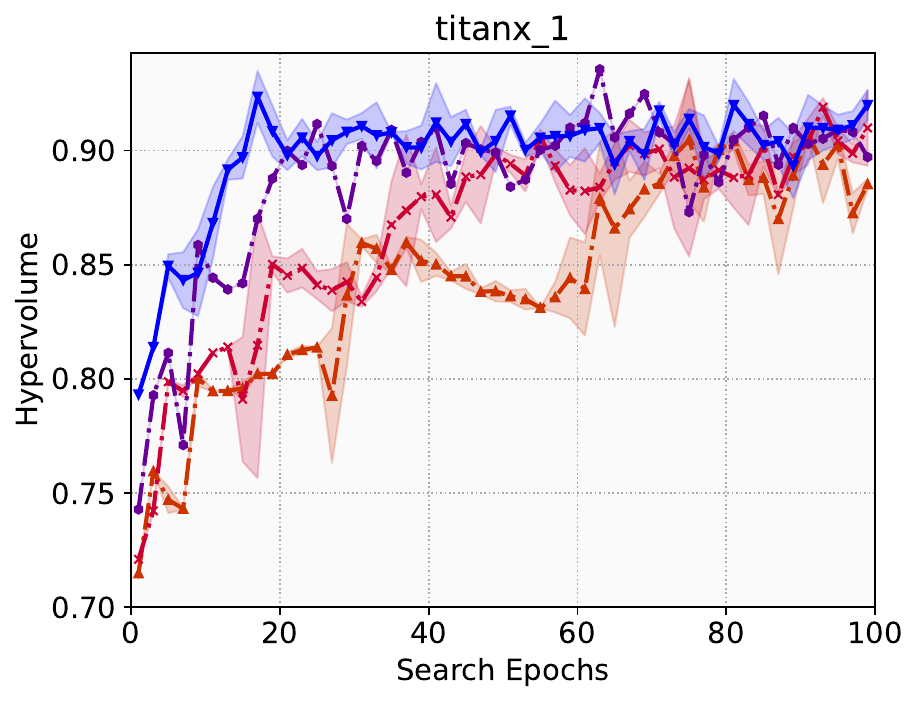}
    \includegraphics[width=.24\linewidth]{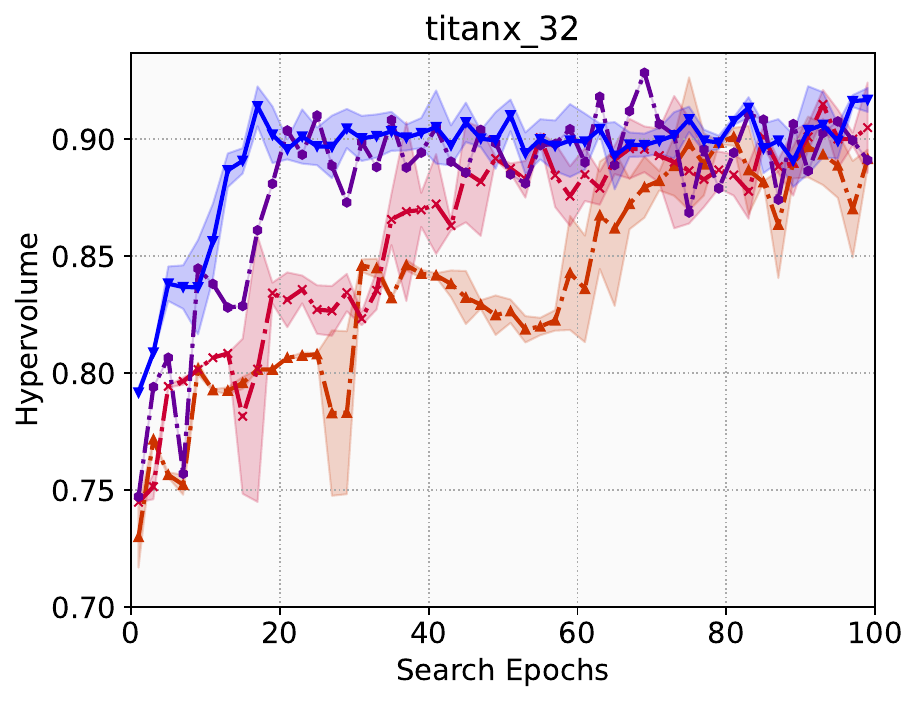}
    \includegraphics[width=.24\linewidth]{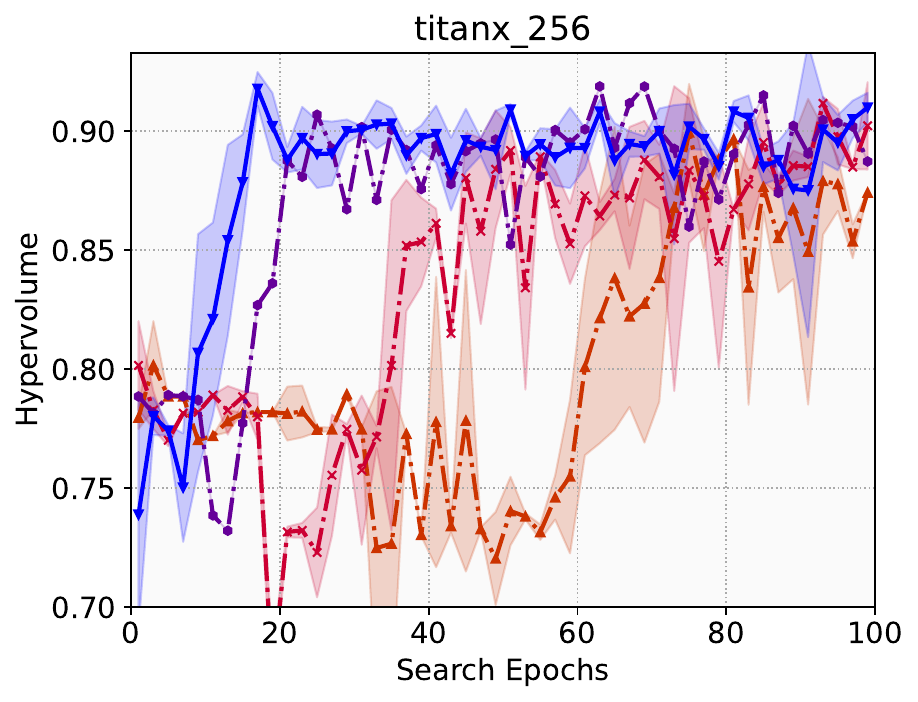}
    \caption{HV over time on NAS-Bench-201 of MODNAS with different number of devices during search. For number of devices less than 13 (default one) we randomly select a subset from these 13 devices.}
    \label{fig:ndevices_hv_full}
\end{figure}

In Figure~\ref{fig:pareto_nb201_full}, we present the Pareto fronts obtained by our method in comparison to different baselines on the NAS-Bench-201 search space. In Figure~\ref{fig:radar_plot_gd_igd}, we present different additional metrics, such as GD and IGD (see Section~\ref{sec:metrics}), to evaluate the quality of the Pareto fronts obtained on NAS-Bench-201. Figure~\ref{fig:fronts_constraints_full} presents the Pareto front MODNAS yields when applying different latency constraints during the search phase. Figure~\ref{fig:radar_hv_gdas} compares our method using the ReinMax gradient estimator to the GDAS estimator~\citep{dong-cvpr19}. As we can see, ReinMax obtains a qualitatively better hypervolume coverage compared to GDAS. Figure~\ref{fig:3d_plot_eyeriss} presents the 3D Pareto front and hypervolume obtained by \methodname compared to other baselines when optimizing for accuracy, latency and energy usage on NAS-Bench-201. Figure~\ref{fig:gradschemes_hv_full} presents the comparison of \methodname with MGD to other gradient aggregation schemes, such as mean, sequential and MC sampling (see Section~\ref{sec:nb201_experiments}), across multiple hardware devices. Finally, in Figure~\ref{fig:ndevices_hv_full} we present the robustness of \methodname to the fraction of devices used for the predictor training and the search phase.  In addition, in Figure~\ref{fig:pixel3-c100} and \ref{fig:edgegpu-c100}, we compare MODNAS against different MO baselines on the CIFAR-100 dataset on two different devices. 

\subsubsection{Predicted v/s Ground-Truth Latencies}

In Figure~\ref{fig:lat_pred}, we present the scatter plots of the predictions of our hardware-aware \metapredictor vs. the ground-truth latencies of different architectures. In the figure title we also report the kendall-tau correlation coefficient for every device. As observed, our predictor achieves high kendall-$\tau$ correlation coefficient across all devices.

\begin{figure}[t!]
\centering
\begin{minipage}{0.39\textwidth}
  \centering
\includegraphics[width=\textwidth]{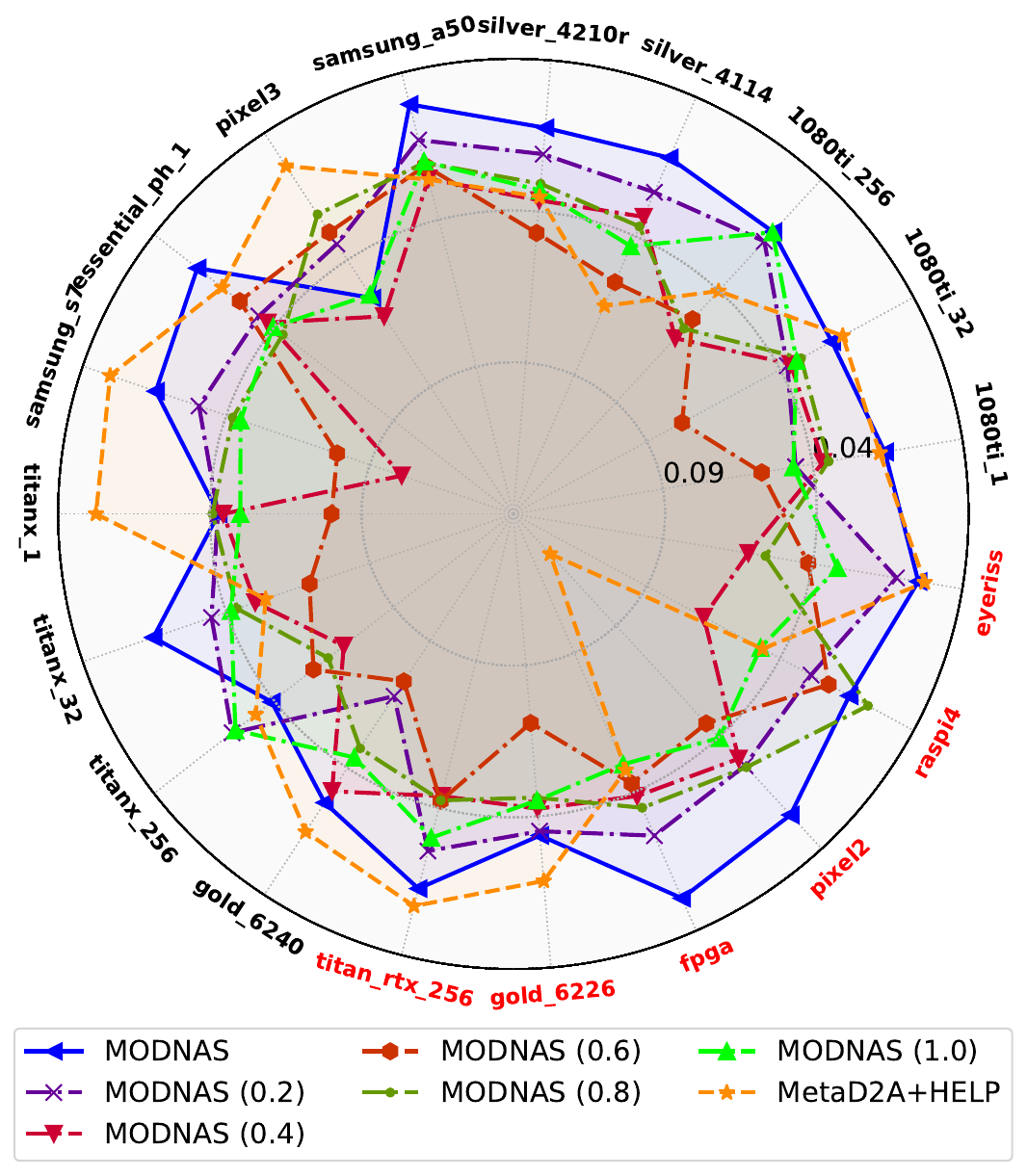}
\subcaption[gd_c]{GD}\label{fig:radar_gd_c}
\end{minipage}%
\hspace{1cm}
\begin{minipage}{0.39\textwidth}
  \centering
\includegraphics[width=\textwidth]{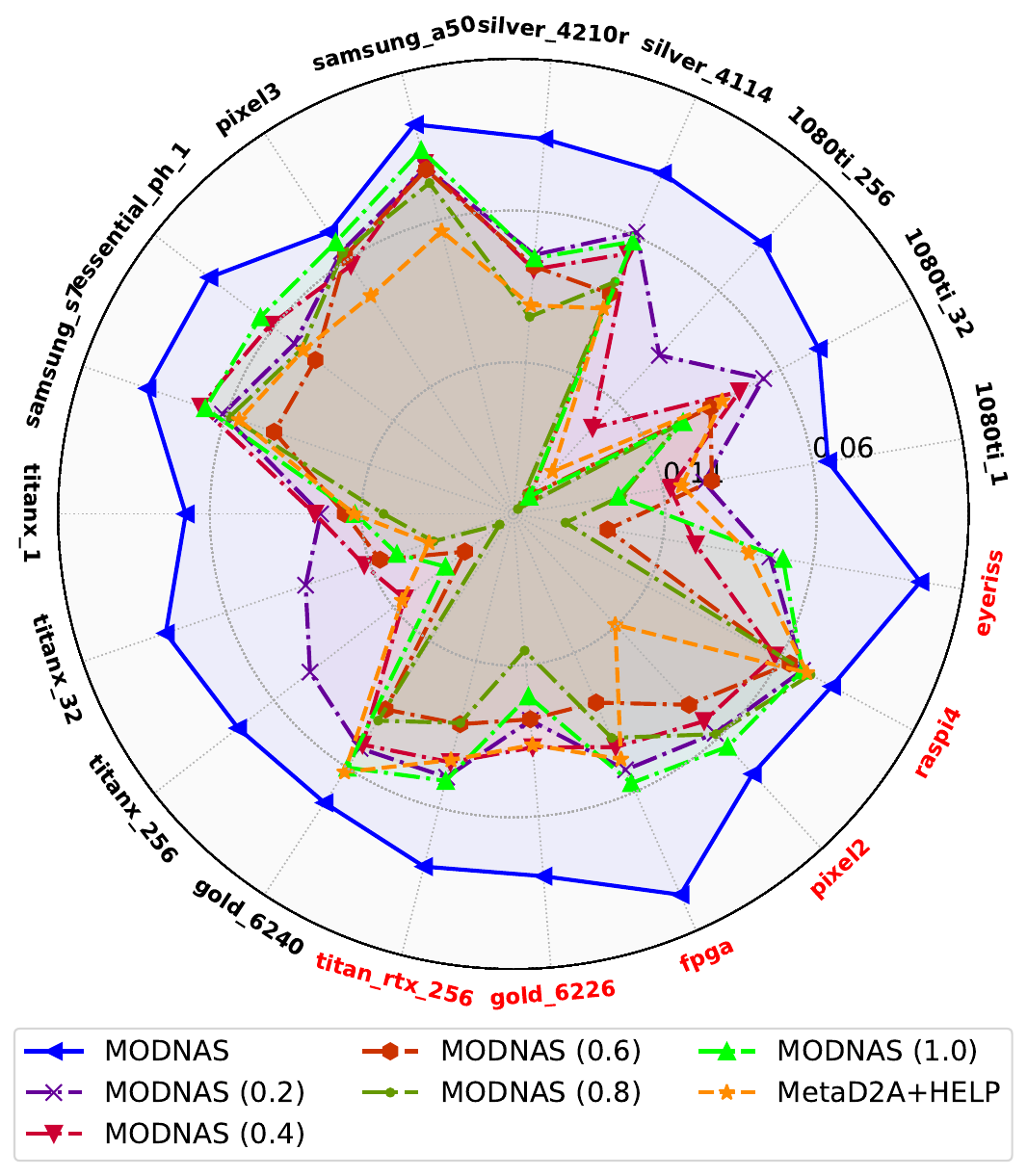}
\subcaption[igd_c]{$\text{IGD}$}\label{fig:radar_igd_c}
\end{minipage}
\vspace{10pt}
\begin{minipage}{0.39\textwidth}
  \centering
\includegraphics[width=\textwidth]{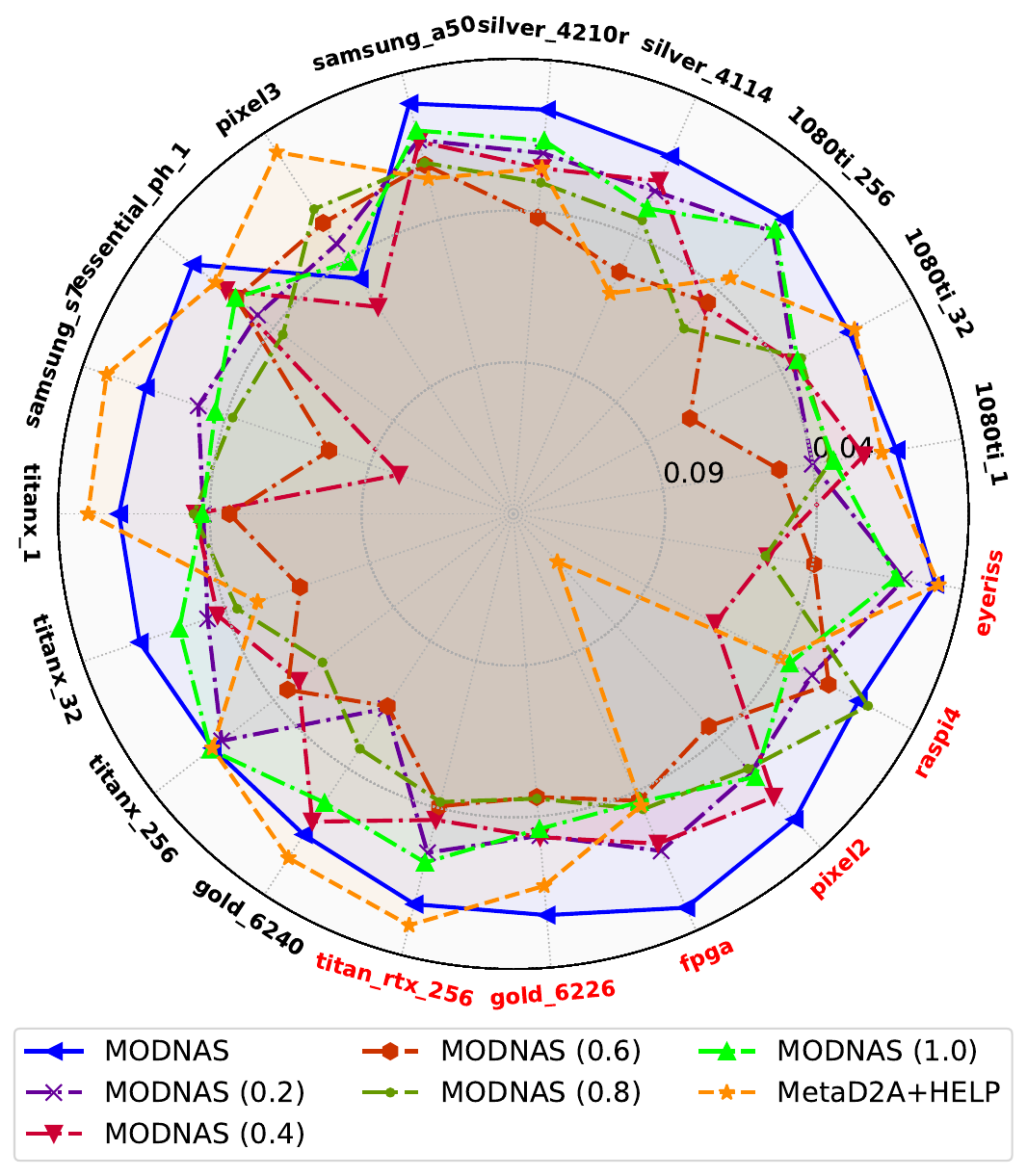}
\subcaption[gdplus_c]{GD+}\label{fig:radar_gdplus_c}
\end{minipage}
\hspace{1cm}
\begin{minipage}{0.39\textwidth}
  \centering
\includegraphics[width=\textwidth]{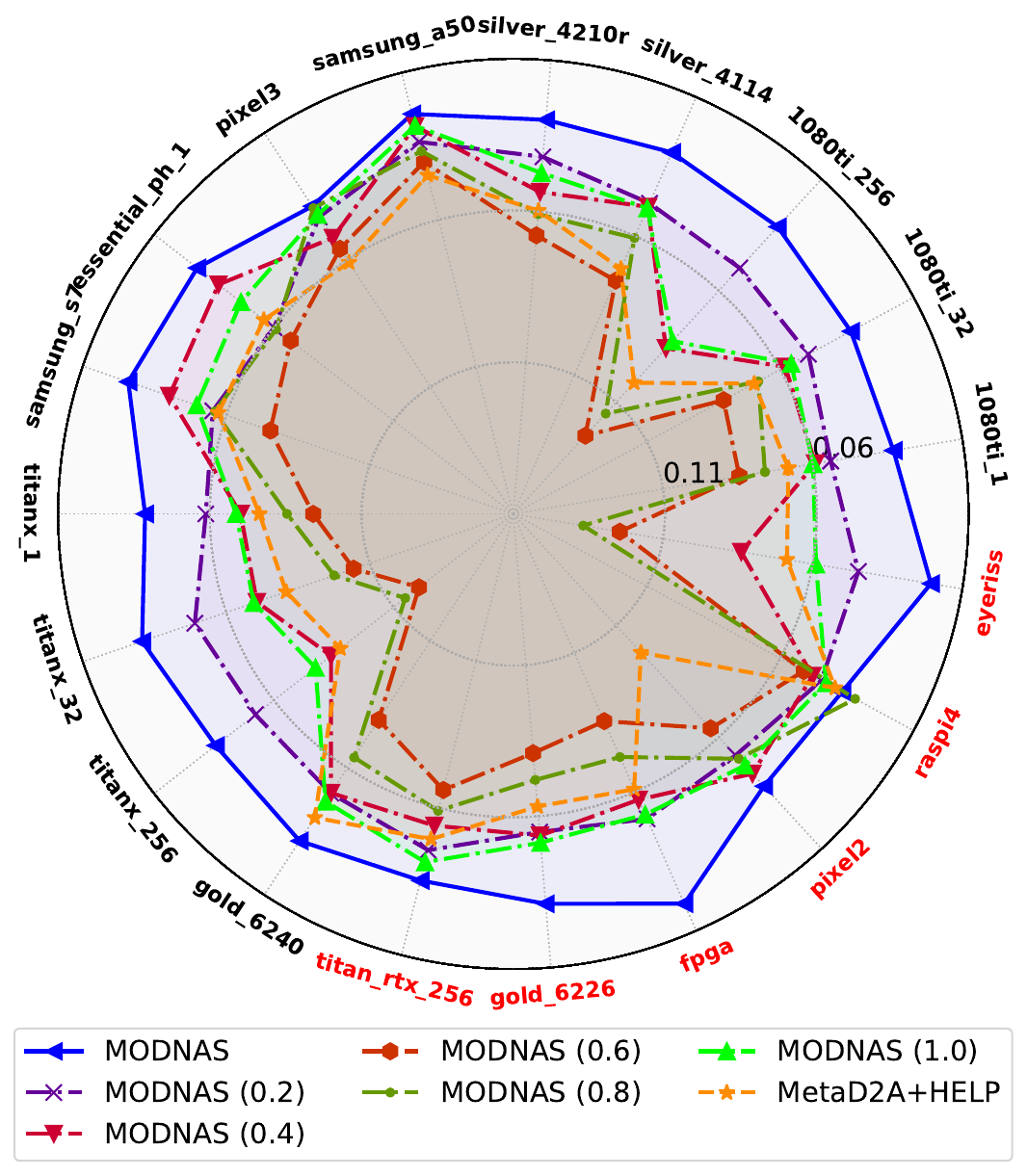}
\subcaption[igdplus_c]{$\text{IGD+}$}\label{fig:radar_igdplus_c}
\end{minipage}
\caption{GD, GD+, IGD and IGD+ of MODNAS with different latency constraints during search across 19 devices on NAS-Bench-201. Higher area in the radar indicates better performance for every metric. Test devices are colored in red around the radar plot.} 
\label{fig:radar_plot_gd_igd_const}
\end{figure}

\begin{figure}[t!]
    \centering
    \includegraphics[width=.24\linewidth]{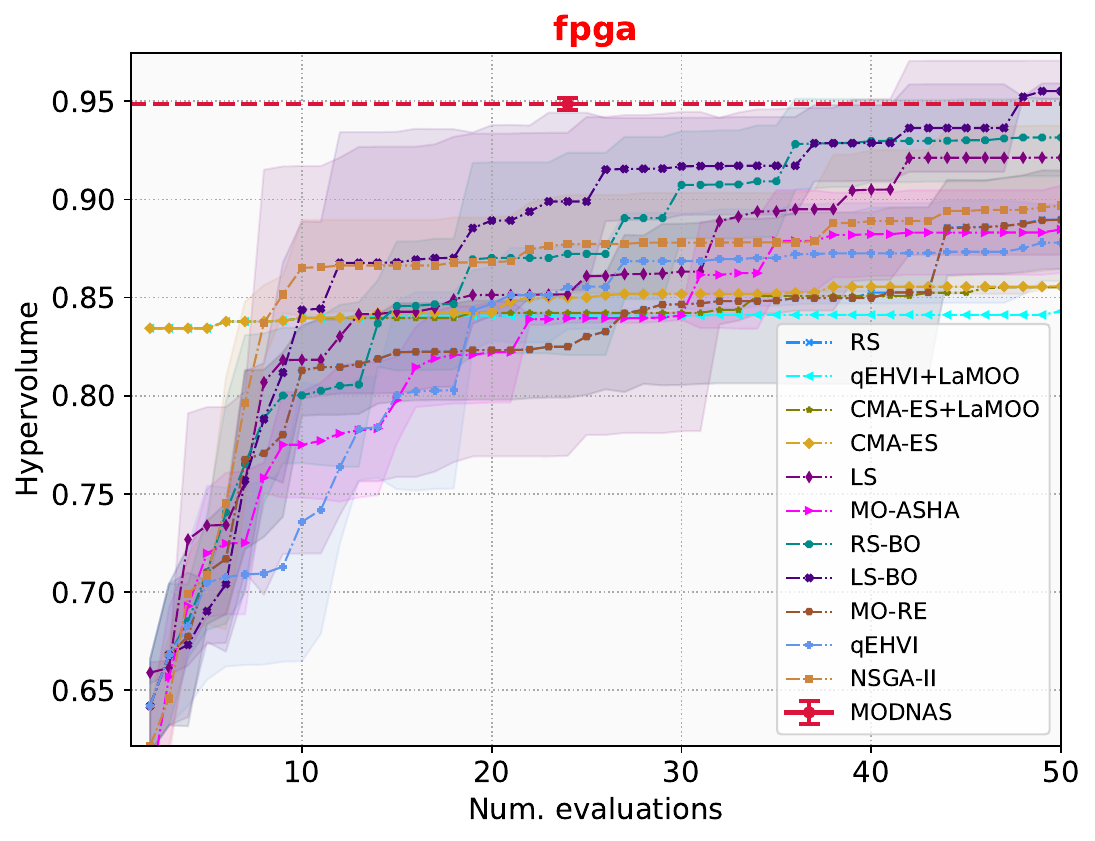}
    \includegraphics[width=.24\linewidth]{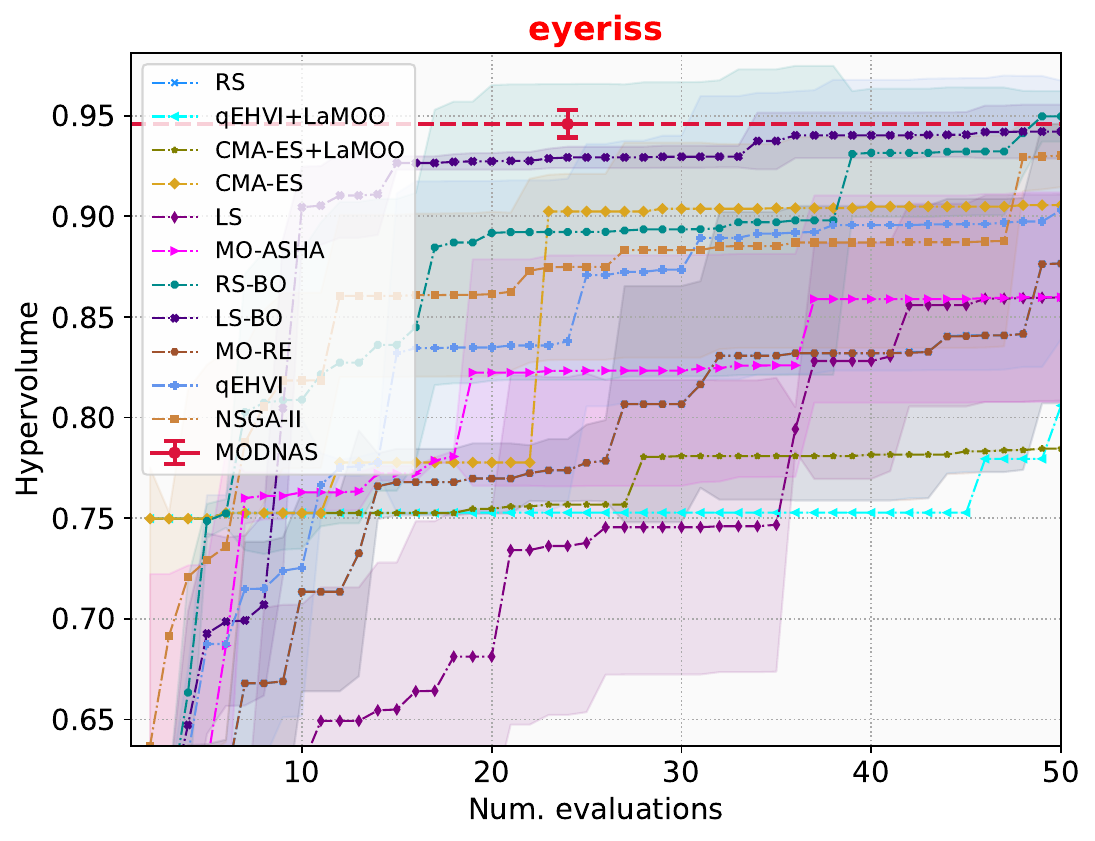}
    \includegraphics[width=.24\linewidth]{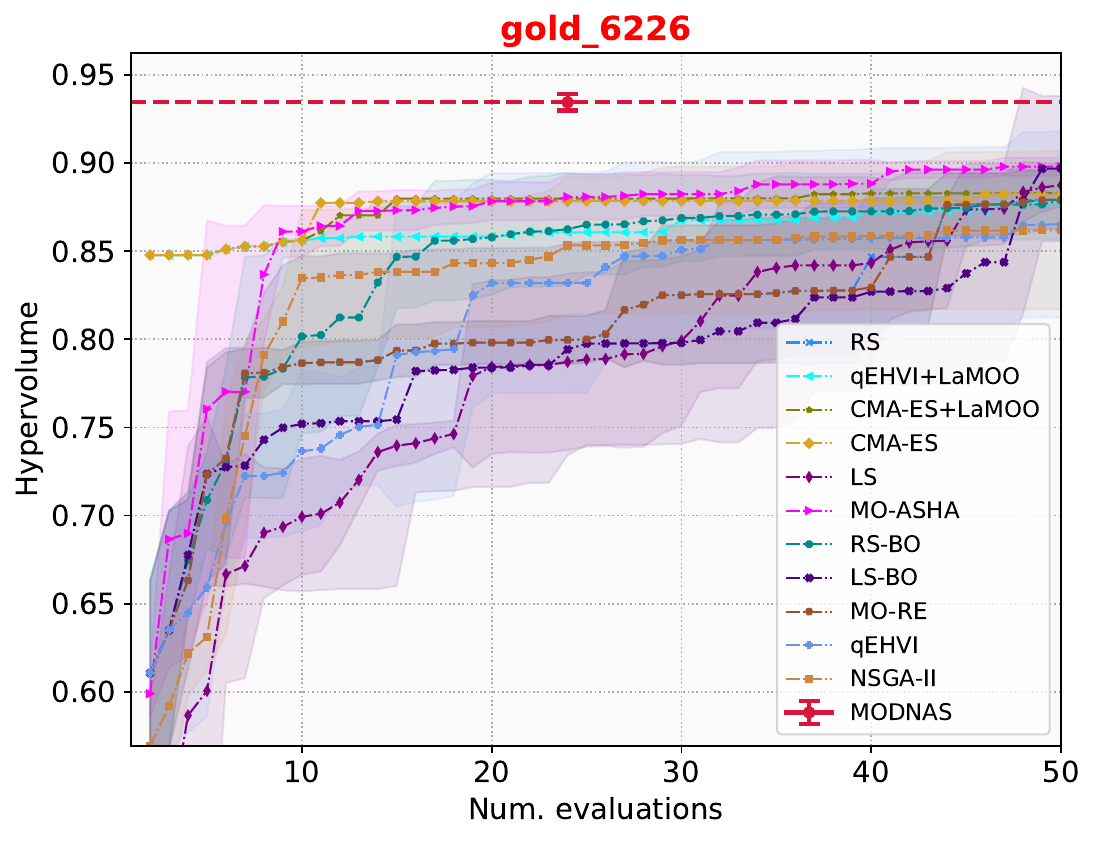}
    \includegraphics[width=.24\linewidth]{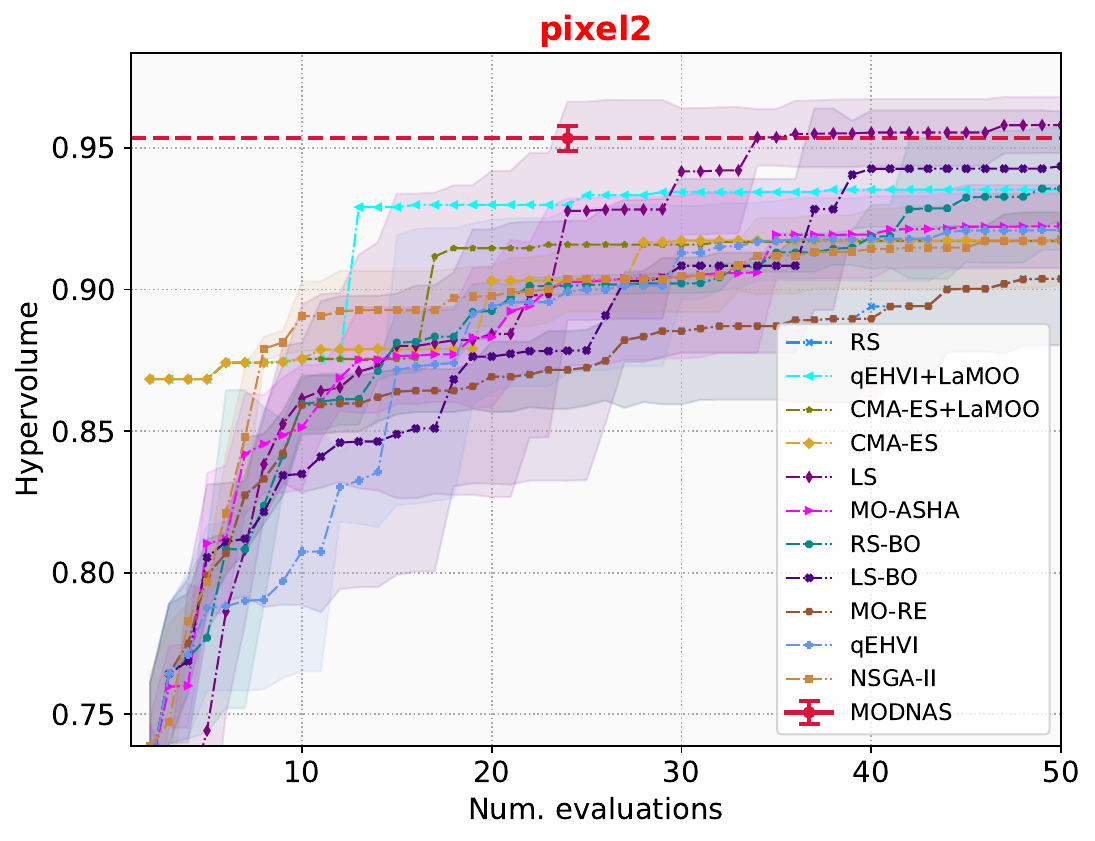} \\
    \includegraphics[width=.24\linewidth]{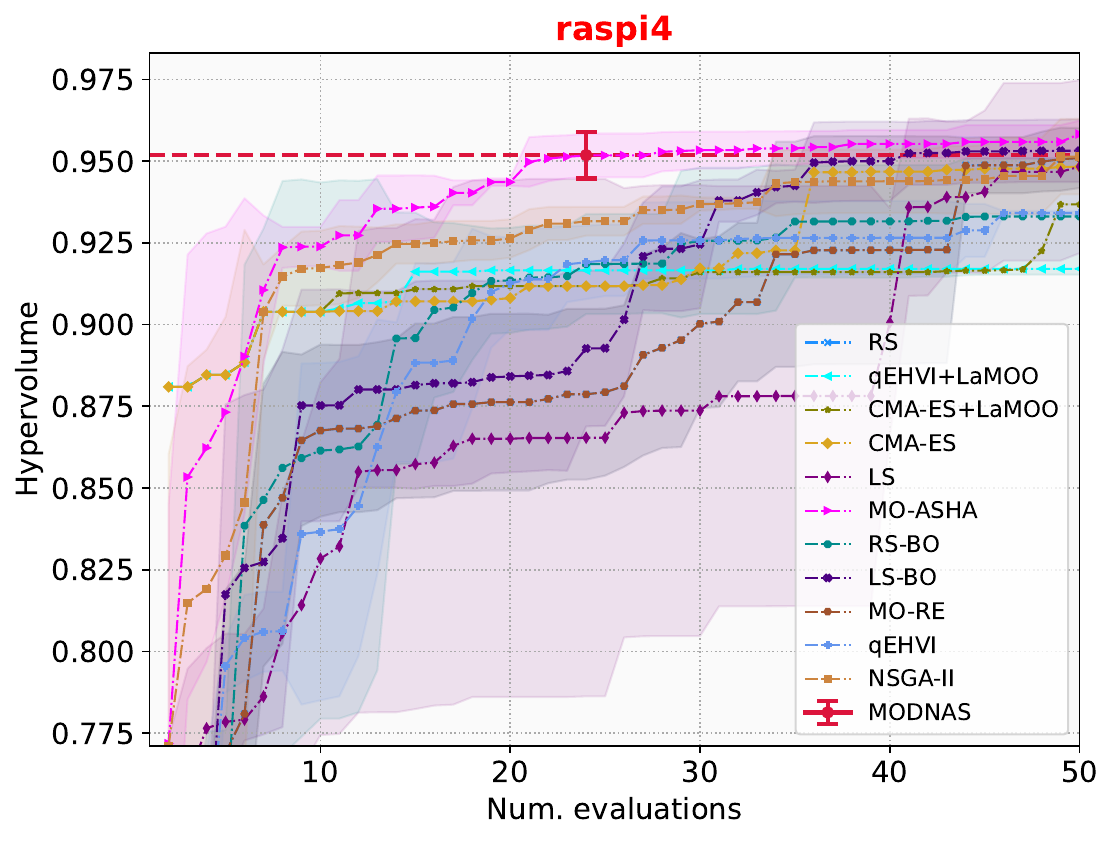}
    \includegraphics[width=.24\linewidth]{figures/hv_vs_budget/hv_titan_rtx_256.pdf}
    \includegraphics[width=.24\linewidth]{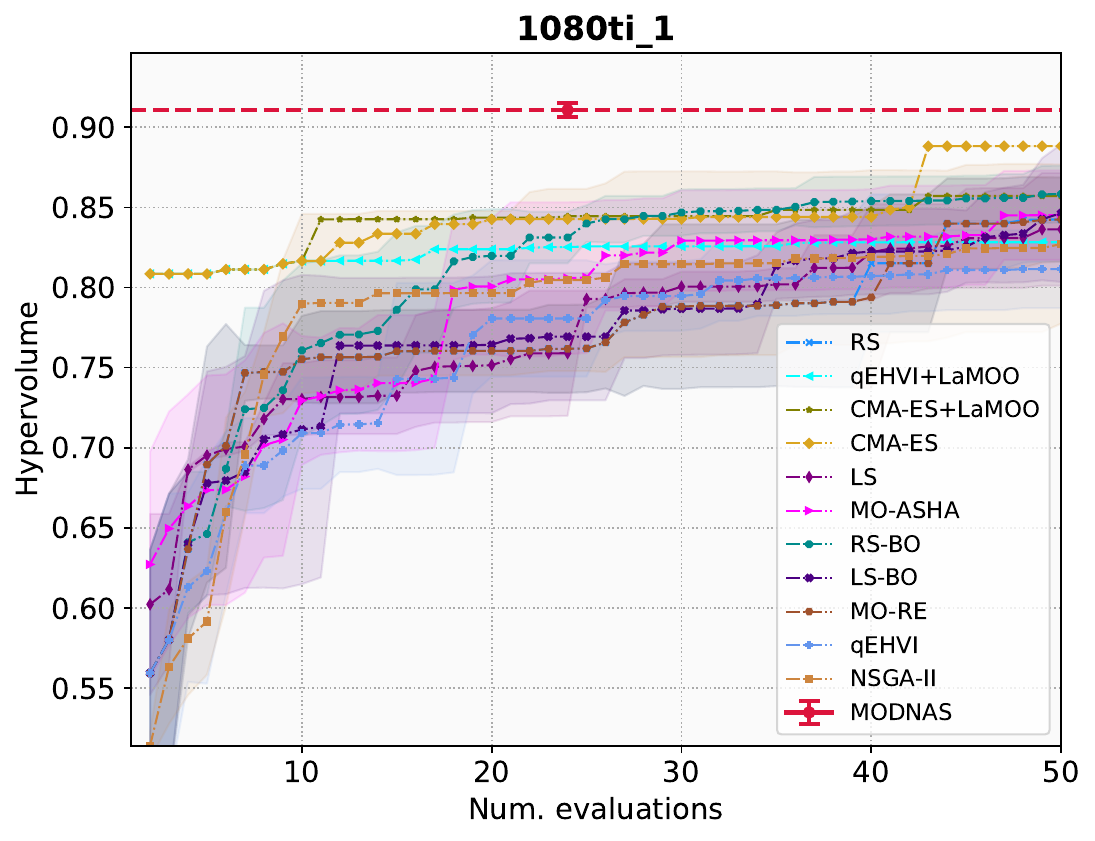}
    \includegraphics[width=.24\linewidth]{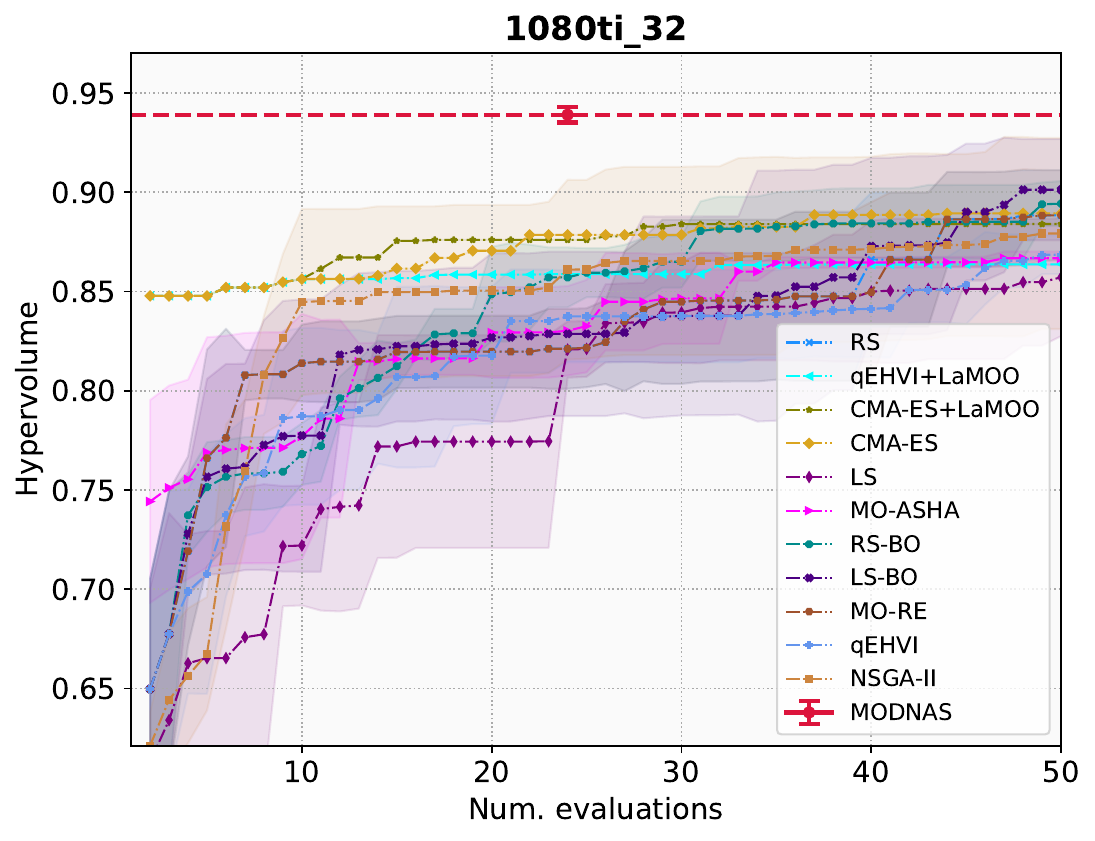}\\
    \includegraphics[width=.24\linewidth]{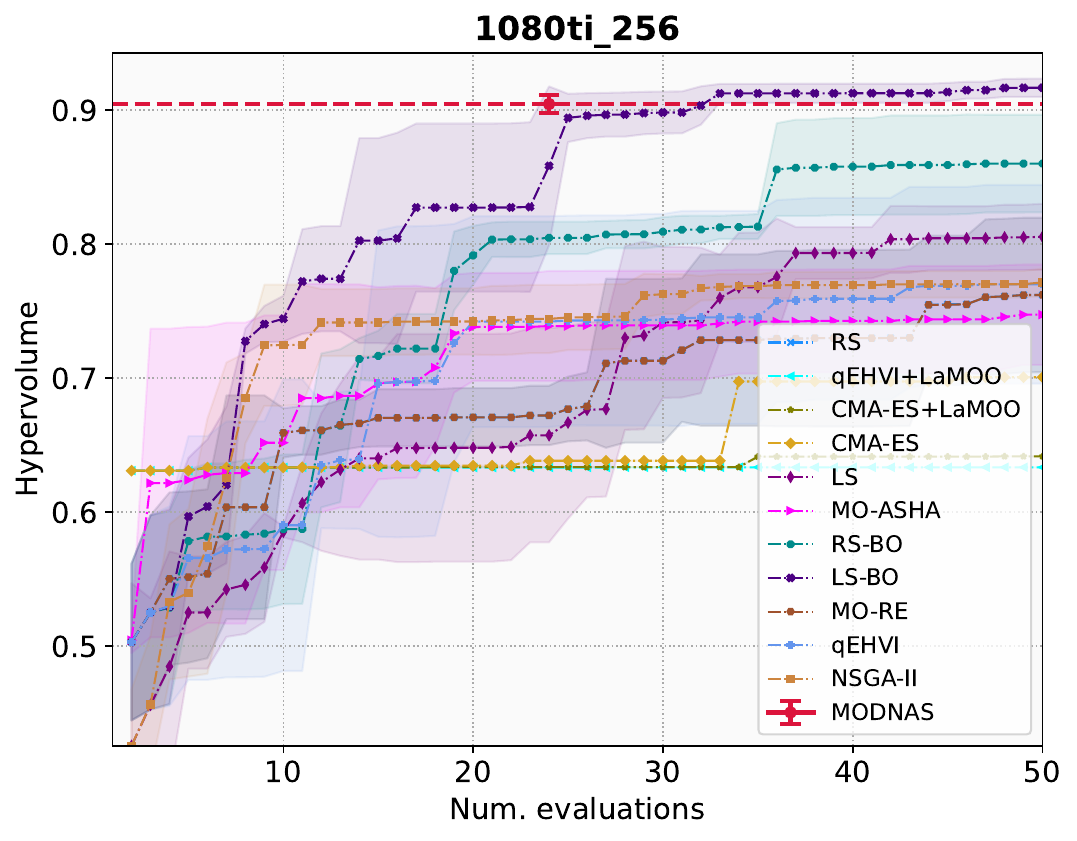}
    \includegraphics[width=.24\linewidth]{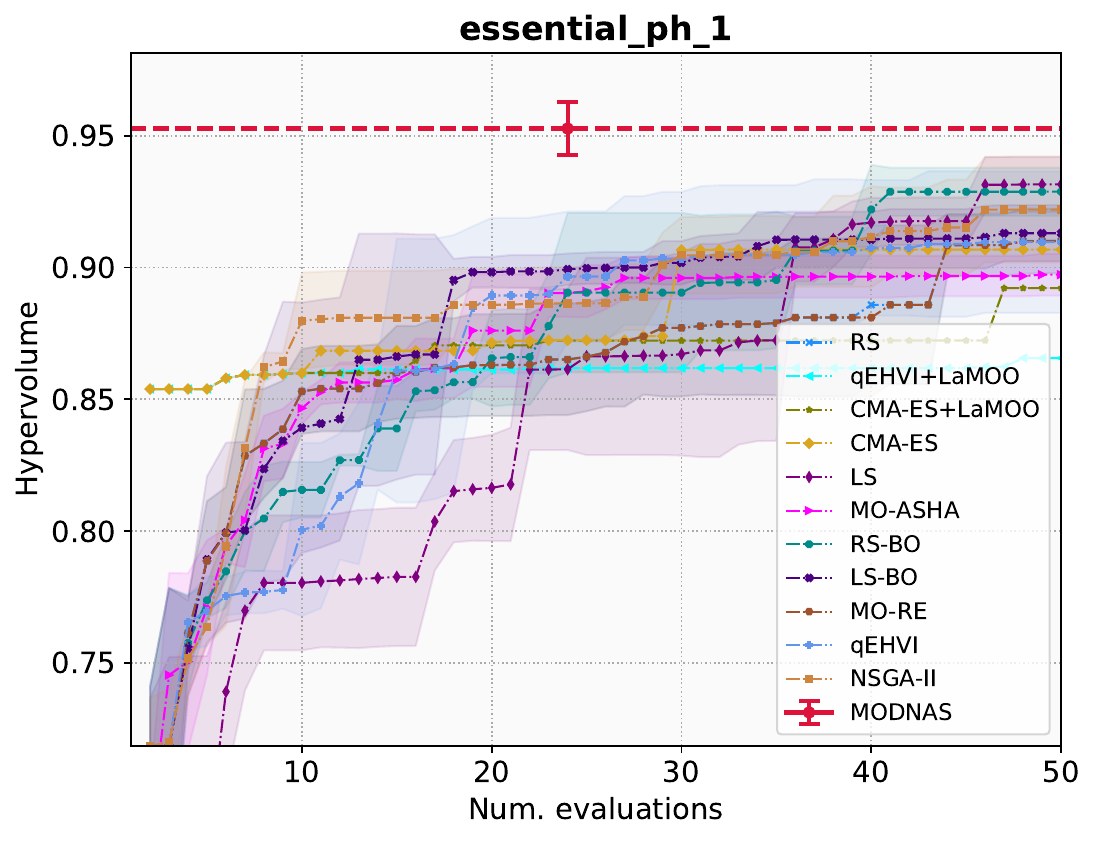}
    \includegraphics[width=.24\linewidth]{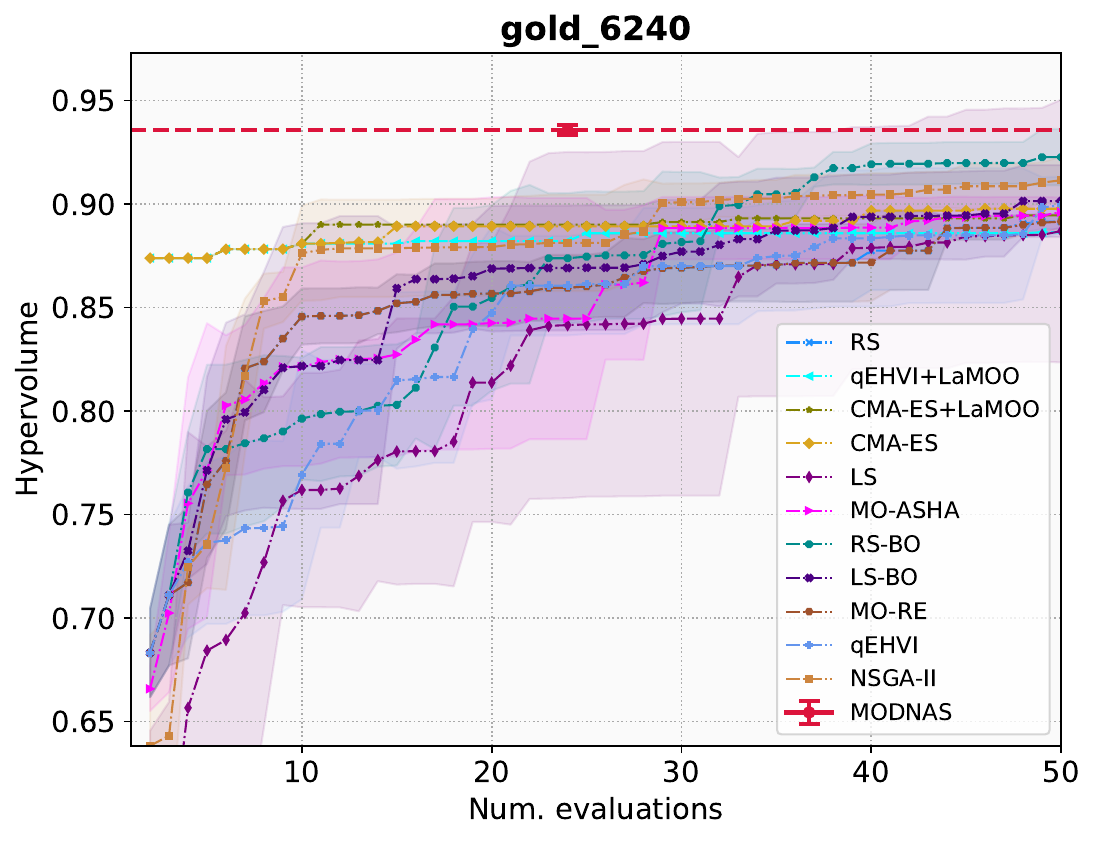}
    \includegraphics[width=.24\linewidth]{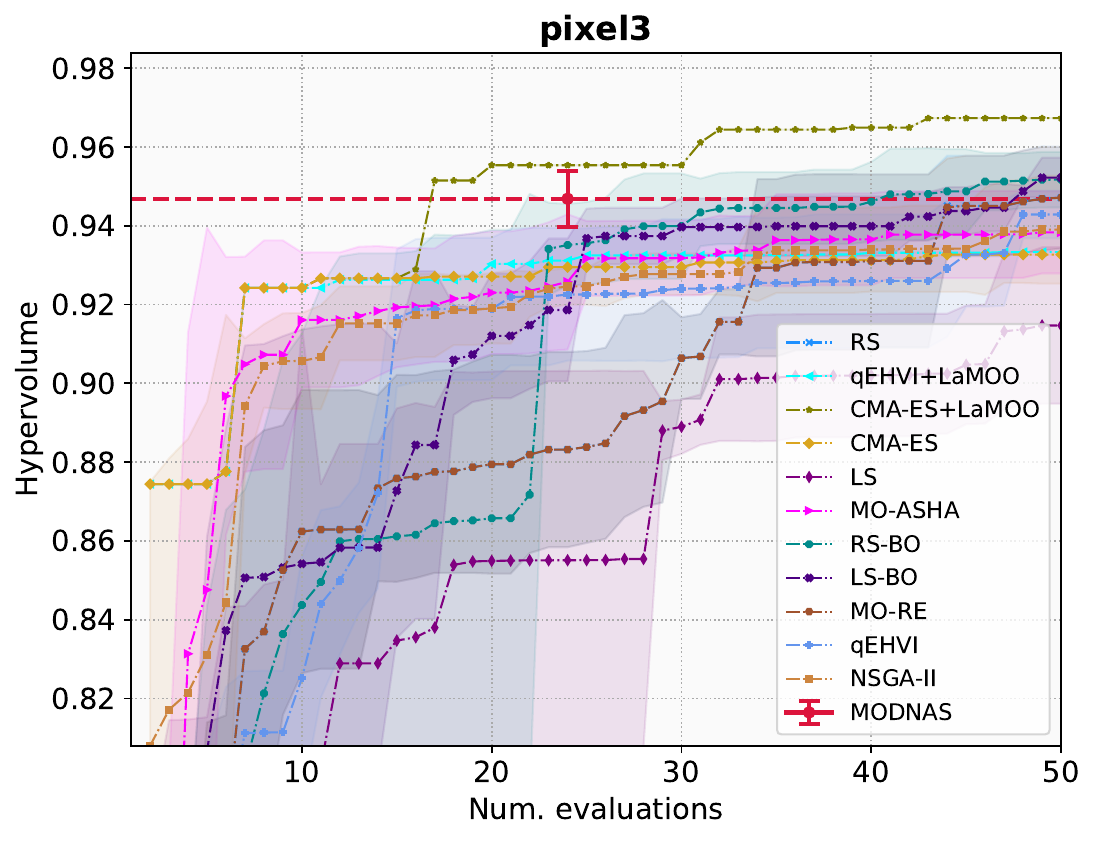} \\
    \includegraphics[width=.24\linewidth]{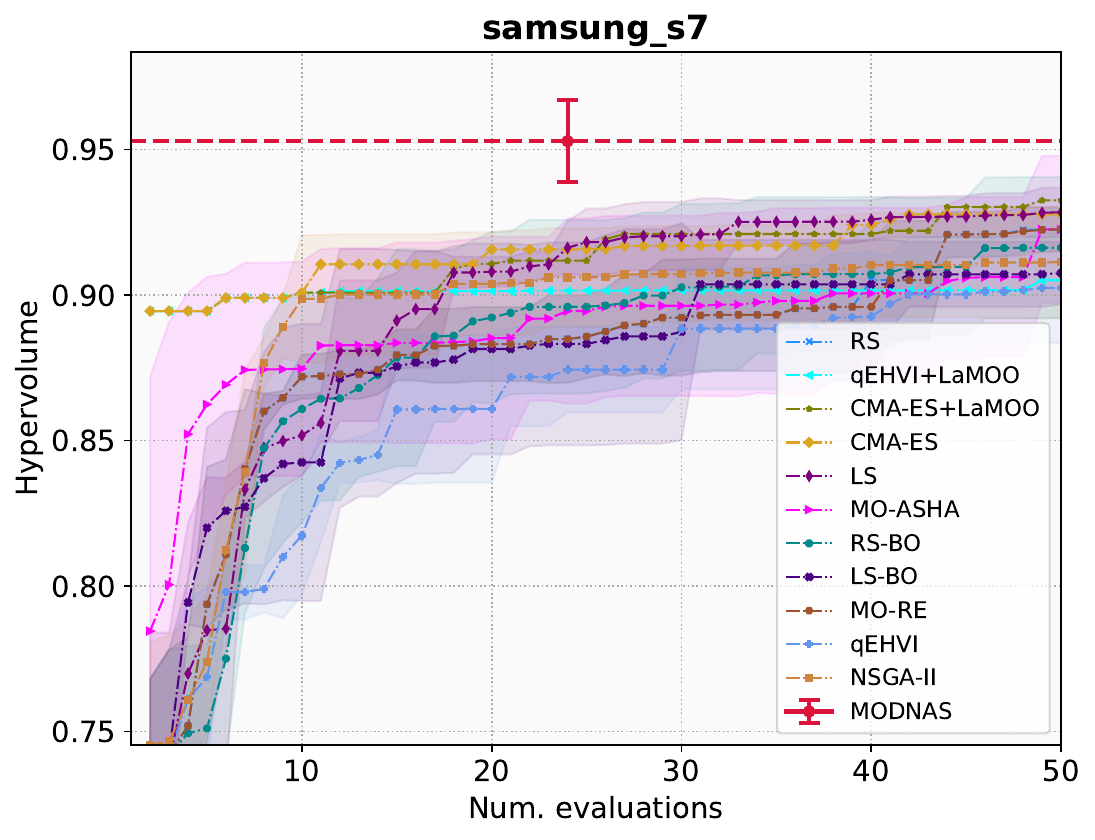}
    \includegraphics[width=.24\linewidth]{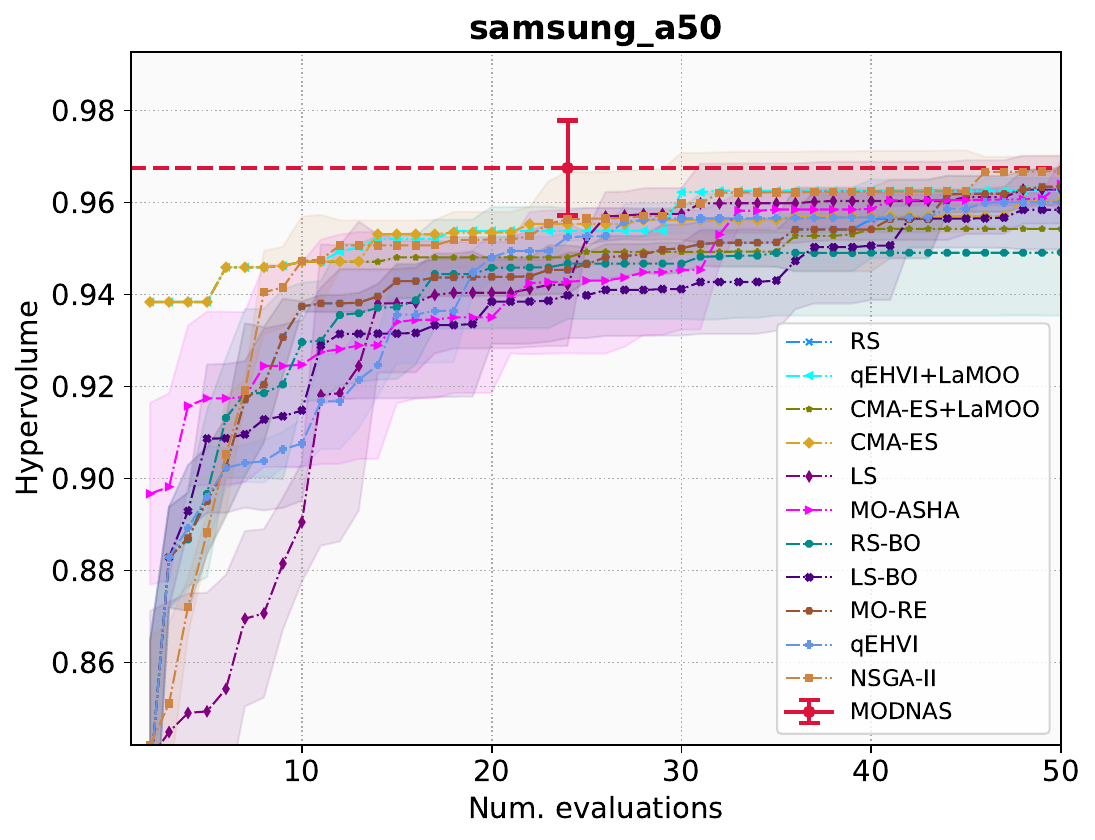}
    \includegraphics[width=.24\linewidth]{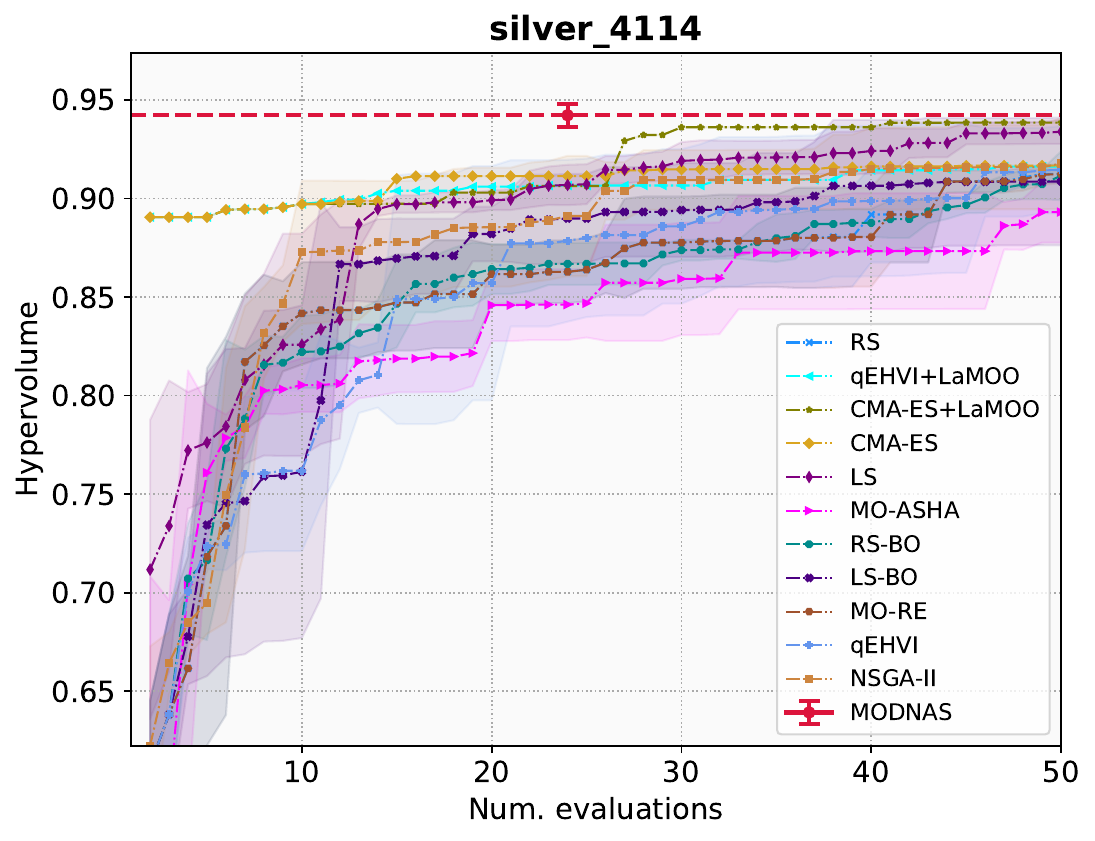}
    \includegraphics[width=.24\linewidth]{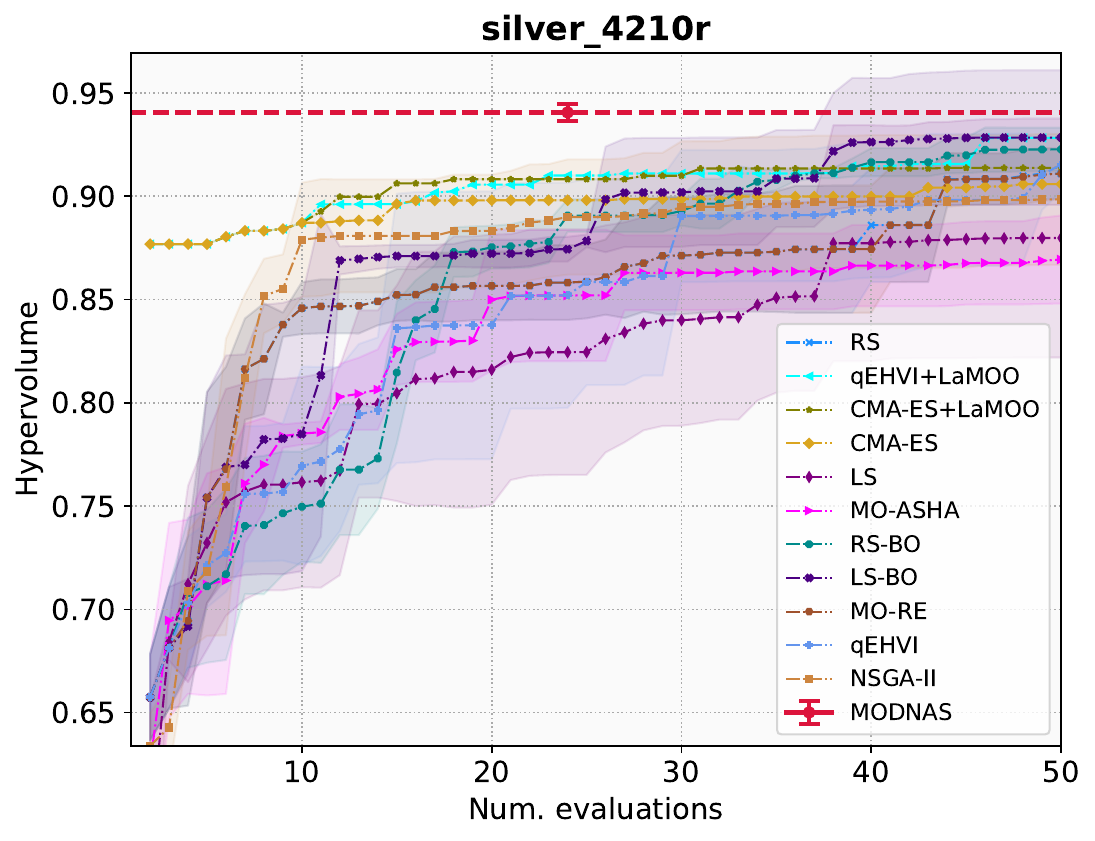}\\
    \includegraphics[width=.24\linewidth]{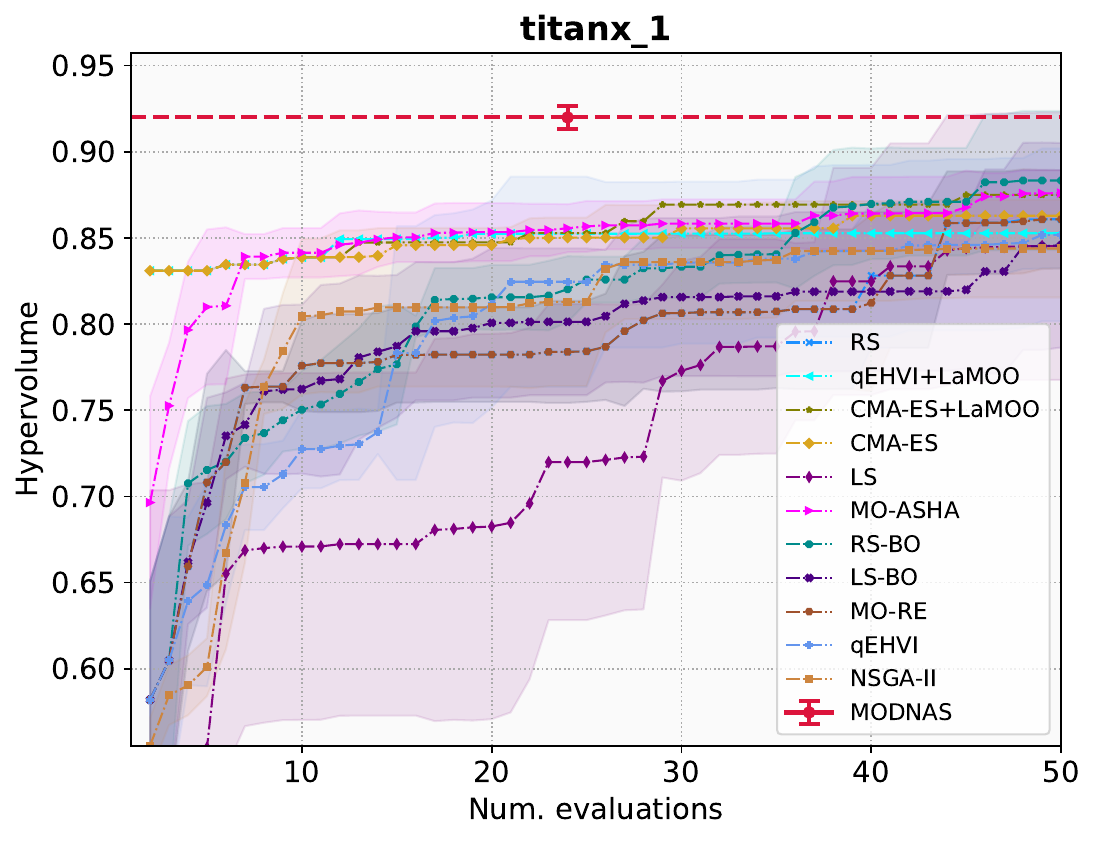}
    \includegraphics[width=.24\linewidth]{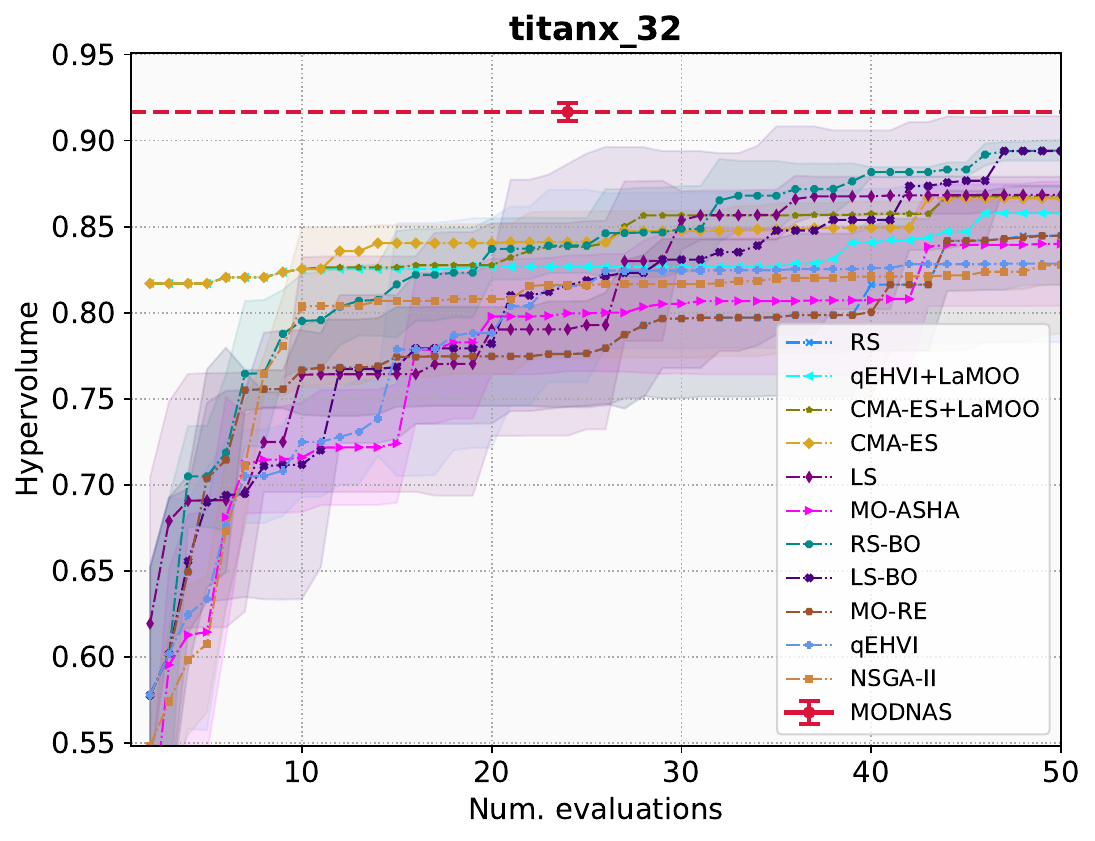}
    \includegraphics[width=.24\linewidth]{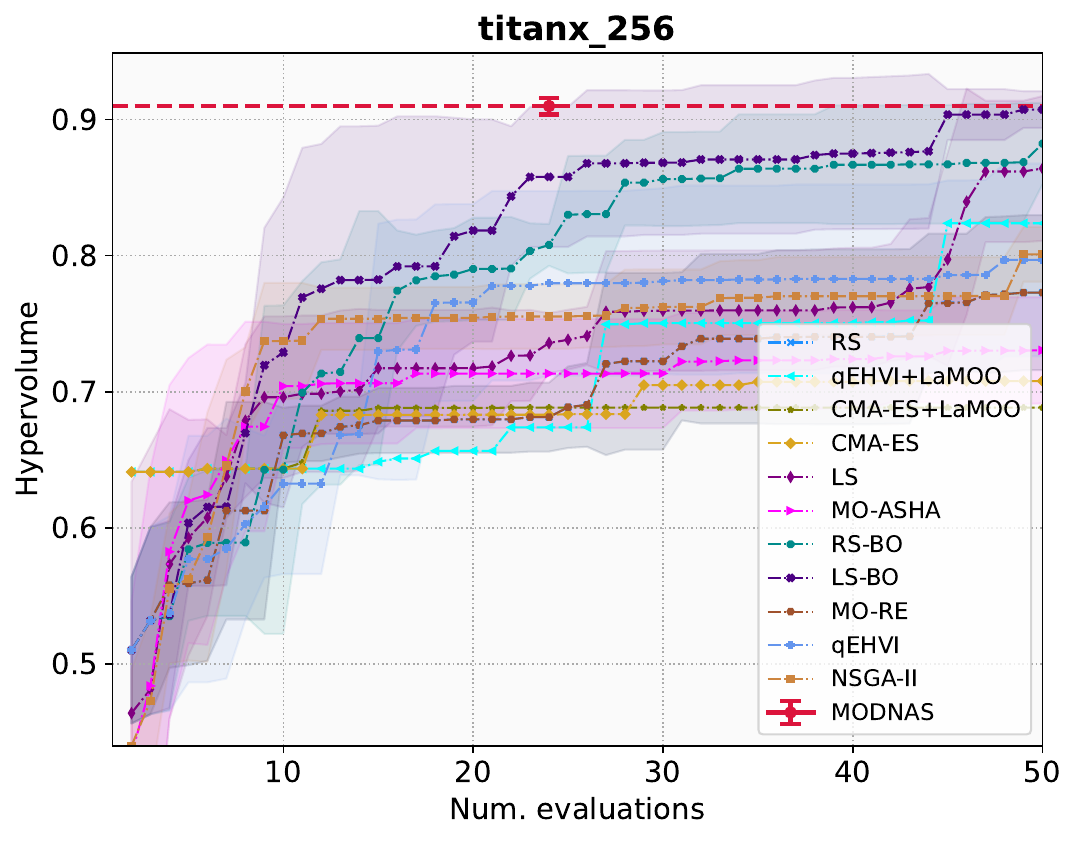}
    \caption{HV over number of evaluated architectures on NAS-Bench-201 of MODNAS and the blackbox MOO baselines. Note that for MODNAS we only have 24 evaluations in the end.}
    \label{fig:hv_budget_all}
\end{figure}

\clearpage
\newpage
\subsubsection{Alignment of Preference Vectors with Pareto Front}

\begin{wrapfigure}[18]{R}{.58\textwidth}
    \vspace{-3ex}
    \centering
    \includegraphics[width=.97\linewidth]{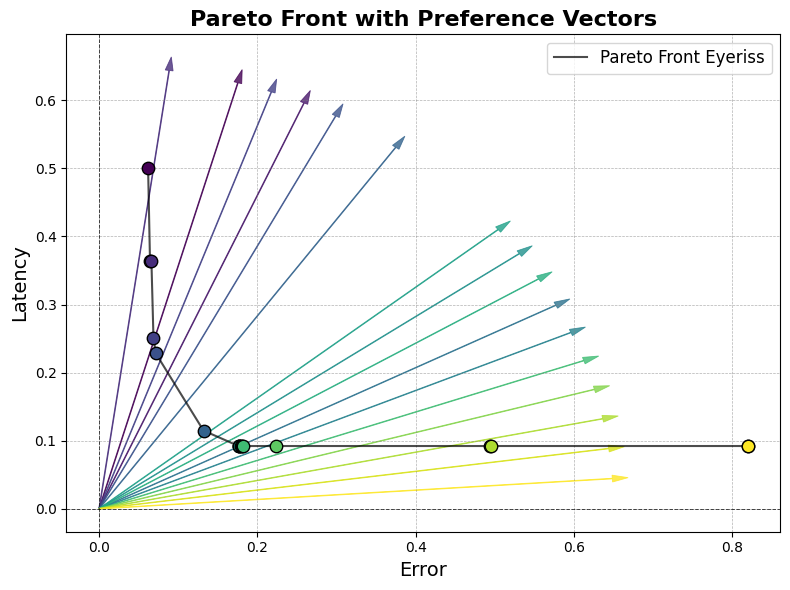}
    \vspace{-1ex}
    \caption{Pareto front and preference vectors on the normalized Eyeriss latency and test error of NAS-Bench-201.} 
    \label{fig:rays_eyeriss}
\end{wrapfigure}

In this section, we provide empirical evidence that the solutions generated using the \metahypernet align well with the preference vectors. To this end, we utilize one of our runs on the NAS-Bench-201 test devices, namely Eyeriss. In Figure~\ref{fig:rays_eyeriss}, we show the Pareto front of the normalized test error and latency on Eyeriss. Note that of the 24 sampled preference vectors, 17 generate solutions that are in the Pareto set. Each point in the Pareto front with a certain color corresponds to the preference vector with the same color. In the figure, there are actually 17 points in the Pareto front; however, some of them are really close to each other or are the same, since the function mapping preference vectors to architectures is a many-to-one function. Nevertheless, we can visually notice that the preference vectors starting from the origin align very well with the generated solutions. The missing vectors are mainly in the center, where there are not many solutions available for this particular device.

\subsubsection{Training and Validation Loss Curves}

\begin{figure}[ht]
    \centering
    \begin{minipage}{0.5\linewidth}
      \centering
      \includegraphics[width=.95\linewidth]{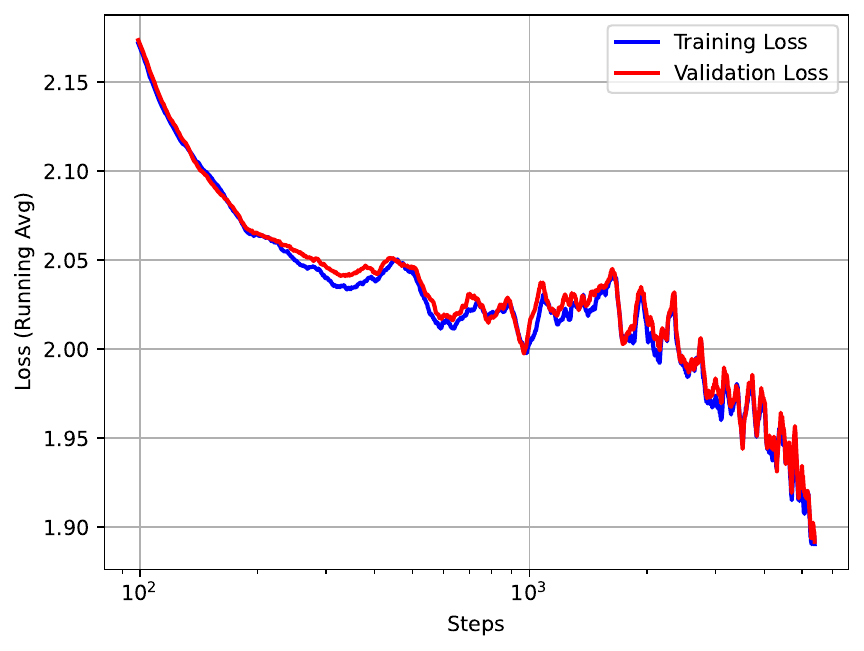}
    \end{minipage}%
    \begin{minipage}{0.47\linewidth}
      \centering
      \vspace{-2ex}
      \includegraphics[width=.9\linewidth]{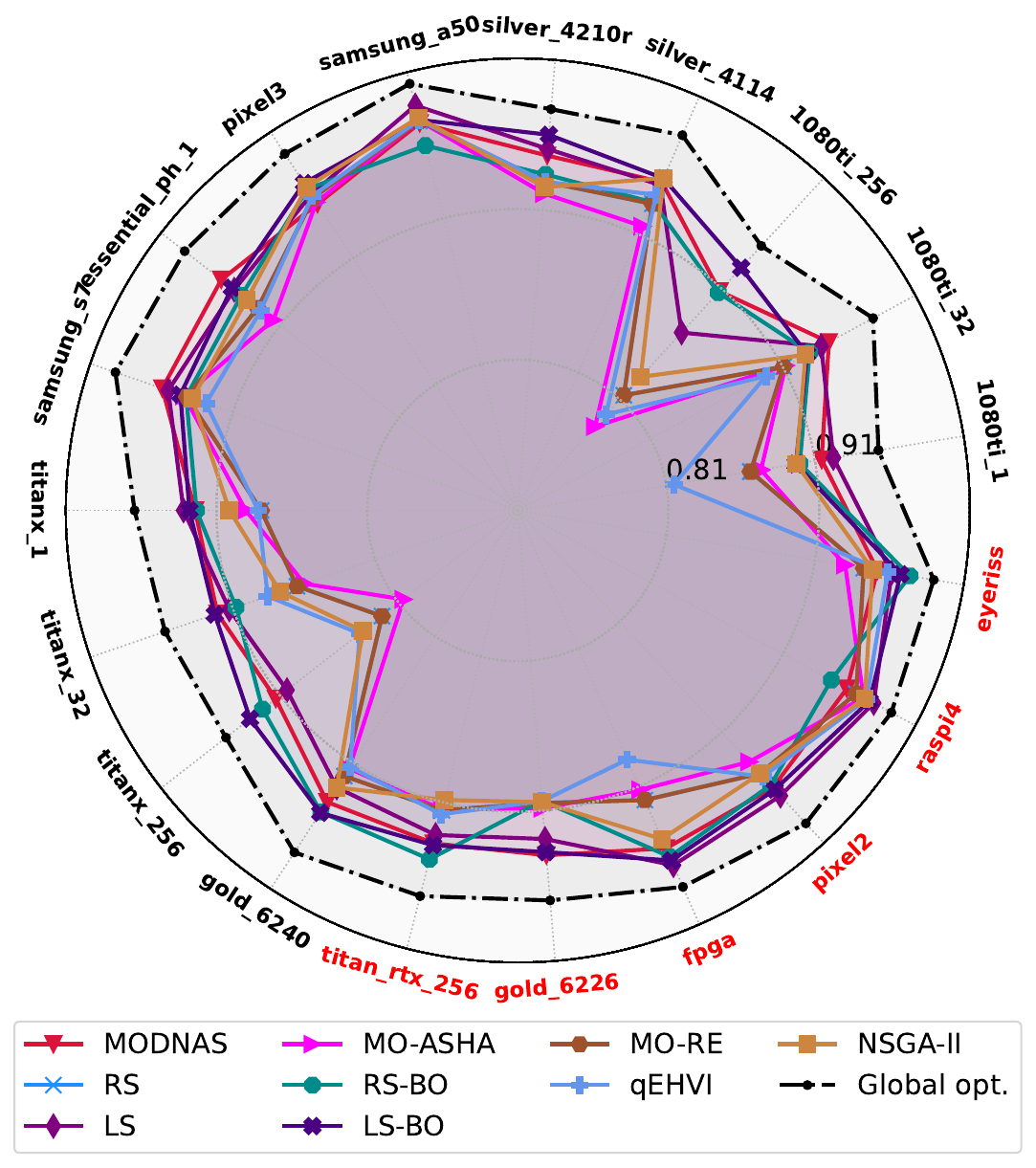}
    \end{minipage}%
    \caption{(\textit{left}) Average training and validation cross-entropy loss across devices during the MODNAS search on NAS-Bench-201. (\textit{right}) HV of MODNAS and baselines across 19 devices on NAS-Bench-201. For every device we optimize for 2 objectives, namely \textit{latency (ms)} and \textit{test accuracy} on CIFAR-10. For method, metric and device we report the mean of 3 independent search runs. Higher area in the radar indicates better performance for every metric. Test devices are colored in red around the radar plot. Here we allocate 4 times the budget to baselines, i.e. we run all baselines for 100 function evaluations.}
    \label{fig:radar_hv_budget_loss}
\end{figure}

In addition to the hypervolume indicator, in this section, we provide the training and validation loss curves in Figure~\ref{fig:radar_hv_budget_loss} (\textit{left}). At each mini-batch iteration we plot the average cross entropy loss across all devices. As expected both training and validation cross-entropy go down and we do not notice any overfitting. The high noise is common for sample-based NAS optimizers, since a sampled different architecture is activate at each mini-batch iteration. In the plot, for visualization purposes, we have used a running average with a window size of 100 to smooth out the noise.

\subsubsection{Multi-objective Optimization Baselines with More Budget}

Black-box multi-objective optimizers can potentially reach the global Pareto front if the compute resources are not a concern and given enough time. However, it is not practical to train or even evaluate these architectures, especially for larger model sizes (e.g. Transformer spaces from HW-GPT-Bench). Sometimes in practice, the user wants to get a quick estimation of the Pareto front instead the global optimum, and this is the use-case where MODNAS shines. Given enough budget, even a random search (RS) will find a near-optimal solution. For example, in NAS-Bench-201, the size of the search space is $K=15625$ architectures. The optimal theoretical number of RS steps $n$ to achieve a success probability $\alpha$ is approximately: $n \geq K ln(1/1-\alpha)$, therefore, for random search to have a success probability higher than 0.5 it requires $n \geq 10781$ iterations in theory. For the other guided search methods, this number is even smaller, though similar to MODNAS, they have the same limitation that they can converge to a local minimum. We conducted the same experiment as the one in Figure~\ref{fig:radar_plot_hv}, but this time with baselines given 4 times more budget than MODNAS. We show the result in Figure~\ref{fig:radar_hv_budget_loss} (\textit{right}). As we can see, some of the methods such as LS-BO can reach results closer to the global Pareto front compared to MODNAS.

\subsection{Additional Results on Hardware-aware Transformers (En-De)}
\begin{figure}[ht]
    \centering
    \includegraphics[width=.3\linewidth]{figures/hat_en_de/pareto_en_de_cpu_raspberrypi_bleu.pdf}
    \includegraphics[width=.3\linewidth]{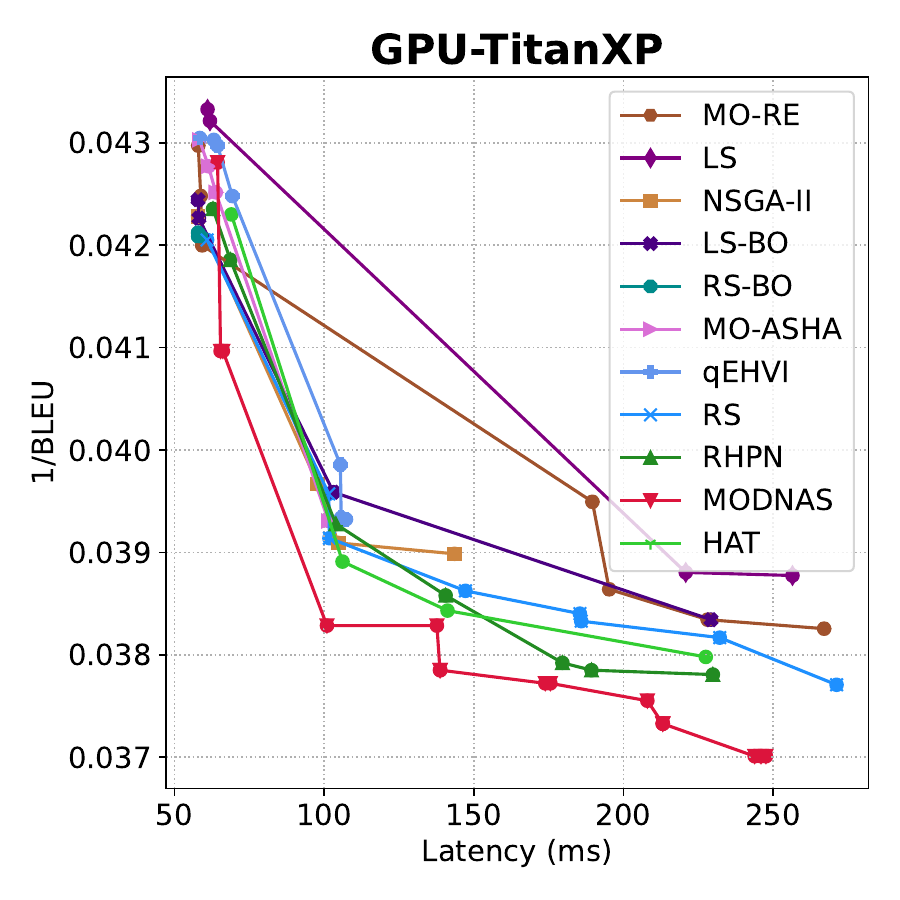}
    \includegraphics[width=.3\linewidth]{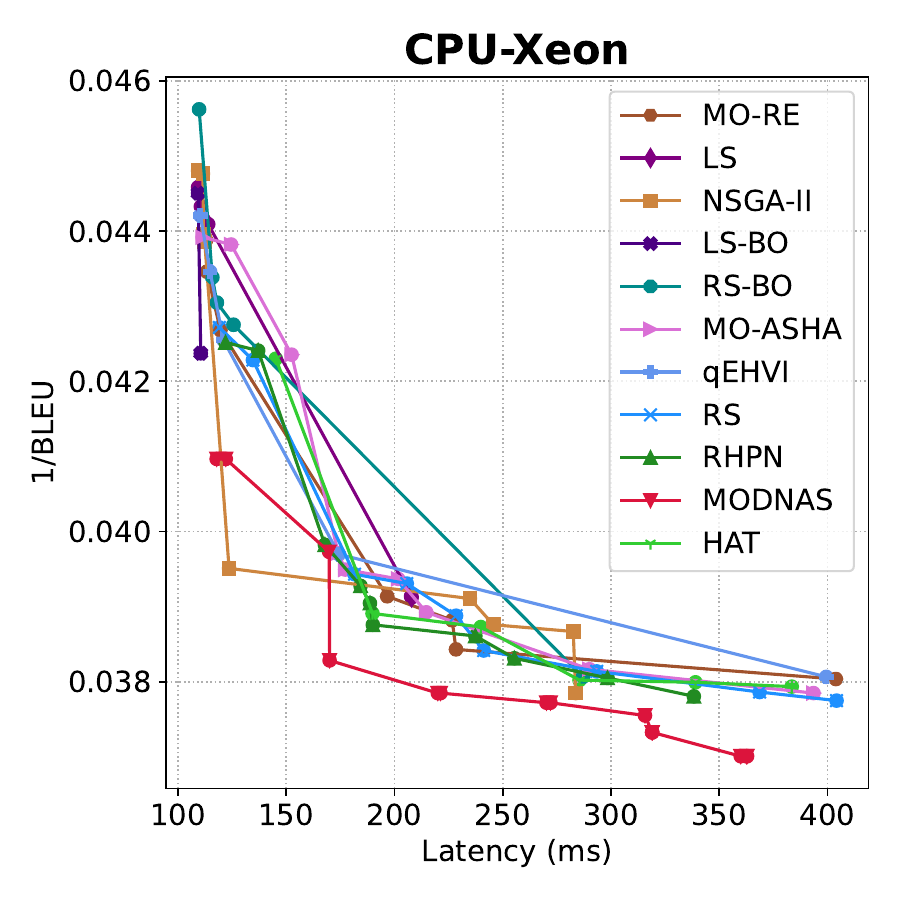}
    \includegraphics[width=.3\linewidth]{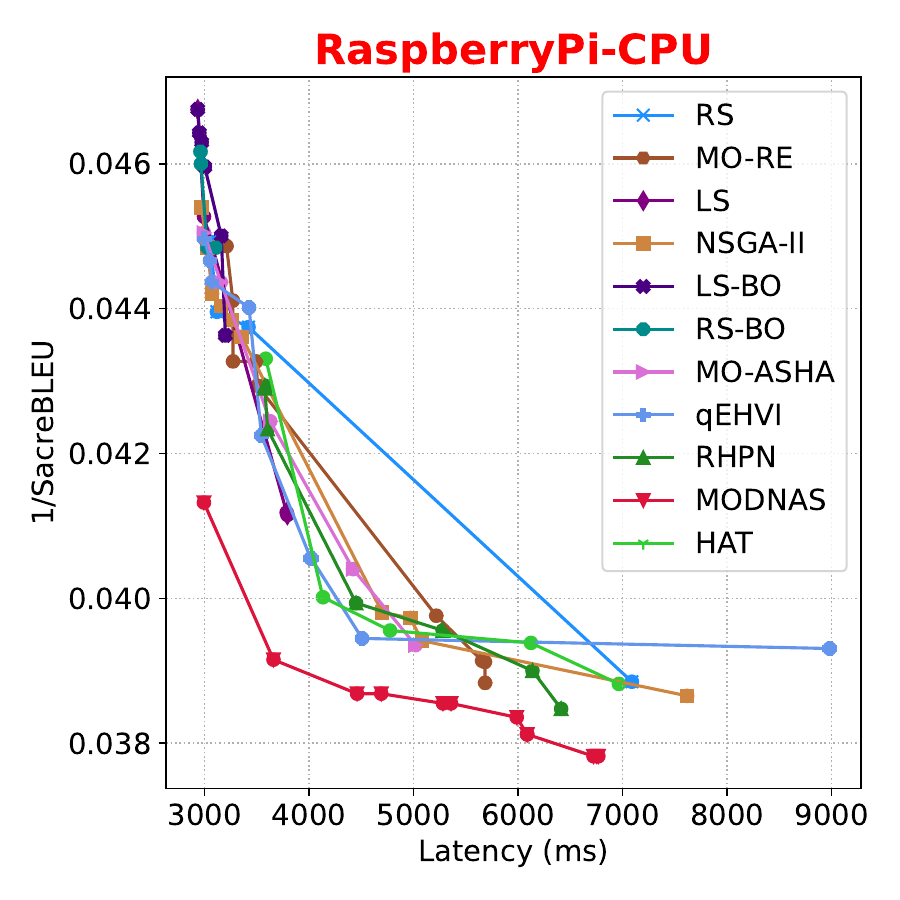}
    \includegraphics[width=.3\linewidth]{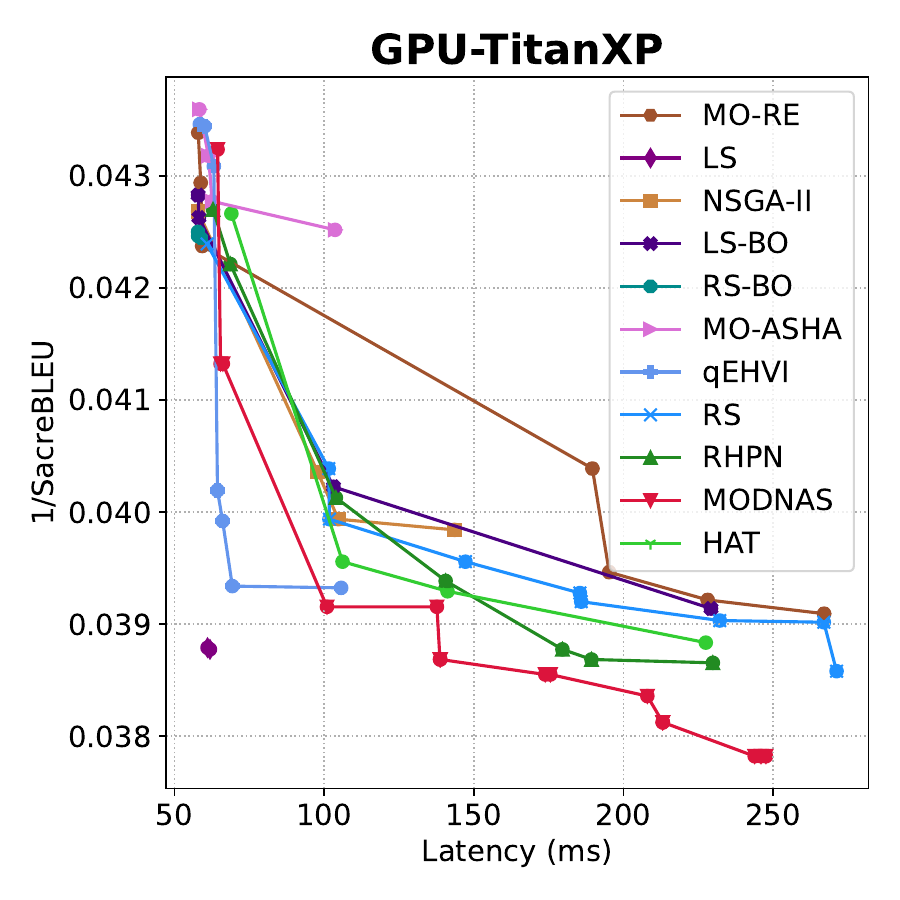}
    \includegraphics[width=.3\linewidth]{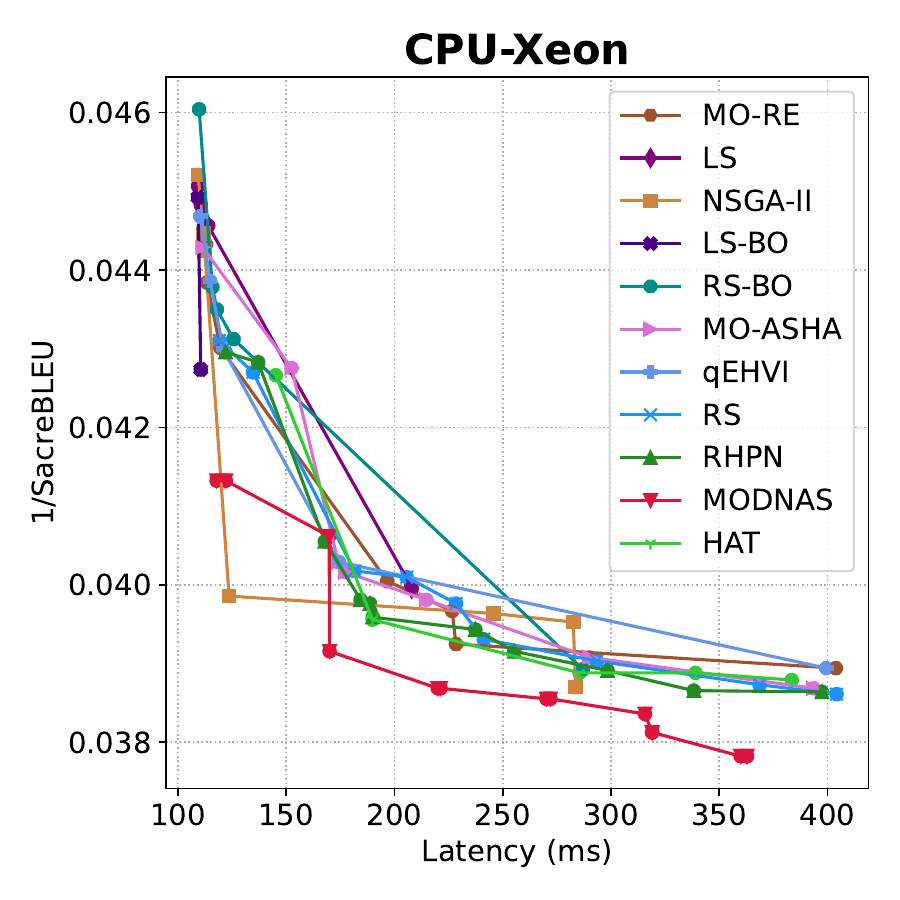}
    \caption{Pareto fronts of MODNAS and baselines on the HAT space for the WMT' En-De task. All performance metrics are obtained from the inherited supernet weights.}
    \label{fig:pareto_hat_full}
\end{figure}

\begin{figure}[ht]
    \centering
    \includegraphics[width=.32\linewidth]{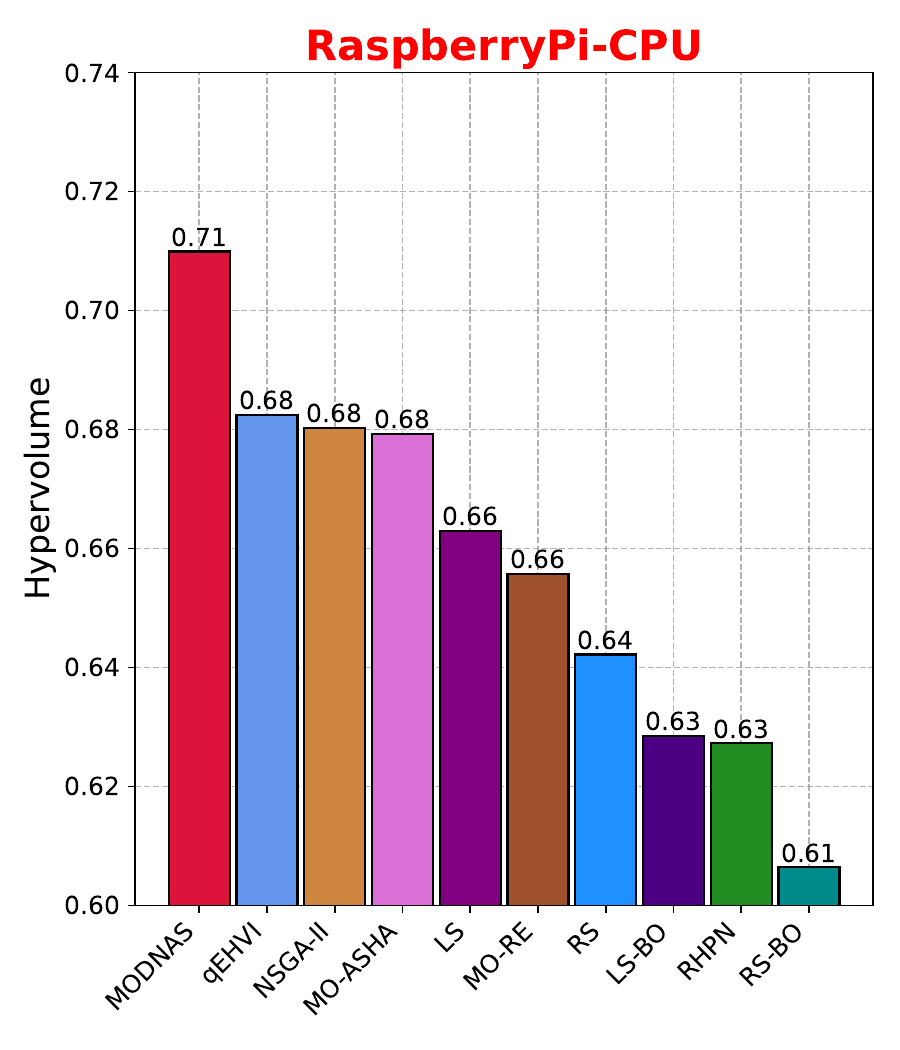}
    \includegraphics[width=.32\linewidth]{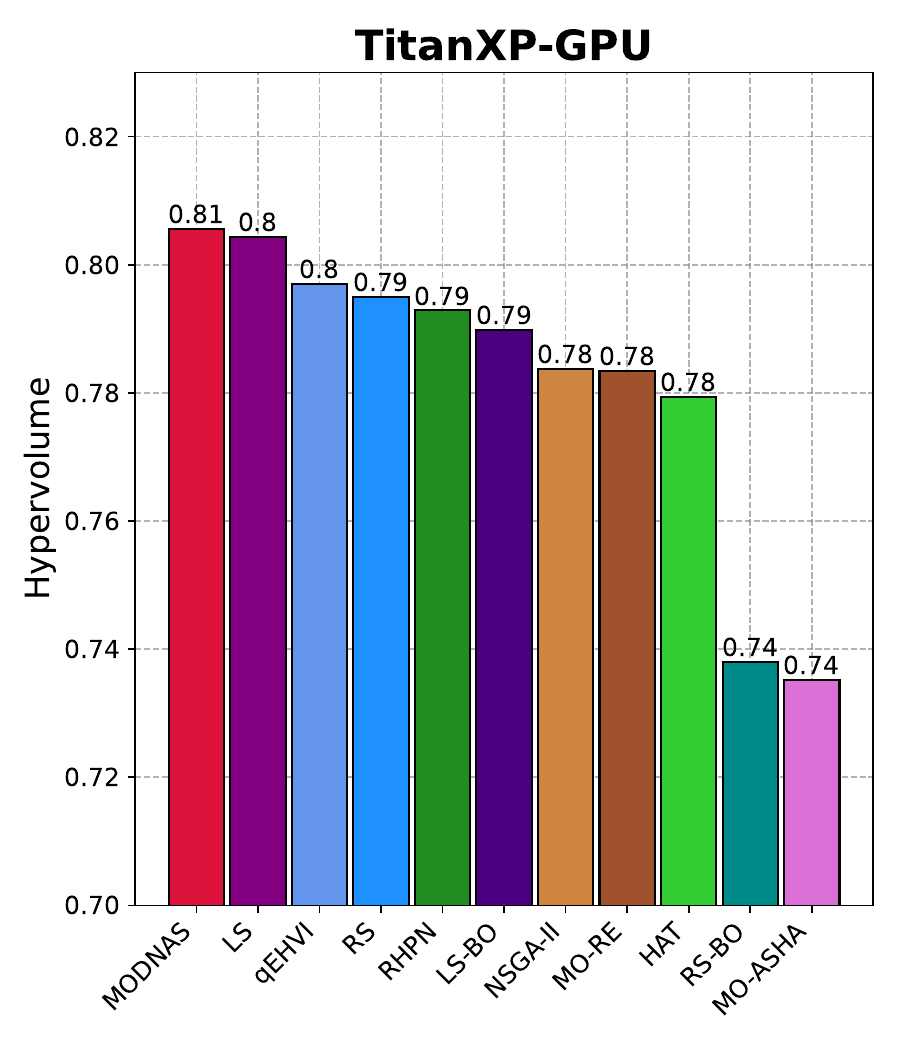}
    \includegraphics[width=.32\linewidth]{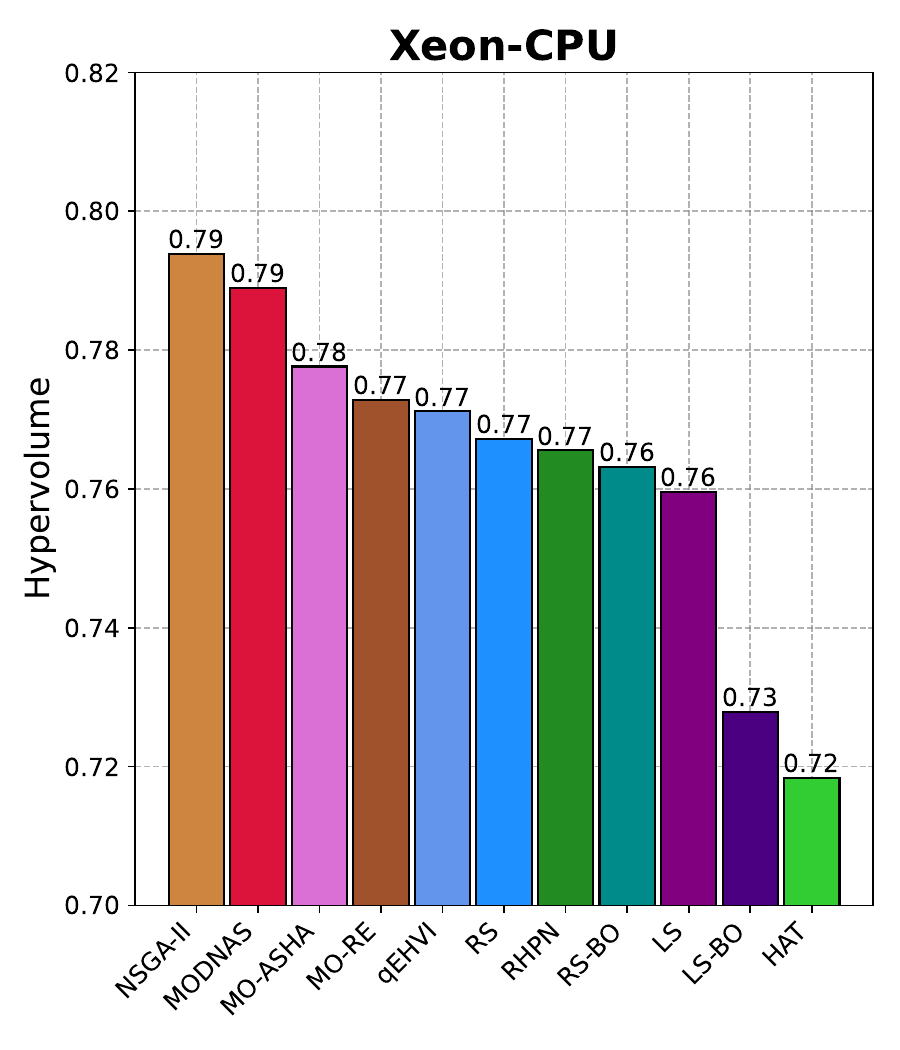}
    \caption{Hypervolume (HV) of MODNAS and baselines across devices on the HAT space. The objectives used to compute the HV are latency and SacreBLEU score. Leftmost plot is for the test device. MODNAS is the best or on par to the baselines across all three devices.}
    \label{fig:hat_barplot_sacre}
\end{figure}

We show the Pareto fronts of \methodname compared to baselines for the Transformer space in Figure~\ref{fig:pareto_hat_full}, as well as their comparison with respect to hypervolume for the SacreBLEU metric in Figure~\ref{fig:hat_barplot_sacre}. These results demonstrate the superior performance of our method compared to the other baselines on this benchmark. All evaluations are done by inheriting the weights of a pretrained supernet.

\begin{figure}[ht]
    \centering
    \includegraphics[width=.32\linewidth]{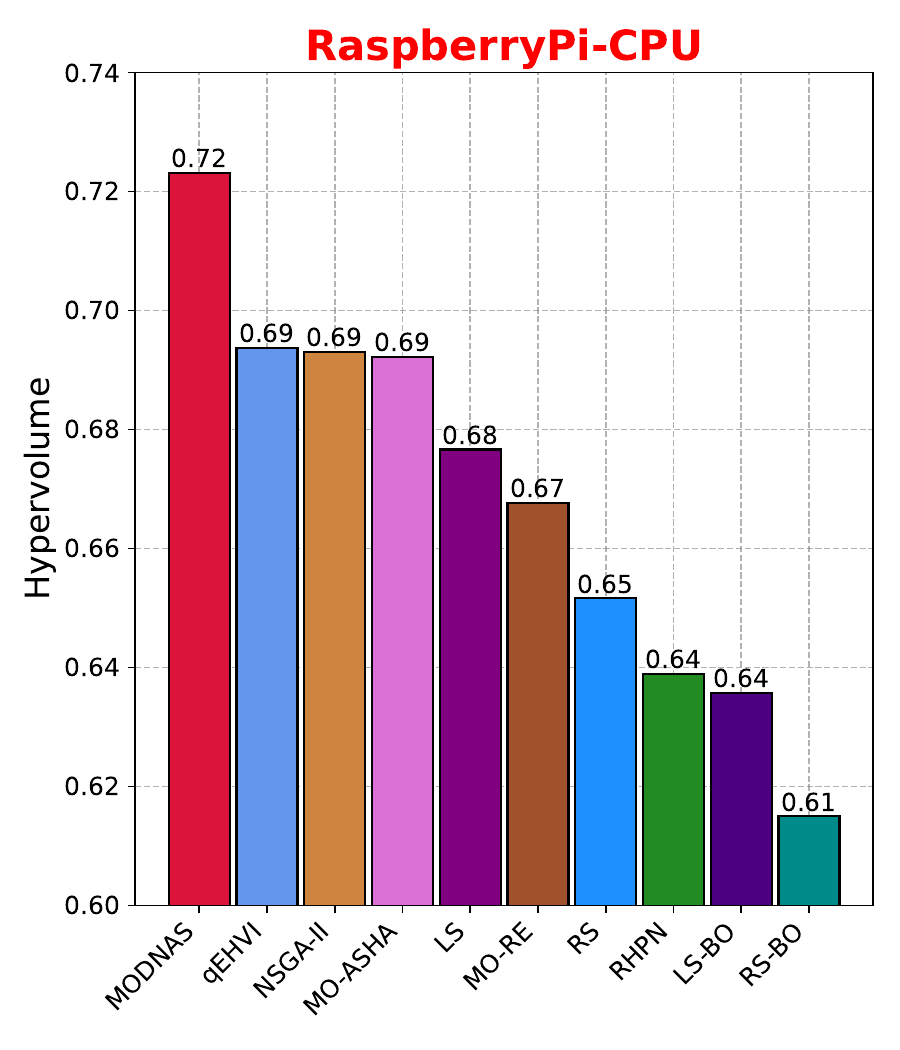}
    \includegraphics[width=.32\linewidth]{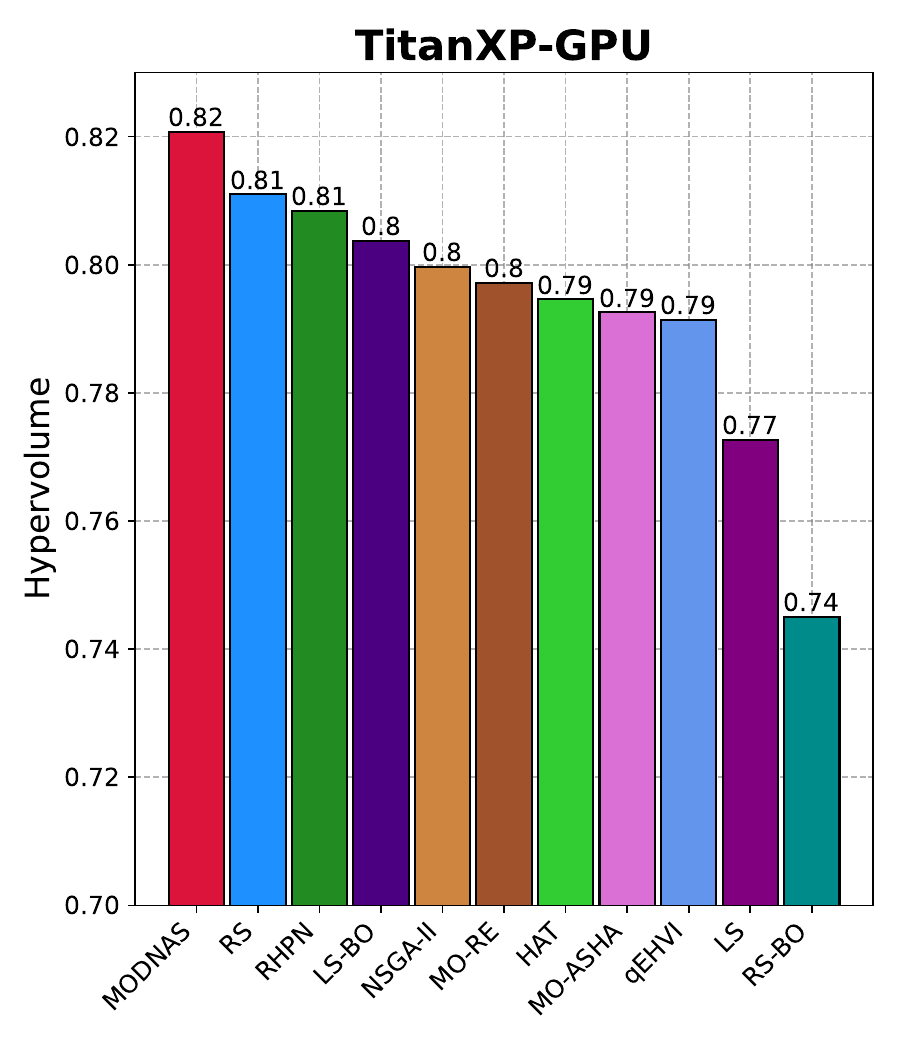}
    \includegraphics[width=.32\linewidth]{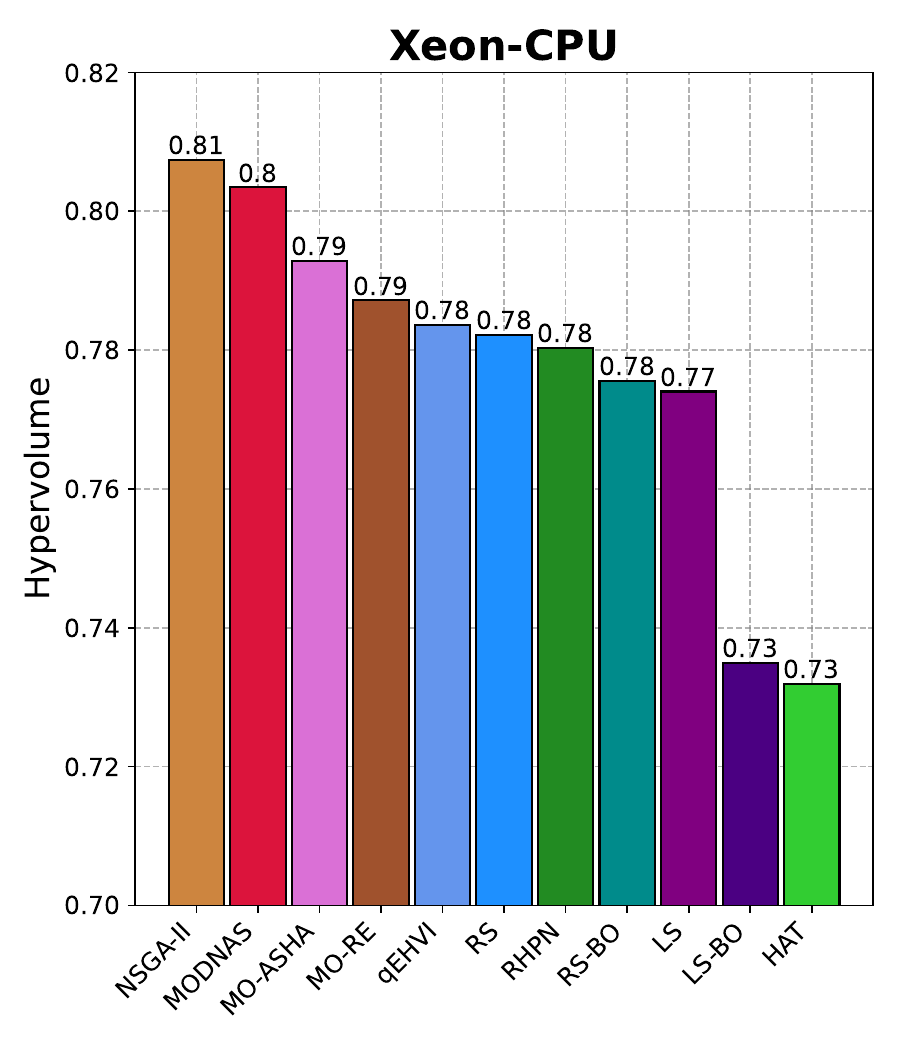}
    \vspace{-1ex}
    \caption{Hypervolume (HV) of MODNAS and baselines across devices on the HAT space. The objectives used to compute the HV are latency and BLEU score. Leftmost plot is for the test device.}
    \label{fig:hat_barplot}
\end{figure}

\subsection{Additional Results on the HW-GPT space}
\begin{figure}[ht]
    \centering
    \includegraphics[width=.24\linewidth]{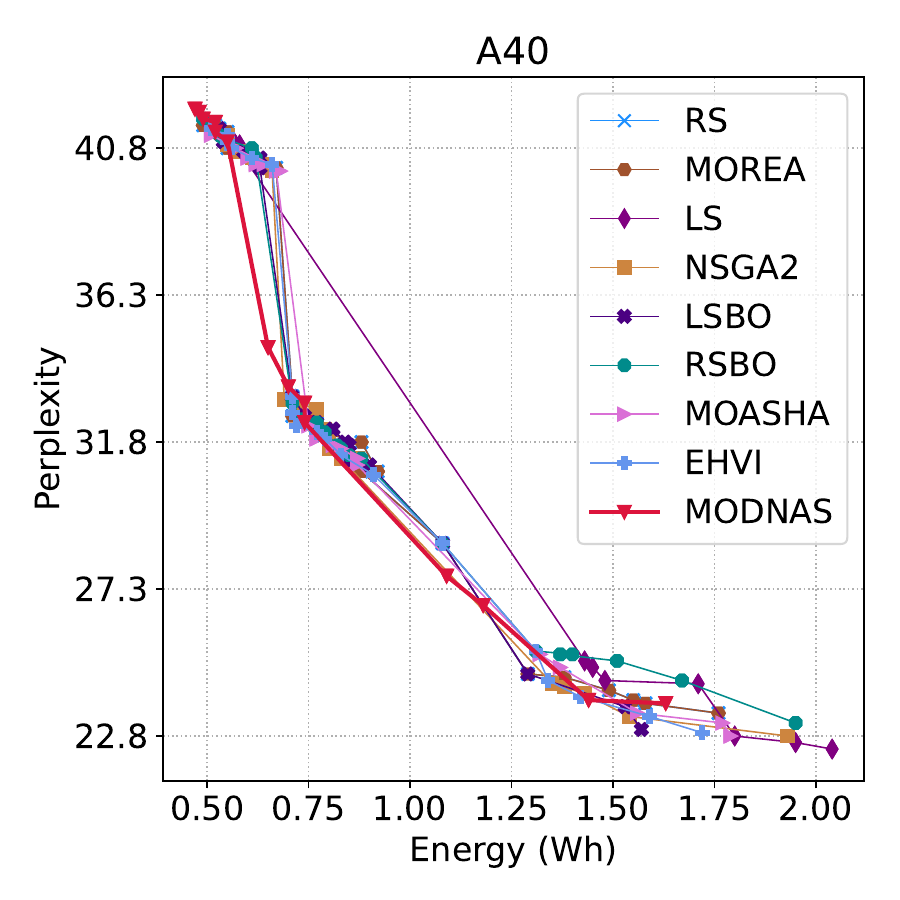}
    \includegraphics[width=.24\linewidth]{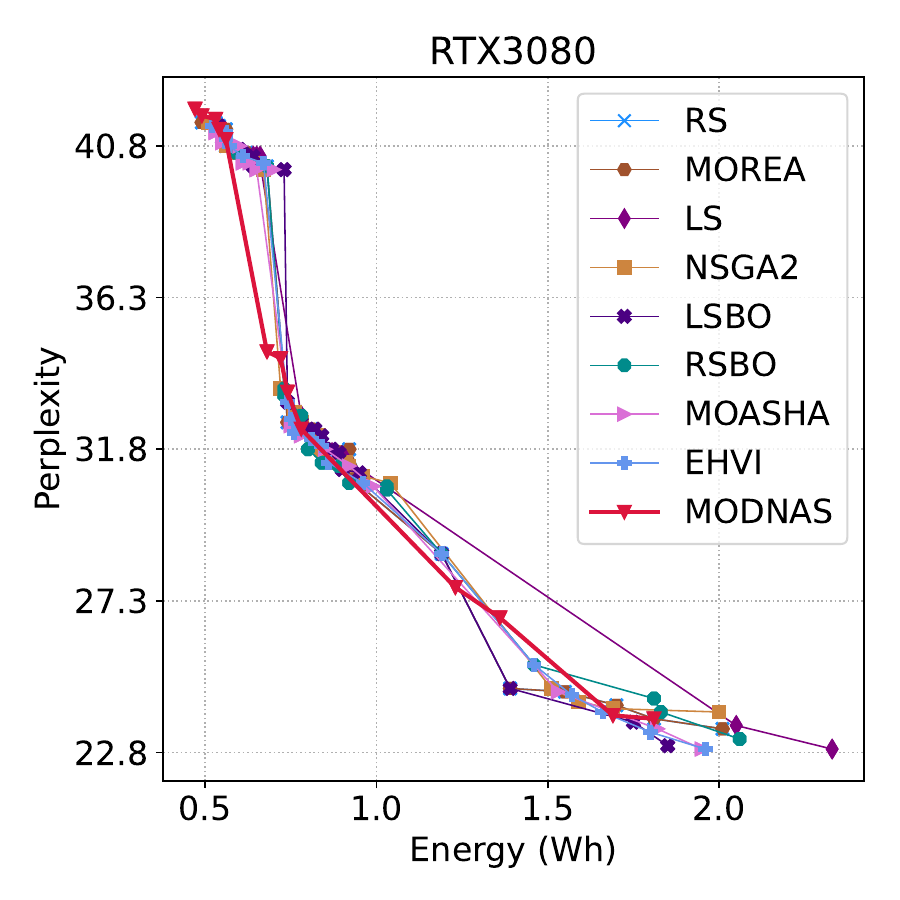}
    \includegraphics[width=.24\linewidth]{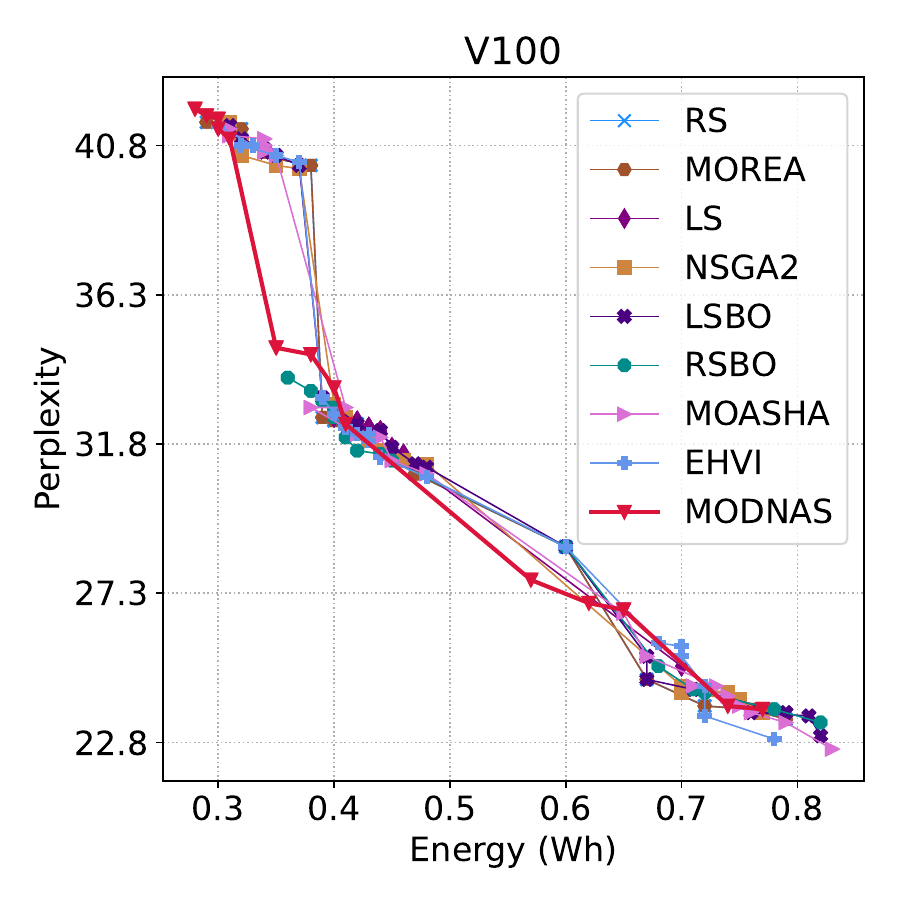}
    \includegraphics[width=.24\linewidth]{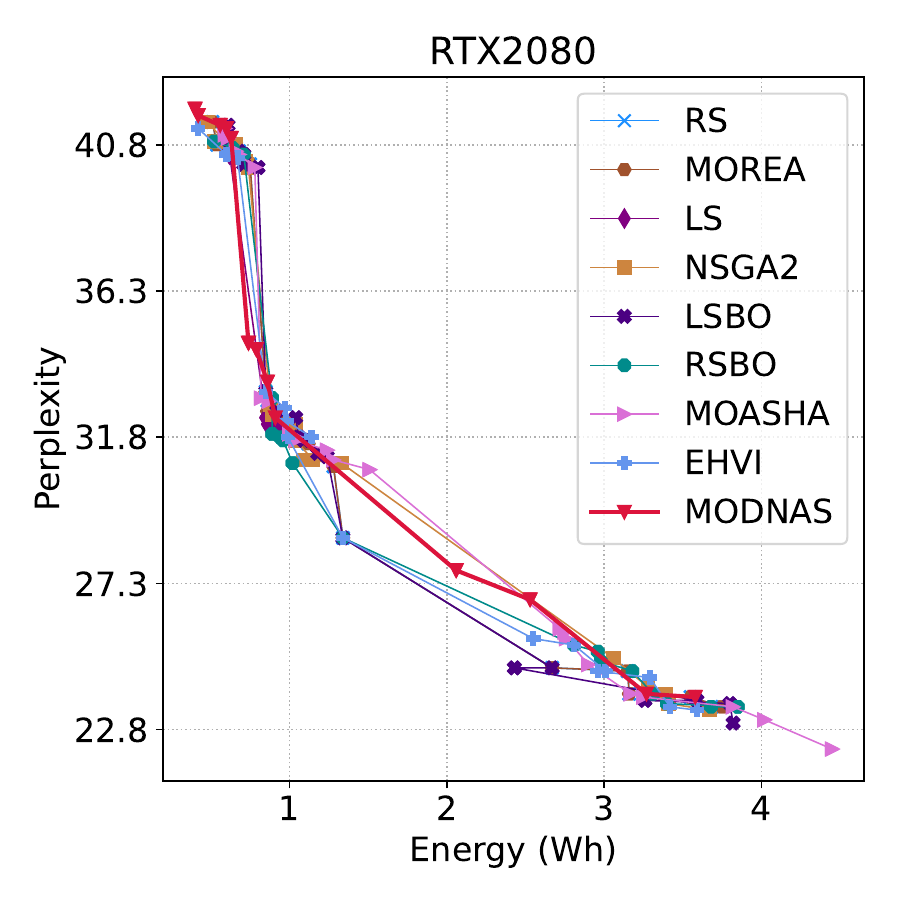}\\
    \includegraphics[width=.24\linewidth]{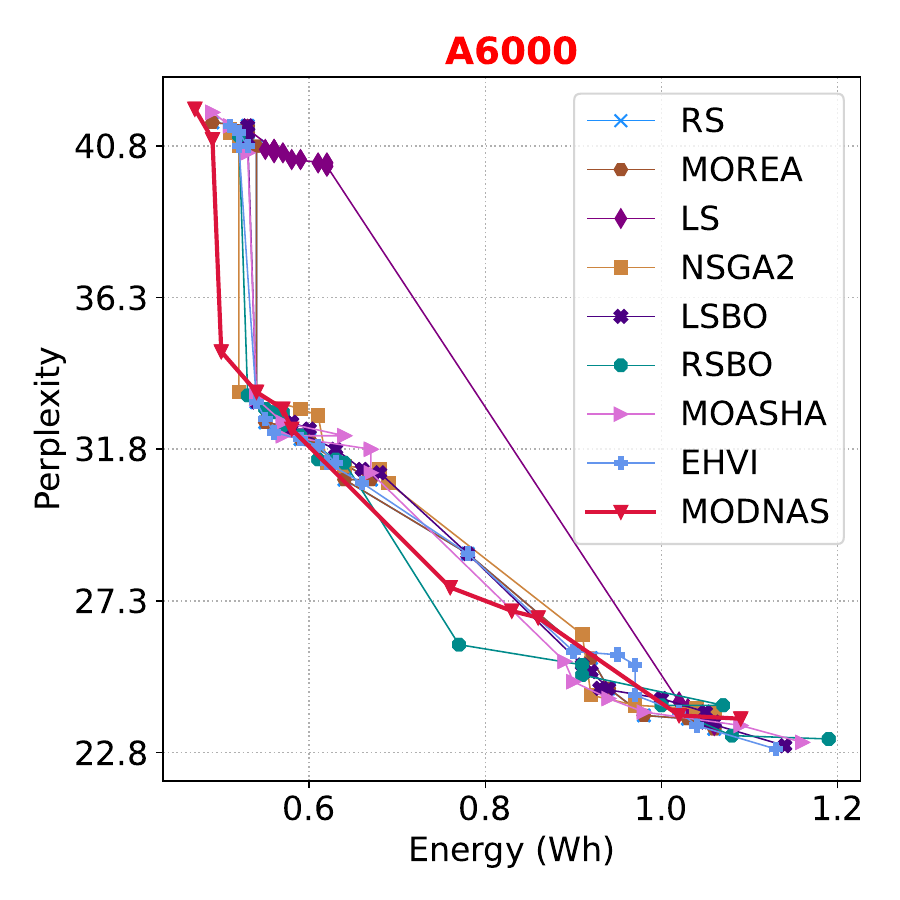}
    \includegraphics[width=.24\linewidth]{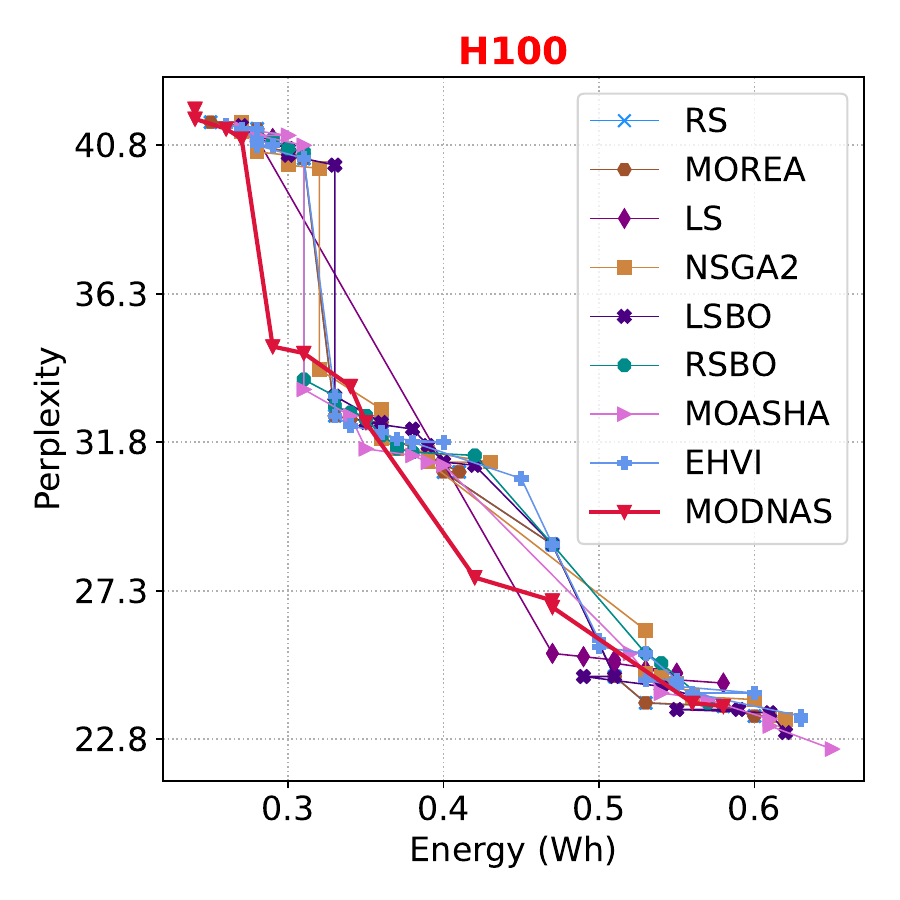}
    \includegraphics[width=.24\linewidth]{figures/paretos_gpt/pareto_a100.pdf}
    \includegraphics[width=.24\linewidth]{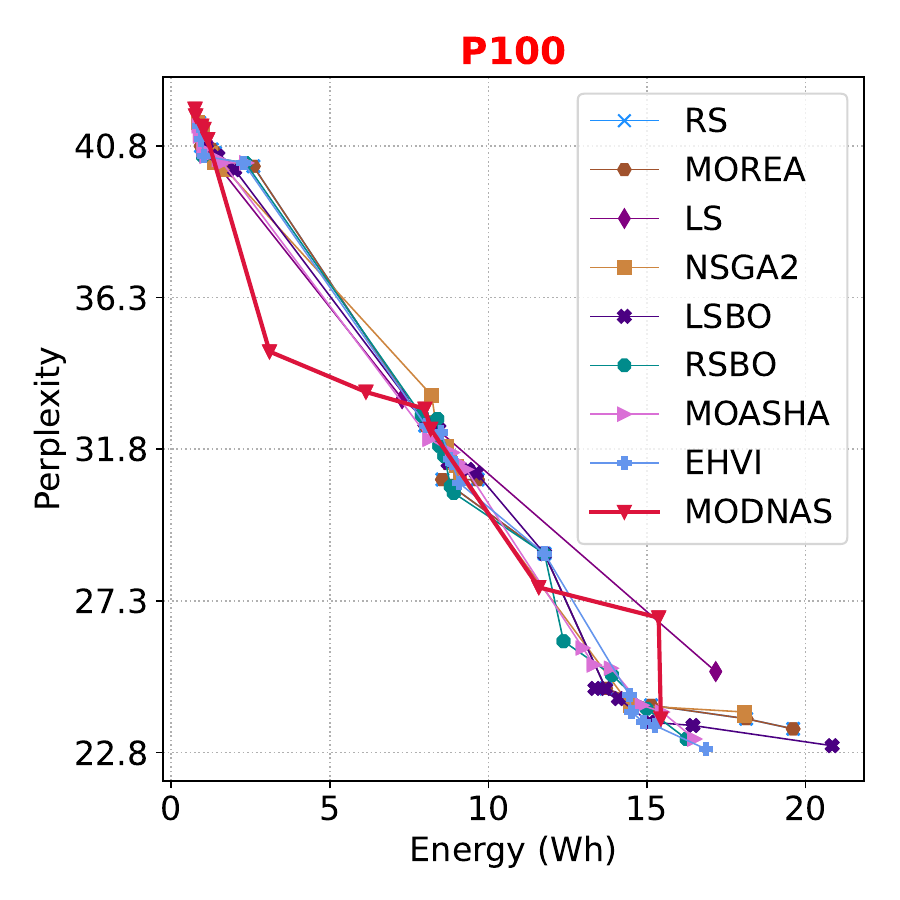}\\
    \caption{Pareto fronts of MODNAS and baselines optimizing for GPU energy consumption (Wh) and perplexity on the HW-GPT-Bench space.}
    \label{fig:pareto_gpt_full}
\end{figure}

In figure \ref{fig:pareto_gpt_full}, we present the Pareto fronts on all the 8 GPU types for MODNAS and different baselines. The Pareto fronts are obtained using the perplexity and energy predictors trained on data collected in the HW-GPT-Bench \citep{Sukthanker2024HWGPTBench}. 

\clearpage
\subsubsection{Experiments on Perplexity and Memory usage Objectives}

\begin{wrapfigure}[14]{R}{.4\textwidth}
    \vspace{-3.5ex}
    \centering
    \includegraphics[width=.9\linewidth]{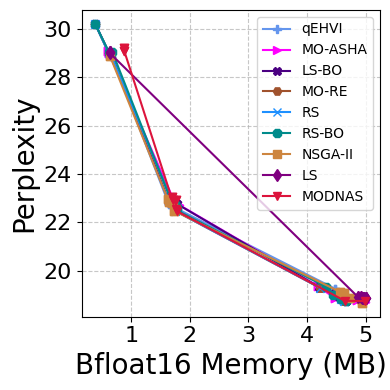}
    \vspace{-2ex}
    \caption{MODNAS vs. baselines on optimizing memory usage and perplexity on GPT-L (774M) of HW-GPT-Bench.}
    \label{fig:memory_ppl}
\end{wrapfigure}

In this section, we showcase the application of MODNAS for optimizing memory usage (using Bfloat16 precision and context size of 1024) and perplexity on OpenWebtext within the HW-GPT-Bench~\citep{Sukthanker2024HWGPTBench} GPT-L search space, featuring Transformer models up to 774M parameters. Since memory usage does not depend on the device type, our approach does not utilize the MGD updates in Algorithm~\ref{alg:modnas_algo} for computing the common gradient descent direction, instead leveraging only preference vectors to calculate the scalarized objective. This highlights once again the flexibility of MODNAS across diverse settings, even the ones it was not designed for. Despite this adjustment, MODNAS remains competitive, delivering a Pareto front comparable to leading black-box MOO baselines. We show the results in Figure~\ref{fig:memory_ppl}. 

\subsection{Additional Results on MobileNetV3}
\begin{figure}[ht]
    \centering
    \includegraphics[width=.24\linewidth]{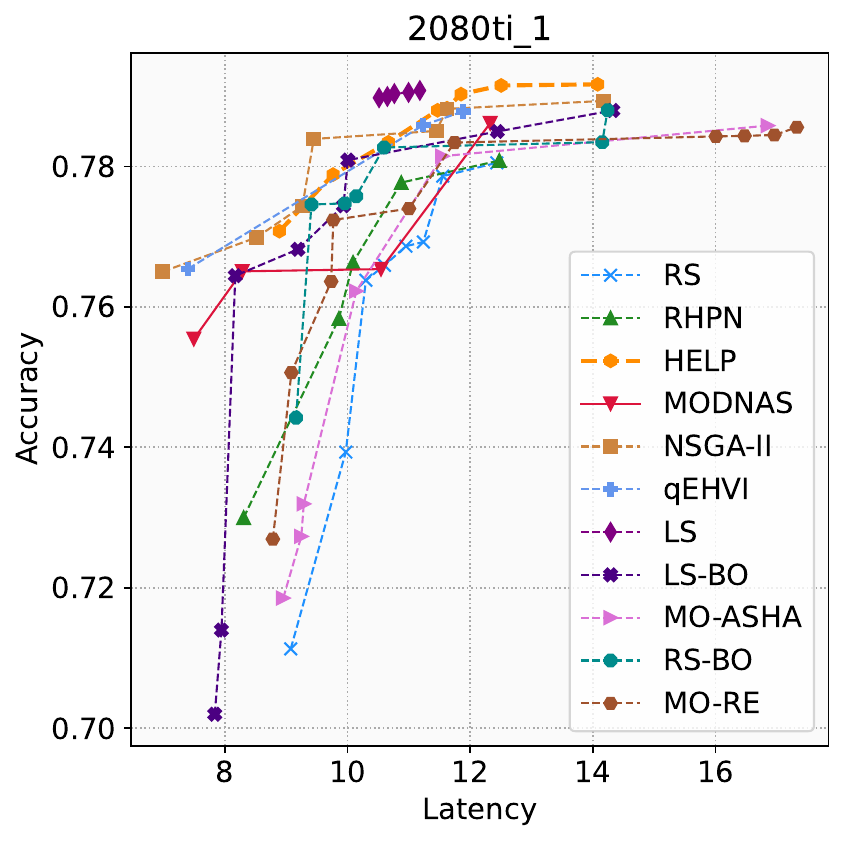}
    \includegraphics[width=.24\linewidth]{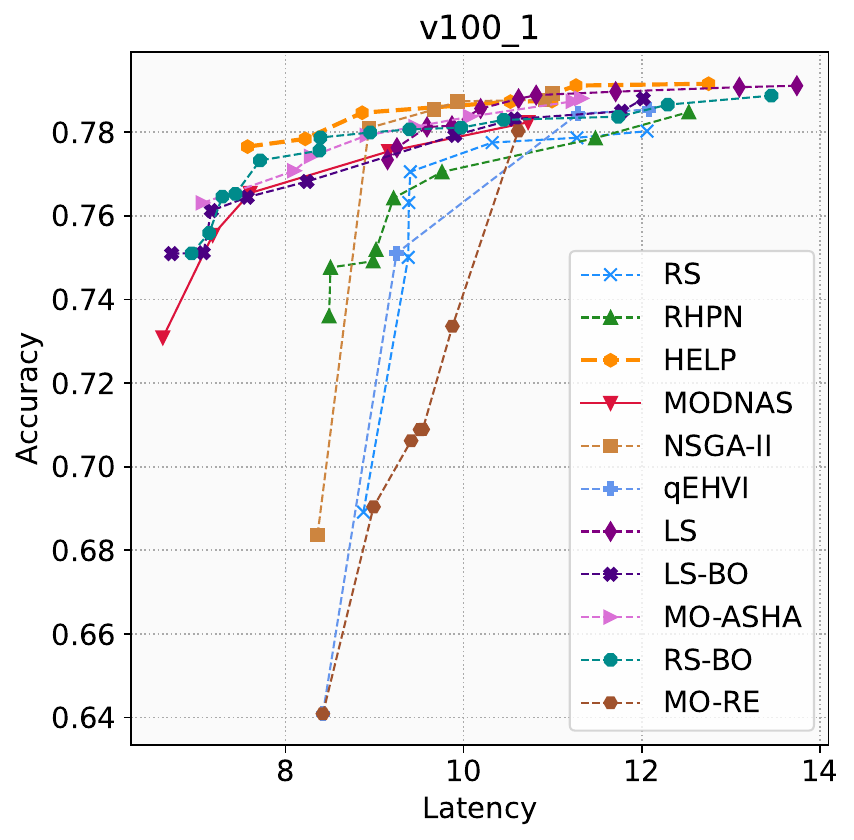}
    \includegraphics[width=.24\linewidth]{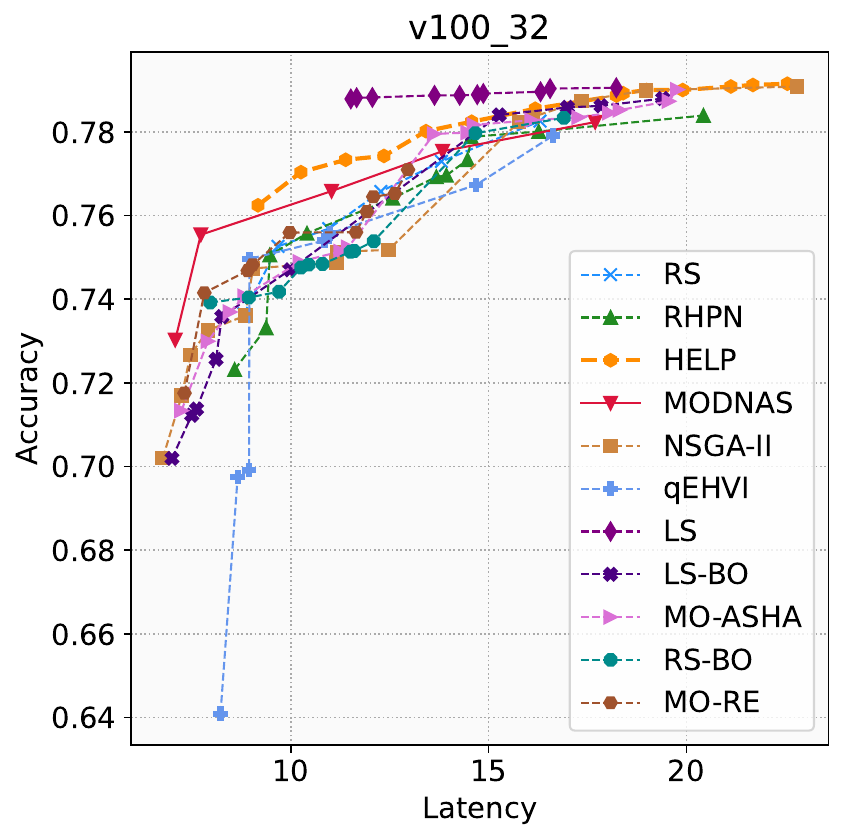}
    \includegraphics[width=.24\linewidth]{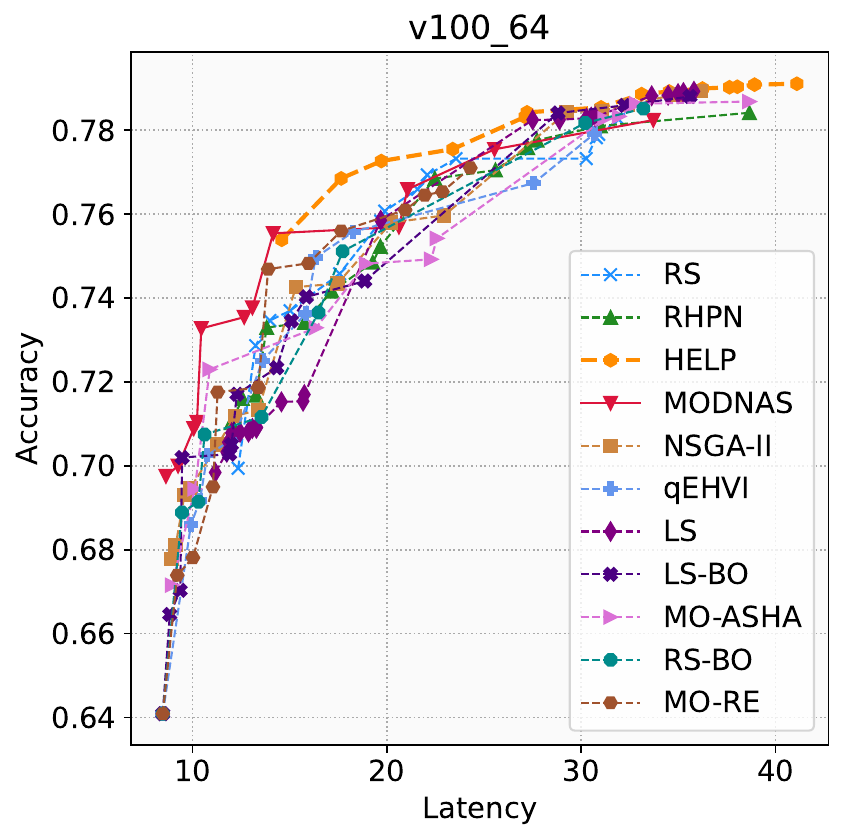}\\
    \includegraphics[width=.24\linewidth]{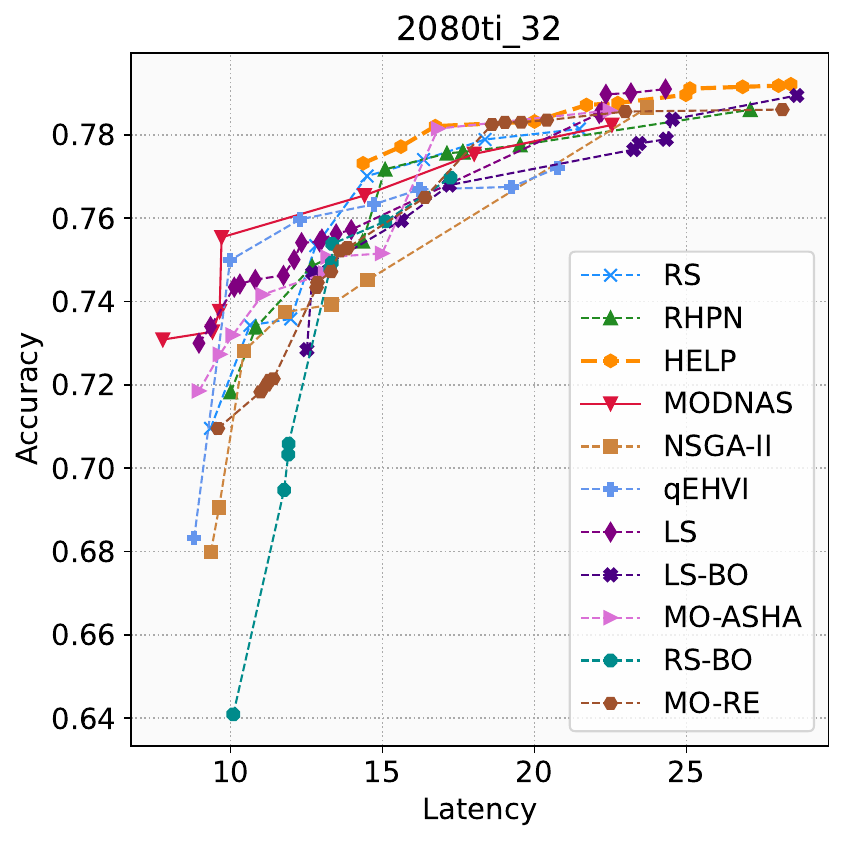}
    \includegraphics[width=.24\linewidth]{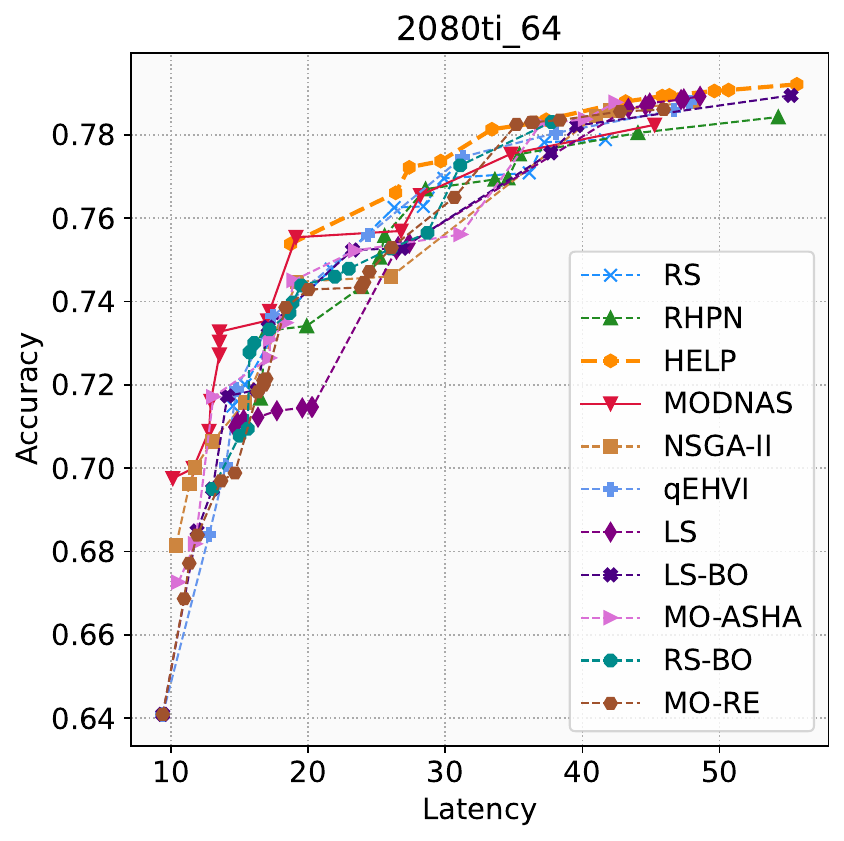}
    \includegraphics[width=.24\linewidth]{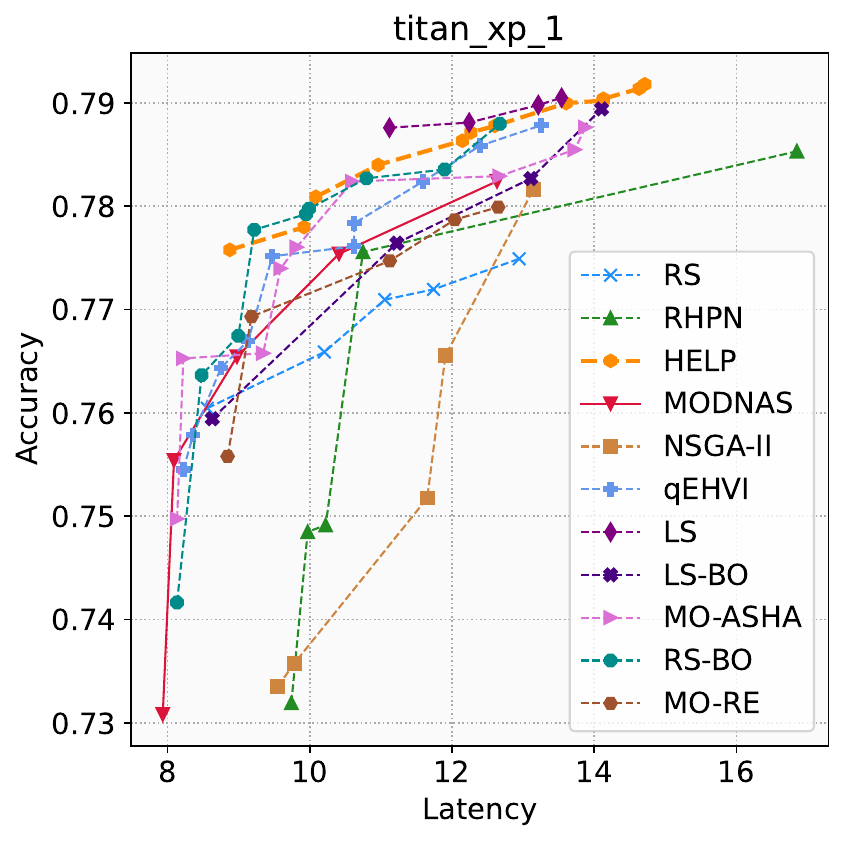}
    \includegraphics[width=.24\linewidth]{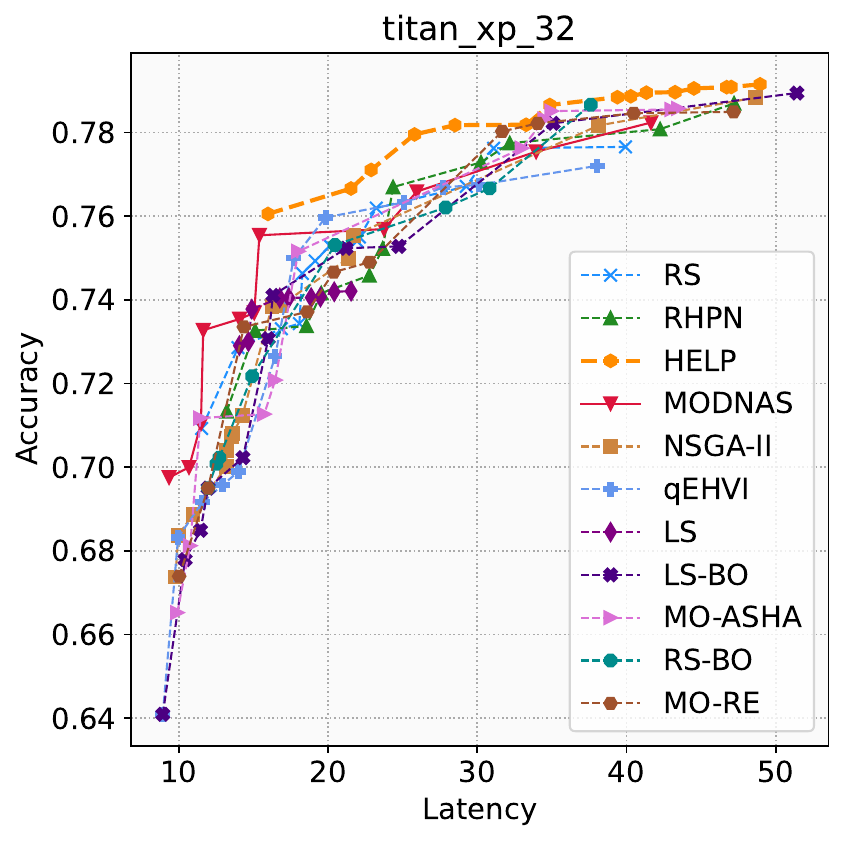}\\
    \includegraphics[width=.24\linewidth]{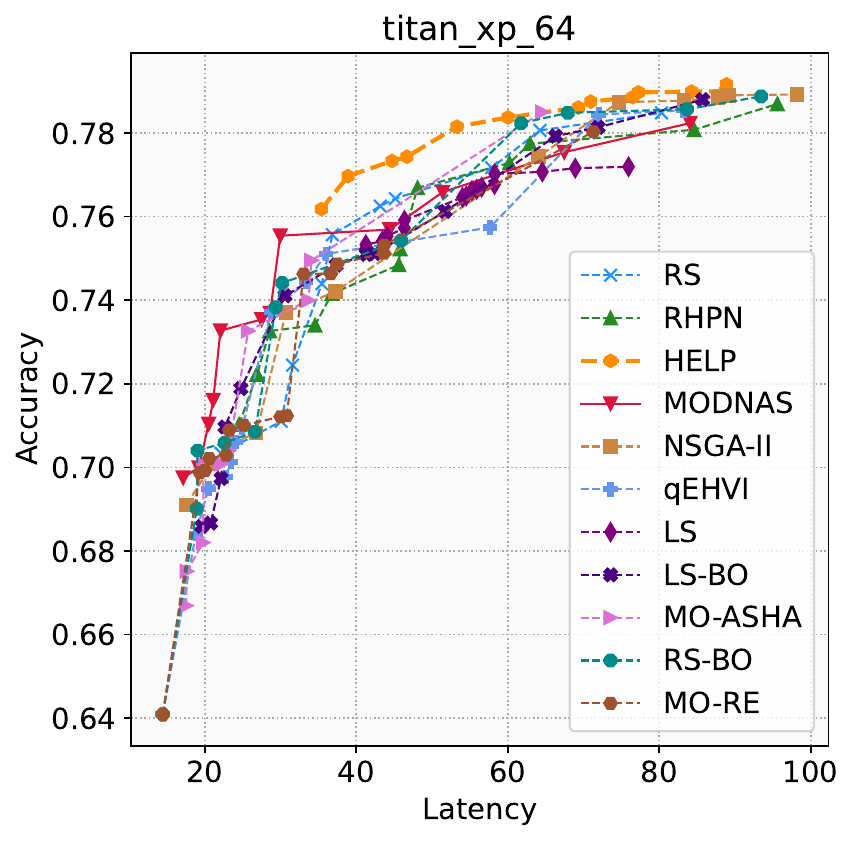}
    \includegraphics[width=.24\linewidth]{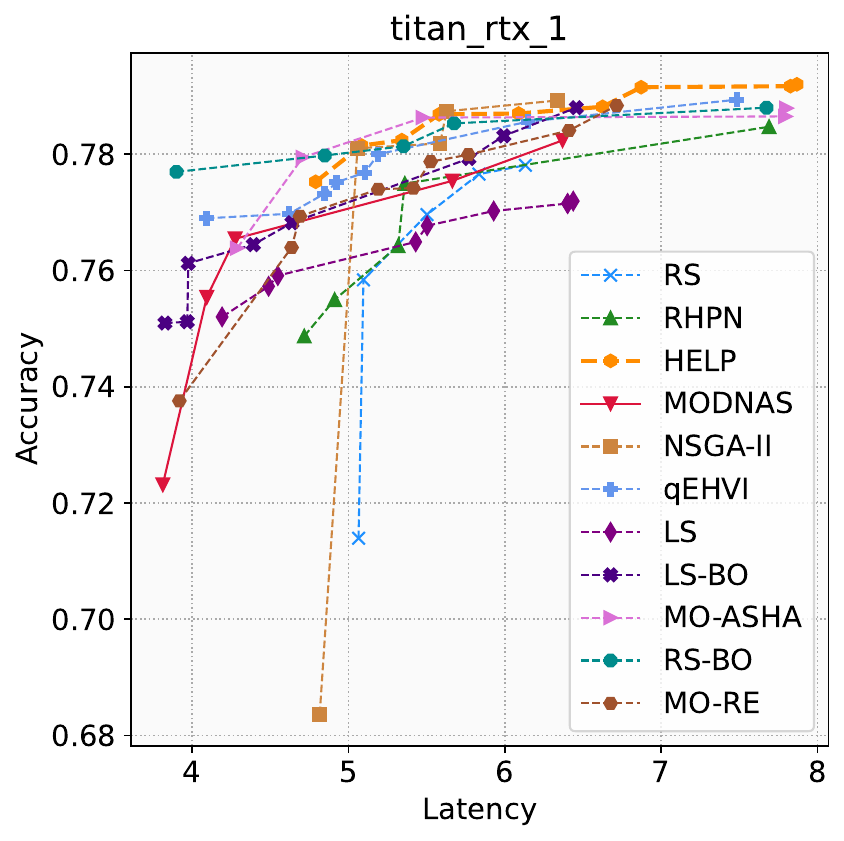}
    \includegraphics[width=.24\linewidth]{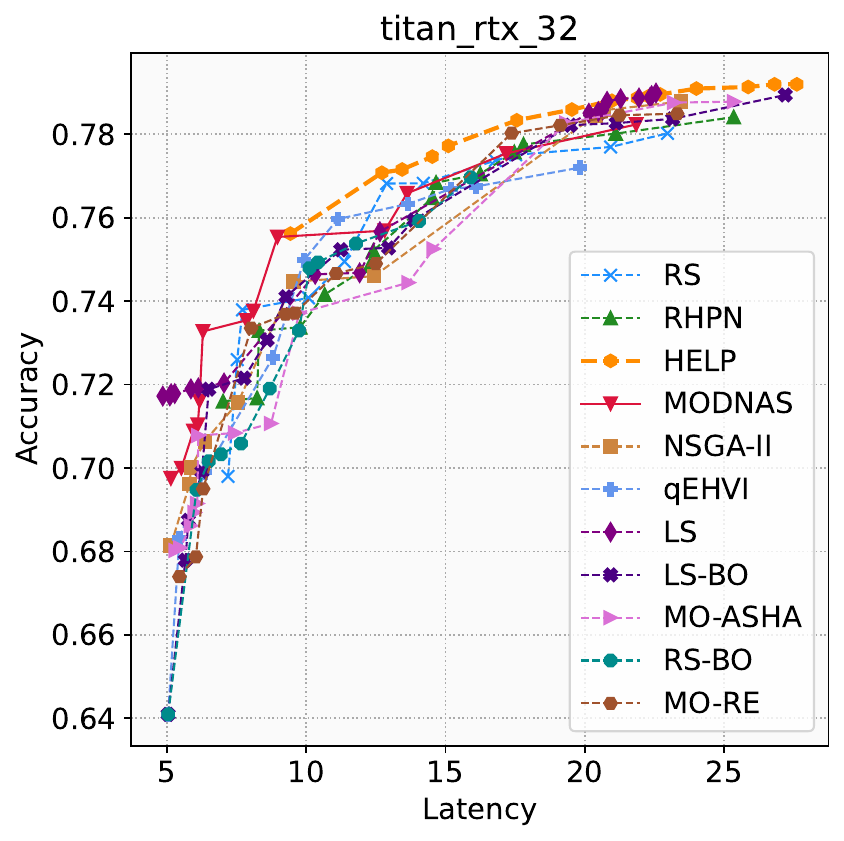}
    \includegraphics[width=.24\linewidth]{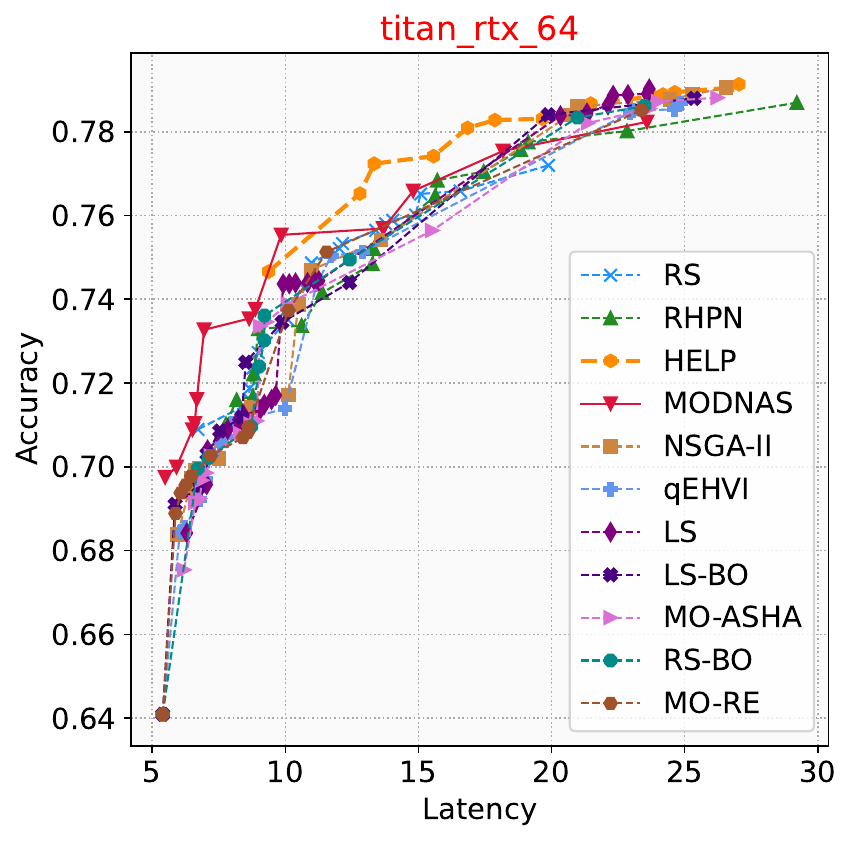}
    \caption{Pareto fronts of MODNAS and baselines on the MobileNetV3 space.}
    \label{fig:pareto_ofa_full}
\end{figure}

In Figure~\ref{fig:pareto_ofa_full}, we present the Pareto fronts of our method compared to different baselines for 12 different hardware devices on the MobileNetV3 space. We show as well the Pareto front of OFA+HELP~\citep{lee2021help}, ran with the original setting.

\begin{table}[ht]
\centering
\caption{HV of MODNAS and baselines on the OFA search space. For every device we optimize for 2 objectives: \emph{latency (ms)} and validation accuracy on ImageNet-1k.}
\vspace{-2mm}
\label{tab:ofa-table-zico}
\resizebox{0.999\linewidth}{!}{%
\begin{tabular}{lcccccccccccc}
\toprule
\textbf{Device Name} & \textbf{RS} & \textbf{RHPN} & \textbf{HELP} & \textbf{EHVI} & \textbf{LS} & \textbf{LS-BO} & \textbf{MO-ASHA} & \textbf{RS-BO} & \textbf{MO-REA} & \textbf{NSGA-II} & \textbf{Zico-NSGA-II} & \textbf{MODNAS} \\
\midrule
v100\_64 & 0.677 & 0.683 & 0.638 & 0.748 & 0.697 & 0.749 & 0.740 & 0.747 & 0.750 & 0.744 & 0.689 & \textbf{0.757} \\
titan\_rtx\_64 & 0.722 & 0.698 & 0.663 & 0.751 & 0.734 & 0.755 & 0.736 & 0.753 & 0.752 & 0.744 & 0.690 & \textbf{0.763} \\
\bottomrule
\end{tabular}
}
\end{table}

\begin{figure}[t!]
    \centering
    \includegraphics[width=.32\linewidth]{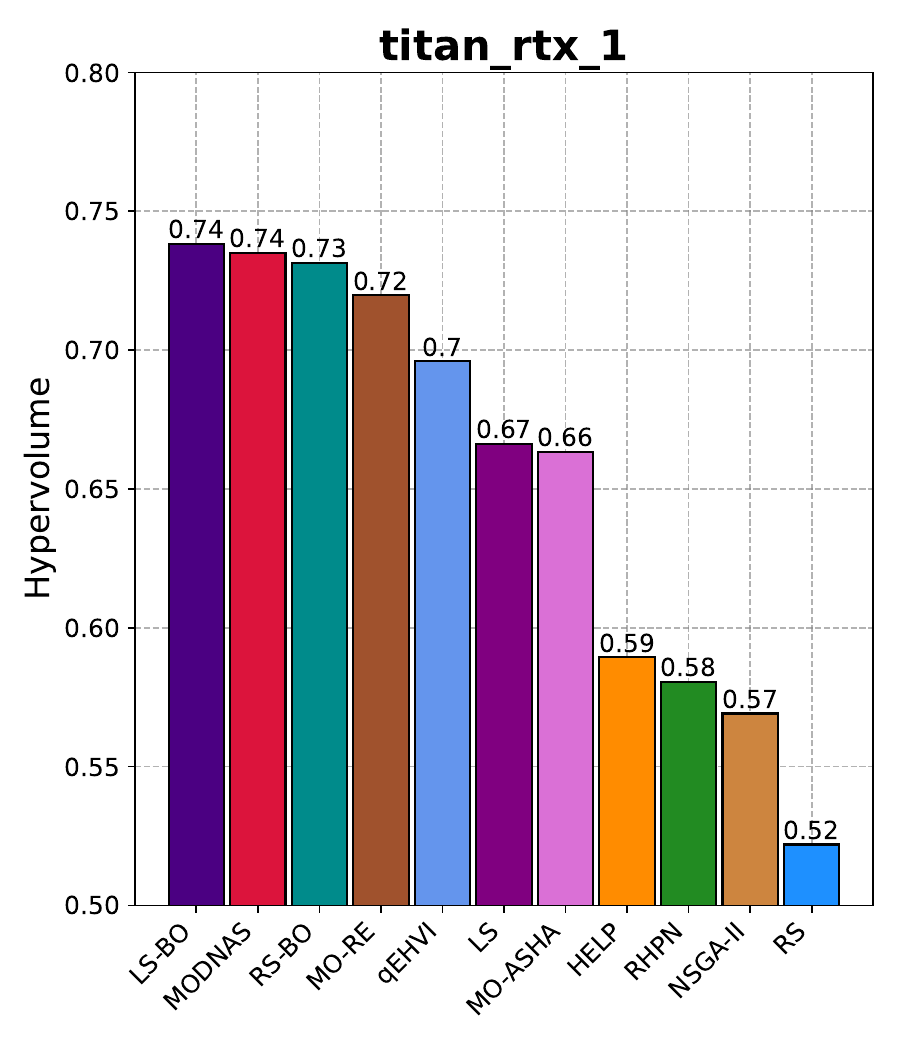}
    \includegraphics[width=.32\linewidth]{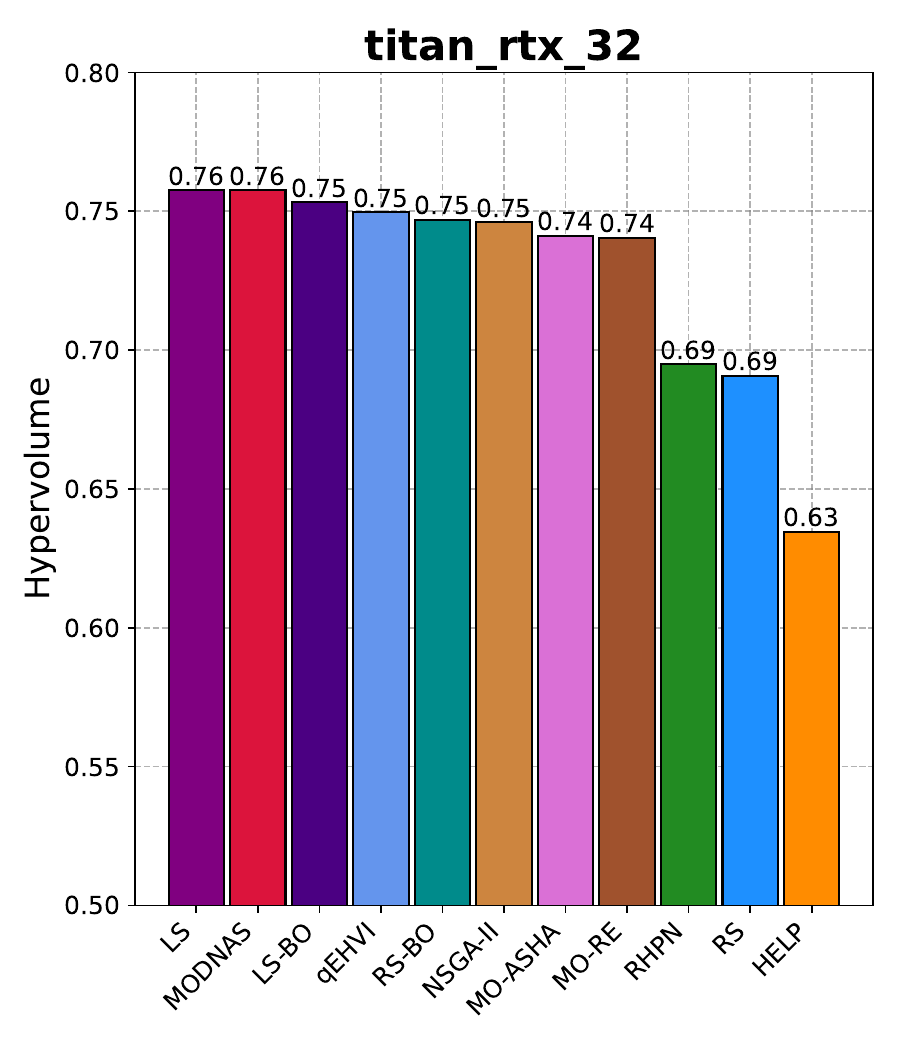}
    \includegraphics[width=.32\linewidth]{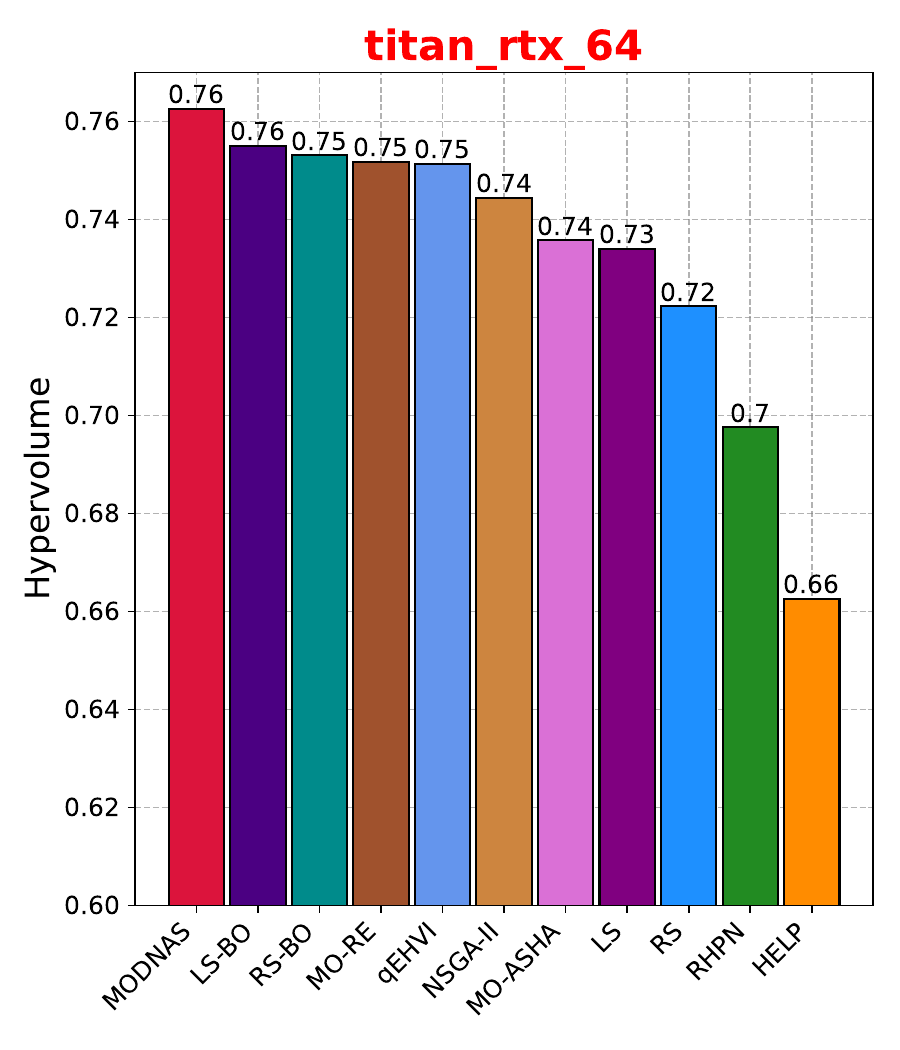}\\
    \includegraphics[width=.32\linewidth]{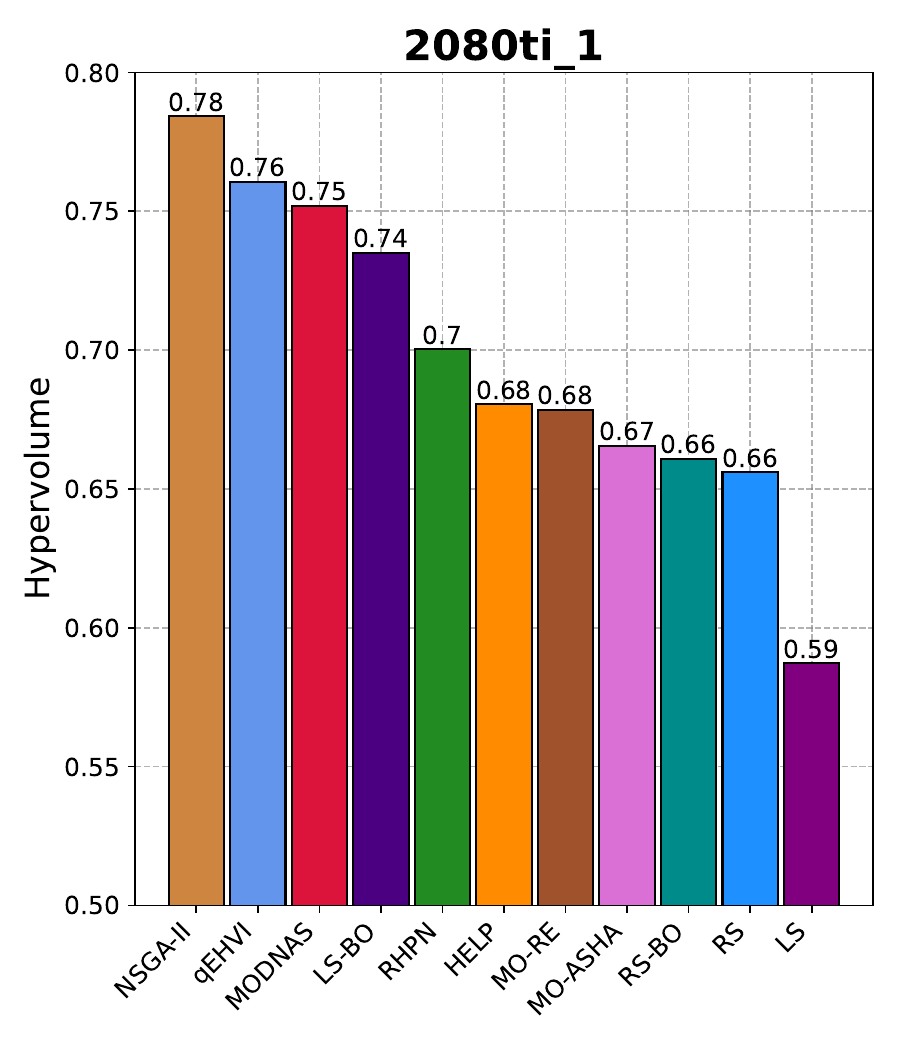}
    \includegraphics[width=.32\linewidth]{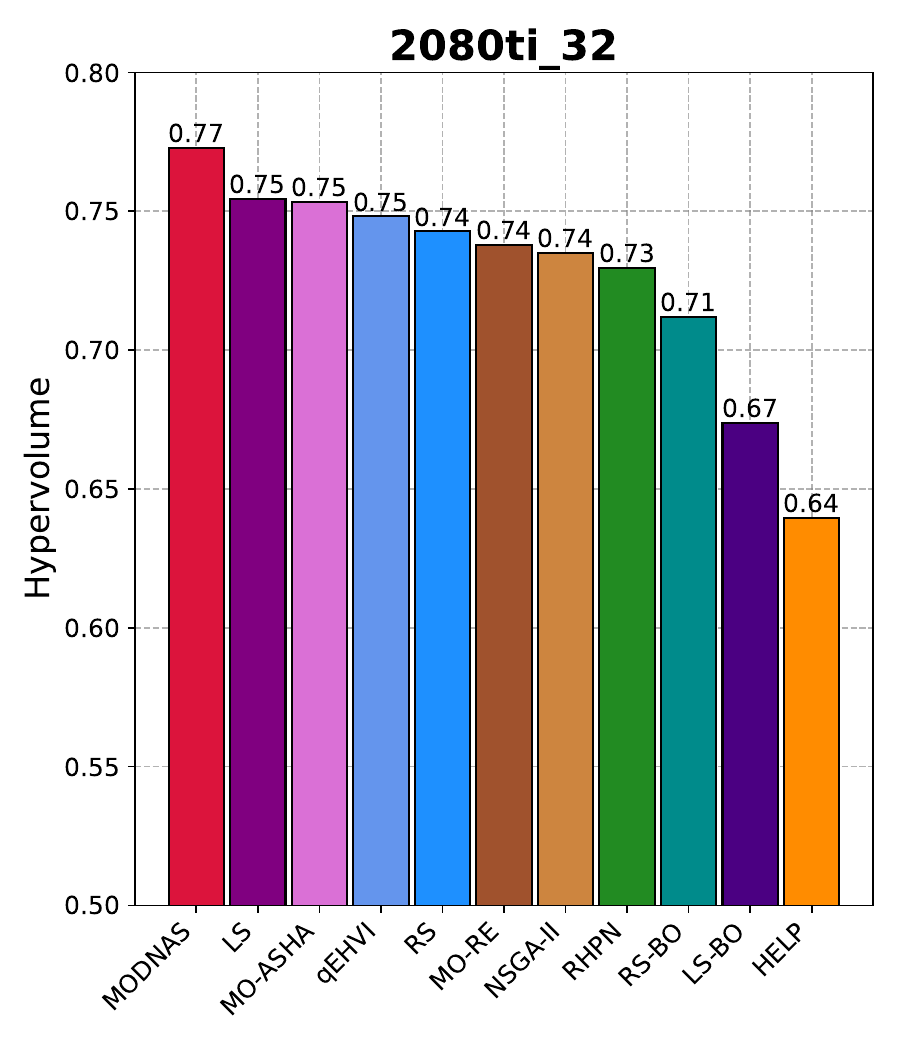}
    \includegraphics[width=.32\linewidth]{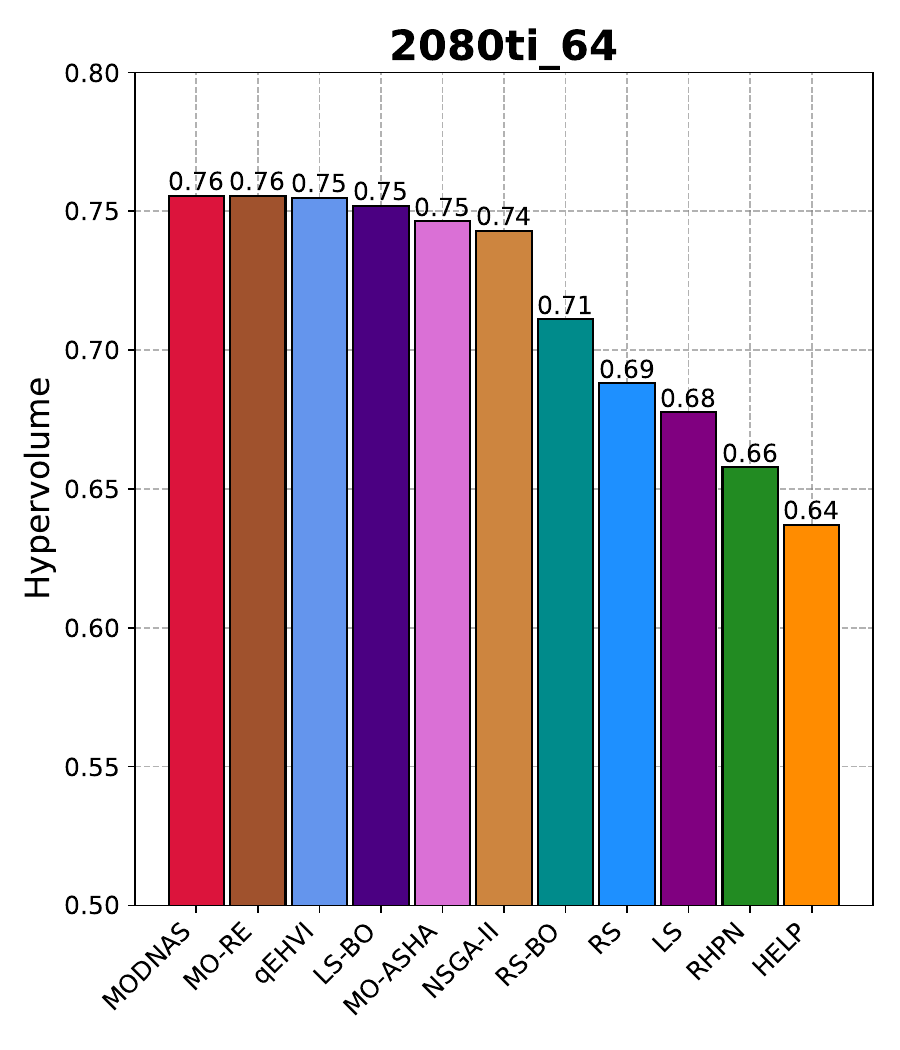}\\
    \includegraphics[width=.32\linewidth]{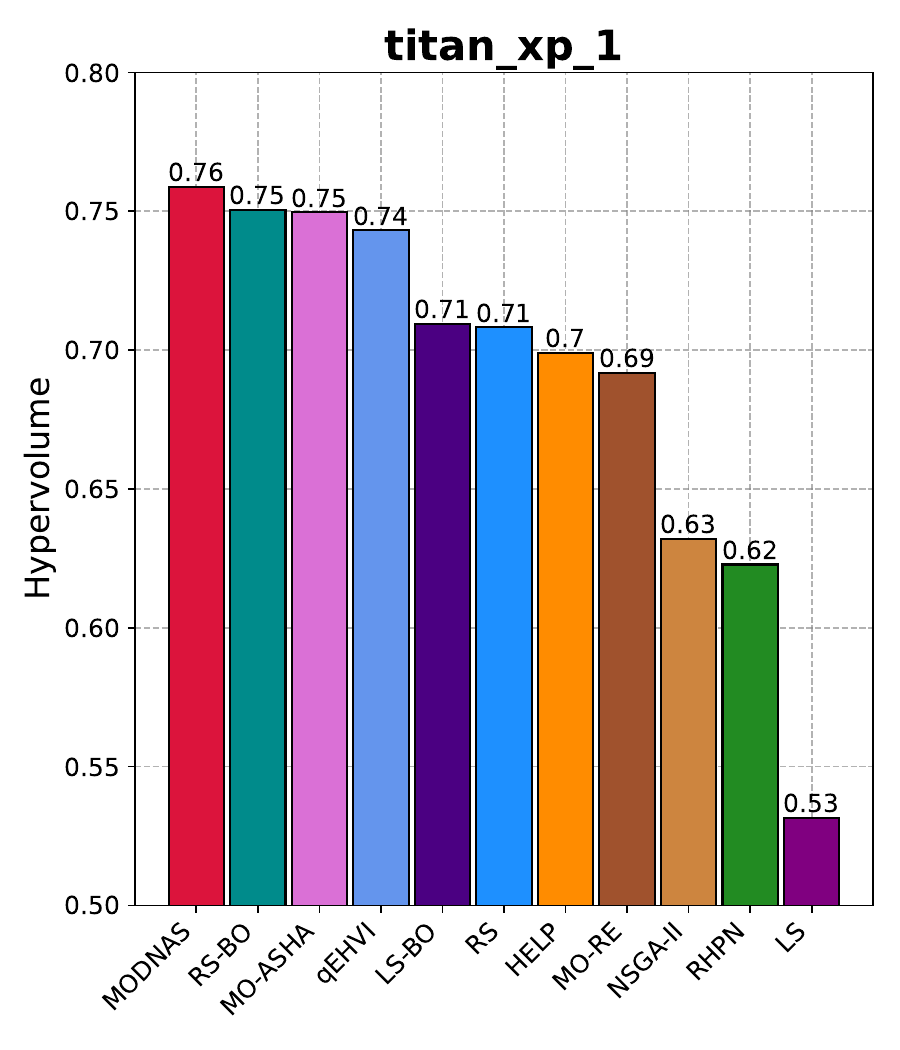}
    \includegraphics[width=.32\linewidth]{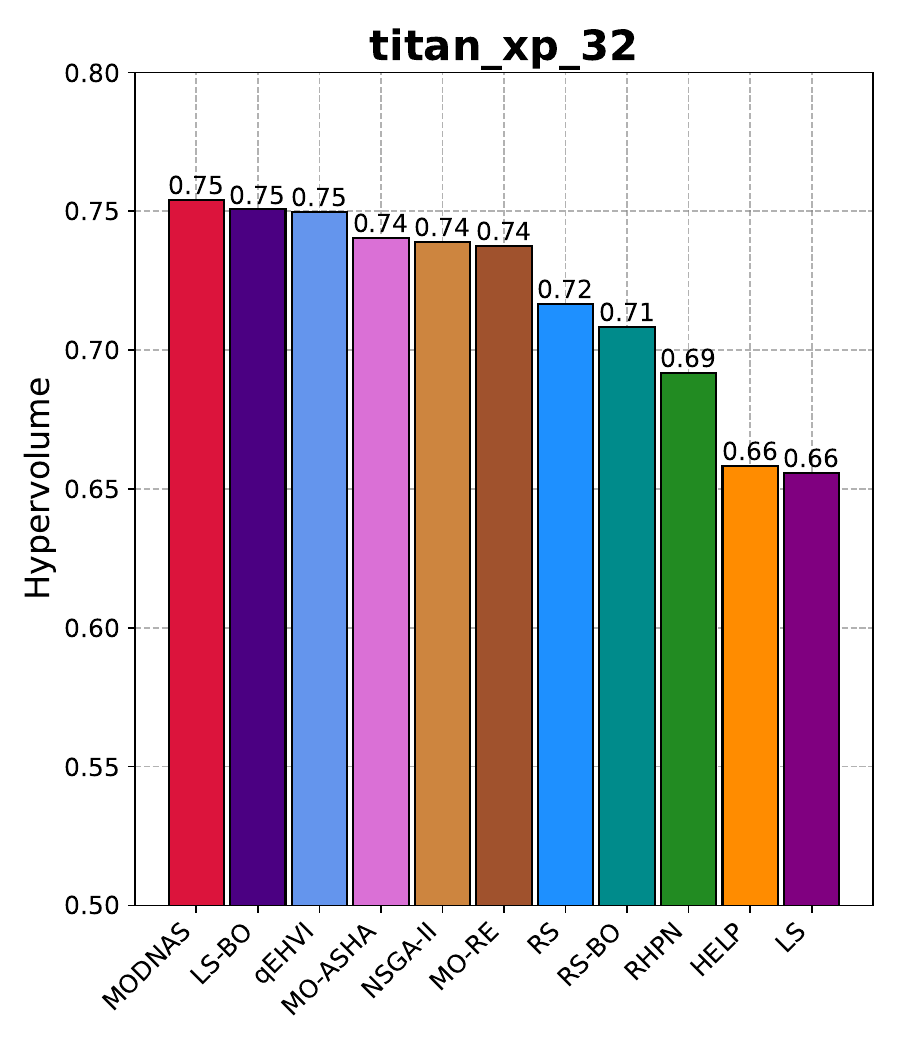}
    \includegraphics[width=.32\linewidth]{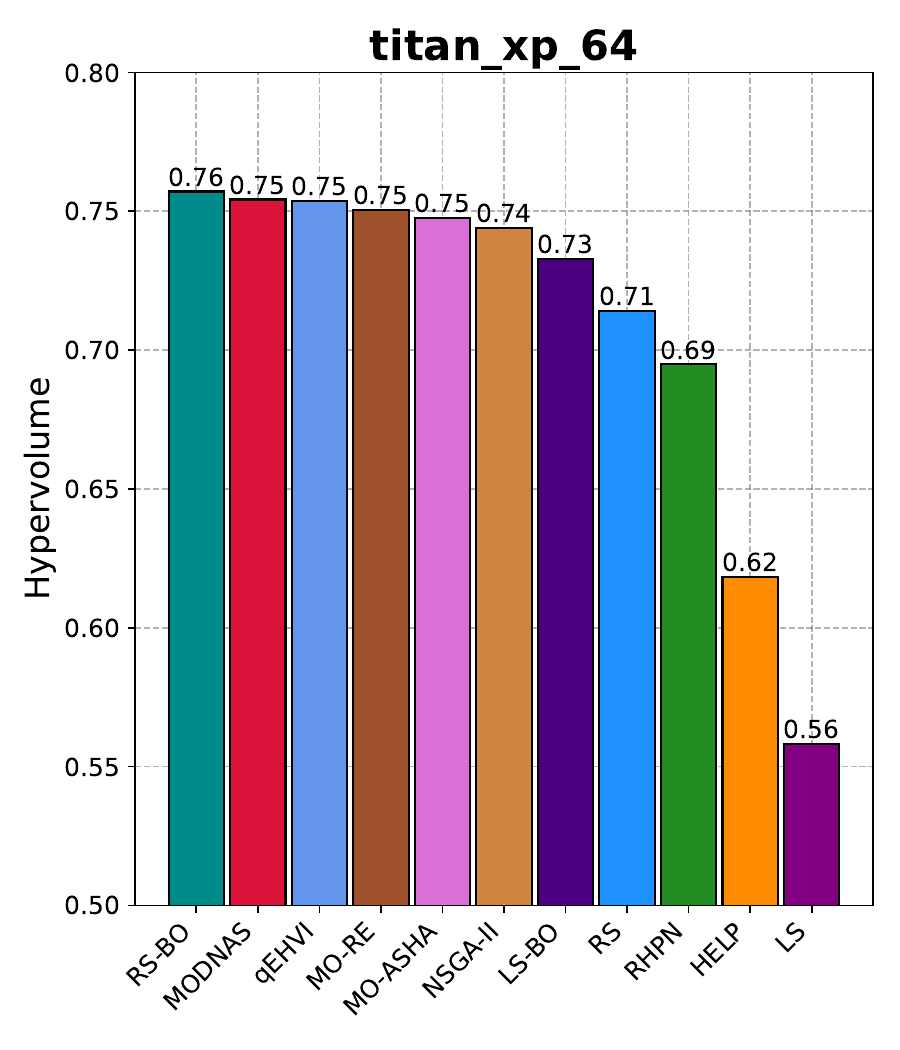}\\
    \includegraphics[width=.32\linewidth]{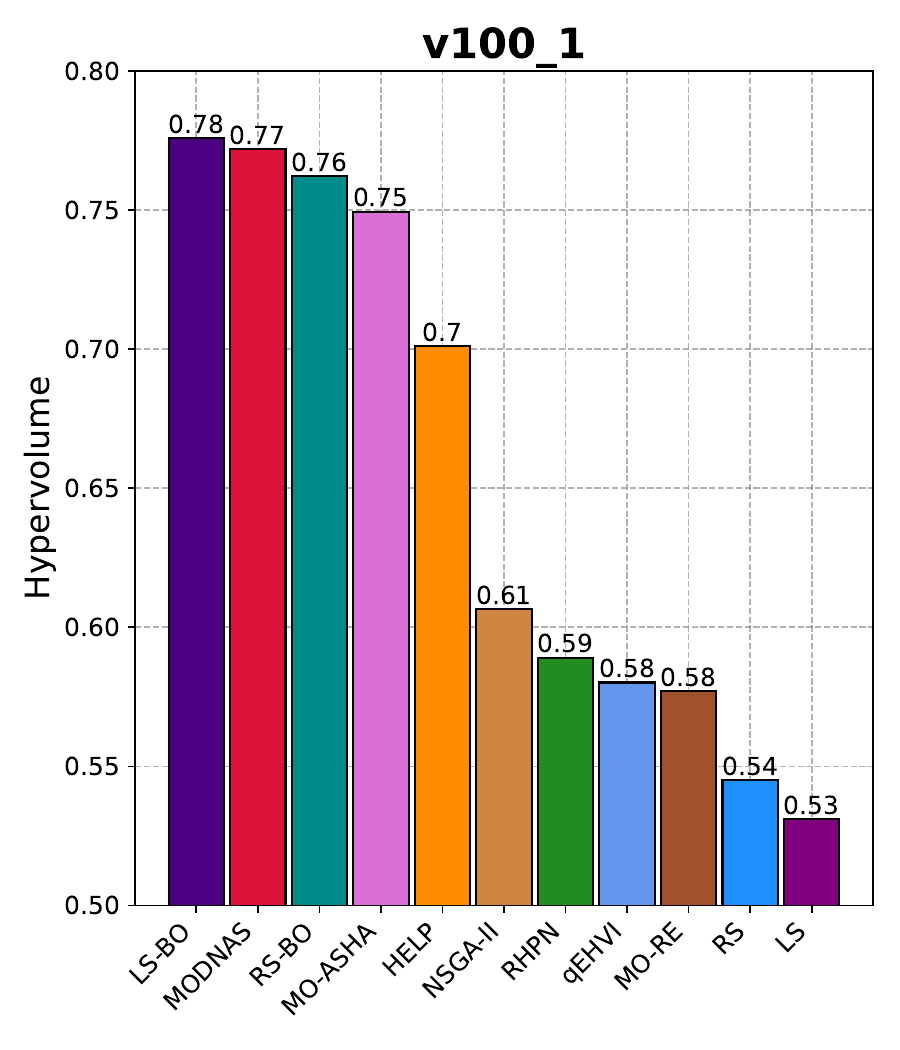}
    \includegraphics[width=.32\linewidth]{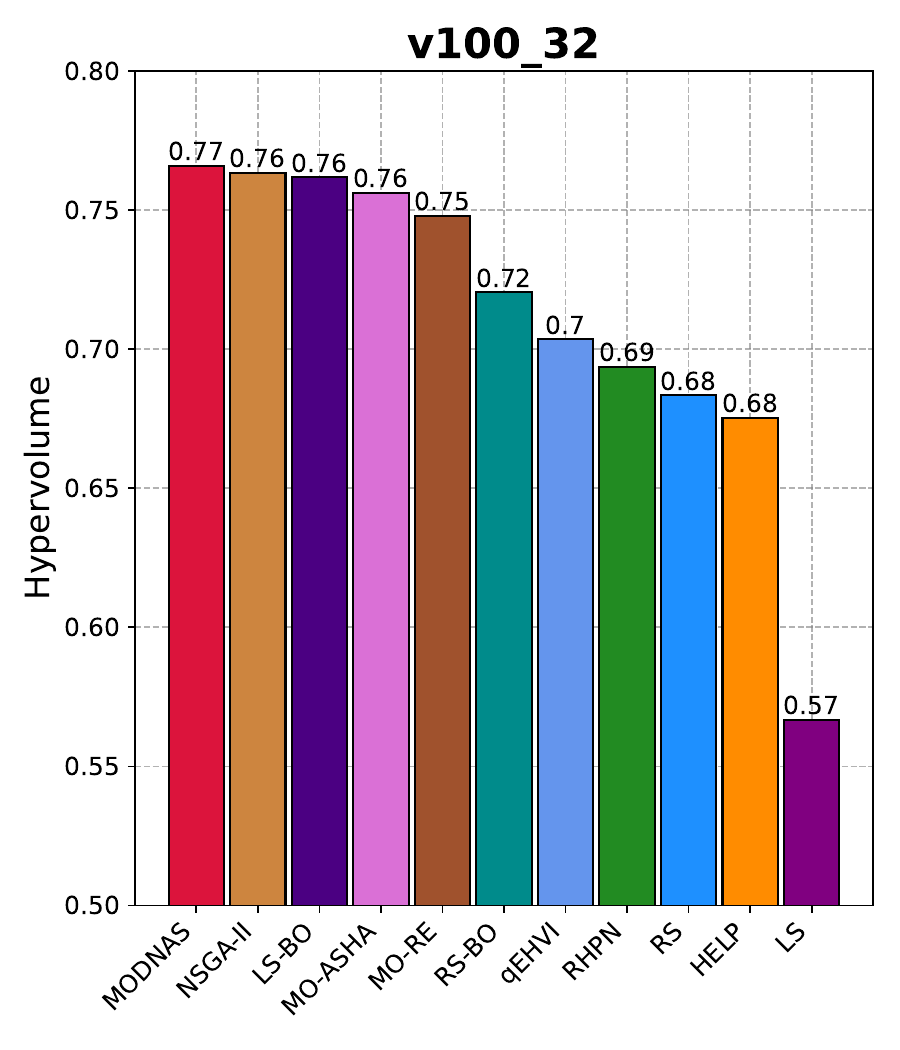}
    \includegraphics[width=.32\linewidth]{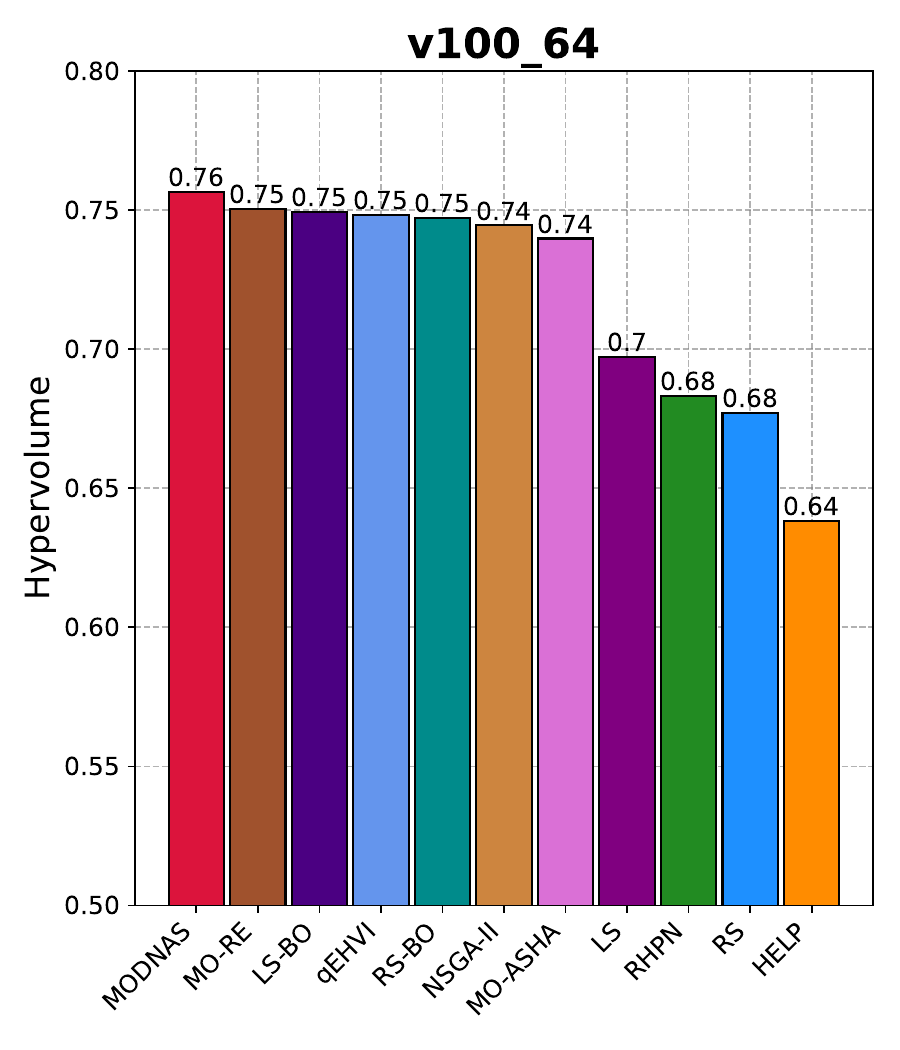}
    \caption{Hypervolume across devices on the MobileNetV3 search space of MODNAS and baselines. Here the Nvidia Titan RTX is the test device.}
    \label{fig:ofa_barplot_all}
\end{figure}

\clearpage
\newpage

\section{Further Discussion on the Robustness of MODNAS}

Initially observed by ~\citep{zela-iclr20a}, differentiable NAS methods can be very sensitive to their hyperparameter choices, especially the regularization ones responsible for the loss landscape in the upper level problem. In our experiments, there were three crucial components that made MODNAS robust and to work reliably across benchmarks:
\begin{enumerate}[leftmargin=*]
    \item \textbf{Choice of \metahypernet update scheme:} this played a pivotal role in the performance of MODNAS. Although other gradient update strategies underperformed or started diverging (Figure~\ref{fig:gradschemes_hv}), MGD converged relatively quickly to a hypervolume close to that of the global Pareto front. The convergence of MGD to a pareto stationary point is discussed in \citet{desideri2012mgd} and more recently in \citet{Zhang2024MGDACU}. The convergence of MGD in bilevel optimization is an open research topic (see recent results from \citet{ye24bilevel} and \citet{Yang2024GradientbasedAF}). One potential scenario when MGD could fail is when the gradient directions of the objectives it is optimizing point in different opposing directions; however, this becomes practically unlikely, especially as the number of objectives grows (in our case we use it to find the common gradient across devices, which is for instance 13 on NAS-Bench-201).
    \item \textbf{Choice of gradient estimation method in the \architect}: In Section~\ref{sec:methodology}, we discuss our choice for the method that enables gradient estimation through discrete variables (since architectures are discrete variables). We noticed that the ReinMax~\citep{liu2023bridging} estimator always outperformed previous estimators such as the one in GDAS~\citep{dong-cvpr19} (Figure~\ref{fig:radar_hv_gdas}), so we believe this choice is crucial.
    \item \textbf{Weight entanglement vs. weight sharing in the \supernet}: In early experiments on NB201 we noticed that weight sharing in the \supernet, was not only more expensive, but much more unstable as well when compared to weight entanglement~\citep{sukthanker2023weight,cai-iclr2020}, even yielding diverging solutions quite often (common pattern seen in differentiable NAS with shared weights as you mention; see \citet{zela-iclr20a} for instance).
\end{enumerate}

We hypothesize that all design choices mentioned above play an implicit regularization effect on the upper level optimization in the bi-level problem, leading to a faster convergence and robustness~\citep{zela-iclr20a, chen2020stabilizing, smith2021on}.

\section{Extended Related Work}
\label{sec:related_work}

\textbf{Multi-objective optimization.}
Multi-objective optimization (MOO) \citep{gunantara2018review} is a crucial field in optimization theory, tackling decision-making scenarios with multiple conflicting objectives. MOO techniques can be categorized into gradient-based and gradient-free approaches. \textit{Gradient-free} MOO approaches, such as evolutionary algorithms and dominance-based methods like NSGA-II \citep{deb2000fast}, often suffer from sample inefficiency and are typically unsuitable for deep learning applications.
On the other hand, \textit{gradient-based} MOO methods leverage gradients. The foundational work by \citet{desideri2012mgd} has been significantly extended in multi-task learning contexts, demonstrating considerable potential \citep{Sener2018MultiTaskLA, lin-neurips19, mahapatra-icml20a, Liu2019TheSM}. However, these methods are primarily applied to fixed architectures, and adapting them to architecture search spaces is complex. This adaptation would require retraining each architecture with multiple objectives, which is impractically expensive for large search spaces. Another major challenge in MOO is balancing the different objectives. To address this, preference vectors have been proposed to guide the prioritization of objectives on the Pareto Front \citep{ye2022pareto, momma2022multi}. An emerging approach to mitigate the retraining issue involves hypernetworks, which determine the weights of the main network in MOO scenarios \citep{Lin2020ControllablePM}, often incorporating preference vector \citep{navon2020learning, hoang2023improving, phan2022stochastic}.

\textbf{Neural Architecture Search.}
A major challenge in the automated design of neural network architectures is the efficient exploration of vast search spaces. Early NAS methods relied on Reinforcement Learning \citep{zoph-iclr17a}, evolutionary algorithms \citep{deb2002fast,lu2020nsganetv2,elsken2018efficient}, and other black-box optimization techniques \citep{daulton2022multi} to train and evaluate numerous architectures from scratch. The advent of one-shot NAS introduced weight sharing among architectures by training an over-parameterized network, known as a supernet, to expedite the evaluation of individual networks within the search space \citep{saxena2016convolutional, bender-icml18a, pham-icml18, liu-iclr19}. Differentiable one-shot NAS methods \citep{wu2019fbnet,cai2018proxylessnas,wu2021trilevel,He2020MiLeNASEN,fu2020autogan} further improved efficiency by applying a continuous relaxation to the search space, enabling the use of gradient descent to identify optimal sub-models within the supernet. In contrast, two-stage NAS methods initially train a supernet, often through random sampling of subnetworks, and subsequently employ black-box optimization to identify optimal subnetworks \citep{bender-icml18a, li-uai20, guo2020single}.
 
\textbf{Hardware-aware and Multi-objective Neural Architecture Search.}
Early NAS methods primarily focused on maximizing accuracy for a given task. In contrast, hardware-aware NAS aims to optimize architectures for efficient performance on specific hardware devices \citep{benmeziane-arxiv21a, zhang2020fast, lee2020s3nas, shaw2019squeezenas}, naturally leading to multi-objective NAS \citep{hsu2018monas,kim2021mdarts,tan-cvpr19}. Two-stage NAS methods can be adapted to this context by incorporating a multi-objective search in the second stage \citep{cai2018proxylessnas,ito2023ofa}. However, most two-stage methods depend on random sampling during supernet training, which doesn't prioritize promising architectures.
Differentiable NAS methods, such as those in \citet{wu2019fbnet,cai2018proxylessnas,wu2021trilevel,fu2020autogan, xu2020latency, jiang2021eh, wang2021attentivenas}, use latency proxies like layer-wise latencies and FLOPS \citep{dudziak2020brp} to evaluate hardware performance, combining task and hardware objectives with fixed weighting to find a single optimal solution. However, changing the objective weighting requires a complete search rerun, which is computationally demanding.

In contrast, our proposed search algorithm offers the entire Pareto Front of objectives in a single run, making it more efficient. While our focus is on multi-objective NAS for hardware constraints, our technique is applicable to other objectives such as fairness \citep{martinez2020minimax, dooley2023rethinking, das2023fairer}, suggesting promising avenues for future research. 

\paragraph{Application to Other Tasks.}
Finally, we want to briefly discuss the potential application of MODNAS to other tasks not mentioned in this work. Object detection is a very important application since it is probably one of the most important use cases of neural networks on embedded devices (e.g., in self-driving cars)~\citep{Lazarevich2023YOLOBenchBE,lee23virtuoso,Baller2021DeepEdgeBenchBD}. An interesting benchmark is YOLOBench~\citep{Lazarevich2023YOLOBenchBE}, where the authors benchmark more than 550 architectures on four datasets and four different hardware platforms. One way to leverage the \supernet here would be to parameterize the search space via the AutoDeepLab~\citep{Liu2019AutoDeepLabHN} supernetwork model, which parameterizes the resolution too. This parameterization would not only leverage MODNAS to work for object detection, but for other computer vision tasks such as semantic segmentation, disparity estimation, etc~\citep{mohan23survey}.

\end{document}